\documentclass[final,5p,times,twocolumn]{elsarticle}
\usepackage[british]{babel}

\usepackage{amsmath}
\usepackage{amssymb}
\usepackage{balance}
\usepackage{amsfonts}
\usepackage{mathrsfs}

\usepackage{graphicx}
\usepackage{epstopdf}
\usepackage{float}

\usepackage[figuresright]{rotating}

\usepackage{adjustbox}

\usepackage{multirow}

\usepackage{booktabs}

\usepackage{longtable}

\usepackage{array}

\usepackage{threeparttable}

\usepackage{diagbox}

\usepackage{caption}

\usepackage{subcaption}
\usepackage{cuted}

\usepackage[section]{placeins}

\usepackage{algorithm}

\usepackage{algorithmicx}

\usepackage{algpseudocode}

\usepackage{enumitem}
\usepackage{comment}

\usepackage{listings}

\usepackage{url}

\usepackage[hidelinks]{hyperref}

\journal{}

\usepackage{cuted}

\makeatletter
\AtBeginDocument{%
	\@ifpackageloaded{balance}{%
		\global\let\@@utputdblcol\@outputdblcol
	}{}%
}
\makeatother

\begin{document}

	
	\begin{frontmatter}
		
		
		\title{IWOO: Wave Optics Optimizer Improved with Hybrid Strategies for Numerical Optimization and Engineering Applications}

		
		\author[aff1]{Rui Du}
		
		\author[aff1]{Yaming Zhang\corref{cor1}}
		\ead{zhangym333@163.com}
		
		\author[aff1]{Yongpeng Wen}
		
		\author[aff1]{Linghong Cai}
		
		\author[aff1]{Nanqing Liu}

		
		\cortext[cor1]{Corresponding author}

		
		\affiliation[aff1]{
			organization={School of Information, Yunnan Normal University},
			addressline={No.~768 Juxian Street, Chenggong District},
			city={Kunming},
			postcode={650500},
			state={Yunnan},
			country={China}
		}

		
		\begin{abstract}
			The Wave Optics Optimizer (WOO) introduces the diffraction and interference phenomena of light into the optimization process and has demonstrated strong competitiveness in benchmark testing and engineering applications. Nevertheless, WOO still suffers from several limitations, including slow convergence, susceptibility to local optima, and degraded search performance in the later stages of optimization under different problem conditions. To address these issues, this paper proposes an Improved Wave Optics Optimizer (IWOO), in which multiple strategies are collaboratively integrated to enhance the overall optimization performance. First, the integration of the good point set and opposition-based learning introduces the GPS--OBL initialization strategy to improve the quality and diversity of the initial population. Second, incorporating an E-DM nonlinear perturbation mechanism into the temporary triangle-based search mechanism enables the construction of the TriEDM multi-path search strategy, thereby strengthening the search capability of the algorithm. Third, we propose the Correlation-Guided Groupwise Orthogonal Learning (CGOL) strategy, which exploits correlations among variables for groupwise orthogonal learning, thereby improving search efficiency while reducing computational overhead. Finally, we design an adaptive stagnation-aware Runge--Kutta correction strategy to reduce the risk of the algorithm becoming trapped in local optima. To comprehensively evaluate the performance of IWOO, experiments are conducted on the IEEE CEC2019, CEC2022, and CEC2017 benchmark suites. IWOO is compared with representative and high-performing optimizers, and ablation experiments are conducted to further verify the effectiveness of each improvement strategy. Experimental results show that the Friedman ranks obtained by IWOO on CEC2019, CEC2022, and CEC2017 are 1.8000, 1.5833, and 1.4483, respectively, within the representative algorithm group, demonstrating strong competitiveness across different benchmark suites. Furthermore, IWOO is applied to six practical engineering optimization problems to evaluate its capability in solving real-world constrained optimization tasks. The results demonstrate that IWOO achieves favorable overall performance in terms of numerical optimization, stability, and engineering applicability.
		\end{abstract}

		

	\end{frontmatter}

	
	\section{Introduction}
	
\begin{figure}[t] 
		\centering 
		\includegraphics[ 
		width=\linewidth, 
		trim=250 100 250 50, 
		clip 
		]{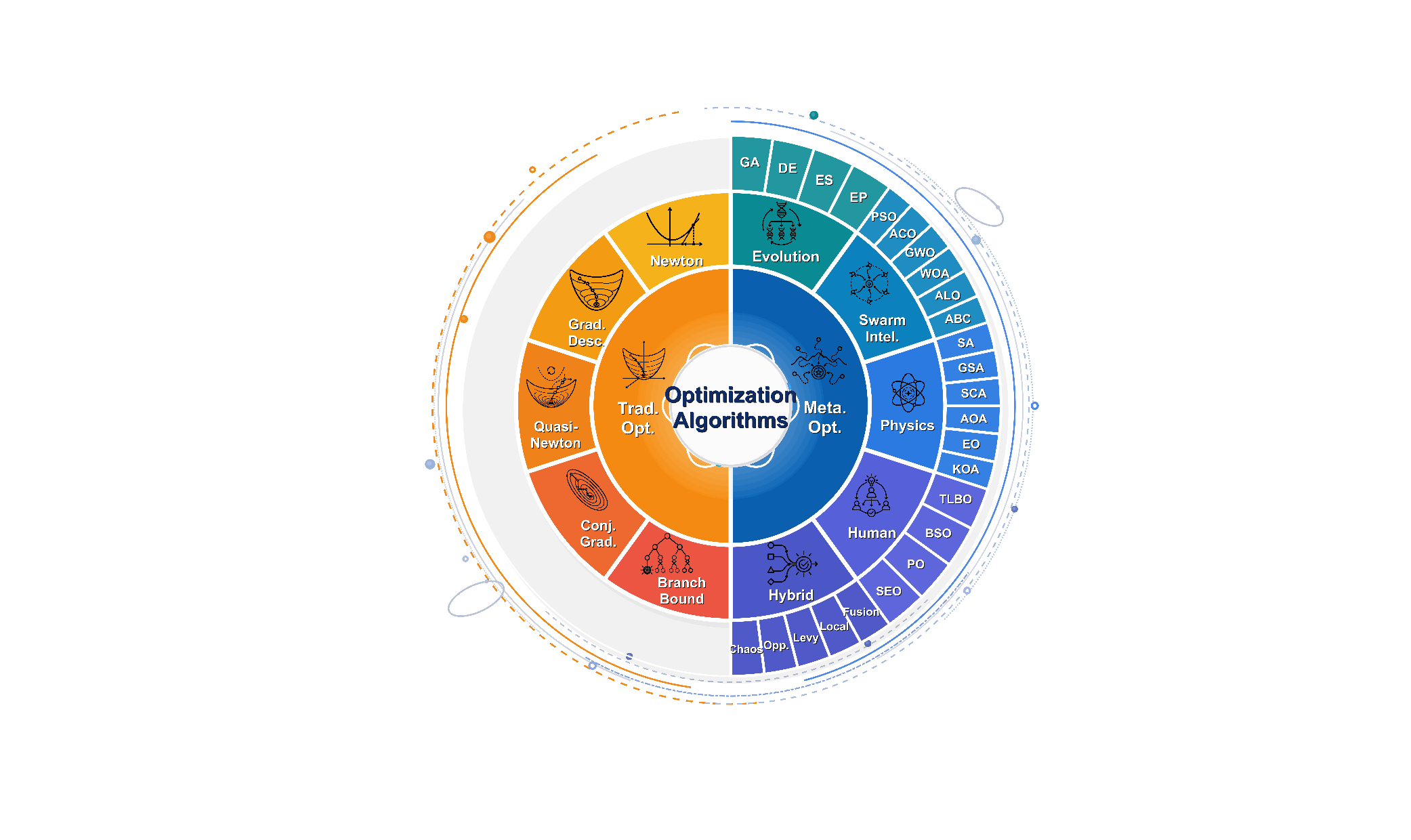} 
		\caption{Taxonomy of optimization algorithms} 
		\label{fig:al_summary} 
	\end{figure}

	Optimization algorithms have been extensively investigated and widely applied as effective tools for solving complex optimization problems across numerous scientific and engineering fields~\cite{Yang2025BeamSlab,Azizi2022CryStAl,Yin2025PathPlanning,Samra2014LicensePlate,Nuaekaew2017ReactivePower,Yang2026Scheduling}. According to their solution mechanisms and dependence on model information, existing optimization methods can be broadly classified into traditional optimization algorithms and metaheuristic optimization algorithms, as illustrated in Fig.~\ref{fig:al_summary}. Traditional optimization algorithms mainly include Newton's method, gradient descent, quasi-Newton methods, conjugate gradient methods, and branch-and-bound methods, and generally exhibit high computational efficiency when dealing with continuous, smooth, and well-structured problems. However, when addressing high-dimensional, nonlinear, multimodal, non-differentiable, and complex constrained problems, their applicability is often limited by the requirements for gradient information or explicit mathematical models~\cite{Zelinka2012HandbookOptimization}. In contrast, metaheuristic algorithms generally do not rely on gradient information of the objective function or exact mathematical models. According to their search mechanisms and sources of inspiration, they can be further classified into evolutionary-based, swarm intelligence-based, physics-based, human-based, and hybrid or improved algorithms~\cite{Hussain2019MetaheuristicSurvey}. Representative algorithms include the Genetic Algorithm (GA)~\cite{Forrest1996GeneticAlgorithms}, Differential Evolution (DE)~\cite{Storn1997DifferentialEvolution}, Simulated Annealing (SA)~\cite{Bertsimas1993SimulatedAnnealing}, Particle Swarm Optimization (PSO)~\cite{Kennedy1995ParticleSwarmOptimization}, Grey Wolf Optimizer (GWO)~\cite{Mirjalili2014GreyWolfOptimizer}, and Kepler Optimization Algorithm (KOA)~\cite{AbdelBasset2023KOA}. Owing to their flexible structures, ease of implementation, and broad applicability, metaheuristic algorithms have become important tools for solving complex optimization problems. 
	
	The Wave Optics Optimizer (WOO)~\cite{Peng2026WOO}, proposed by Peng et al., is a metaheuristic optimization algorithm inspired by two wave-optics phenomena: Fraunhofer single-slit diffraction and optical interference. WOO models population individuals as propagating light waves and constructs its search update strategies by simulating the diffraction and interference effects of light, thereby enabling global exploration and local exploitation of the search space. Its effectiveness has also been experimentally validated on engineering optimization problems, including welded beam and tension/compression spring design, where it exhibits favorable optimization performance. 
	
	To address the unsupervised feature selection problem for unbalanced incomplete multi-view data, Cai et al. proposed the Fractal Wave Optics Optimizer (FWOO), which enhances the search performance of WOO by introducing fractal phase-shift perturbation and an adaptive light-intensity attenuation update mechanism~\cite{Cai2026WaveOpticsTensor}. This study demonstrates the extensibility of WOO to specific learning tasks. However, FWOO is primarily designed for multi-view feature selection, leaving room for further systematic improvements to WOO for general continuous optimization problems. 
	
	Despite its favorable global exploration and local exploitation capabilities, WOO still suffers from several limitations, including relatively slow convergence, susceptibility to local optima, and declining population diversity in the later stages of the search. Therefore, systematic improvements based on the underlying search mechanism of WOO remain necessary. 
	
	To address these limitations, this paper proposes an Improved Wave Optics Optimizer (IWOO), which enhances the original WOO from four complementary aspects. The main contributions are summarized as follows: 
	
	\begin{enumerate} 
		\item GPS--OBL initialization combines the good point set with opposition-based learning to improve the spatial coverage uniformity, quality, and diversity of the initial population. 
		
		\item TriEDM multi-path search incorporates the E-DM nonlinear perturbation mechanism into the temporary triangle-based search mechanism to expand the available search directions and maintain population diversity. 
		
		\item Correlation-Guided Groupwise Orthogonal Learning (CGOL) exploits correlations among decision variables to strengthen collaborative exploitation, improve search accuracy, and reduce the function-evaluation overhead of orthogonal learning. 
		
		\item Adaptive Stagnation-Aware Runge--Kutta (ASRK) correction performs multi-stage directional adjustments on stagnated individuals to enhance fine-grained exploitation and alleviate persistent stagnation during the middle and later stages of the search. 
	\end{enumerate} 
	
	The remainder of this paper is organized as follows. Section~2 introduces the fundamental principles of the Wave Optics Optimizer. Section~3 presents the proposed Improved Wave Optics Optimizer in detail. Section~4 presents the experimental results and discussion. Section~5 investigates the application of IWOO to engineering optimization problems. Finally, Section~6 concludes the paper and discusses future work.

	\section{Principles of the Wave Optics Optimizer}
	\label{sec:woo}
	
	The Wave Optics Optimizer (WOO) is a physics-based metaheuristic algorithm inspired by Fraunhofer single-slit diffraction and optical interference~\cite{Peng2026WOO}. In WOO, each parallel light wave corresponds to a candidate solution. The diffraction effect is used to simulate the diffusion of candidate solutions in the search space, while the interference effect and variations in light intensity regulate information exchange among population individuals. Consequently, the search process gradually transitions from broad global exploration to local exploitation in promising regions.
	
	For a continuous minimization problem, the mathematical model can be formulated as
	\begin{equation}
		\min_{\boldsymbol{x}\in\Omega} f(\boldsymbol{x}),\qquad
		\Omega=
		\left\{
		\boldsymbol{x}\in\mathbb{R}^{D}
		\,\middle|\,
		L_i\leq x_i\leq U_i,\;
		i=1,\ldots,D
		\right\}
	\end{equation}
	where $f(\boldsymbol{x})$ denotes the objective function, $\Omega$ is the search space, and $\boldsymbol{x}\in\mathbb{R}^{D}$ is a $D$-dimensional decision vector. The lower- and upper-bound vectors of the search space are denoted by
	$\boldsymbol{L}=[L_1,\ldots,L_D]^{\top}$
	and
	$\boldsymbol{U}=[U_1,\ldots,U_D]^{\top}$,
	respectively.
	
	Let the population size be $N$. At the $t$-th iteration, the position vector of the $i$-th individual is denoted by
	\begin{equation}
		\boldsymbol{X}_{i}(t)
		=
		\left[
		x_{i,1}(t),
		x_{i,2}(t),
		\ldots,
		x_{i,D}(t)
		\right]^{\top},
		\qquad
		i=1,\ldots,N
		\label{eq:woo_individual}
	\end{equation}
	where $i$ denotes the individual index. The fitness value of
	$\boldsymbol{X}_{i}(t)$ is defined as
	$F_i(t)=f(\boldsymbol{X}_{i}(t))$.
	The global best solution obtained up to the $t$-th iteration and its corresponding fitness value are denoted by
	$\boldsymbol{X}_{\mathrm{best}}(t)$
	and
	$F_{\mathrm{best}}(t)=f(\boldsymbol{X}_{\mathrm{best}}(t))$,
	respectively.
	
	\subsection{Optical model and diffraction-based initialization}
	\label{subsec:woo_optical_model}
	
	In Fraunhofer single-slit diffraction, the secondary waves originating from different positions within the slit are superposed, forming alternating bright and dark diffraction fringes on the observation screen. The $m$-th dark fringe and the $m$-th secondary bright fringe in single-slit diffraction approximately satisfy
	\begin{align}
		d\sin\theta_{m}^{\mathrm{d}}
		&=
		m\lambda
		\label{eq:woo_dark_fringe}
		\\
		d\sin\theta_{m}^{\mathrm{b}}
		&\approx
		\left(m+\frac{1}{2}\right)\lambda,
		\qquad
		m=1,2,\ldots
		\label{eq:woo_bright_fringe}
	\end{align}
	where $d$ denotes the slit width, $\lambda$ is the wavelength of the incident light, and
	$\theta_{m}^{\mathrm{d}}$ and $\theta_{m}^{\mathrm{b}}$
	denote the diffraction angles corresponding to the $m$-th dark fringe and the $m$-th secondary bright fringe, respectively. The superscripts $\mathrm{d}$ and $\mathrm{b}$ denote the dark fringe and bright fringe, respectively.
	
	The relative light intensity corresponding to an arbitrary diffraction direction $\theta$ can be expressed as
	\begin{equation}
		I(\theta)
		=
		I_0
		\left(
		\frac{\sin\beta(\theta)}{\beta(\theta)}
		\right)^2,
		\qquad
		\beta(\theta)
		=
		\frac{\pi d\sin\theta}{\lambda}
		\label{eq:woo_diffraction_intensity}
	\end{equation}
	where $I_0$ denotes the maximum relative light intensity at the central bright fringe, and $\beta(\theta)$ is the diffraction phase parameter. When $\beta(\theta)=0$, $I(0)=I_0$.
	
	WOO maps the above optical relationships to the optimization search process and combines the uncertainty principle with the de Broglie relation to construct the diffusion-control parameter
	\begin{equation}
		U_{\mathrm{a}}
		=
		\frac{3N\lambda}{4\pi d}
		\label{eq:woo_ua}
	\end{equation}
	where $U_{\mathrm{a}}>0$ regulates the diffusion scale during diffraction-based initialization and the subsequent search process.
	
	WOO first randomly generates $N$ base individuals within the search space. The $i$-th base individual is denoted by
	$\boldsymbol{S}_{i}=[s_{i,1},\ldots,s_{i,D}]^{\top}$,
	and its $j$-th component is generated as
	\begin{equation}
		s_{i,j}
		=
		L_j+r_{i,j}(U_j-L_j),
		\qquad
		r_{i,j}\sim\mathcal{U}(0,1)
		\label{eq:woo_random_initialization}
	\end{equation}
	where $\mathcal{U}(0,1)$ denotes the continuous uniform distribution over $[0,1]$, and $r_{i,j}$ is the corresponding independent random variable.
	
	To map different individuals to the central bright fringe, dark fringes, or secondary bright fringes, the diffraction mapping coefficient of the $i$-th individual is defined as
	\begin{equation}
		\chi_i
		=
		\begin{cases}
			0,
			& i=1
			\\[0.5em]
			\dfrac{m_{\mathrm{d}}(i)\lambda}{d},
			& i>1,\ \operatorname{mod}(i,2)=0
			\\[1em]
			\dfrac{\left[2m_{\mathrm{b}}(i)+1\right]\lambda}{2d},
			& i>1,\ \operatorname{mod}(i,2)\neq0
		\end{cases}
		\label{eq:woo_diffraction_angle}
	\end{equation}
	where
	$m_{\mathrm{d}}(i)$ and
	$m_{\mathrm{b}}(i)$
	denote the orders of the dark fringe and secondary bright fringe, respectively, and
	$\operatorname{mod}(\cdot,\cdot)$ denotes the modulo operation.
	When the optical parameters lie within their physically admissible ranges, $\chi_i$ corresponds to $\sin\theta_i$; in WOO, it primarily serves as a numerical mapping coefficient for controlling the diffraction displacement of each individual.
	The corresponding optical path difference is defined as
	\begin{equation}
		\delta_i=d\chi_i
		\label{eq:woo_path_difference}
	\end{equation}
	
	Furthermore, define
	\begin{equation}
		s_i^{\max}
		=
		\max_{1\leq j\leq D}s_{i,j},
		\qquad
		s_i^{\min}
		=
		\min_{1\leq j\leq D}s_{i,j},
		\qquad
		\kappa_i
		=
		\begin{cases}
			1, & i=1\\
			\chi_i, & i>1
		\end{cases}
		\label{eq:woo_initial_auxiliary}
	\end{equation}
	The upper and lower wavefront position vectors corresponding to the $i$-th base individual are then constructed as
	\begin{align}
		\boldsymbol{S}_{i}^{+}
		&=
		\boldsymbol{S}_{i}
		+
		\frac{
			d\left[
			\xi_{i}^{+}
			\left(s_i^{\max}-s_i^{\min}\right)
			+s_i^{\min}
			\right]
		}{
			\kappa_i
		}
		\boldsymbol{1}_{D}
		\label{eq:woo_upper_wavefront}
		\\
		\boldsymbol{S}_{i}^{-}
		&=
		\boldsymbol{S}_{i}
		-
		\frac{
			d\left[
			\xi_{i}^{-}
			\left(s_i^{\max}-s_i^{\min}\right)
			+s_i^{\min}
			\right]
		}{
			\kappa_i
		}
		\boldsymbol{1}_{D}
		\label{eq:woo_lower_wavefront}
	\end{align}
	where $\xi_i^{+}$ and $\xi_i^{-}$ are mutually independent random variables following $\mathcal{U}(0,1)$, and $\boldsymbol{1}_{D}$ denotes a $D$-dimensional all-ones vector.
	
	Accordingly, the initial position of the $i$-th individual after diffraction mapping is expressed as
	\begin{equation}
		\boldsymbol{X}_{i}(0)
		=
		\frac{
			\boldsymbol{S}_{i}^{+}
			+
			\boldsymbol{S}_{i}^{-}
		}{2}
		+
		\operatorname{sgn}
		\left(
		\xi_i^{(0)}-\frac{1}{2}
		\right)
		\xi_i^{(\delta)}
		\delta_i
		\boldsymbol{1}_{D}
		\label{eq:woo_diffraction_initialization}
	\end{equation}
	where $\xi_i^{(0)}$ and $\xi_i^{(\delta)}$ are mutually independent uniformly distributed random variables, and $\operatorname{sgn}(\cdot)$ denotes the sign function. For $i=1$, $\delta_1=0$, and the corresponding individual is therefore mapped to the central bright fringe region. Any out-of-bound components generated after initialization are handled using the boundary repair strategy given in Eq.~\eqref{eq:woo_boundary_handling}.
	
	\subsection{Adaptive search control and stochastic diffusion}
	\label{subsec:woo_control}
	
	After initialization, WOO sorts the population in ascending order of fitness, such that
	\begin{equation}
		F_1(t)
		\leq
		F_2(t)
		\leq
		\cdots
		\leq
		F_N(t)
		\label{eq:woo_fitness_ranking}
	\end{equation}
	A linearly decreasing parameter is then employed to regulate the balance between global exploration and local exploitation:
	\begin{equation}
		C(t)
		=
		1-\frac{t}{T_{\max}},
		\qquad
		t=1,\ldots,T_{\max}
		\label{eq:woo_control_parameter}
	\end{equation}
	where $T_{\max}$ denotes the maximum number of iterations, and
	$C(t)\in[0,1)$ is the search control parameter at the $t$-th iteration.
	
	Based on this parameter, two stage-control quantities are defined as
	\begin{equation}
		\tau_1(t)
		=
		I_0
		\left[
		1+
		\frac{U_{\mathrm{a}}}{3}
		\left(
		C(t)-1
		\right)
		\right],
		\qquad
		\tau_2(t)
		=
		\frac{\exp\!\left(\tau_1(t)\right)}{3}
		\label{eq:woo_stage_parameters}
	\end{equation}
	where $\tau_1(t)$ regulates the candidate-generation mode within the Reinforcement strategy, while $\tau_2(t)$ controls the activation of the stochastic diffusion branch.
	
	To restrict the selection range of rank-based reference individuals, WOO constructs a dynamic fitness-ranking interval:
	\begin{align}
		T_{\mathrm{r}}
		&=
		\left|
		\operatorname{round}
		\left[
		\left(
		1-\sin\frac{\pi U_{\mathrm{a}}}{2}
		\right)N
		\right]
		\right|
		\label{eq:woo_rank_control}
		\\
		\ell_{\mathrm{r}}
		&=
		\operatorname{clip}_{[1,N]}
		\left(
		T_{\mathrm{r}}+1
		\right)
		\label{eq:woo_rank_lower}
		\\
		u_{\mathrm{r}}(t)
		&=
		\operatorname{clip}_{[1,N]}
		\left[
		\operatorname{round}
		\left(
		N-
		\frac{
			C(t)T_{\mathrm{r}}
		}{
			2U_{\mathrm{a}}^{2}
		}
		\right)
		\right]
		\label{eq:woo_rank_upper}
		\\
		\mathcal{I}_{\mathrm{r}}(t)
		&=
		\left\{
		k\in\{1,\ldots,N\}
		\,\middle|\,
		\min\!\left(
		\ell_{\mathrm{r}},
		u_{\mathrm{r}}(t)
		\right)
		\leq k
		\leq
		\max\!\left(
		\ell_{\mathrm{r}},
		u_{\mathrm{r}}(t)
		\right)
		\right\}
		\label{eq:woo_rank_interval}
	\end{align}
	where $T_{\mathrm{r}}$ is the ranking-range control quantity,
	$\ell_{\mathrm{r}}$ and $u_{\mathrm{r}}(t)$ are the two endpoints of the ranking interval, and
	$\mathcal{I}_{\mathrm{r}}(t)$ is the corresponding set of ranking indices.
	The operator $\operatorname{round}(\cdot)$ denotes rounding to the nearest integer, and
	\[
	\operatorname{clip}_{[1,N]}(x)
	=
	\min\{N,\max\{1,x\}\}.
	\]
	
	When updating the $i$-th individual, a ranking index
	$R_i^{\mathrm{r}}(t)$ is sampled uniformly from
	$\mathcal{I}_{\mathrm{r}}(t)$, i.e.,
	\begin{equation}
		\Pr
		\left[
		R_i^{\mathrm{r}}(t)=k
		\right]
		=
		\frac{1}{
			\left|
			\mathcal{I}_{\mathrm{r}}(t)
			\right|
		},
		\qquad
		k\in\mathcal{I}_{\mathrm{r}}(t)
		\label{eq:woo_rank_sampling}
	\end{equation}
	Based on this sampled reference, the rank-guided direction is constructed as
	\begin{equation}
		\boldsymbol{u}_{i}^{\mathrm{r}}(t)
		=
		w_{\mathrm{g}}(t)
		\left[
		\boldsymbol{X}_{R_i^{\mathrm{r}}(t)}(t)
		-
		\boldsymbol{X}_{i}(t)
		\right],
		\qquad
		w_{\mathrm{g}}(t)
		=
		\sin
		\left[
		\left(
		1-e^{-t}
		\right)
		\frac{\pi}{2}
		\right]
		\label{eq:woo_interference_vector}
	\end{equation}
	where $\boldsymbol{u}_{i}^{\mathrm{r}}(t)$ denotes the rank-guided direction, and
	$w_{\mathrm{g}}(t)$ is an iteration-dependent guidance weight.
	
	Let
	\begin{equation}
		N_{+}
		=
		\operatorname{round}
		\left(
		\frac{N}{2}
		\right),
		\qquad
		\mathcal{I}^{+}
		=
		\{1,\ldots,N_{+}\},
		\qquad
		\mathcal{I}^{-}
		=
		\{N_{+}+1,\ldots,N\}
		\label{eq:woo_subpopulations}
	\end{equation}
	where $\mathcal{I}^{+}$ and $\mathcal{I}^{-}$ denote the ranking-index sets of the better-fitness subgroup and the remaining subgroup, respectively.
	According to the subgroup to which an individual belongs and a random condition, the differential learning direction is defined as
	\begin{equation}
		\boldsymbol{u}_{i}^{\mathrm{d}}(t)
		=
		\begin{cases}
			\boldsymbol{X}_{R_i^{+}(t)}(t)
			-
			\boldsymbol{X}_{R_{i,1}(t)}(t),
			&
			i\in\mathcal{I}^{+},
			\ q_i(t)>C(t)
			\\[0.9em]
			\boldsymbol{X}_{R_i^{+}(t)}(t)
			-
			\boldsymbol{X}_{R_i^{-}(t)}(t),
			&
			i\in\mathcal{I}^{-},
			\ q_i(t)>C(t)
			\\[0.9em]
			\boldsymbol{X}_{R_{i,1}(t)}(t)
			-
			\boldsymbol{X}_{R_{i,2}(t)}(t),
			&
			\text{otherwise}
		\end{cases}
		\label{eq:woo_learning_vector}
	\end{equation}
	where $\boldsymbol{u}_{i}^{\mathrm{d}}(t)$ denotes the differential learning direction.
	The indices $R_i^{+}(t)$ and $R_i^{-}(t)$ are randomly sampled from
	$\mathcal{I}^{+}$ and $\mathcal{I}^{-}$, respectively, while
	$R_{i,1}(t)$ and $R_{i,2}(t)$ are randomly sampled from
	$\{1,\ldots,N\}$, and
	$q_i(t)\sim\mathcal{U}(0,1)$.
	All the above random variables are independently generated for different individuals and iterations.
	
	In the original WOO, $\boldsymbol{u}_{i}^{\mathrm{r}}(t)$ and
	$\boldsymbol{u}_{i}^{\mathrm{d}}(t)$ are combined in a piecewise manner according to the current search stage, relative light intensity, and historical search state.
	To avoid repeatedly expanding the multiple conditional branches of the original algorithm, the overall combination process is compactly represented as
	\begin{equation}
		\boldsymbol{d}_{i}^{\mathrm{s}}(t)
		=
		\mathcal{G}_{\mathrm{WOO}}
		\left(
		\boldsymbol{u}_{i}^{\mathrm{r}}(t),
		\boldsymbol{u}_{i}^{\mathrm{d}}(t),
		\boldsymbol{X}_{i}(t),
		\boldsymbol{X}_{\mathrm{best}}(t);
		t
		\right)
		\label{eq:woo_diffusion_direction}
	\end{equation}
	where $\boldsymbol{d}_{i}^{\mathrm{s}}(t)$ denotes the final stochastic diffusion direction, and
	$\mathcal{G}_{\mathrm{WOO}}(\cdot)$ represents the stage-dependent interference-combination operator of the original WOO.
	
	Let $\zeta_{i,0}(t)\sim\mathcal{U}(0,1)$.
	When $\zeta_{i,0}(t)<\tau_2(t)$, the stochastic diffusion branch is activated, and the corresponding candidate solution is generated as
	\begin{equation}
		\widetilde{\boldsymbol{X}}_{i}(t+1)
		=
		\boldsymbol{X}_{i}(t)
		+
		\boldsymbol{\omega}_{i}(t)
		\odot
		\boldsymbol{Z}_{i}(t)
		\odot
		\boldsymbol{d}_{i}^{\mathrm{s}}(t)
		\label{eq:woo_random_update}
	\end{equation}
	where $\boldsymbol{\omega}_{i}(t)\in\mathbb{R}^{D}$ is the search-amplitude adjustment vector,
	$\boldsymbol{Z}_{i}(t)\in\{0,1\}^{D}$ is the dimension-selection mask, and
	$\odot$ denotes element-wise multiplication.
	
	\subsection{Search intensification}
	\label{subsec:woo_reinforcement_search}
	
	When $\zeta_{i,0}(t)\geq\tau_2(t)$, WOO enters the Reinforcement strategy for search intensification and generates a candidate solution from three update modes according to the current search state:
	\begin{align}
		\boldsymbol{Y}_{i}^{(1)}(t)
		&=
		\boldsymbol{V}_{i}^{(1)}(t)
		\odot
		\boldsymbol{X}_{\mathrm{best}}(t)
		\label{eq:woo_reinforcement_best}
		\\
		\boldsymbol{Y}_{i}^{(2)}(t)
		&=
		\boldsymbol{V}_{i}^{(2)}(t)
		\odot
		\frac{
			\boldsymbol{X}_{\mathrm{best}}(t)
			+
			\boldsymbol{X}_{i}(t)
		}{2}
		\label{eq:woo_reinforcement_midpoint}
		\\
		\boldsymbol{Y}_{i}^{(3)}(t)
		&=
		\boldsymbol{X}_{i}(t)
		+
		\alpha_{i}^{\mathrm{d}}(t)
		\boldsymbol{u}_{i}^{\mathrm{d}}(t)
		+
		\Delta\boldsymbol{X}_{i}(t)
		\label{eq:woo_reinforcement_difference}
	\end{align}
	where $\boldsymbol{Y}_{i}^{(k)}(t)$,
	$k\in\{1,2,3\}$, denotes the $k$-th intensified candidate solution for the $i$-th individual;
	$\boldsymbol{V}_{i}^{(1)}(t)$ and
	$\boldsymbol{V}_{i}^{(2)}(t)$ are two mode-dependent stochastic beam-modulation vectors; and
	$\alpha_{i}^{\mathrm{d}}(t)$ denotes the differential search step size.
	
	The stochastic beam-modulation vectors can be uniformly expressed as
	\begin{equation}
		\boldsymbol{V}_{i}^{(k)}(t)
		=
		\gamma_{i}^{(k)}(t)
		\boldsymbol{\eta}_{i}^{(k)}(t),
		\qquad
		\boldsymbol{\eta}_{i}^{(k)}(t)
		\sim
		\mathcal{N}
		\left(
		\boldsymbol{0},
		\boldsymbol{I}_{D}
		\right),
		\qquad
		k\in\{1,2\}
		\label{eq:woo_beam_modulation}
	\end{equation}
	where $\gamma_i^{(k)}(t)$ is a mode-dependent scalar modulation coefficient determined by the light-intensity focusing rule of WOO, and
	$\mathcal{N}(\boldsymbol{0},\boldsymbol{I}_{D})$
	denotes a $D$-dimensional standard normal distribution with zero mean and covariance matrix $\boldsymbol{I}_{D}$.
	
	The position-correction vector is defined as
	\begin{equation}
		\Delta\boldsymbol{X}_{i}(t)
		=
		z_i(t)
		\alpha_{i}^{\mathrm{c}}(t)
		\left[
		\frac{
			\boldsymbol{X}_{\mathrm{best}}(t)
			+
			\boldsymbol{X}_{i}(t)
		}{2}
		-
		\boldsymbol{X}_{R_i^{\mathrm{ref}}(t)}(t)
		\right]
		\label{eq:woo_reinforcement_correction}
	\end{equation}
	where $R_i^{\mathrm{ref}}(t)$ denotes the index of a reference individual randomly selected from
	$\{1,\ldots,N\}$,
	$\alpha_i^{\mathrm{c}}(t)$ is the correction step size, and
	$z_i(t)\sim\operatorname{Bernoulli}(1/2)$
	is the activation variable for the correction term.
	
	In WOO, the two stochastic step-size coefficients are defined as
	\begin{equation}
		\alpha_i^{\mathrm{c}}(t)
		=
		\frac{
			\left|
			\eta_{3,i}(t)
			\right|
			\xi_{3,i}(t)
			\left[
			1-\xi_{3,i}(t)
			\right]
		}{i},
		\qquad
		\alpha_i^{\mathrm{d}}(t)
		=
		\exp
		\left(
		-\frac{\eta_{4,i}(t)}{i}
		\right)
		\xi_{4,i}(t)
		\label{eq:woo_reinforcement_steps}
	\end{equation}
	where
	$\eta_{3,i}(t),\eta_{4,i}(t)\sim\mathcal{N}(0,1)$,
	$\xi_{3,i}(t),\xi_{4,i}(t)\sim\mathcal{U}(0,1)$,
	and all random variables are mutually independent.
	
	Let
	$\zeta_{i,1}(t),\zeta_{i,2}(t)\sim\mathcal{U}(0,1)$.
	The raw candidate solution generated by the Reinforcement strategy is then determined as
	\begin{equation}
		\widetilde{\boldsymbol{X}}_{i}(t+1)
		=
		\begin{cases}
			\boldsymbol{Y}_{i}^{(1)}(t),
			&
			\zeta_{i,1}(t)>\tau_1(t),
			\ \zeta_{i,2}(t)>C(t)
			\\[0.6em]
			\boldsymbol{Y}_{i}^{(2)}(t),
			&
			\zeta_{i,1}(t)>\tau_1(t),
			\ \zeta_{i,2}(t)\leq C(t)
			\\[0.6em]
			\boldsymbol{Y}_{i}^{(3)}(t),
			&
			\zeta_{i,1}(t)\leq\tau_1(t)
		\end{cases}
		\label{eq:woo_reinforcement_selection}
	\end{equation}
	The three update modes generate candidate solutions using the global best position, midpoint information, and the differential learning direction, respectively, thereby maintaining diversity among the intensified search directions.
	
	\subsection{Success-history adaptation and greedy selection}
	\label{subsec:woo_adjustment}
	
	Regardless of whether the raw candidate solution is generated by stochastic diffusion or search intensification, WOO employs the same boundary-repair and greedy-selection mechanisms.
	If the $j$-th component of
	$\widetilde{\boldsymbol{X}}_{i}(t+1)$
	falls outside the feasible region, it is randomly regenerated within the corresponding search interval:
	\begin{equation}
		\widehat{x}_{i,j}(t+1)
		=
		\begin{cases}
			L_j+\rho_{i,j}(t)(U_j-L_j),
			&
			\substack{
				\widetilde{x}_{i,j}(t+1)<L_j
				\\[-0.3mm]
				\text{or }\widetilde{x}_{i,j}(t+1)>U_j
			}
			\\[0.4em]
			\widetilde{x}_{i,j}(t+1),
			&
			\text{otherwise}
		\end{cases}
		\label{eq:woo_boundary_handling}
	\end{equation}
	where $\widehat{x}_{i,j}(t+1)$ denotes the component after boundary repair, and
	$\rho_{i,j}(t)\sim\mathcal{U}(0,1)$
	is the random variable used for boundary repair.
	All repaired components form the candidate vector
	$\widehat{\boldsymbol{X}}_{i}(t+1)$,
	whose fitness value is evaluated as
	\begin{equation}
		\widehat{F}_{i}(t+1)
		=
		f
		\left(
		\widehat{\boldsymbol{X}}_{i}(t+1)
		\right)
		\label{eq:woo_repaired_fitness}
	\end{equation}
	
	When
	$\widehat{F}_{i}(t+1)<F_i(t)$,
	the candidate solution is regarded as a successful update.
	For each successful update, the relative fitness improvement, the magnitude of individual displacement, and the distance from the current individual to the global best solution are defined as
	\begin{align}
		\Delta F_i(t)
		&=
		\left|
		\frac{
			F_i(t)-\widehat{F}_i(t+1)
		}{
			F_i(t)-F_{\mathrm{best}}(t)+\varepsilon
		}
		\right|
		\label{eq:woo_delta_f}
		\\
		\Delta S_i(t)
		&=
		\sqrt{
			\frac{1}{D}
			\sum_{j=1}^{D}
			\left[
			x_{i,j}(t)
			-
			\widehat{x}_{i,j}(t+1)
			\right]^2
		}
		\label{eq:woo_delta_position}
		\\
		d_{i,\mathrm{best}}(t)
		&=
		\sqrt{
			\frac{1}{D}
			\sum_{j=1}^{D}
			\left[
			x_{i,j}(t)
			-
			x_{\mathrm{best},j}(t)
			\right]^2
		}
		\label{eq:woo_delta_best}
	\end{align}
	where $\varepsilon>0$ is a sufficiently small constant introduced to avoid division by zero, and
	$d_{i,\mathrm{best}}(t)$ denotes the distance between the current individual and the global best solution.
	WOO records these statistics, together with the search parameters associated with successful updates, in the success history, which is subsequently used to regulate the diffusion scale and information-interaction intensity in later iterations.
	
	A greedy selection criterion is then employed to determine whether the candidate replaces the current individual:
	\begin{equation}
		\boldsymbol{X}_{i}(t+1)
		=
		\begin{cases}
			\widehat{\boldsymbol{X}}_{i}(t+1),
			&
			\widehat{F}_{i}(t+1)<F_i(t)
			\\[0.5em]
			\boldsymbol{X}_{i}(t),
			&
			\text{otherwise}
		\end{cases}
		\label{eq:woo_final_selection}
	\end{equation}
	If the fitness value of the newly accepted individual is further better than
	$F_{\mathrm{best}}(t)$,
	the global best solution and its corresponding fitness value are updated accordingly.
	
	Let $\mathcal{C}_{f}(D)$ denote the computational complexity of one objective-function evaluation.
	Population initialization requires
	$\mathcal{O}(ND)$
	operations.
	At each iteration, the computational complexities of position updating, fitness ranking, and objective-function evaluation are
	$\mathcal{O}(ND)$,
	$\mathcal{O}(N\log N)$,
	and
	$\mathcal{O}(N\mathcal{C}_{f}(D))$,
	respectively.
	Therefore, the overall time complexity of WOO can be expressed as
	\begin{equation}
		\mathcal{O}
		\left(
		ND
		+
		T_{\max}
		\left[
		ND
		+
		N\log N
		+
		N\mathcal{C}_{f}(D)
		\right]
		\right)
		\label{eq:woo_complexity}
	\end{equation}
	
	If only the intrinsic search overhead of the algorithm is considered, excluding the internal computational cost of the objective function, Eq.~\eqref{eq:woo_complexity} can be simplified to
	\begin{equation}
		\mathcal{O}
		\left(
		ND
		+
		T_{\max}N
		\left(
		D+\log N
		\right)
		\right)
		\label{eq:woo_algorithmic_complexity}
	\end{equation}
	When $D\gg\log N$ and $T_{\max}$ is sufficiently large, the dominant term can be approximated as
	$\mathcal{O}(T_{\max}ND)$.
	
	In summary, WOO evolves the population through diffraction-based initialization, adaptive search control, stochastic diffusion, search intensification, success-history adaptation, and greedy selection.
	Based on this fundamental framework, the next section introduces the proposed Improved Wave Optics Optimizer (IWOO).

	\section{The Proposed Improved Wave Optics Optimizer}
	\label{sec:iwoo}
	
	This paper proposes an Improved Wave Optics Optimizer (IWOO), which incorporates four enhancement mechanisms into the original WOO framework: an initialization strategy combining the Good Point Set (GPS) with Opposition-Based Learning (OBL), TriEDM multi-path search, Correlation-Guided Groupwise Orthogonal Learning (CGOL), and Adaptive Stagnation-Aware Runge--Kutta (ASRK). These mechanisms are designed to improve the quality of the initial population, enhance search diversity, strengthen local exploitation, and increase the algorithm's ability to escape from local optima.
	
	\subsection{GPS--OBL initialization} 
	\label{subsec:gps_obl} 
	
	The distribution characteristics of the initial population directly affect the search efficiency and convergence performance of an optimization algorithm. An initial population with a uniform distribution and sufficient diversity can provide better coverage of the search space and more favorable starting points for subsequent iterations. WOO first randomly generates initial individuals and then adjusts their positions through diffraction mapping. However, under a limited population size, random initialization may still lead to local clustering and uneven coverage. 
	
	The Good Point Set (GPS) constructs deterministic low-discrepancy point sets based on number-theoretic principles, thereby improving the distribution uniformity of finite samples in multidimensional spaces~\cite{Keng1981NumberTheory}. Opposition-Based Learning (OBL) expands the candidate range by simultaneously considering the current solutions and their opposite solutions, thereby increasing the probability of obtaining high-quality initial solutions~\cite{Tizhoosh2005OppositionBasedLearning}. Previous studies have shown that both GPS and OBL can effectively improve population initialization and search performance~\cite{Yang2024DynamicDE,Yang2023SalpSwarmOBL,Ma2025ZebraOptimization}. Therefore, this paper combines the uniform-coverage property of GPS with the opposition-based expansion capability of OBL to construct the GPS--OBL initialization strategy. 
	
	First, the GPS generating vector is defined as 
	$\boldsymbol{\eta}=(\eta_1,\eta_2,\ldots,\eta_D)^{\mathrm T}$, 
	whose $j$-th component is given by 
	\begin{equation} 
		\eta_j 
		= 
		2\cos\left(\frac{2\pi j}{p}\right), 
		\qquad 
		j=1,2,\ldots,D 
		\label{eq:gps_generator} 
	\end{equation} 
	where $p$ is the smallest prime number satisfying $p\geq 2D+3$. 
	The normalized coordinate of the $j$-th dimension of the $i$-th GPS point is defined as 
	\begin{equation} 
		g_{i,j} 
		= 
		\operatorname{frac}(i\eta_j) 
		= 
		i\eta_j-\left\lfloor i\eta_j\right\rfloor, 
		\qquad 
		i=1,2,\ldots,N, 
		\quad 
		j=1,2,\ldots,D 
		\label{eq:gps_point} 
	\end{equation} 
	where $\operatorname{frac}(\cdot)$ denotes the fractional-part operator, 
	$\lfloor\cdot\rfloor$ denotes the floor operator, and 
	$g_{i,j}\in[0,1)$ is the corresponding normalized coordinate. 
	Mapping these coordinates to the actual search space gives 
	\begin{equation} 
		x_{i,j}^{\mathrm{GPS}} 
		= 
		L_j+g_{i,j}(U_j-L_j), 
		\qquad 
		i=1,2,\ldots,N, 
		\quad 
		j=1,2,\ldots,D 
		\label{eq:gps_mapping} 
	\end{equation} 
	where $x_{i,j}^{\mathrm{GPS}}$ denotes the $j$-th component of the $i$-th GPS individual. 
	Let 
	$\boldsymbol{X}_{i}^{\mathrm{GPS}} 
	=[x_{i,1}^{\mathrm{GPS}},x_{i,2}^{\mathrm{GPS}},\ldots,x_{i,D}^{\mathrm{GPS}}]^{\mathrm T}$. 
	The GPS population is then expressed as 
	$\mathcal{P}_{\mathrm{GPS}} 
	=\{\boldsymbol{X}_{i}^{\mathrm{GPS}}\}_{i=1}^{N}$. 
	
	Standard OBL is subsequently applied to the GPS individuals. 
	The opposite individual corresponding to the $i$-th GPS individual is defined as 
	\begin{equation} 
		\boldsymbol{X}_{i}^{\mathrm{OBL}} 
		= 
		\boldsymbol{L} 
		+ 
		\boldsymbol{U} 
		- 
		\boldsymbol{X}_{i}^{\mathrm{GPS}}, 
		\qquad 
		i=1,2,\ldots,N 
		\label{eq:gps_obl_vector} 
	\end{equation} 
	where $\boldsymbol{X}_{i}^{\mathrm{OBL}}$ is the opposite individual of 
	$\boldsymbol{X}_{i}^{\mathrm{GPS}}$ 
	with respect to the center of the search space, 
	$(\boldsymbol{L}+\boldsymbol{U})/2$. 
	All opposite individuals form the opposite population 
	$\mathcal{P}_{\mathrm{OBL}} 
	=\{\boldsymbol{X}_{i}^{\mathrm{OBL}}\}_{i=1}^{N}$. 
	The two populations are then merged, and the better individuals are selected according to their objective-function values: 
	\begin{equation} 
		\mathcal{P}_{\mathrm{cand}} 
		= 
		\mathcal{P}_{\mathrm{GPS}} 
		\cup 
		\mathcal{P}_{\mathrm{OBL}}, 
		\qquad 
		\mathcal{P}_{0} 
		= 
		\operatorname{Best}_{N} 
		\left(\mathcal{P}_{\mathrm{cand}};f\right) 
		= 
		\left\{\boldsymbol{X}_{i}(0)\right\}_{i=1}^{N} 
		\label{eq:gps_obl_selection} 
	\end{equation} 
	where $\mathcal{P}_{\mathrm{cand}}$ denotes the candidate population obtained by merging the two populations and contains $2N$ candidate individuals. 
	The operator $\operatorname{Best}_{N}(\cdot;f)$ selects the $N$ individuals with the smallest objective-function values according to $f$, and 
	$\mathcal{P}_{0}$ denotes the resulting initial population. 
	Since both the GPS individuals and their opposite individuals are evaluated once, this procedure requires a total of $2N$ objective-function evaluations. 
	
\begin{figure*}[t] 
		\centering 
		
		\begin{tabular}{@{}cccc@{}}
			\includegraphics[width=0.23\textwidth,height=0.17\textheight,keepaspectratio]{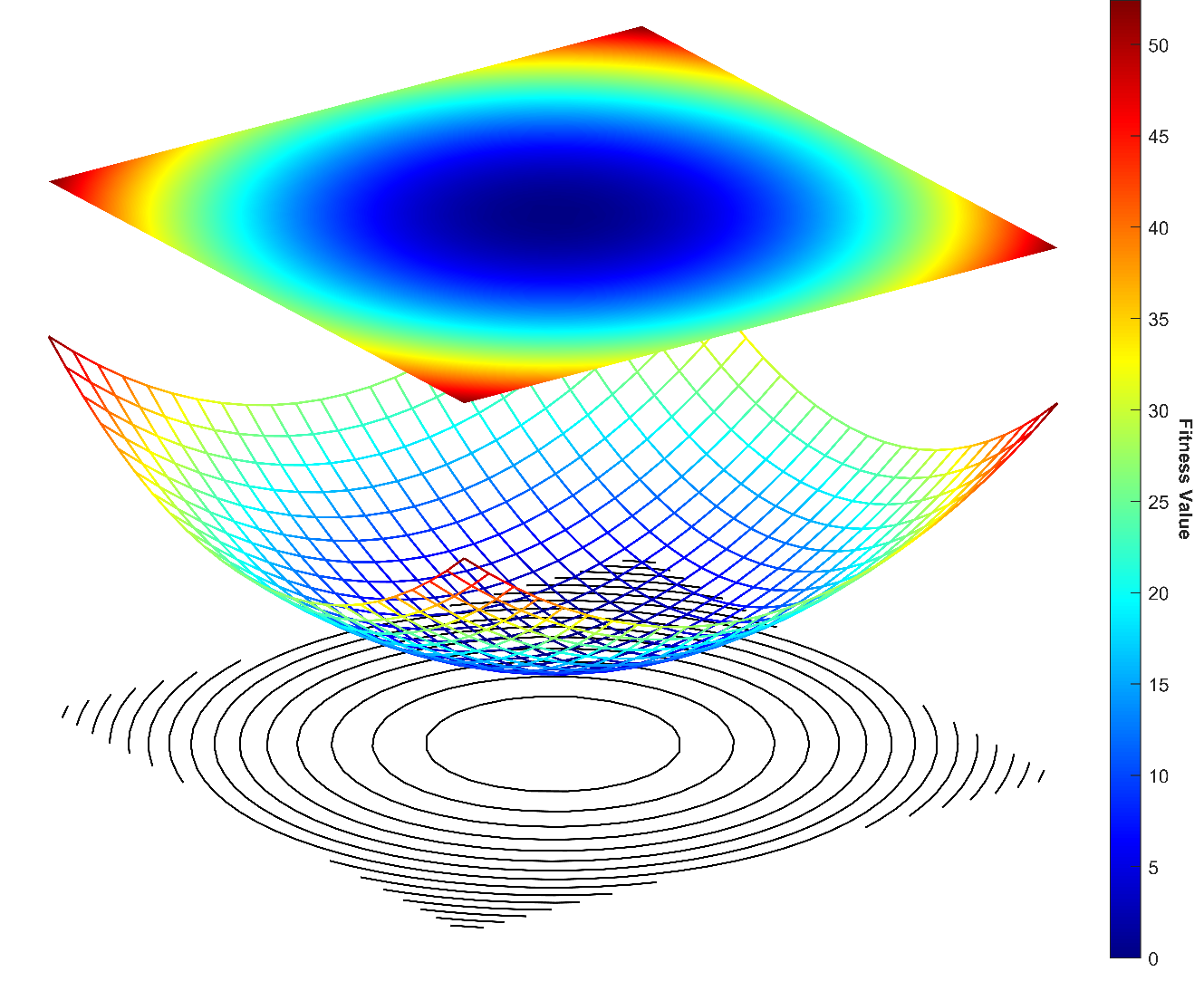} &
			\includegraphics[width=0.23\textwidth,height=0.17\textheight,keepaspectratio]{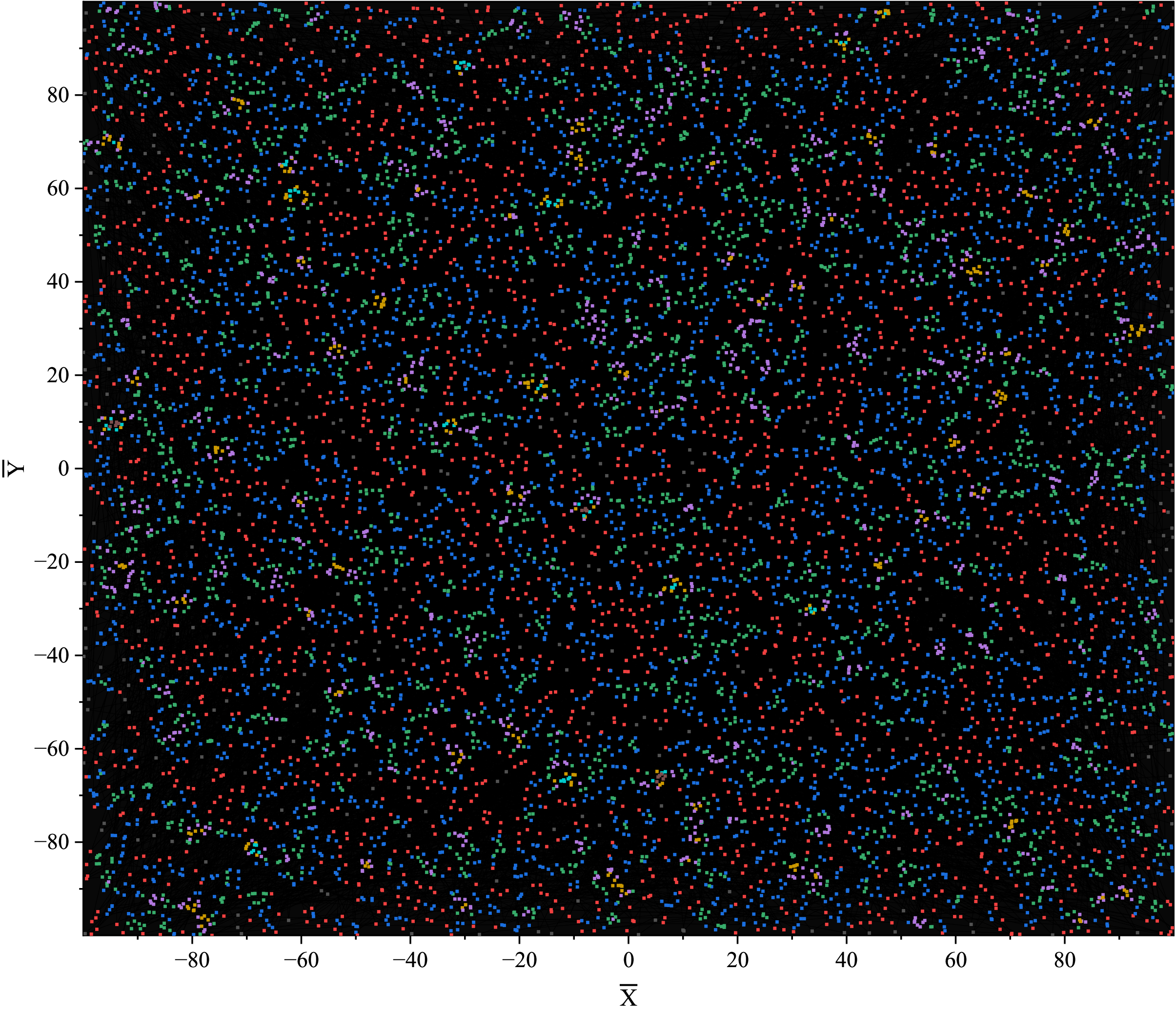} &
			\includegraphics[width=0.23\textwidth,height=0.17\textheight,keepaspectratio]{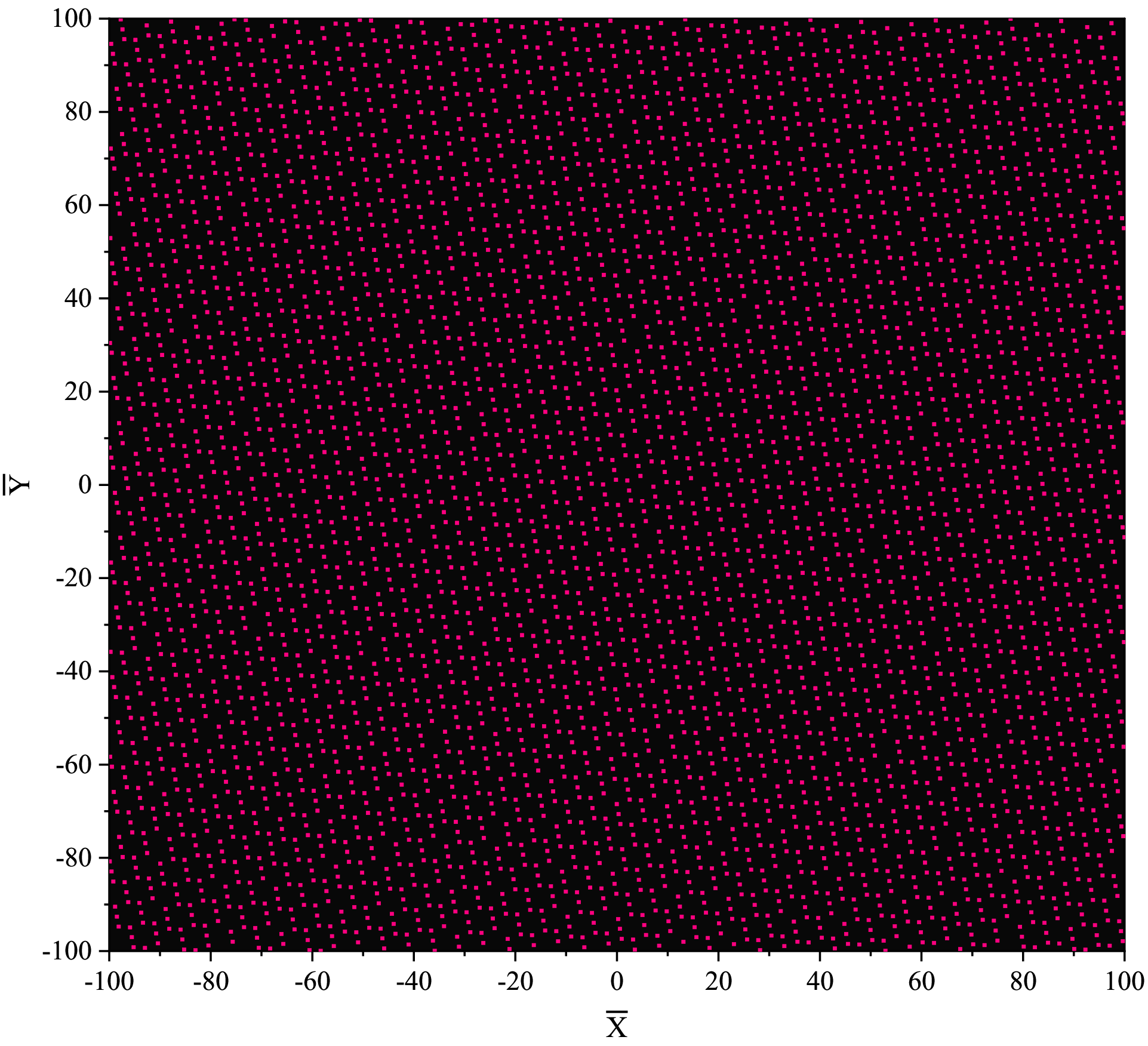} &
			\includegraphics[width=0.23\textwidth,height=0.17\textheight,keepaspectratio]{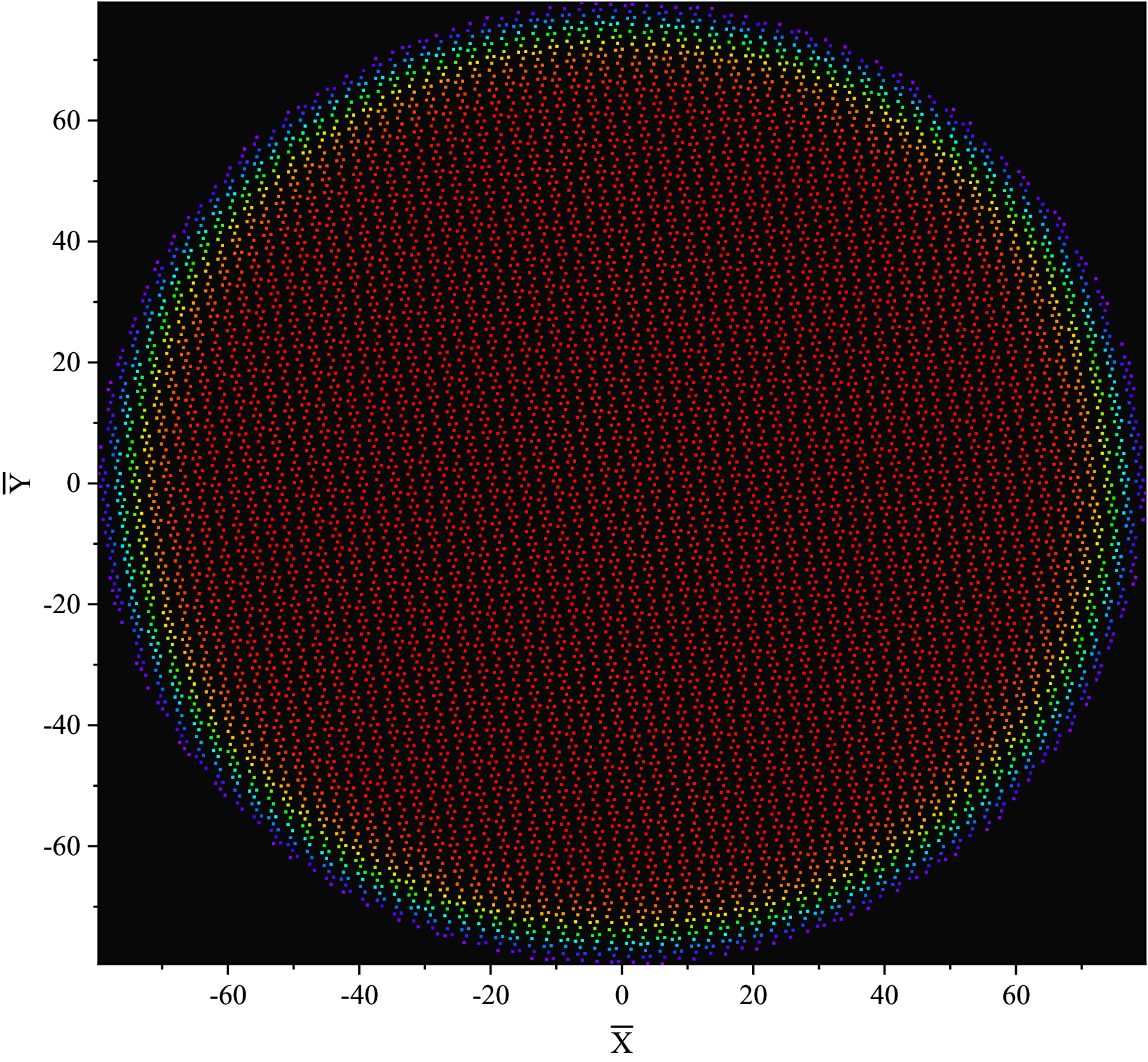} \\
			{\scriptsize (a)} & {\scriptsize (b)} & {\scriptsize (c)} & {\scriptsize (d)}
		\end{tabular}
		
		\caption{Initialization distributions on the Sphere function 
			(a) Sphere function 
			(b) Random initialization 
			(c) GPS initialization 
			(d) GPS--OBL initialization} 
		\label{fig:sphere_initialization_compare} 
	\end{figure*}

	To visually compare the distribution characteristics of different initialization methods, the Sphere function is employed as an illustrative example and is defined as 
	\begin{equation} 
		f(\boldsymbol{x}) 
		= 
		\sum_{j=1}^{D}x_j^2 
		\label{eq:sphere_function} 
	\end{equation} 
	Figure~\ref{fig:sphere_initialization_compare} shows the Sphere function together with the individual distributions generated by Random, GPS, and GPS--OBL initialization. Random initialization exhibits local clustering and sparsely covered regions, whereas the individuals generated by GPS are distributed more uniformly and provide more extensive coverage of the search space. After introducing opposite individuals and applying fitness-based selection, the retained individuals become further concentrated in regions with lower objective-function values. These results indicate that GPS--OBL improves the quality of the initial individuals while maintaining a relatively regular spatial distribution. 
	
	
	The three initialization strategies are further compared on the CEC2017 benchmark suite. To reduce the influence of differences in numerical scale among the benchmark functions, the initial best fitness obtained by each initialization method is normalized. 
	As shown in Fig.~\ref{fig:gps_obl_cec2017}, GPS--OBL achieves lower normalized initial best fitness values than Random and GPS initialization on most benchmark functions, further demonstrating its effectiveness in improving the quality of the initial population. 
	
	In summary, GPS--OBL first employs the structured point set generated by GPS to improve the spatial coverage of candidate individuals, and then uses opposition-based expansion and fitness-based selection to enhance the quality of the initial solutions. In this way, IWOO is provided with an initial population that achieves a favorable balance between spatial distribution regularity and solution quality. 
	
	
	\noindent
	\begin{minipage}{\linewidth}
		\centering
		
		\includegraphics[
		width=0.95\linewidth,
		keepaspectratio
		]{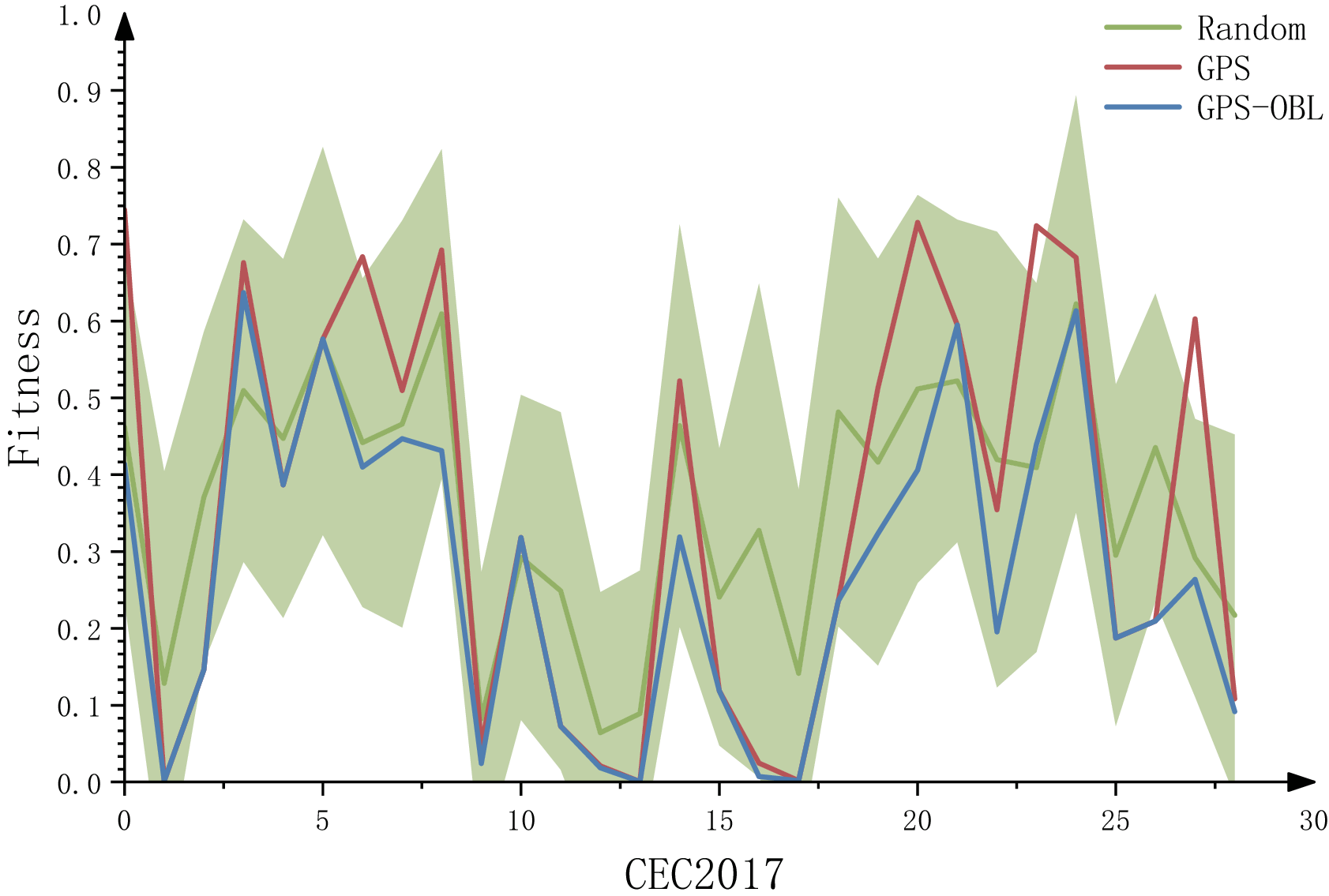}
		
		\vspace{1mm}
		
		\captionof{figure}{Normalized initial fitness comparison on CEC2017}
		\label{fig:gps_obl_cec2017}
	\end{minipage}
	
	\vspace{2mm}
	
	\subsection{TriEDM multi-path search}
	\label{subsec:triedm}
	
	In the reinforcement strategy of WOO, only one update mode is randomly activated at each update to generate a single candidate solution, resulting in a relatively limited search direction for an individual update. Inspired by the temporary triangle-based search mechanism of Chaos Game Optimization (CGO)~\cite{Talatahari2021CGO,Mehrabi2024HybridClassifier,Talatahari2022CGOCrossover}, this paper proposes a TriEDM multi-path search strategy that integrates the temporary triangle with the exponential discrete memristor (E-DM) map. Specifically, three geometric search paths are first constructed based on the temporary triangle, and a fourth nonlinear search path is subsequently generated using the E-DM~\cite{Dong2025ChaoticEvolutionOptimization}. The individual is finally updated through competitive selection among the candidate solutions. The detailed procedure is described as follows.
	
	\textbf{(1) Temporary triangle and geometric search paths.}
	For the individual $\boldsymbol{X}_{i}(t)$ at the $t$-th iteration, a subgroup
	$\mathcal{S}_{i}(t)\subseteq\{\boldsymbol{X}_{k}(t)\}_{k=1}^{N}$
	is randomly selected from the current population.
	Its size
	$n_i(t)=|\mathcal{S}_{i}(t)|$
	is uniformly sampled from
	$\{1,2,\ldots,N\}$,
	and the corresponding mean-guided point is defined as
	\begin{equation}
		\boldsymbol{M}_{i}(t)
		=
		\frac{1}{n_i(t)}
		\sum_{\boldsymbol{X}_{k}(t)\in\mathcal{S}_{i}(t)}
		\boldsymbol{X}_{k}(t)
		\label{eq:triedm_mean_point}
	\end{equation}
	
	A temporary triangle is formed by
	$\boldsymbol{X}_{i}(t)$,
	$\boldsymbol{X}_{\mathrm{best}}(t)$,
	and
	$\boldsymbol{M}_{i}(t)$.
	The scaling-vector pool is defined as
	\begin{equation}
		\mathcal{A}_{i}(t)
		=
		\left\{
		\begin{aligned}
			&\boldsymbol{r}_{i,1}(t),
			\quad
			2\boldsymbol{r}_{i,2}(t)-\boldsymbol{1}_{D},
			\\[-0.5mm]
			&\delta_{i,1}(t)\boldsymbol{r}_{i,3}(t)
			+\boldsymbol{1}_{D},
			\\[-0.5mm]
			&\delta_{i,2}(t)\boldsymbol{r}_{i,4}(t)
			+
			\left[1-\delta_{i,2}(t)\right]
			\boldsymbol{1}_{D}
		\end{aligned}
		\right.
		\label{eq:triedm_alpha_pool}
	\end{equation}
	where
	$\boldsymbol{r}_{i,h}(t)\sim\mathcal{U}(0,1)^D$
	$(h=1,2,3,4)$
	are mutually independent random vectors whose components independently follow the uniform distribution over $[0,1]$.
	The variables
	$\delta_{i,1}(t)$
	and
	$\delta_{i,2}(t)$
	are mutually independent and follow
	$\operatorname{Bernoulli}(0.5)$.
	Three scaling vectors are randomly selected from
	$\mathcal{A}_{i}(t)$
	and denoted by
	$\boldsymbol{\alpha}_{i,1}(t)$,
	$\boldsymbol{\alpha}_{i,2}(t)$,
	and
	$\boldsymbol{\alpha}_{i,3}(t)$,
	respectively.
	The candidate solutions generated by the three geometric search paths are
	\begin{align}
		\boldsymbol{P}_{i,1}(t)
		&=
		\boldsymbol{X}_{i}(t)
		+
		\boldsymbol{\alpha}_{i,1}(t)
		\odot
		\left[
		\kappa_{1}\boldsymbol{X}_{\mathrm{best}}(t)
		-
		\kappa_{2}\boldsymbol{M}_{i}(t)
		\right]
		\label{eq:triedm_path1}
		\\
		\boldsymbol{P}_{i,2}(t)
		&=
		\boldsymbol{X}_{\mathrm{best}}(t)
		+
		\boldsymbol{\alpha}_{i,2}(t)
		\odot
		\left[
		\kappa_{3}\boldsymbol{M}_{i}(t)
		-
		\kappa_{4}\boldsymbol{X}_{i}(t)
		\right]
		\label{eq:triedm_path2}
		\\
		\boldsymbol{P}_{i,3}(t)
		&=
		\boldsymbol{M}_{i}(t)
		+
		\boldsymbol{\alpha}_{i,3}(t)
		\odot
		\left[
		\kappa_{5}\boldsymbol{X}_{\mathrm{best}}(t)
		-
		\kappa_{6}\boldsymbol{X}_{i}(t)
		\right]
		\label{eq:triedm_path3}
	\end{align}
	where
	$\kappa_{1},\ldots,\kappa_{6}$
	are mutually independent discrete random coefficients, each uniformly sampled from the set $\{1,2\}$.
	The three paths use the current individual, global best solution, and mean-guided point as their respective base points to generate candidate solutions along different search directions.
	
	\textbf{(2) E-DM-based nonlinear search path.}
	The population is sorted in ascending order of fitness, and $i$ denotes the resulting rank index throughout this subsection.
	Each rank position maintains a pair of continuously evolving E-DM state vectors, initialized as
	\begin{equation}
		\boldsymbol{c}_{i}^{x}(0)
		\sim
		\mathcal{U}(-0.5,0.5)^D,
		\qquad
		\boldsymbol{c}_{i}^{y}(0)
		\sim
		\mathcal{U}(-0.25,0.25)^D
		\label{eq:triedm_state_initialization}
	\end{equation}
	When TriEDM is triggered at rank position $i$, the E-DM states are updated as
	\begin{align}
		\boldsymbol{c}_{i}^{x}(s_i+1)
		&=
		2.66
		\left\{
		\exp\!\left[
		-\cos\!\left(
		\pi\boldsymbol{c}_{i}^{y}(s_i)
		\right)
		\right]
		-
		\boldsymbol{1}_{D}
		\right\}
		\odot
		\boldsymbol{c}_{i}^{x}(s_i)
		\label{eq:triedm_chaos_x}
		\\
		\boldsymbol{c}_{i}^{y}(s_i+1)
		&=
		\boldsymbol{c}_{i}^{y}(s_i)
		+
		\boldsymbol{c}_{i}^{x}(s_i)
		\label{eq:triedm_chaos_y}
	\end{align}
	where $s_i$ denotes the cumulative number of times TriEDM has been triggered at rank position $i$, and both
	$\exp(\cdot)$
	and
	$\cos(\cdot)$
	are applied element-wise.
	
	The component-wise lower and upper bounds of the current population are defined as
	\begin{equation}
		\boldsymbol{\ell}(t)
		=
		\min_{1\leq k\leq N}\boldsymbol{X}_{k}(t),
		\qquad
		\boldsymbol{u}(t)
		=
		\max_{1\leq k\leq N}\boldsymbol{X}_{k}(t)
		\label{eq:triedm_dynamic_bounds}
	\end{equation}
	where the vector minimum and maximum are calculated independently for each dimension.
	The updated E-DM state is then mapped to the hyper-rectangular region spanned by the current population:
	\begin{equation}
		\boldsymbol{E}_{i}(t)
		=
		\boldsymbol{\ell}(t)
		+
		\frac{
			\boldsymbol{c}_{i}^{x}(s_i+1)
			-
			\ell_c\boldsymbol{1}_{D}
		}{
			u_c-\ell_c
		}
		\odot
		\left[
		\boldsymbol{u}(t)
		-
		\boldsymbol{\ell}(t)
		+
		\varepsilon\boldsymbol{1}_{D}
		\right]
		\label{eq:triedm_chaos_mapping}
	\end{equation}
	where 
	$\ell_c=-0.5$ 
	and 
	$u_c=0.5$ 
	define the reference interval used for affine state mapping rather than hard bounds on the evolving E-DM states, and 
	$\varepsilon=2^{-52}$ 
	is adopted in this study.
	Therefore, the updated E-DM state may fall outside this reference interval, in which case Eq.~\eqref{eq:triedm_chaos_mapping} performs affine extrapolation.
	The mapped vector is subsequently clipped component-wise to the global search bounds:
	\begin{equation}
		\widehat{\boldsymbol{E}}_{i}(t)
		=
		\min\!\left\{
		\max\!\left[
		\boldsymbol{E}_{i}(t),
		\boldsymbol{L}
		\right],
		\boldsymbol{U}
		\right\}
		\label{eq:triedm_chaos_clip}
	\end{equation}
	
\begin{figure}[!htbp]
		\centering
		\includegraphics[width=0.88\linewidth]{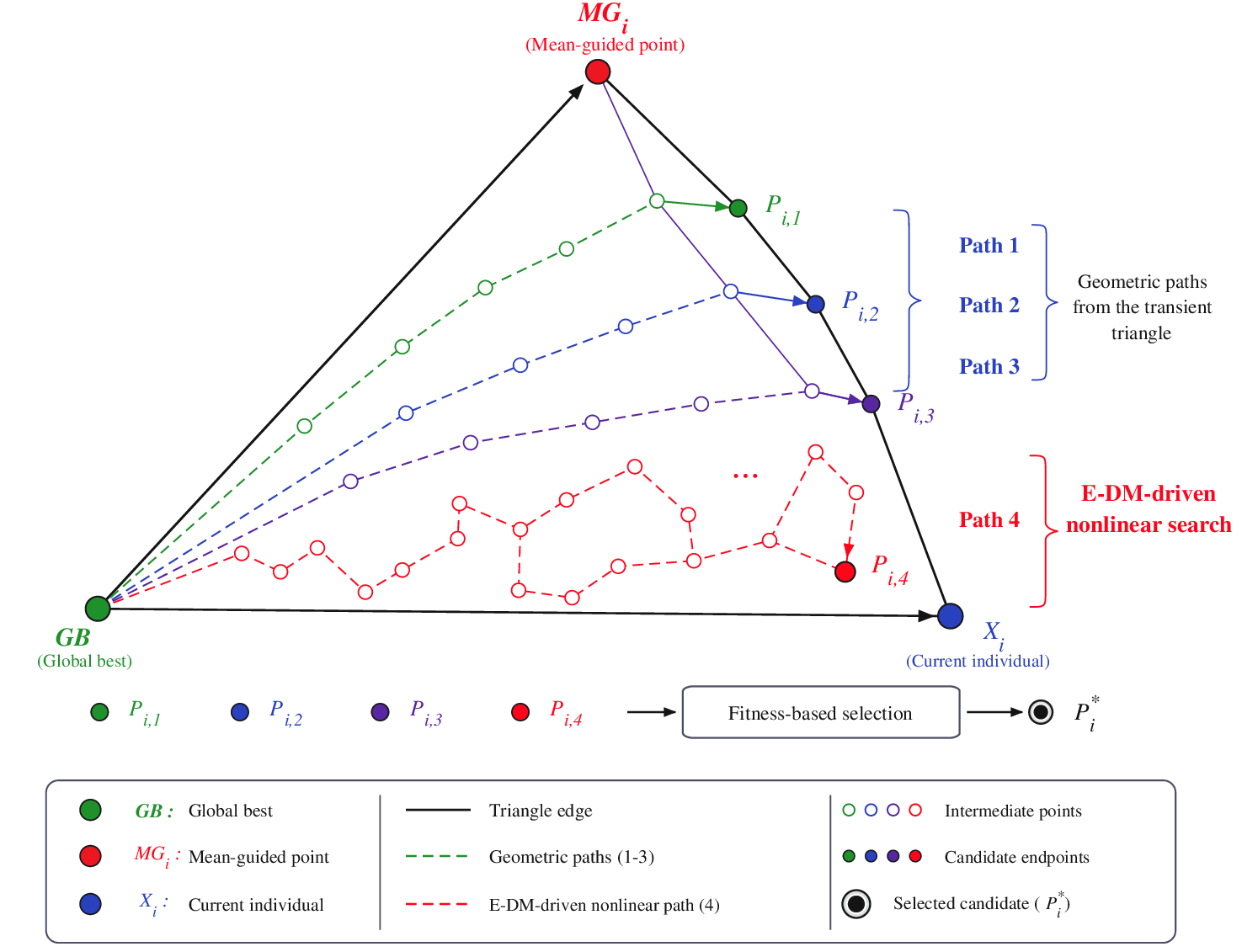}
		\caption{TriEDM multi-path search framework}
		\label{fig:triedm_paths}
	\end{figure}

	A binary mask
	$\boldsymbol{B}_{i}(t)
	=[b_{i,1}(t),\ldots,b_{i,D}(t)]^{\mathrm T}
	\in\{0,1\}^{D}$
	is then generated, where
	$b_{i,j}(t)\sim\operatorname{Bernoulli}(0.5)$
	independently across dimensions.
	Using the current individual and the global best solution as dual anchors, an intermediate mutant vector
	$\boldsymbol{V}_{i}(t)
	=[v_{i,1}(t),\ldots,v_{i,D}(t)]^{\mathrm T}$
	is constructed, whose $j$-th component is given by
	\begin{equation}
		v_{i,j}(t)
		=
		\begin{cases}
			x_{i,j}(t)
			+
			r_{i,j}^{\mathrm{m}}(t)
			\left[
			\widehat{e}_{i,j}(t)-x_{i,j}(t)
			\right],
			& b_{i,j}(t)=1
			\\[0.6em]
			x_{\mathrm{best},j}(t)
			+
			r_{i,j}^{\mathrm{m}}(t)
			\left[
			\widehat{e}_{i,j}(t)-x_{i,j}(t)
			\right],
			& b_{i,j}(t)=0
		\end{cases}
		\label{eq:triedm_mutant}
	\end{equation}
	where
	$r_{i,j}^{\mathrm{m}}(t)\sim\mathcal{U}(0,1)$
	is a random mutation step coefficient, and
	$\widehat{e}_{i,j}(t)$
	denotes the $j$-th component of
	$\widehat{\boldsymbol{E}}_{i}(t)$.
	A binomial crossover operator is subsequently employed to generate the fourth candidate solution:
	\begin{equation}
		p_{i,4,j}(t)
		=
		\begin{cases}
			v_{i,j}(t),
			&
			r_{i,j}^{\mathrm{c}}(t)<CR_i(t)
			\ \lor\
			j=j_{\mathrm{rand}}
			\\[0.5em]
			x_{i,j}(t),
			& \text{otherwise}
		\end{cases}
		\label{eq:triedm_path4}
	\end{equation}
	where
	$\boldsymbol{P}_{i,4}(t)
	=[p_{i,4,1}(t),\ldots,p_{i,4,D}(t)]^{\mathrm T}$
	denotes the fourth candidate solution;
	$CR_i(t)\sim\mathcal{U}(0,1)$
	is the crossover probability;
	$r_{i,j}^{\mathrm{c}}(t)\sim\mathcal{U}(0,1)$
	is the crossover random variable, independent of
	$r_{i,j}^{\mathrm{m}}(t)$;
	and
	$j_{\mathrm{rand}}$
	is a dimension index uniformly sampled from
	$\{1,\ldots,D\}$,
	which guarantees that at least one component of
	$\boldsymbol{P}_{i,4}(t)$
	is inherited from
	$\boldsymbol{V}_{i}(t)$.
	The first three paths provide geometric guidance, whereas the fourth path introduces E-DM-driven nonlinear perturbations.
	The overall architecture of TriEDM is illustrated in Fig.~\ref{fig:triedm_paths}.
	
	\textbf{(3) Candidate competition and update.}
	The candidate solutions generated by the four search paths form the candidate set
	\begin{equation}
		\mathcal{C}_{i}(t)
		=
		\left\{
		\boldsymbol{P}_{i,1}(t),
		\boldsymbol{P}_{i,2}(t),
		\boldsymbol{P}_{i,3}(t),
		\boldsymbol{P}_{i,4}(t)
		\right\}
		\label{eq:triedm_candidate_set}
	\end{equation}
	Let $\mathrm{FEs}$ and $\mathrm{FEs}_{\max}$ denote the current number of function evaluations and the maximum function-evaluation budget, respectively.
	The number of candidate solutions that can be evaluated in the current update is
	\begin{equation}
		K_i(t)
		=
		\min\!\left\{
		4,
		\max\!\left[0,\mathrm{FEs}_{\max}-\mathrm{FEs}\right]
		\right\}.
		\label{eq:triedm_eval_budget}
	\end{equation}
	When $K_i(t)>0$, the index set of the paths to be evaluated is defined as
	$\mathcal{Q}_{i}(t)=\{1,2,\ldots,K_i(t)\}$.
	When fewer than four function evaluations remain, the candidate paths are
	evaluated in their predefined order until the remaining budget is exhausted;
	this deterministic rule is used only for budget completion at the end of the search.
	The corresponding candidate solutions are processed using the boundary-repair rule described previously:
	\begin{equation}
		\widehat{\boldsymbol{P}}_{i,q}(t)
		=
		\operatorname{Bound}
		\left(
		\boldsymbol{P}_{i,q}(t);
		\boldsymbol{L},
		\boldsymbol{U}
		\right),
		\qquad
		q\in\mathcal{Q}_{i}(t)
		\label{eq:triedm_bounded_candidate}
	\end{equation}
	where
	$\operatorname{Bound}(\cdot;\boldsymbol{L},\boldsymbol{U})$
	denotes the boundary-repair operation defined in
	Eq.~\eqref{eq:woo_boundary_handling}.
	The fitness of each evaluated candidate, the best path index, and the corresponding best candidate solution are given by
	\begin{align}
		F_{i,q}^{\mathrm{Tri}}(t)
		&=
		f\!\left(
		\widehat{\boldsymbol{P}}_{i,q}(t)
		\right),
		\qquad q\in\mathcal{Q}_{i}(t)
		\label{eq:triedm_candidate_fitness}
		\\
		q_i^{*}
		&=
		\arg\min_{q\in\mathcal{Q}_{i}(t)}
		F_{i,q}^{\mathrm{Tri}}(t)
		\label{eq:triedm_best_path_index}
		\\
		\boldsymbol{P}_{i}^{*}(t)
		&=
		\widehat{\boldsymbol{P}}_{i,q_i^{*}}(t),
		\qquad
		F_i^{*}(t)=F_{i,q_i^{*}}^{\mathrm{Tri}}(t)
		\label{eq:triedm_best_candidate}
	\end{align}
	where
	$F_{i,q}^{\mathrm{Tri}}(t)$
	denotes the fitness value of the candidate solution generated by the $q$-th path,
	$q_i^{*}$
	is the best path index, and
	$\boldsymbol{P}_{i}^{*}(t)$
	and
	$F_i^{*}(t)$
	denote the best candidate solution and its corresponding fitness value, respectively.
	
	Let
	$\boldsymbol{X}_{i,\mathrm{main}}(t+1)$
	denote the primary search result produced by TriEDM.
	A greedy selection criterion is employed to update the individual:
	\begin{equation}
		\boldsymbol{X}_{i,\mathrm{main}}(t+1)
		=
		\begin{cases}
			\boldsymbol{P}_{i}^{*}(t),
			&
			F_i^{*}(t)<F_i(t)
			\\[0.5em]
			\boldsymbol{X}_{i}(t),
			& \text{otherwise}
		\end{cases}
		\label{eq:triedm_greedy_update}
	\end{equation}
	with the corresponding fitness value
	$F_{i,\mathrm{main}}(t+1)
	=f(\boldsymbol{X}_{i,\mathrm{main}}(t+1))$.
	When $K_i(t)=0$, the individual is retained directly, i.e.,
	$\boldsymbol{X}_{i,\mathrm{main}}(t+1)
	=\boldsymbol{X}_{i}(t)$
	and
	$F_{i,\mathrm{main}}(t+1)=F_i(t)$.
	
	A full activation of TriEDM requires four function evaluations.
	When the remaining evaluation budget is insufficient, only the candidate solutions permitted by the available budget are evaluated.
	In summary, the proposed TriEDM strategy extends the original single-candidate update into a multi-directional search through competitive selection among four search paths, thereby enhancing the complementarity among different search directions.
	
	The detailed procedure of TriEDM is presented in Algorithm~\ref{alg:triedm}.
	
	\begingroup
	\setlength{\intextsep}{3pt}
	\setlength{\abovecaptionskip}{2pt}
	\setlength{\belowcaptionskip}{2pt}
	
	\begin{algorithm}[H]
		\caption{Pseudocode of the TriEDM multi-path search mechanism}
		\label{alg:triedm}
		\footnotesize
		\linespread{0.92}\selectfont
		
		\begin{algorithmic}[1]
			\Require Fitness-sorted population
			$\{\boldsymbol{X}_{k}(t),F_k(t)\}_{k=1}^{N}$;
			global best solution $\boldsymbol{X}_{\mathrm{best}}(t)$;
			E-DM states $\boldsymbol{c}_{i}^{x}(s_i)$ and
			$\boldsymbol{c}_{i}^{y}(s_i)$;
			objective function $f$;
			bounds $\boldsymbol{L},\boldsymbol{U}$;
			$\mathrm{FEs}$ and $\mathrm{FEs}_{\max}$
			
			\Ensure Updated individual
			$\boldsymbol{X}_{i,\mathrm{main}}(t+1)$ and fitness
			$F_{i,\mathrm{main}}(t+1)$;
			updated E-DM states and $\mathrm{FEs}$
			
			\State Compute $K_i(t)$ using
			Eq.~\eqref{eq:triedm_eval_budget}
			
			\If{$K_i(t)=0$}
			\State
			$\boldsymbol{X}_{i,\mathrm{main}}(t+1)
			\gets\boldsymbol{X}_{i}(t)$,
			$F_{i,\mathrm{main}}(t+1)\gets F_i(t)$
			\State \Return
			$\boldsymbol{X}_{i,\mathrm{main}}(t+1)$,
			$F_{i,\mathrm{main}}(t+1)$,
			E-DM states, and $\mathrm{FEs}$
			\EndIf
			
			\State Randomly select $\mathcal{S}_{i}(t)$ and compute
			$\boldsymbol{M}_{i}(t)$ using
			Eq.~\eqref{eq:triedm_mean_point}
			
			\State Generate $\mathcal{A}_{i}(t)$ and
			$\boldsymbol{P}_{i,1}(t)$--$\boldsymbol{P}_{i,3}(t)$ using
			Eqs.~\eqref{eq:triedm_alpha_pool}--\eqref{eq:triedm_path3}
			
			\State Update the E-DM states using
			Eqs.~\eqref{eq:triedm_chaos_x}--\eqref{eq:triedm_chaos_y}
			
			\State Compute $\boldsymbol{\ell}(t)$ and
			$\boldsymbol{u}(t)$, and obtain
			$\widehat{\boldsymbol{E}}_{i}(t)$ using
			Eqs.~\eqref{eq:triedm_dynamic_bounds}--%
			\eqref{eq:triedm_chaos_clip}
			
			\State Generate $\boldsymbol{V}_{i}(t)$ and
			$\boldsymbol{P}_{i,4}(t)$ using
			Eqs.~\eqref{eq:triedm_mutant}--\eqref{eq:triedm_path4}
			
			\State Form $\mathcal{C}_{i}(t)$ using
			Eq.~\eqref{eq:triedm_candidate_set}
			
			\State
			$\mathcal{Q}_{i}(t)
			\gets\{1,2,\ldots,K_i(t)\}$
			
			\For{$q\in\mathcal{Q}_{i}(t)$}
			\State
			$\widehat{\boldsymbol{P}}_{i,q}(t)
			\gets
			\operatorname{Bound}\!\left(
			\boldsymbol{P}_{i,q}(t);
			\boldsymbol{L},\boldsymbol{U}
			\right)$
			
			\State
			$F_{i,q}^{\mathrm{Tri}}(t)
			\gets
			f\!\left(
			\widehat{\boldsymbol{P}}_{i,q}(t)
			\right)$;
			$\mathrm{FEs}\gets\mathrm{FEs}+1$
			\EndFor
			
			\State
			$q_i^{*}
			\gets
			\arg\min_{q\in\mathcal{Q}_{i}(t)}
			F_{i,q}^{\mathrm{Tri}}(t)$
			
			\State
			$\boldsymbol{P}_{i}^{*}(t)
			\gets
			\widehat{\boldsymbol{P}}_{i,q_i^{*}}(t)$,
			$F_i^{*}(t)
			\gets
			F_{i,q_i^{*}}^{\mathrm{Tri}}(t)$
			
			\If{$F_i^{*}(t)<F_i(t)$}
			\State
			$\boldsymbol{X}_{i,\mathrm{main}}(t+1)
			\gets\boldsymbol{P}_{i}^{*}(t)$,
			$F_{i,\mathrm{main}}(t+1)
			\gets F_i^{*}(t)$
			\Else
			\State
			$\boldsymbol{X}_{i,\mathrm{main}}(t+1)
			\gets\boldsymbol{X}_{i}(t)$,
			$F_{i,\mathrm{main}}(t+1)
			\gets F_i(t)$
			\EndIf
			
			\State $s_i\gets s_i+1$
			
			\State \Return
			$\boldsymbol{X}_{i,\mathrm{main}}(t+1)$,
			$F_{i,\mathrm{main}}(t+1)$,
			E-DM states, and $\mathrm{FEs}$
		\end{algorithmic}
	\end{algorithm}
	
	\endgroup
	
	\FloatBarrier
	
	\subsection{Correlation-Guided Groupwise Orthogonal Learning}
	\label{subsec:cgol}
	
	Orthogonal Learning (OL) exploits the balanced sampling property of orthogonal arrays to evaluate different combinations of candidate levels with a relatively small number of trials. Zhan et al. introduced OL into particle swarm optimization to construct learning exemplars~\cite{Zhan2011OrthogonalLearningPSO}, and subsequent studies further improved the candidate levels through auxiliary vectors and multi-source guidance information~\cite{Qin2015OrthogonalDesignPSO,Ma2021OrthogonalLearningBrainStorm,Bakhshpoori2023OrthogonalLearning}. However, conventional OL generally treats each decision variable as an independent factor. As the problem dimensionality increases, both the size of the orthogonal array and the function-evaluation overhead increase substantially, while the cooperative relationships among variables are difficult to exploit. To address these limitations, this paper proposes Correlation-Guided Groupwise Orthogonal Learning (CGOL), which integrates an interpolated reference solution, correlation-based variable grouping~\cite{Zhang2023PearsonGrouping}, and groupwise orthogonal combination to improve search accuracy while reducing the evaluation overhead.
	
	\textbf{(1) Interpolated reference solution.}
	
	For an individual $\boldsymbol{X}_{i}(t)$ that enters CGOL at the $t$-th iteration, an index
	$r_i\in\{1,\ldots,N\}$
	is randomly selected from the current population. A dimension-wise interpolation is then performed between the corresponding individual
	$\boldsymbol{X}_{r_i}(t)$
	and the global best solution
	$\boldsymbol{X}_{\mathrm{best}}(t)$:
	
	\begin{equation}
		\boldsymbol{X}_{i}^{\mathrm{ref}}(t)
		=
		\boldsymbol{X}_{r_i}(t)
		+
		\boldsymbol{\gamma}_{i}(t)
		\odot
		\left[
		\boldsymbol{X}_{\mathrm{best}}(t)
		-
		\boldsymbol{X}_{r_i}(t)
		\right],
		\qquad
		\boldsymbol{\gamma}_{i}(t)\sim\mathcal{U}(0,1)^D
		\label{eq:cgol_reference}
	\end{equation}
	where
	$\boldsymbol{X}_{i}^{\mathrm{ref}}(t)$
	denotes the interpolated reference solution, and
	$\boldsymbol{\gamma}_{i}(t)\in\mathbb{R}^{D}$
	is a random interpolation vector.
	The current individual
	$\boldsymbol{X}_{i}(t)$
	and the interpolated reference solution
	$\boldsymbol{X}_{i}^{\mathrm{ref}}(t)$
	jointly define the two candidate levels used in orthogonal learning.
	
	\textbf{(2) Correlation-aware variable grouping.}
	
	CGOL exploits the historical trajectory of the population center to characterize the cooperative variations among decision variables.
	The population center at the $\ell$-th iteration and the history matrix available at the $t$-th iteration are defined as
	
	\begin{align}
		\boldsymbol{\mu}(\ell)
		&=
		\frac{1}{N}
		\sum_{i=1}^{N}
		\boldsymbol{X}_{i}(\ell),
		\qquad
		\ell=1,\ldots,t-1
		\label{eq:cgol_population_center}
		\\
		\boldsymbol{H}(t)
		&=
		\begin{bmatrix}
			\boldsymbol{\mu}^{\mathrm T}(1)\\
			\boldsymbol{\mu}^{\mathrm T}(2)\\
			\vdots\\
			\boldsymbol{\mu}^{\mathrm T}(t-1)
		\end{bmatrix}
		\in\mathbb{R}^{(t-1)\times D}
		\label{eq:cgol_history_matrix}
	\end{align}
	where
	$\boldsymbol{\mu}(\ell)$
	denotes the population center and
	$\boldsymbol{H}(t)$
	is the population-center history matrix.
	When sufficient historical samples are available, let
	$\mu_p(\ell)$
	denote the $p$-th component of
	$\boldsymbol{\mu}(\ell)$
	and define
	$\bar{\mu}_{p}(t)
	=(t-1)^{-1}\sum_{\ell=1}^{t-1}\mu_p(\ell)$.
	The Pearson correlation coefficient between variables $p$ and $q$,
	where
	$p,q\in\{1,\ldots,D\}$,
	is then calculated as
	
	\begin{equation}
		\rho_{p,q}(t)
		=
		\frac{
			\displaystyle
			\sum_{\ell=1}^{t-1}
			\left[\mu_p(\ell)-\bar{\mu}_{p}(t)\right]
			\left[\mu_q(\ell)-\bar{\mu}_{q}(t)\right]
		}{
			\displaystyle
			\sqrt{
				\sum_{\ell=1}^{t-1}
				\left[\mu_p(\ell)-\bar{\mu}_{p}(t)\right]^2
			}
			\sqrt{
				\sum_{\ell=1}^{t-1}
				\left[\mu_q(\ell)-\bar{\mu}_{q}(t)\right]^2
			}
		}
		\label{eq:cgol_correlation}
	\end{equation}
	where
	$\rho_{p,q}(t)\in[-1,1]$,
	and a larger absolute value indicates a stronger correlation between the historical evolutionary trends of the two variables.
	All correlation coefficients form the correlation matrix
	$\boldsymbol{R}(t)=[\rho_{p,q}(t)]_{D\times D}$.
	If the historical variance of either variable is zero, the corresponding off-diagonal correlation coefficient is set to zero, while the diagonal element is set to one.
	
	To balance the combination granularity and evaluation overhead, the number of variable groups is set to
	$G=\min\{31,D\}$.
	When $D>31$ and sufficient historical samples are available, each row of
	$\boldsymbol{R}(t)$
	is treated as a correlation feature vector of the corresponding variable, and K-means clustering is employed to obtain the grouping labels
	
	\begin{equation}
		\boldsymbol{c}(t)
		=
		\operatorname{KMeans}
		\left(
		\boldsymbol{R}(t),G
		\right),
		\qquad
		\boldsymbol{c}(t)
		=
		[c_1(t),\ldots,c_D(t)]^{\mathrm T}
		\label{eq:cgol_grouping}
	\end{equation}
	where
	$c_j(t)\in\{1,\ldots,G\}$
	denotes the group index of the $j$-th variable.
	When $D\leq31$, each variable forms an individual group, i.e.,
	$c_j(t)=j$.
	When insufficient historical samples are available, balanced grouping is adopted.
	After sufficient historical information has been accumulated, the correlation matrix and grouping results are updated every 20 iterations without consuming the function-evaluation budget.
	
	\textbf{(3) Groupwise orthogonal learning and individual update.}
	
	Each variable group is treated as a two-level orthogonal factor.
	For $q$ factors, the required number of orthogonal trials and the number of function evaluations for one OL invocation are respectively given by
	
	\begin{equation}
		M(q)
		=
		2^{\left\lceil\log_{2}(q+1)\right\rceil},
		\qquad
		C_{\mathrm{FE}}(q)
		=
		M(q)+1,
		\qquad
		q\in\{D,G\}
		\label{eq:cgol_cost}
	\end{equation}
	where
	$q=D$
	and
	$q=G$
	correspond to full-dimensional OL and CGOL, respectively.
	The additional function evaluation is used for the predicted solution.
	Since $G<D$ in most high-dimensional cases, groupwise processing reduces the scale of the required orthogonal experiments.
	
	Let
	$\boldsymbol{O}
	=[o_{m,g}]_{M_G\times G}
	\in\{0,1\}^{M_G\times G}$
	denote a groupwise two-level orthogonal array, where
	$M_G=M(G)$.
	The $m$-th orthogonal trial generates a candidate solution
	$\boldsymbol{X}_{i,m}^{\mathrm{OA}}(t)$,
	whose $j$-th component is defined as
	
	\begin{equation}
		x_{i,m,j}^{\mathrm{OA}}(t)
		=
		\begin{cases}
			x_{i,j}(t),
			& o_{m,c_j(t)}=0,\\[0.4em]
			x_{i,j}^{\mathrm{ref}}(t),
			& o_{m,c_j(t)}=1,
		\end{cases}
		\qquad
		m=1,\ldots,M_G
		\label{eq:cgol_oa_candidate}
	\end{equation}
	where all variables belonging to the same group share the same candidate level.
	After applying the boundary-repair operation defined in
	Eq.~\eqref{eq:woo_boundary_handling},
	the repaired trial solution and its fitness value are denoted by
	$\widehat{\boldsymbol{X}}_{i,m}^{\mathrm{OA}}(t)$
	and
	$F_{i,m}^{\mathrm{OA}}(t)
	=f(\widehat{\boldsymbol{X}}_{i,m}^{\mathrm{OA}}(t))$,
	respectively.
	
	For the $g$-th group, the average fitness corresponding to level
	$b\in\{0,1\}$
	is defined as
	
	\begin{equation}
		\phi_{i,g}^{(b)}(t)
		=
		\frac{1}{\left|\mathcal{I}_{g}^{(b)}\right|}
		\sum_{m\in\mathcal{I}_{g}^{(b)}}
		F_{i,m}^{\mathrm{OA}}(t),
		\qquad
		\mathcal{I}_{g}^{(b)}
		=
		\left\{m\mid o_{m,g}=b\right\}
		\label{eq:cgol_level_fitness}
	\end{equation}
	where
	$\mathcal{I}_{g}^{(b)}$
	denotes the index set of orthogonal trials in which the $g$-th group adopts level $b$, and
	$\phi_{i,g}^{(b)}(t)$
	is the corresponding average fitness value.
	For a minimization problem, the level with the lower average fitness is selected for each group to construct the predicted solution
	$\boldsymbol{X}_{i}^{\mathrm{OL}}(t)$:
	
	\begin{equation}
		x_{i,j}^{\mathrm{OL}}(t)
		=
		\begin{cases}
			x_{i,j}(t),
			& \phi_{i,c_j(t)}^{(0)}(t)
			\leq
			\phi_{i,c_j(t)}^{(1)}(t),\\[0.4em]
			x_{i,j}^{\mathrm{ref}}(t),
			& \text{otherwise}
		\end{cases}
		\label{eq:cgol_prediction}
	\end{equation}
	
	The predicted solution after boundary repair is denoted by
	$\widehat{\boldsymbol{X}}_{i}^{\mathrm{OL}}(t)$,
	with fitness
	$F_i^{\mathrm{OL}}(t)
	=f(\widehat{\boldsymbol{X}}_{i}^{\mathrm{OL}}(t))$.
	A greedy selection criterion is subsequently employed to update the individual
	
	\begin{equation}
		\boldsymbol{X}_{i}(t+1)
		=
		\begin{cases}
			\widehat{\boldsymbol{X}}_{i}^{\mathrm{OL}}(t),
			& F_i^{\mathrm{OL}}(t)<F_i(t),\\[0.4em]
			\boldsymbol{X}_{i}(t),
			& \text{otherwise}
		\end{cases}
		\label{eq:cgol_greedy_update}
	\end{equation}
	
	and
	$F_i(t+1)=\min\{F_i^{\mathrm{OL}}(t),F_i(t)\}$.
	
	To control the function-evaluation overhead during the early search stage, CGOL employs a linearly increasing activation probability:
	
	\begin{equation}
		p_{\mathrm{OL}}(t)
		=
		0.03
		+
		0.22\frac{t}{T_{\max}}
		\label{eq:cgol_probability}
	\end{equation}
	where
	$p_{\mathrm{OL}}(t)$
	denotes the probability of activating CGOL at the $t$-th iteration, such that the mechanism primarily operates during the middle and later stages of the search.
	If the remaining function-evaluation budget is smaller than
	$C_{\mathrm{FE}}(G)$,
	the current CGOL invocation is skipped.
	The execution procedure of CGOL is summarized in Algorithm~\ref{alg:cgol}.
	
	\par\vspace{1mm}
	
	\noindent
	\begin{minipage}{\linewidth}
		
		\footnotesize
		\setlength{\parskip}{0pt}
		
		\noindent\rule{\linewidth}{0.5pt}
		
		\vspace{-1.5mm}
		
		\captionof{algorithm}{Pseudocode of the Correlation-Guided Groupwise Orthogonal Learning mechanism}
		\label{alg:cgol}
		
		\vspace{-1.5mm}
		
		\noindent\rule{\linewidth}{0.5pt}
		
		\vspace{0.5mm}
		
		\begin{algorithmic}[1]
			
			\Require Population
			$\{\boldsymbol{X}_{k}(t),F_k(t)\}_{k=1}^{N}$;
			global best solution $\boldsymbol{X}_{\mathrm{best}}(t)$;
			grouping vector $\boldsymbol{c}(t)$ and group number $G$;
			objective function $f$;
			bounds $\boldsymbol{L},\boldsymbol{U}$;
			$\mathrm{FEs}$ and $\mathrm{FEs}_{\max}$
			
			\Ensure Updated individual
			$\boldsymbol{X}_{i}(t+1)$
			and fitness
			$F_i(t+1)$;
			updated $\mathrm{FEs}$
			
			\State Compute $p_{\mathrm{OL}}(t)$ using
			Eq.~\eqref{eq:cgol_probability}
			
			\State Generate
			$r_{\mathrm{OL}}\sim\mathcal{U}(0,1)$
			and set $M_G\gets M(G)$
			
			\If{$r_{\mathrm{OL}}>p_{\mathrm{OL}}(t)
				\ \lor\
				\mathrm{FEs}_{\max}-\mathrm{FEs}<M_G+1$}
			
			\State
			$\boldsymbol{X}_{i}(t+1)
			\gets
			\boldsymbol{X}_{i}(t)$,
			$F_i(t+1)\gets F_i(t)$
			
			\State
			\Return
			$\boldsymbol{X}_{i}(t+1)$,
			$F_i(t+1)$,
			and $\mathrm{FEs}$
			
			\EndIf
			
			\State Randomly select
			$r_i\in\{1,\ldots,N\}$
			and construct
			$\boldsymbol{X}_{i}^{\mathrm{ref}}(t)$
			using Eq.~\eqref{eq:cgol_reference}
			
			\State Construct the two-level orthogonal array
			$\boldsymbol{O}$
			
			\For{$m=1$ to $M_G$}
			
			\State Generate
			$\boldsymbol{X}_{i,m}^{\mathrm{OA}}(t)$
			using Eq.~\eqref{eq:cgol_oa_candidate}
			
			\State Repair
			$\boldsymbol{X}_{i,m}^{\mathrm{OA}}(t)$
			and evaluate
			$F_{i,m}^{\mathrm{OA}}(t)$
			
			\State
			$\mathrm{FEs}\gets\mathrm{FEs}+1$
			
			\EndFor
			
			\For{$g=1$ to $G$}
			
			\State Compute
			$\phi_{i,g}^{(0)}(t)$
			and
			$\phi_{i,g}^{(1)}(t)$
			using Eq.~\eqref{eq:cgol_level_fitness}
			
			\EndFor
			
			\State Construct
			$\boldsymbol{X}_{i}^{\mathrm{OL}}(t)$
			using Eq.~\eqref{eq:cgol_prediction}
			
			\State Repair
			$\boldsymbol{X}_{i}^{\mathrm{OL}}(t)$
			and evaluate
			$F_i^{\mathrm{OL}}(t)$;
			$\mathrm{FEs}\gets\mathrm{FEs}+1$
			
			\State Update
			$\boldsymbol{X}_{i}(t+1)$
			using Eq.~\eqref{eq:cgol_greedy_update}
			
			\State
			$F_i(t+1)
			\gets
			\min\{F_i^{\mathrm{OL}}(t),F_i(t)\}$
			
			\State
			\Return
			$\boldsymbol{X}_{i}(t+1)$,
			$F_i(t+1)$,
			and $\mathrm{FEs}$
			
		\end{algorithmic}
		
		\vspace{0.5mm}
		
		\noindent\rule{\linewidth}{0.5pt}
		
	\end{minipage}
	
	\vspace{2mm}
	
	To investigate the effects of the interpolated reference solution and groupwise orthogonal learning, three OL variants are considered:
	V1\_OL represents the original OL method;
	V2\_OL incorporates the interpolated reference solution into the original OL;
	and V3\_OL further introduces Pearson correlation-based grouping into V2\_OL.
	Experiments are conducted on F1, F7, and F15 of the CEC2017 benchmark suite with
	$D=50$
	and
	$N=30$.
	Each variant is independently run 30 times using the same initial populations and random seeds.
	The average convergence curves are shown in Fig.~\ref{fig:cgol_module_convergence}.
	
	\par\vspace{1mm}
	
	\noindent
	\begin{minipage}{\linewidth}
		\centering
		\captionsetup{skip=2pt}
		
		\includegraphics[
		width=\linewidth,
		height=0.17\textheight,
		keepaspectratio
		]{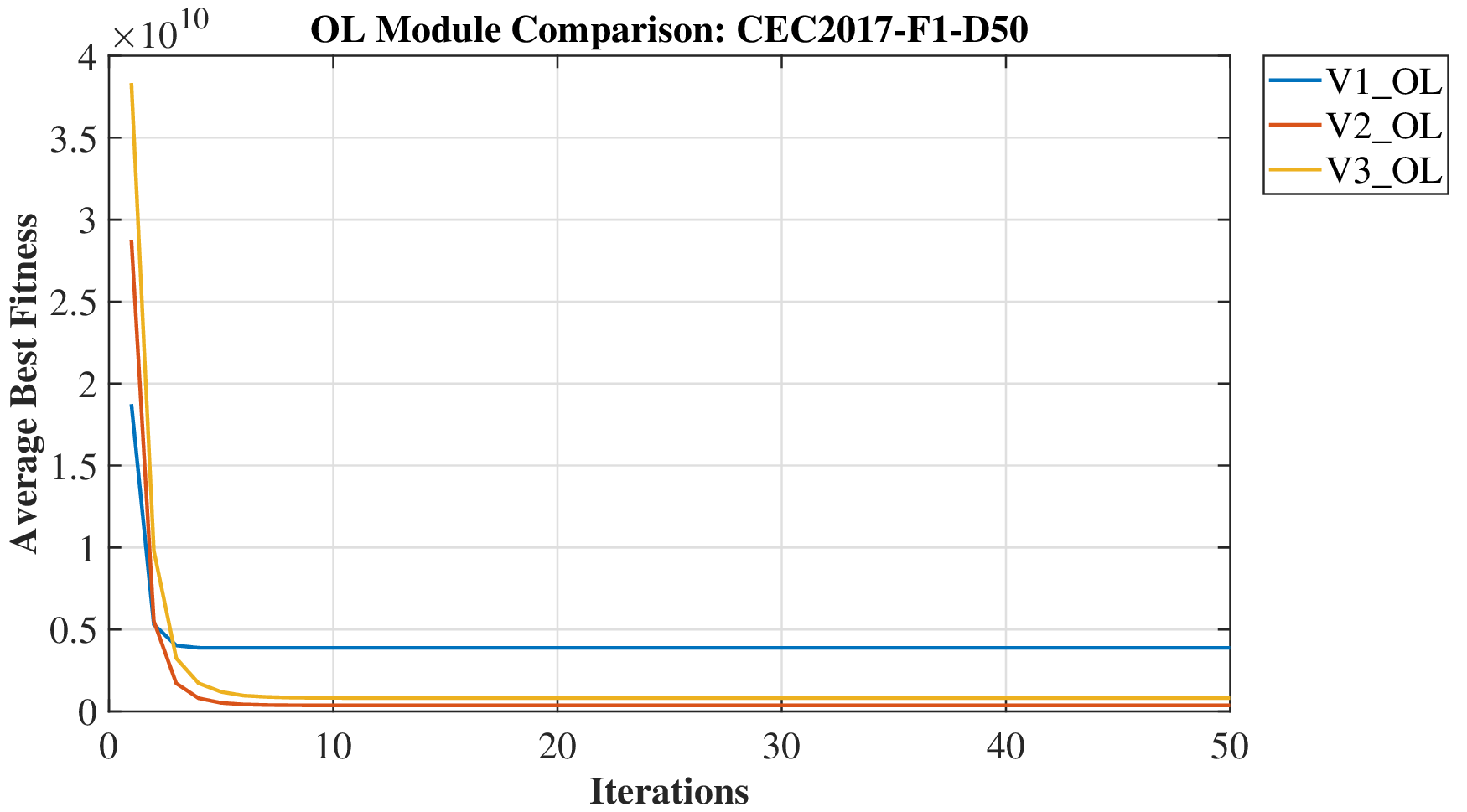}
		
		\vspace{-1mm}
		
		{\footnotesize (a) F1}
		
		\vspace{1mm}
		
		\includegraphics[
		width=\linewidth,
		height=0.17\textheight,
		keepaspectratio
		]{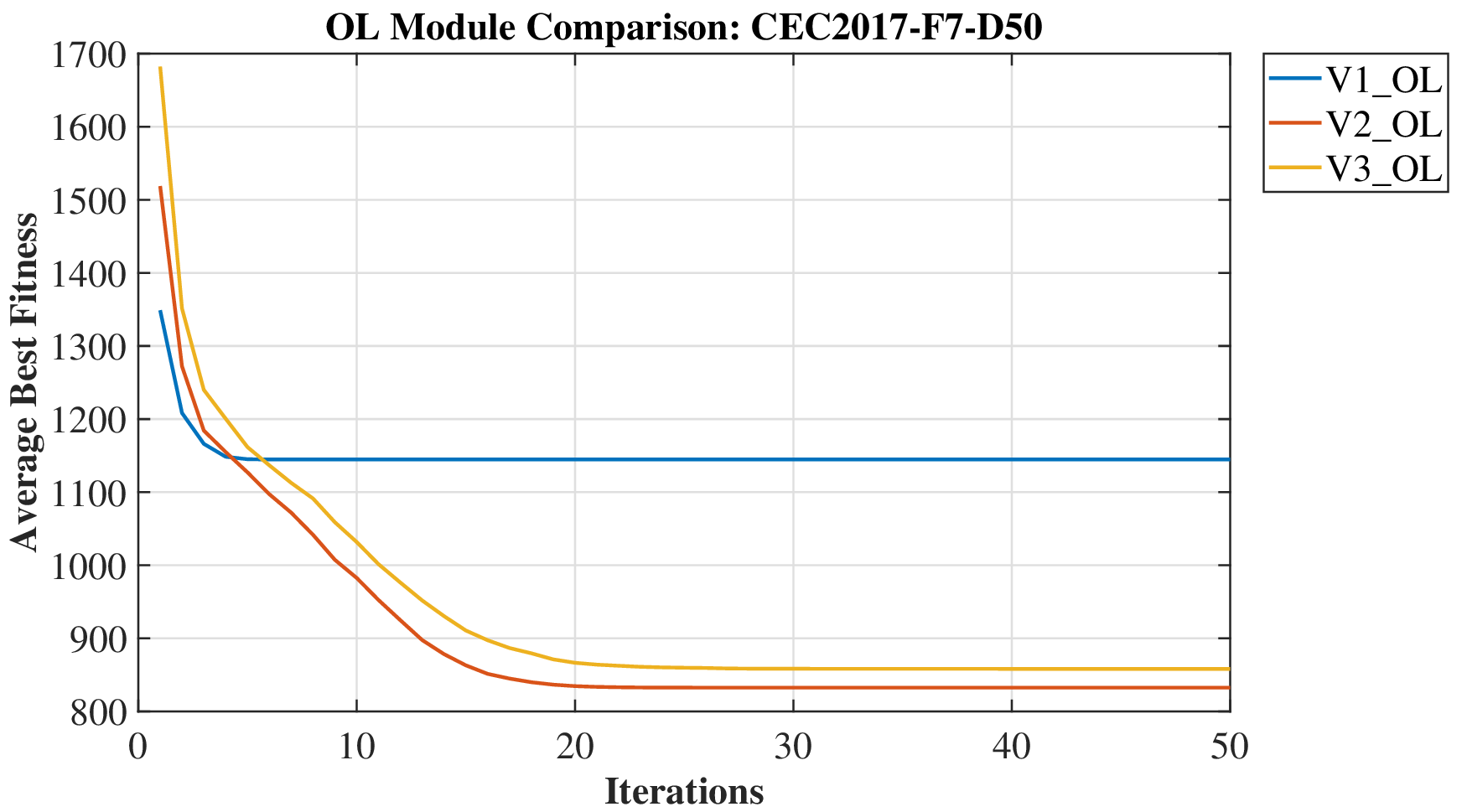}
		
		\vspace{-1mm}
		
		{\footnotesize (b) F7}
		
		\vspace{1mm}
		
		\includegraphics[
		width=\linewidth,
		height=0.17\textheight,
		keepaspectratio
		]{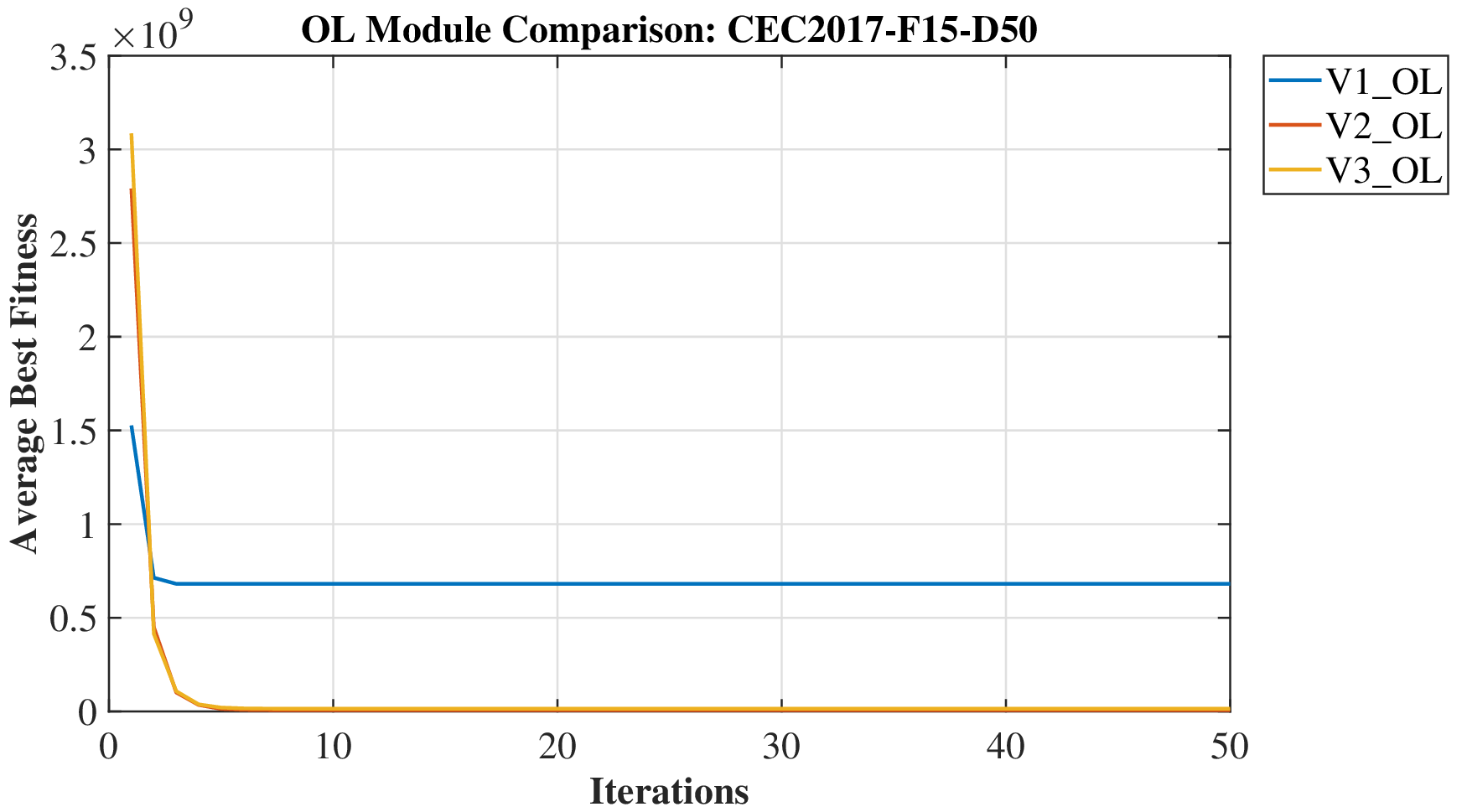}
		
		\vspace{-1mm}
		
		{\footnotesize (c) F15}
		
		\vspace{1mm}
		
		\captionof{figure}{Comparison of convergence curves for different OL variants}
		\label{fig:cgol_module_convergence}
		
	\end{minipage}
	
	\vspace{2mm}
	
	V1\_OL exhibits early stagnation on all three test functions.
	After introducing the interpolated reference solution, V2\_OL achieves higher search accuracy.
	Although the search accuracy of V3\_OL is slightly lower than that of V2\_OL, it remains consistently better than V1\_OL, indicating that the groupwise combination preserves effective search capability despite the reduced combination granularity.
	The function-evaluation overhead of the three OL variants is compared in Fig.~\ref{fig:cgol_module_fes}.
	
	\par\vspace{1mm}
	
	\noindent
	\begin{minipage}{\linewidth}
		\centering
		\captionsetup{skip=2pt}
		
		\includegraphics[
		width=0.96\linewidth,
		keepaspectratio
		]{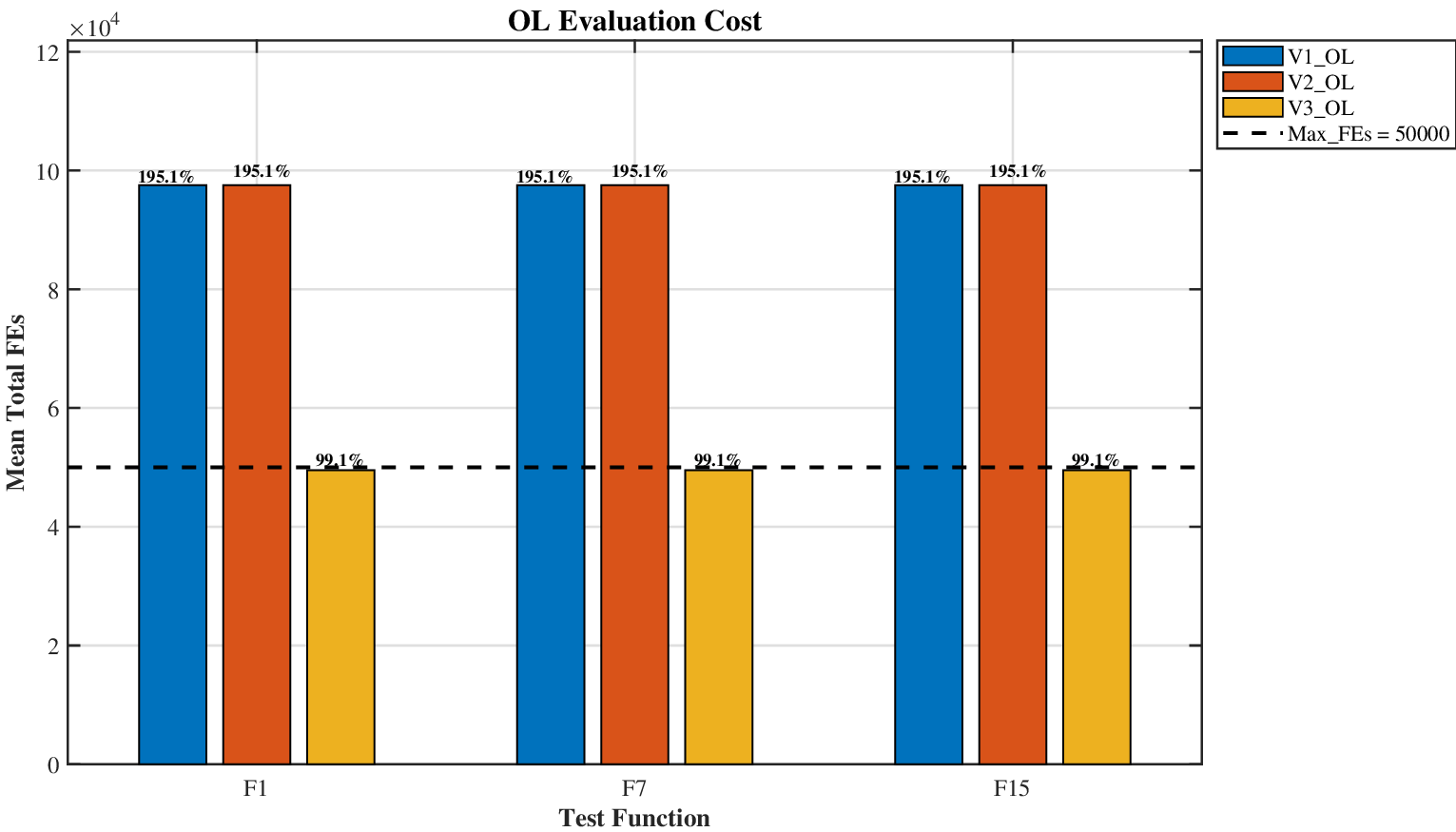}
		
		\vspace{-1mm}
		
		\captionof{figure}{Comparison of function-evaluation overhead for different OL variants}
		\label{fig:cgol_module_fes}
		
	\end{minipage}
	
	\vspace{2mm}
	
	The cumulative numbers of OL function evaluations for V1\_OL and V2\_OL are both
	$97\,500$, whereas V3\_OL requires only
	$49\,500$ evaluations, corresponding to a reduction of approximately
	$49.23\%$.
	These results indicate that the interpolated reference solution mainly improves the quality of the candidate levels, whereas groupwise orthogonal learning substantially reduces the function-evaluation overhead.
	
	\FloatBarrier
	
	\subsection{Adaptive Stagnation-Aware Runge--Kutta}
	\label{subsec:rk_refinement}
	
	The Runge--Kutta (RK) method estimates an update direction through multiple intermediate states and combines them in a weighted manner~\cite{Runge1895NumerischeAufloesung,Kutta1901Integration}. Its multi-stage structure has also been employed to construct gradient-free optimization search mechanisms~\cite{Ahmadianfar2021RUN}. Rather than adopting RK as a complete search framework, this study only draws on its multi-stage direction-estimation principle and proposes an Adaptive Stagnation-Aware Runge--Kutta (ASRK) mechanism to conditionally refine individuals that have entered a stagnation state.
	
	\textbf{(1) Stagnation detection and dynamic triggering.}
	Let
	$\boldsymbol{X}_{i,\mathrm{main}}(t+1)$
	denote the individual obtained after the main search and greedy selection at the $t$-th iteration, with fitness
	$F_{i,\mathrm{main}}(t+1)$.
	Let
	$F_{\mathrm{best}}^{-}(t+1)$
	denote the global best fitness before the global best state is updated using this individual.
	After the main search, the stagnation counter is updated as
	\begin{equation}
		\zeta_{i,\mathrm{main}}(t+1)
		=
		\begin{cases}
			0,
			& F_{i,\mathrm{main}}(t+1)<F_{\mathrm{best}}^{-}(t+1),\\[0.5em]
			\zeta_i(t)+1,
			& \text{otherwise}
		\end{cases}
		\label{eq:rk_stagnation_counter}
	\end{equation}
	where 
	$\zeta_{i,\mathrm{main}}(t+1)$ 
	denotes the global-best-improvement-based stagnation counter after the main search, and 
	$\zeta_i(t)$ 
	is its value before the current update. 
	The counter is reset to zero only when the individual produces a new global best solution; otherwise, it is incremented by one. 
	After this criterion is evaluated, the current global best state is updated according to 
	$\boldsymbol{X}_{i,\mathrm{main}}(t+1)$. 
	
	ASRK employs an iteration-dependent trigger probability that increases as the search proceeds
	\begin{equation}
		\begin{aligned}
			p_{\mathrm{ASRK}}(t)
			&=
			p_{\min}
			+
			\left(p_{\max}-p_{\min}\right)
			\left(\frac{t}{T_{\max}}\right)^2,
			\\[-0.5mm]
			p_{\min}
			&=
			0.05,
			\qquad
			p_{\max}
			=
			0.40
		\end{aligned}
		\label{eq:rk_trigger_probability}
	\end{equation}
	where
	$p_{\min}$
	and
	$p_{\max}$
	denote the minimum and maximum trigger probabilities, respectively.
	The minimum stagnation threshold is set to
	$\zeta_{\min}=3$,
	and
	$\xi_i(t)\sim\mathcal{U}(0,1)$
	is generated.
	ASRK is activated only when all of the following conditions are satisfied:
	\begin{equation}
		\zeta_{i,\mathrm{main}}(t+1)\geq\zeta_{\min},
		\qquad
		\xi_i(t)<p_{\mathrm{ASRK}}(t),
		\qquad
		\mathrm{FEs}<\mathrm{FEs}_{\max}
		\label{eq:rk_trigger_condition}
	\end{equation}
	Otherwise,
	$\boldsymbol{X}_i(t+1)=\boldsymbol{X}_{i,\mathrm{main}}(t+1)$,
	$F_i(t+1)=F_{i,\mathrm{main}}(t+1)$,
	and
	$\zeta_i(t+1)=\zeta_{i,\mathrm{main}}(t+1)$.
	
	\textbf{(2) Multi-stage directional correction.}
	When the condition in Eq.~\eqref{eq:rk_trigger_condition} is satisfied,
	$\boldsymbol{X}_{i,\mathrm{main}}(t+1)$
	is taken as the starting point for correction.
	A decreasing step-size factor is defined as
	\begin{equation}
		h(t)
		=
		1-
		\left(\frac{t}{T_{\max}}\right)^2
		\label{eq:rk_start_step}
	\end{equation}
	
	To introduce dimension-wise stochastic perturbations along the direction guided by the global best solution, a randomly weighted direction operator is defined as
	\begin{equation}
		\boldsymbol{\Phi}
		\left(
		\boldsymbol{z},
		\boldsymbol{y};
		\boldsymbol{r}
		\right)
		=
		\left(
		\boldsymbol{y}-\boldsymbol{z}
		\right)
		\odot
		\left(
		0.1\boldsymbol{1}_{D}
		+
		0.9\boldsymbol{r}
		\right),
		\qquad
		\boldsymbol{r}\sim\mathcal{U}(0,1)^D
		\label{eq:rk_direction_operator}
	\end{equation}
	where
	$\boldsymbol{z},\boldsymbol{y}\in\mathbb{R}^{D}$
	denote the current state and guiding state, respectively, and
	$\boldsymbol{r}$
	is a random weighting vector.
	
	Using the updated global best solution
	$\boldsymbol{X}_{\mathrm{best}}(t+1)$
	as the guiding state, and letting
	$\boldsymbol{r}_{i,\ell}(t)\sim\mathcal{U}(0,1)^D$
	$(\ell=1,2,3,4)$
	be mutually independent random vectors, the four stage directions are defined as
	\begin{equation}
		\left\{
		\begin{aligned}
			\boldsymbol{k}_{i,1}(t)
			&=
			\boldsymbol{\Phi}
			\left(
			\boldsymbol{X}_{i,\mathrm{main}}(t+1),
			\boldsymbol{X}_{\mathrm{best}}(t+1);
			\boldsymbol{r}_{i,1}(t)
			\right)
			\\
			\boldsymbol{k}_{i,2}(t)
			&=
			\boldsymbol{\Phi}
			\left(
			\boldsymbol{X}_{i,\mathrm{main}}(t+1)
			+
			\frac{h(t)}{2}\boldsymbol{k}_{i,1}(t),
			\boldsymbol{X}_{\mathrm{best}}(t+1);
			\boldsymbol{r}_{i,2}(t)
			\right)
			\\
			\boldsymbol{k}_{i,3}(t)
			&=
			\boldsymbol{\Phi}
			\left(
			\boldsymbol{X}_{i,\mathrm{main}}(t+1)
			+
			\frac{h(t)}{2}\boldsymbol{k}_{i,2}(t),
			\boldsymbol{X}_{\mathrm{best}}(t+1);
			\boldsymbol{r}_{i,3}(t)
			\right)
			\\
			\boldsymbol{k}_{i,4}(t)
			&=
			\boldsymbol{\Phi}
			\left(
			\boldsymbol{X}_{i,\mathrm{main}}(t+1)
			+
			h(t)\boldsymbol{k}_{i,3}(t),
			\boldsymbol{X}_{\mathrm{best}}(t+1);
			\boldsymbol{r}_{i,4}(t)
			\right)
		\end{aligned}
		\right.
		\label{eq:rk_four_stages}
	\end{equation}
	
	Following the classical fourth-order Runge--Kutta weighting pattern of
	$1:2:2:1$,
	the combined correction direction is obtained as
	\begin{equation}
		\boldsymbol{d}_{i}^{\mathrm{RK}}(t)
		=
		\frac{
			\boldsymbol{k}_{i,1}(t)
			+
			2\boldsymbol{k}_{i,2}(t)
			+
			2\boldsymbol{k}_{i,3}(t)
			+
			\boldsymbol{k}_{i,4}(t)
		}{6}
		\label{eq:rk_combined_direction}
	\end{equation}
	
	After applying the boundary-repair rule, the ASRK candidate solution is defined as
	\begin{equation}
		\widehat{\boldsymbol{X}}_{i}^{\mathrm{RK}}(t+1)
		=
		\operatorname{Bound}
		\left(
		\boldsymbol{X}_{i,\mathrm{main}}(t+1)
		+
		h(t)\boldsymbol{d}_{i}^{\mathrm{RK}}(t);
		\boldsymbol{L},
		\boldsymbol{U}
		\right)
		\label{eq:rk_candidate}
	\end{equation}
	with fitness
	$F_i^{\mathrm{RK}}(t+1)
	=f(\widehat{\boldsymbol{X}}_{i}^{\mathrm{RK}}(t+1))$.
	
	\textbf{(3) Greedy selection.}
	After evaluating the objective function of the ASRK candidate solution, a greedy selection criterion is employed to determine the updated individual:
	\begin{equation}
		\boldsymbol{X}_{i}(t+1)
		=
		\begin{cases}
			\widehat{\boldsymbol{X}}_{i}^{\mathrm{RK}}(t+1),
			& F_i^{\mathrm{RK}}(t+1)<F_{i,\mathrm{main}}(t+1),\\[0.5em]
			\boldsymbol{X}_{i,\mathrm{main}}(t+1),
			& \text{otherwise}
		\end{cases}
		\label{eq:rk_greedy_update}
	\end{equation}
	and
	$F_i(t+1)
	=\min\{F_i^{\mathrm{RK}}(t+1),F_{i,\mathrm{main}}(t+1)\}$.
	If the ASRK candidate solution is accepted, the stagnation counter is reset as
	$\zeta_i(t+1)=0$;
	otherwise,
	$\zeta_i(t+1)=\zeta_{i,\mathrm{main}}(t+1)$.
	If the final individual further improves upon the current global best solution, the global best state is updated accordingly.
	
	To analyze the activation behavior of ASRK, F1, F3, F7, F15, F21, and F28 from the CEC2017 benchmark suite are selected as representative test functions.
	The Application Rate (AR)~\cite{Cassis2026MHMetLib} and Success Rate (SR)~\cite{Wang2011CoDE} are employed as behavioral indicators:
	\begin{equation}
		\mathrm{AR}
		=
		\frac{N_{\mathrm{ASRK}}}{NT_{\max}},
		\qquad
		\mathrm{SR}
		=
		\frac{N_{\mathrm{succ}}}{N_{\mathrm{ASRK}}}
		\label{eq:rk_behavior_metrics}
	\end{equation}
	where
	$N_{\mathrm{ASRK}}$
	denotes the actual number of ASRK activations, and
	$N_{\mathrm{succ}}$
	denotes the number of ASRK candidate solutions accepted by the greedy selection criterion.
	
	As shown in Fig.~\ref{fig:rk_behavior_statistics}, the AR of ASRK on the six representative functions ranges from
	$8.44\%$ to $15.46\%$,
	while the corresponding SR ranges from
	$11.88\%$ to $96.20\%$.
	The relatively low AR indicates that ASRK is activated only for a subset of stagnated individuals, whereas the comparatively high SR observed on most test functions indicates that the mechanism is effective in generating improved candidate solutions.
	
	\par\vspace{1mm}
	
	\noindent
	\begin{minipage}{\linewidth}
		\centering
		\captionsetup{skip=2pt}
		
		\includegraphics[
		width=0.96\linewidth,
		keepaspectratio
		]{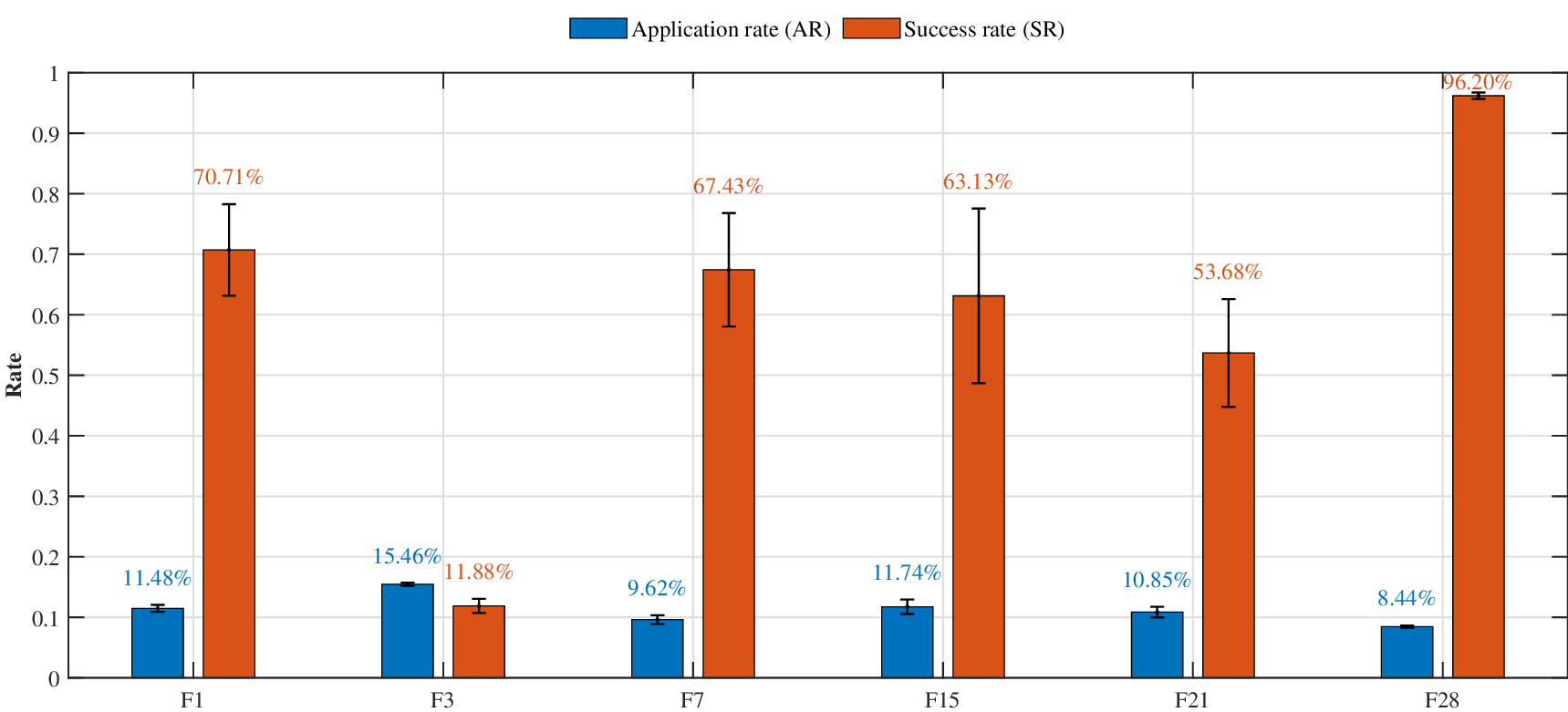}
		
		\vspace{-1mm}
		
		\captionof{figure}{Behavioral statistics of ASRK on the representative test functions}
		\label{fig:rk_behavior_statistics}
	\end{minipage}
	
	\vspace{2mm}
	
	Figure~\ref{fig:rk_events_iterations} shows that the AR generally increases as the iterations proceed, which is consistent with the dynamic triggering scheme defined in Eq.~\eqref{eq:rk_trigger_probability}.
	Meanwhile, most representative functions maintain relatively high SR values over extended stages of the search.
	Overall, ASRK conditionally refines stagnated individuals at a relatively low activation frequency, thereby alleviating search stagnation and promoting continued optimization.
	
	\par\vspace{1mm}
	
	\noindent
	\begin{minipage}{\linewidth}
		\centering
		\captionsetup{skip=2pt}
		
		\begin{minipage}[c]{0.49\linewidth}
			\centering
			\includegraphics[
			width=\linewidth,
			height=0.105\textheight,
			keepaspectratio
			]{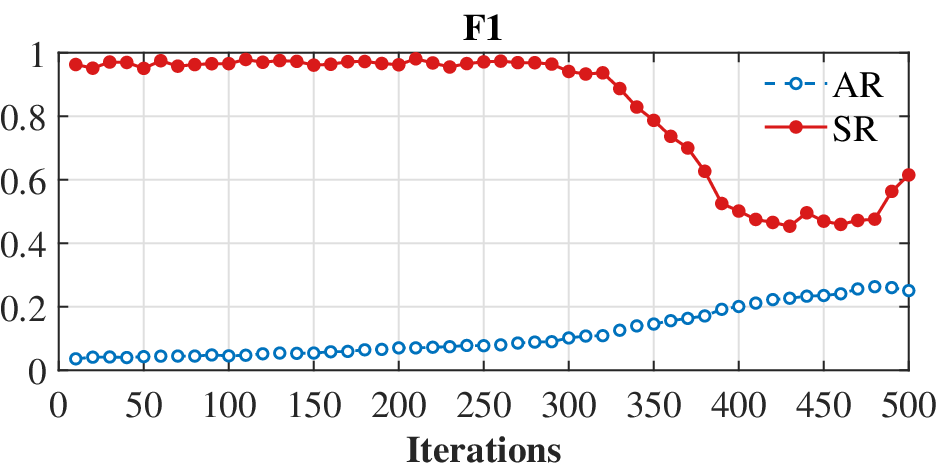}
		\end{minipage}
		\hfill
		\begin{minipage}[c]{0.49\linewidth}
			\centering
			\includegraphics[
			width=\linewidth,
			height=0.105\textheight,
			keepaspectratio
			]{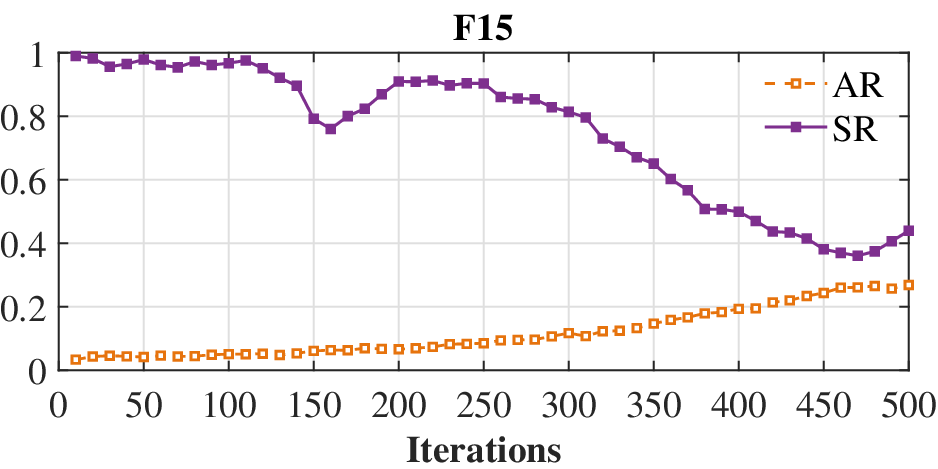}
		\end{minipage}
		
		\vspace{-1mm}
		
		\begin{minipage}[c]{0.49\linewidth}
			\centering
			\includegraphics[
			width=\linewidth,
			height=0.105\textheight,
			keepaspectratio
			]{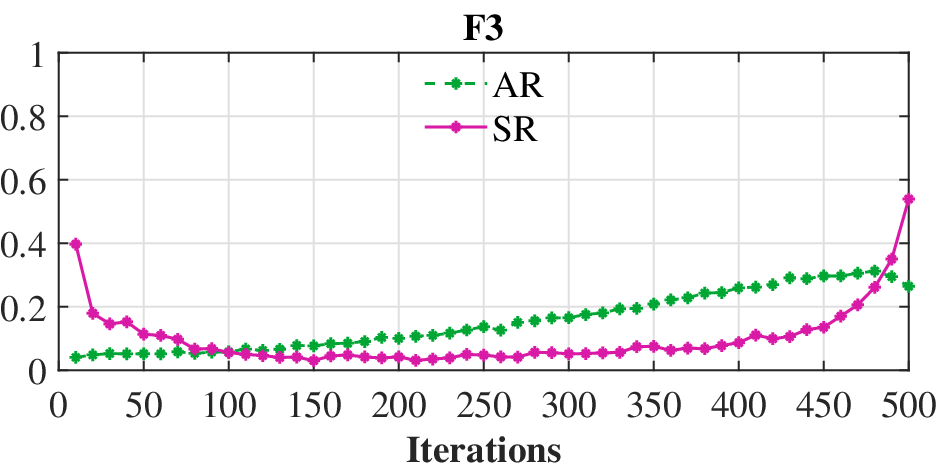}
		\end{minipage}
		\hfill
		\begin{minipage}[c]{0.49\linewidth}
			\centering
			\includegraphics[
			width=\linewidth,
			height=0.105\textheight,
			keepaspectratio
			]{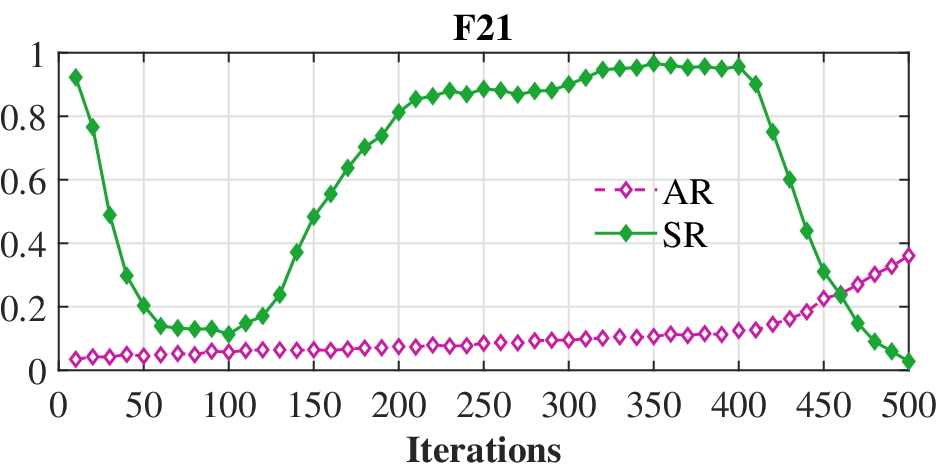}
		\end{minipage}
		
		\vspace{-1mm}
		
		\begin{minipage}[c]{0.49\linewidth}
			\centering
			\includegraphics[
			width=\linewidth,
			height=0.105\textheight,
			keepaspectratio
			]{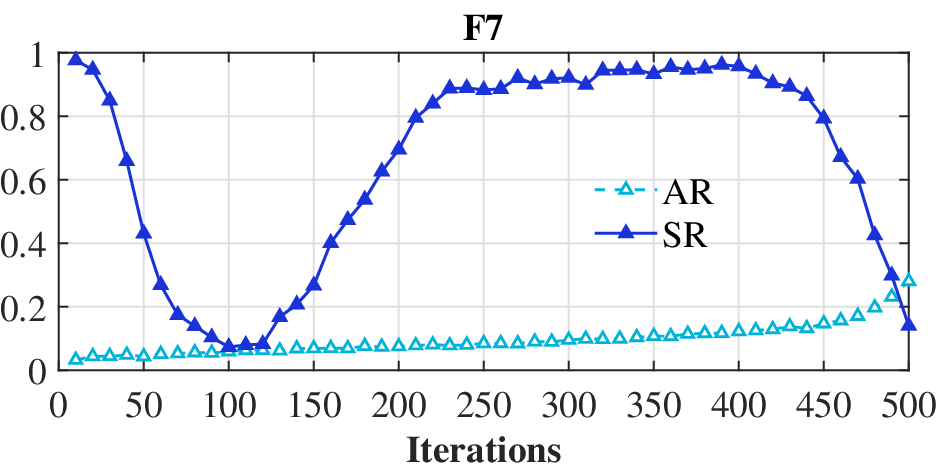}
		\end{minipage}
		\hfill
		\begin{minipage}[c]{0.49\linewidth}
			\centering
			\includegraphics[
			width=\linewidth,
			height=0.105\textheight,
			keepaspectratio
			]{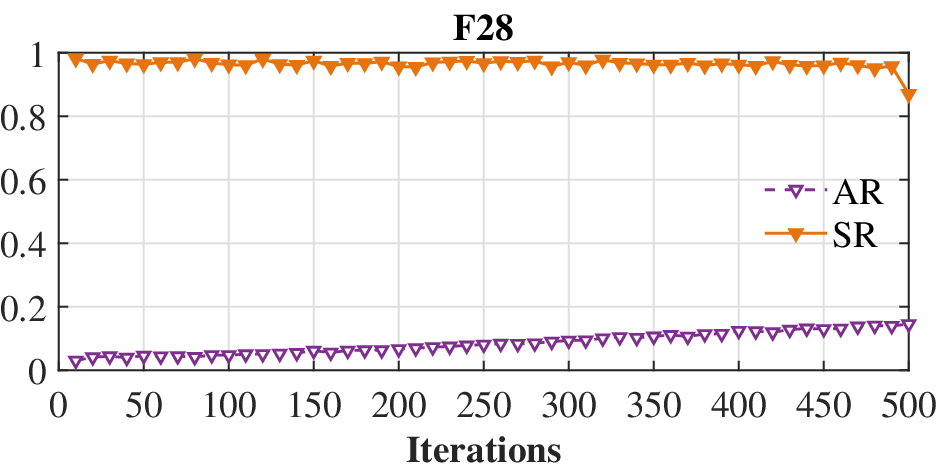}
		\end{minipage}
		
		\vspace{0.5mm}
		
		\captionof{figure}{Evolution of AR and SR for ASRK on the representative test functions}
		\label{fig:rk_events_iterations}
	\end{minipage}
	
	\vspace{2mm}

	\subsection{Overview of IWOO}
	\label{subsec:iwoo_overall_framework}
	
	IWOO integrates four complementary strategies into the original WOO framework.
	The GPS--OBL strategy improves the coverage and solution quality of the initial population through candidate expansion and competitive selection.
	After entering the main iterative process, the algorithm employs the TriEDM search branch to generate candidate solutions along multiple search directions.
	Upon completion of the main search, ASRK-based refinement is conditionally activated, while CGOL further improves individuals through periodic variable grouping and orthogonal learning.
	These strategies act collaboratively during the initialization, main search, and subsequent refinement stages.
	The overall workflow of IWOO and the interactions among its constituent modules are illustrated in Fig.~\ref{fig:iwoo_flowchart}.

	\subsection{Computational complexity of IWOO}
	\label{subsec:iwoo_complexity}
	
	Let $N$ denote the population size of IWOO, $T_{\max}$ the maximum number of iterations, and $D$ the problem dimensionality.
	In addition, let $I_{\mathrm{CGOL}}$ denote the update interval of the correlation matrix and variable grouping in CGOL, which is set to $I_{\mathrm{CGOL}}=20$ in this study.
	
	The main computational overhead of IWOO arises from the subgroup-statistics calculation in TriEDM and the periodic correlation-matrix update in CGOL.
	In the worst case, the computational complexity of TriEDM over the entire iterative process is
	$\mathcal{O}(T_{\max}N^{2}D)$.
	For CGOL, the variable correlation matrix is updated using historical samples, resulting in a cumulative computational complexity of
	$\mathcal{O}(T_{\max}^{2}D^{2}/I_{\mathrm{CGOL}})$.
	The remaining operations, including population initialization, candidate-solution construction, ASRK refinement, population sorting, and boundary handling, do not exceed these dominant terms.
	
	Therefore, after omitting constant factors and lower-order terms, the overall computational complexity of IWOO is
	
	\begin{equation}
		\mathcal{O}\!\left(
		T_{\max}N^{2}D
		+
		\frac{T_{\max}^{2}D^{2}}{I_{\mathrm{CGOL}}}
		\right).
		\label{eq:iwoo_complexity}
	\end{equation}

	\makeatletter
	\newsavebox{\IWOOPageBottomFigure}
	\newdimen\IWOOPageBottomSpace
	\newcommand{\IWOOReservePageBottom}{%
		\par
		\global\setbox\IWOOPageBottomFigure=\vbox{%
			\hbox{\copy\IWOOPageBottomFigure}\kern0pt
		}%
		\global\IWOOPageBottomSpace=\dimexpr
		\ht\IWOOPageBottomFigure+\dp\IWOOPageBottomFigure+10pt\relax
		\global\let\IWOOOriginalCombineFloats\@combinedblfloats
		\gdef\@combinedblfloats{%
			\IWOOOriginalCombineFloats
			\global\let\@combinedblfloats\IWOOOriginalCombineFloats
			\setbox\@outputbox=\vbox to\textheight{%
				\unvbox\@outputbox
				\vfil
				\vskip10pt
				\nointerlineskip
				\copy\IWOOPageBottomFigure
				\kern0pt
			}%
		}%
		\global\advance\@colht-\IWOOPageBottomSpace
		\global\@colroom=\@colht
		\global\vsize=\@colht
	}
	\makeatother
	
	\section{Experimental results and discussion}
	\label{sec:experiments}
	
	This section conducts comparative experiments on benchmark suites from the IEEE Congress on Evolutionary Computation (CEC) competitions.

	\subsection{Experimental environment and parameter settings}
	\label{subsec:experimental_settings}
	
	All experiments are conducted on the same platform, consisting of a 64-bit Windows 11 system equipped with an Intel Core i5-9300H @ 2.40 GHz CPU and 8 GB of RAM.
	All simulations are implemented in MATLAB R2024b.
	In the experiments, the population size is set to 30 and the maximum number of iterations is set to 500.
	In addition, the maximum function-evaluation parameter is set to $D\times10{,}000$, where $D$ denotes the problem dimension.
	Each algorithm is independently run 30 times on each test function, while all settings other than the algorithm-specific parameters are kept identical.

	\subsection{Benchmark suites and compared algorithms}
	\label{subsec:benchmarks_algorithms}
	
	Three benchmark suites, namely CEC2019~\cite{Price2018CEC2019}, CEC2022~\cite{Kumar2021CEC2022}, and CEC2017~\cite{Awad2016CEC2017}, are used for numerical optimization experiments to comprehensively evaluate the overall performance of IWOO.
	
	Among them, CEC2019 contains 10 complex continuous optimization functions with problem dimensions not exceeding 20 and is mainly used to examine the exploration and exploitation capabilities of the algorithm in low-dimensional complex search spaces.
	CEC2022 contains 12 test functions, including unimodal, multimodal, hybrid, and composition functions.
	The problem dimension is uniformly set to 20 to evaluate the overall performance of the algorithms under medium-dimensional conditions.
	For CEC2017, F1 and F3--F30, comprising 29 functions in total, are selected, and the problem dimension is uniformly set to 50 to further examine the stability and scalability of the algorithms on high-dimensional optimization problems.
	
	The comparative experiments involve two groups of algorithms: representative algorithms and high-performing algorithms.
	The representative algorithm group includes
	YDSE~\cite{AbdelBasset2023YDSE},
	EMWPA~\cite{Abdullah2025EMWPA},
	SRA~\cite{Hussein2025SRA},
	RUN~\cite{Ahmadianfar2021RUN},
	FLA~\cite{Ghasemi2024FLA},
	\onecolumn
	
	\noindent
	\begin{minipage}[t][0.98\textheight][t]{\textwidth}
		
		\begin{minipage}{\linewidth}
			\centering
			\captionsetup{skip=2pt}
			
			\raisebox{0pt}[0.30\textheight][0pt]{%
				\includegraphics[
				width=0.95\linewidth,
				height=0.36\textheight,
				keepaspectratio
				]{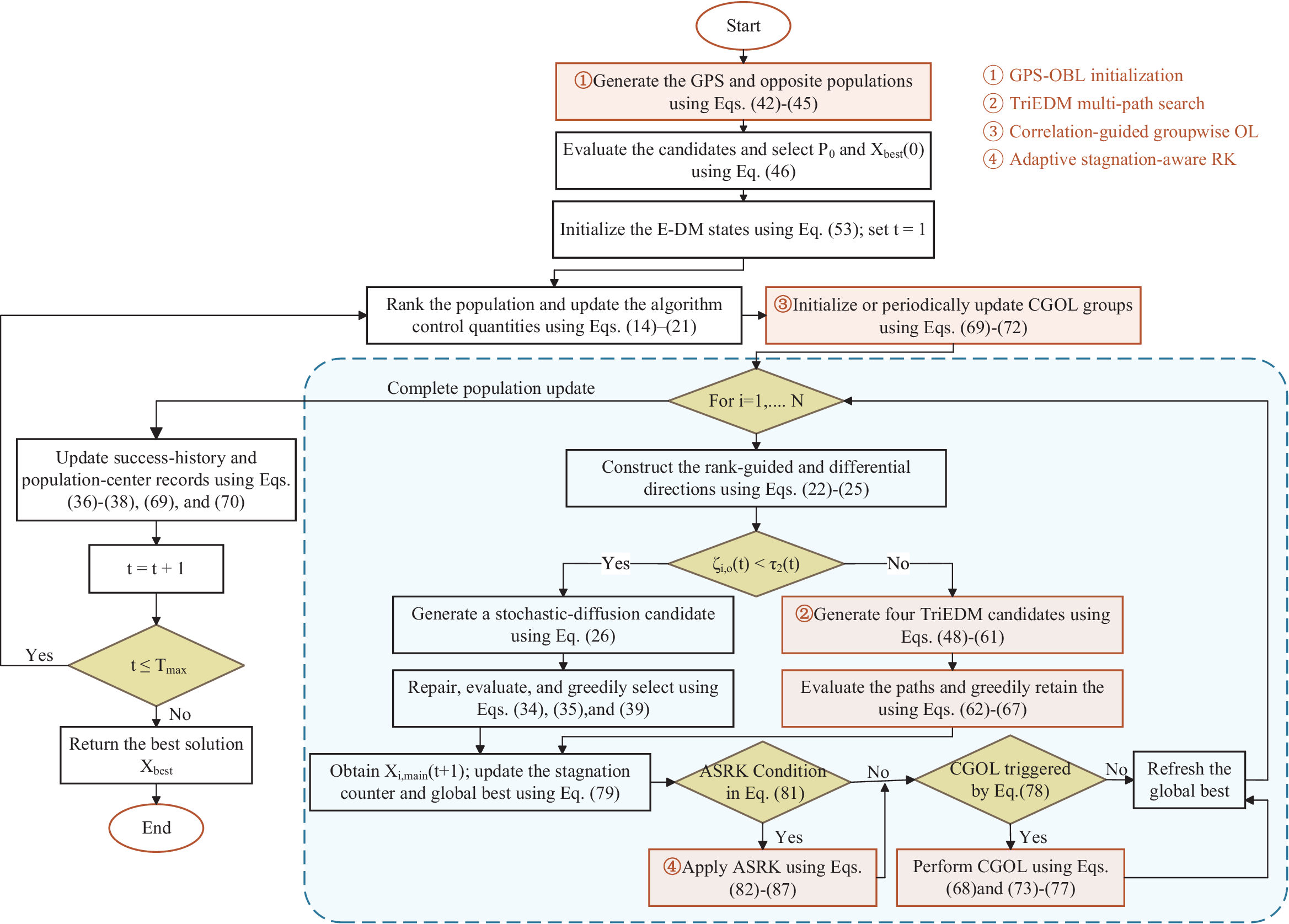}%
			}
			
			\captionof{figure}{Flowchart of IWOO}
			\label{fig:iwoo_flowchart}
		\end{minipage}
		
		\vspace{6pt}
		
		
		\noindent
		\begin{minipage}[t]{\dimexpr0.5\linewidth-0.5\columnsep\relax}
			\noindent RIME~\cite{Su2023RIME},
			FATA~\cite{Qi2024FATA},
			WAA~\cite{Cheng2024WAA},
			ETO~\cite{Luan2024ETO},
			SCHO~\cite{Bai2023SCHO},
			FTO~\cite{Saad2026FTO}, and
			WOO~\cite{Peng2026WOO}.
			
			The high-performing algorithm group includes
			CGO~\cite{Talatahari2021CGO},
			RWGWO~\cite{Gupta2019RWGWO},
			SaDE~\cite{Brest2006SaDE},
			SHADE~\cite{Tanabe2013SHADE},
			L-SHADE~\cite{Tanabe2014LSHADE},
			APSM-jSO~\cite{Li2023APSMjSO},
			NL-SHADE-RSP~\cite{Stanovov2021NLSHADERSP}, and
			L-SRTDE~\cite{Stanovov2024LSRTDE}.
			
			Basic information on the compared algorithms is summarized in Table~\ref{tab:compared_algorithm_information}.
			
			Algorithm-specific parameters follow the recommended settings reported in the original publications or publicly available implementations.
			Parameters that are not explicitly specified are set uniformly according to the experimental settings described in this section.
			In this section, detailed analysis is provided for the experimental results obtained from comparisons between IWOO and the representative algorithm group.
			The experiments involving the high-performing algorithm group are mainly used to evaluate the competitiveness of IWOO against advanced optimizers.
			For this group, only the key comparative results are presented in the main text, while the complete statistical results are provided in the Appendix.
		\end{minipage}%
		\hfill
		\begin{minipage}[t]{\dimexpr0.5\linewidth-0.5\columnsep\relax}
			\subsection{Performance evaluation and statistical analysis}
			\label{subsec:evaluation_metrics}
			
			Algorithm performance is evaluated primarily from three aspects: solution accuracy, run-to-run stability, and overall ranking.
			For each test function, the minimum value (Min), mean value (Mean), and standard deviation (Std) of the objective-function values obtained over 30 independent runs are recorded.
			
			For each test function, the Friedman test is performed using the final objective-function values obtained from the 30 independent runs.
			Each run is treated as a block, and the compared algorithms are ranked in ascending order according to their objective-function values within each run.
			The Friedman rank(Fr-rank) of each algorithm is then obtained by averaging its ranks over the 30 runs, with tied values assigned average ranks.
			A smaller Fr-rank indicates better overall performance.
		\end{minipage}
		\par
		
		\vfill
		\vspace{6pt}
		
		\begin{minipage}{\linewidth}
			\centering
			\captionsetup{skip=3pt}
			\captionof{table}{Basic information on the compared algorithms}
			\label{tab:compared_algorithm_information}
			
			\footnotesize
			\renewcommand{\arraystretch}{0.90}
			\setlength{\tabcolsep}{3pt}
			
			\begin{tabular}{
					@{}
					>{\raggedright\arraybackslash}p{0.16\linewidth}
					>{\raggedright\arraybackslash}p{\dimexpr0.62\linewidth-6\tabcolsep\relax}
					>{\centering\arraybackslash}p{0.15\linewidth}
					>{\centering\arraybackslash}p{0.07\linewidth}
					@{}
				}
				\toprule
				\textbf{Abbreviation}
				& \textbf{Algorithms}
				& \textbf{Group}
				& \textbf{Year} \\
				\midrule
				
				YDSE
				& Young's Double-Slit Experiment Optimizer
				& Representative
				& 2023 \\
				
				EMWPA
				& Electromagnetic Wave Propagation Algorithm
				& Representative
				& 2025 \\
				
				SRA
				& Schrödinger Optimizer Algorithm
				& Representative
				& 2025 \\
				
				RUN
				& Runge Kutta Optimizer
				& Representative
				& 2021 \\
				
				FLA
				& Flood Algorithm
				& Representative
				& 2024 \\
				
				RIME
				& Rime Optimization Algorithm
				& Representative
				& 2023 \\
				
				FATA
				& Fata Morgana Algorithm
				& Representative
				& 2024 \\
				
				WAA
				& Weighted Average Algorithm
				& Representative
				& 2024 \\
				
				ETO
				& Exponential-Trigonometric Optimization Algorithm
				& Representative
				& 2024 \\
				
				SCHO
				& Sinh Cosh Optimizer
				& Representative
				& 2023 \\
				
				FTO
				& Fourier Transform Optimizer
				& Representative
				& 2026 \\
				
				CGO
				& Chaos Game Optimization
				& High-performing
				& 2021 \\
				
				RWGWO
				& Random Walk Grey Wolf Optimizer
				& High-performing
				& 2019 \\
				
				SaDE
				& Self-Adaptive Differential Evolution
				& High-performing
				& 2006 \\
				
				SHADE
				& Success-History Based Parameter Adaptation for Differential Evolution
				& High-performing
				& 2013 \\
				
				L-SHADE
				& Success-History Based Adaptive Differential Evolution with Linear Population Size Reduction
				& High-performing
				& 2014 \\
				
				APSM-jSO
				& jSO with an Adaptive Parameter Selection Mechanism
				& High-performing
				& 2023 \\
				
				NL-SHADE-RSP
				& Nonlinear Population Size Reduction SHADE with Rank-Based Selective Pressure
				& High-performing
				& 2021 \\
				
				L-SRTDE
				& Success Rate-Based Adaptive Differential Evolution with Linear Population Size Reduction
				& High-performing
				& 2024 \\
				
				WOO
				& Wave Optics Optimizer
				& Representative
				& 2026 \\
				
				\bottomrule
			\end{tabular}
		\end{minipage}
	\end{minipage}
	\par
	
	\twocolumn
	
	\begin{lrbox}{\IWOOPageBottomFigure}
		\begin{minipage}{\textwidth}
			\centering
			\captionsetup{skip=2pt}
			\setlength{\tabcolsep}{0pt}
			
			\resizebox{\textwidth}{\height}{%
				\begin{tabular}{@{}cccc@{}}
					
					\includegraphics[width=0.25\textwidth,height=0.12\textheight,keepaspectratio]{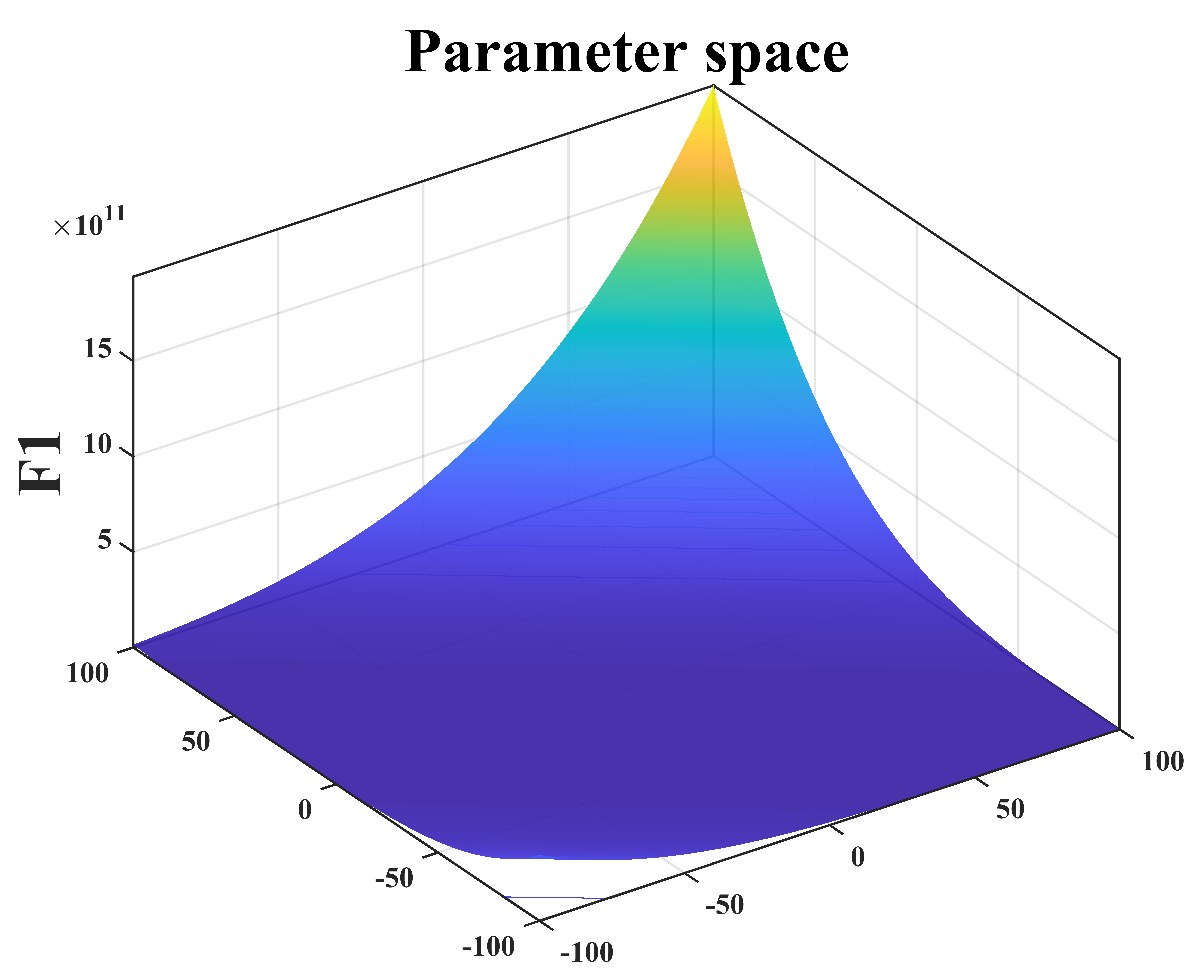}
					&
					\includegraphics[width=0.25\textwidth,height=0.12\textheight,keepaspectratio]{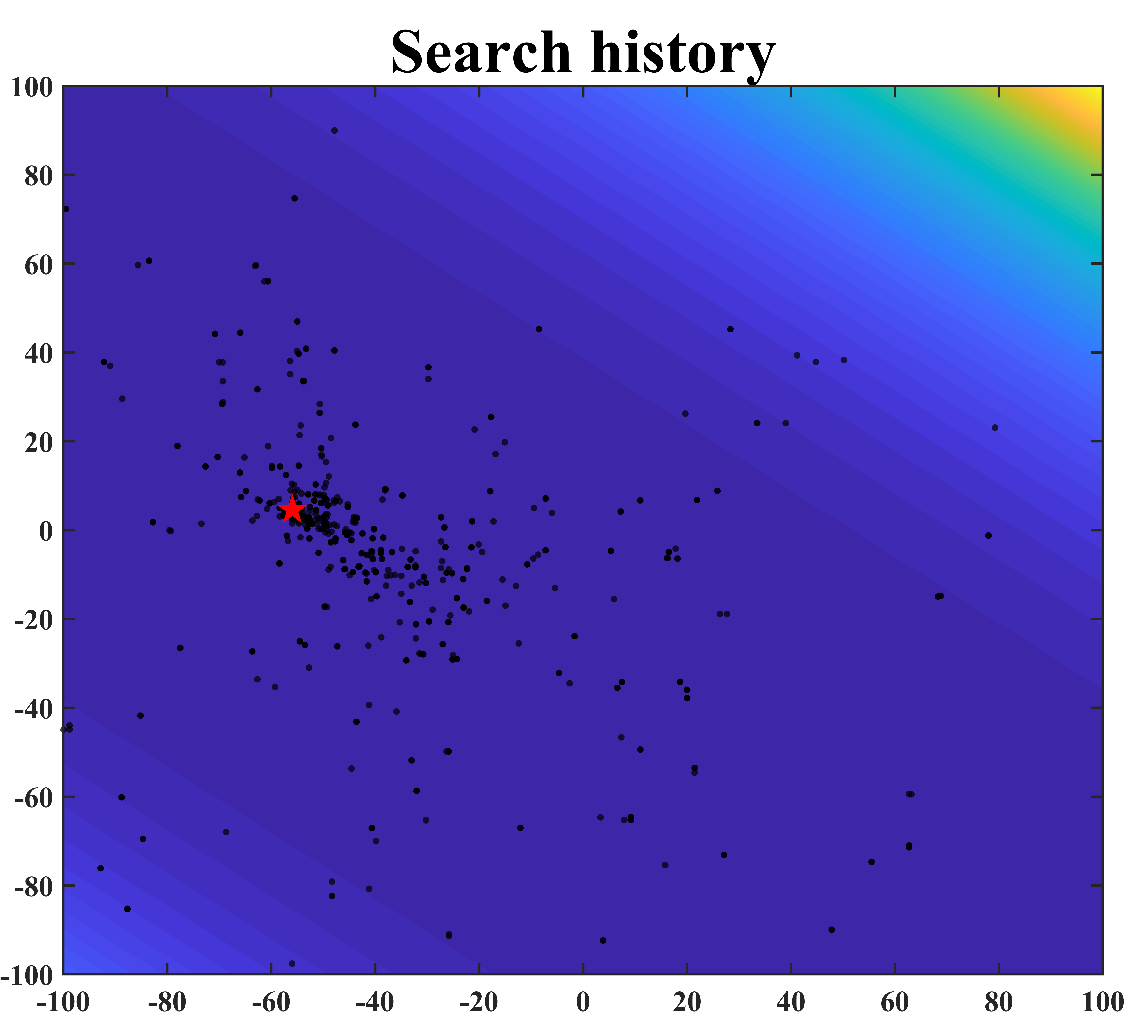}
					&
					\includegraphics[width=0.25\textwidth,height=0.12\textheight,keepaspectratio]{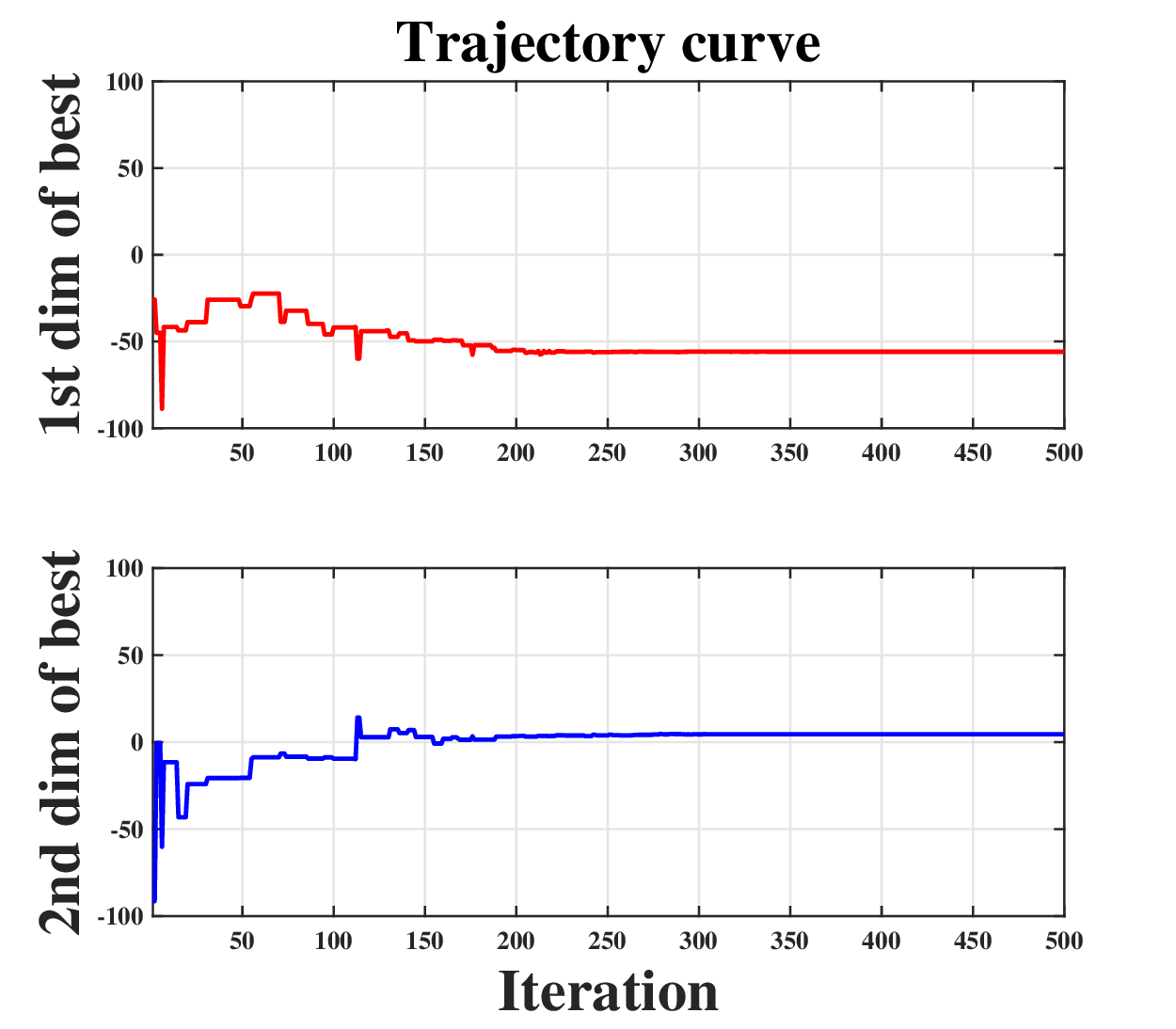}
					&
					\includegraphics[width=0.25\textwidth,height=0.12\textheight,keepaspectratio]{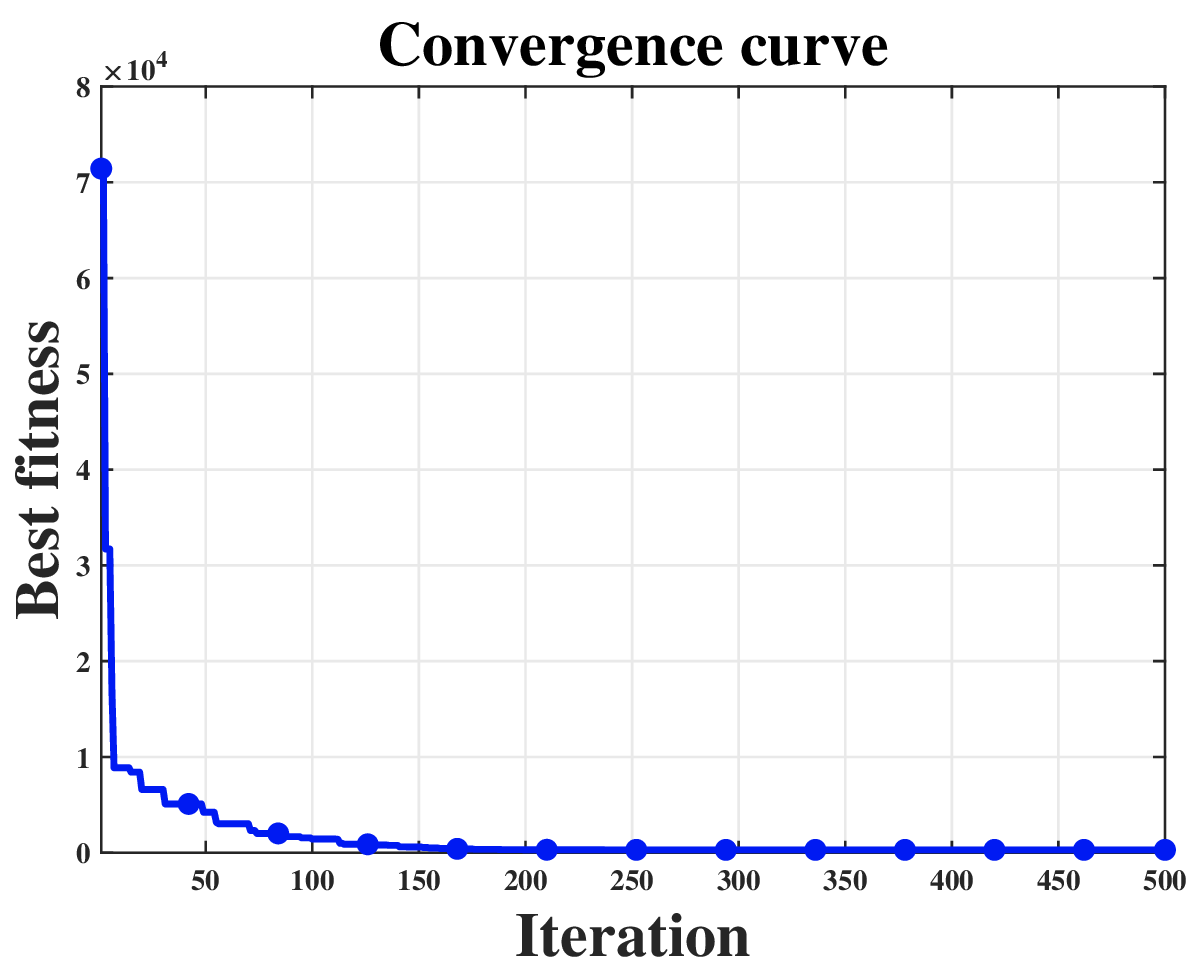}
					\\[0.5mm]
					
					\includegraphics[width=0.25\textwidth,height=0.12\textheight,keepaspectratio]{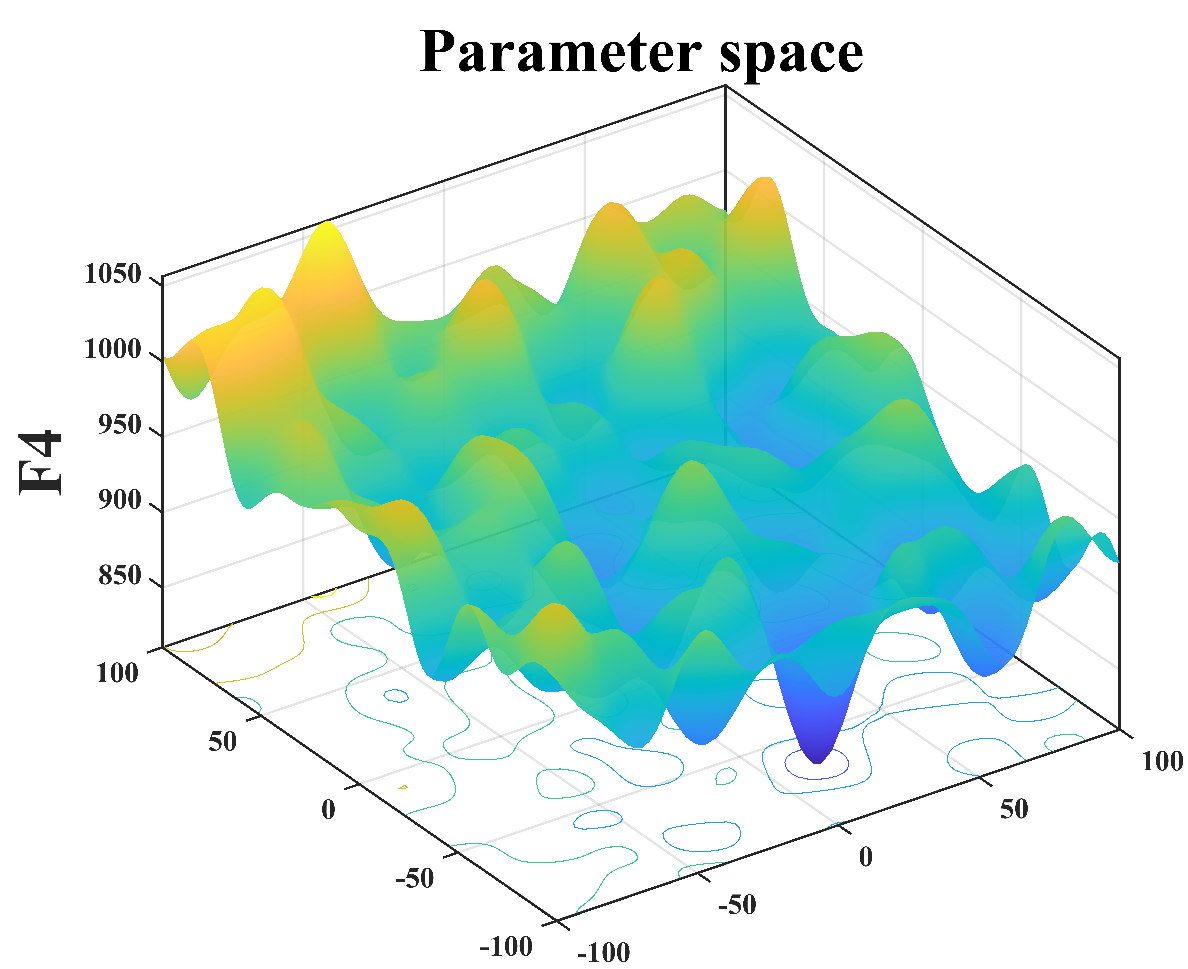}
					&
					\includegraphics[width=0.25\textwidth,height=0.12\textheight,keepaspectratio]{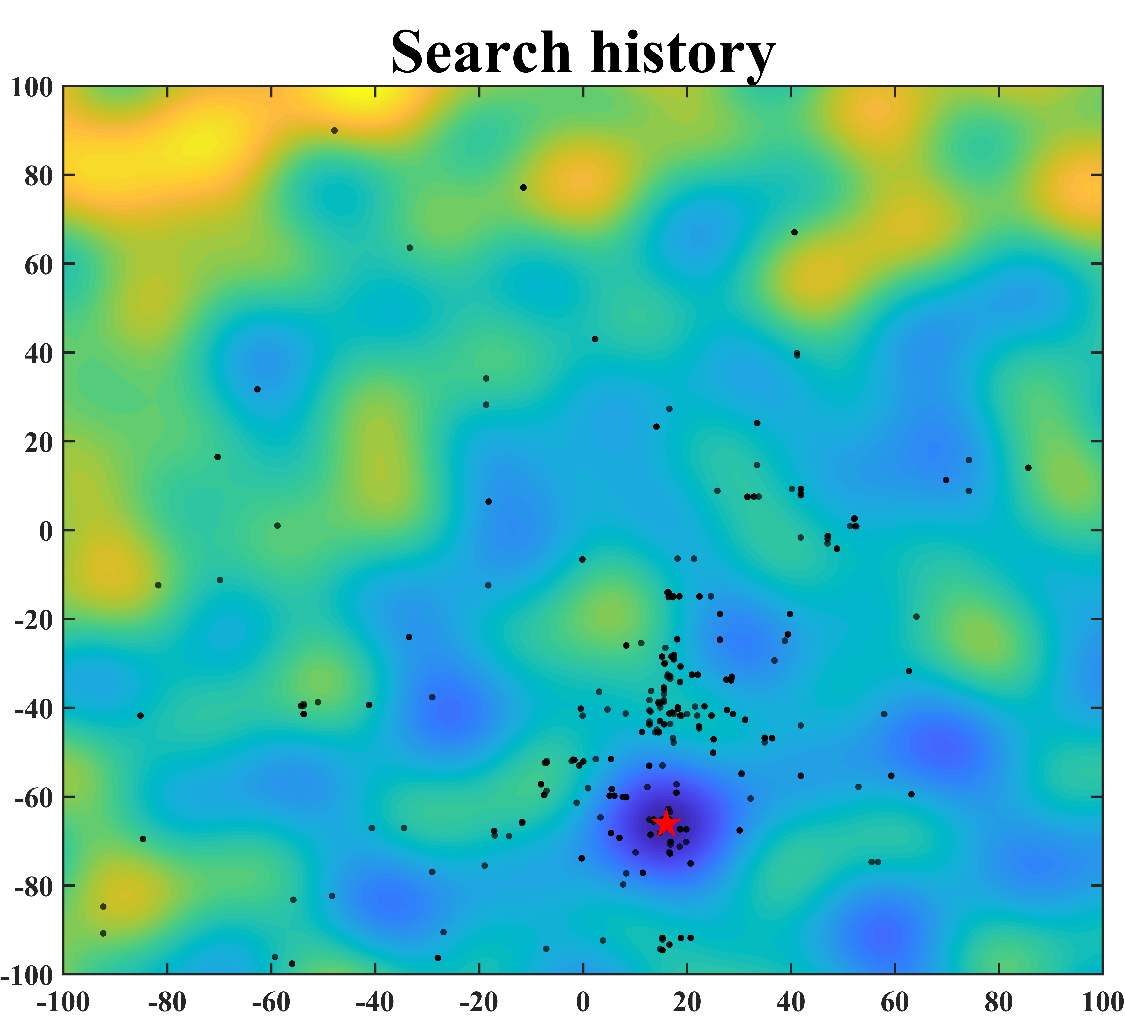}
					&
					\includegraphics[width=0.25\textwidth,height=0.12\textheight,keepaspectratio]{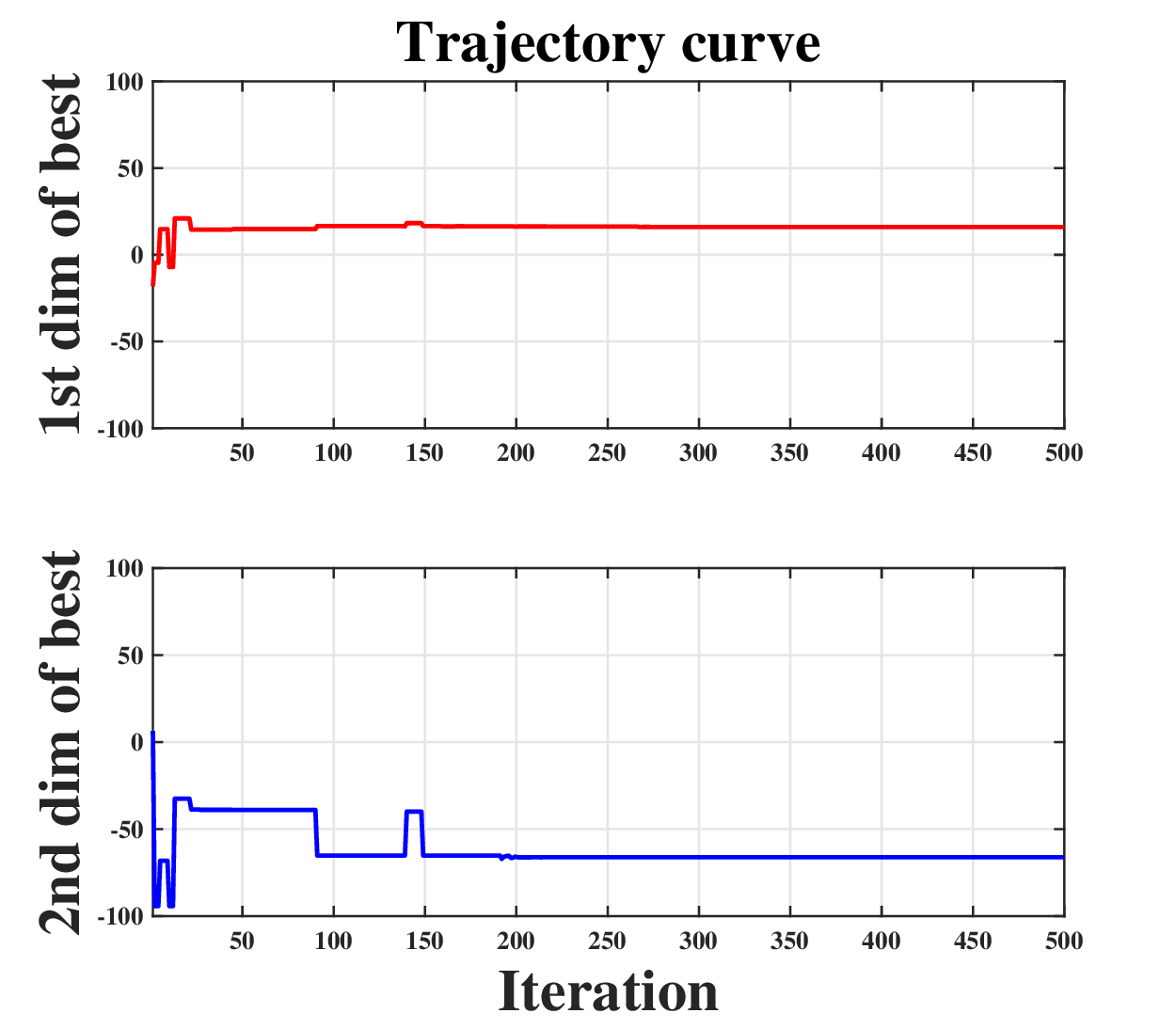}
					&
					\includegraphics[width=0.25\textwidth,height=0.12\textheight,keepaspectratio]{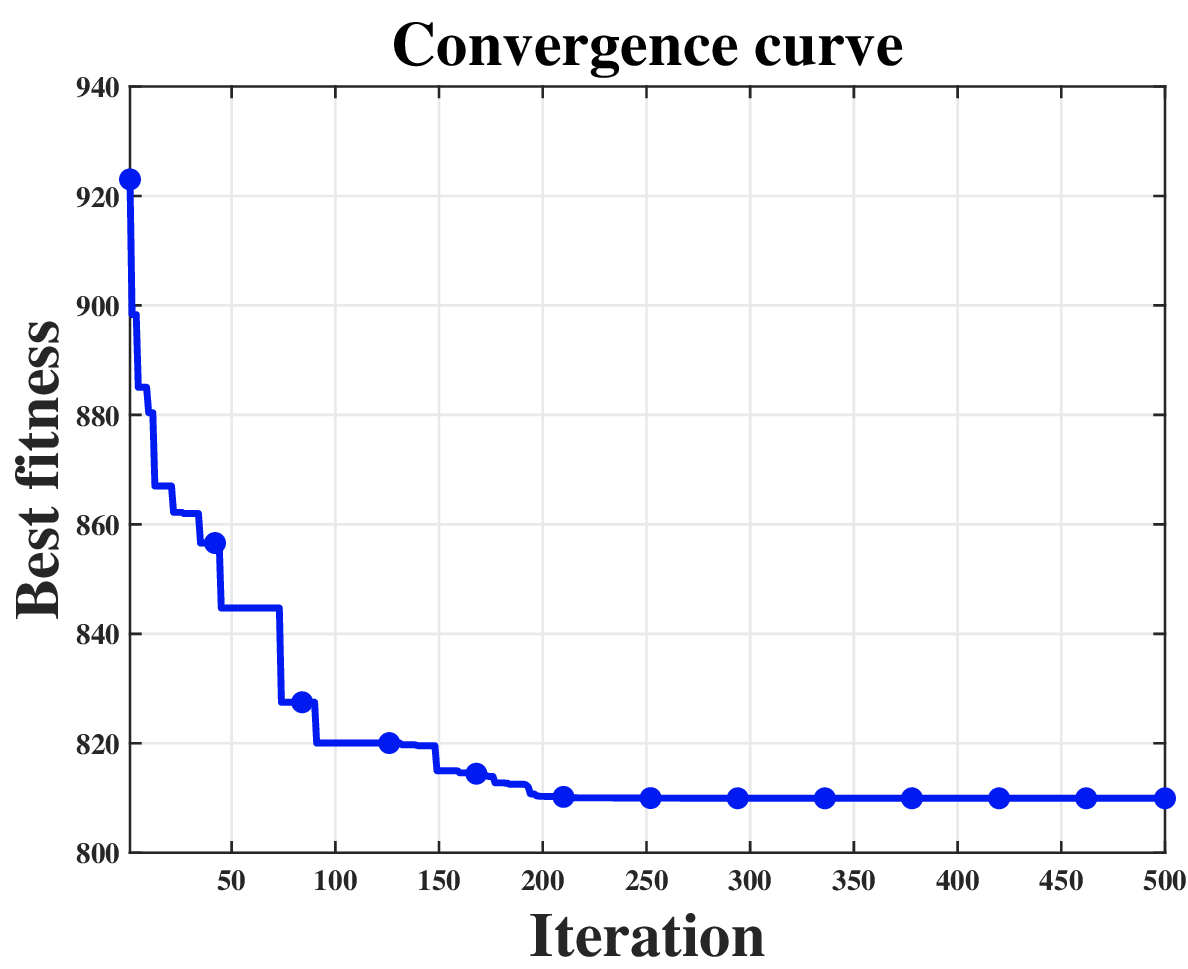}
					\\[0.5mm]
					
					\includegraphics[width=0.25\textwidth,height=0.12\textheight,keepaspectratio]{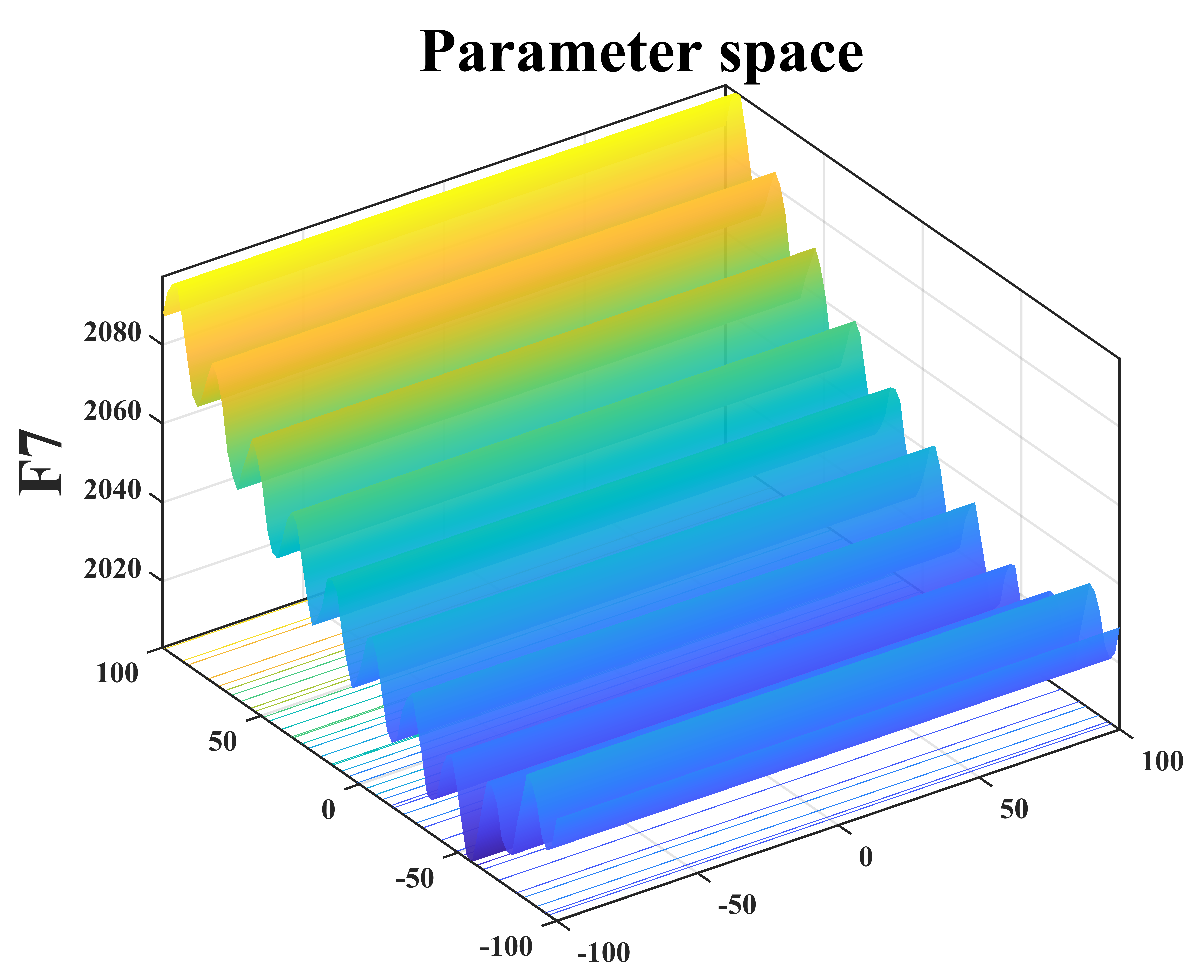}
					&
					\includegraphics[width=0.25\textwidth,height=0.12\textheight,keepaspectratio]{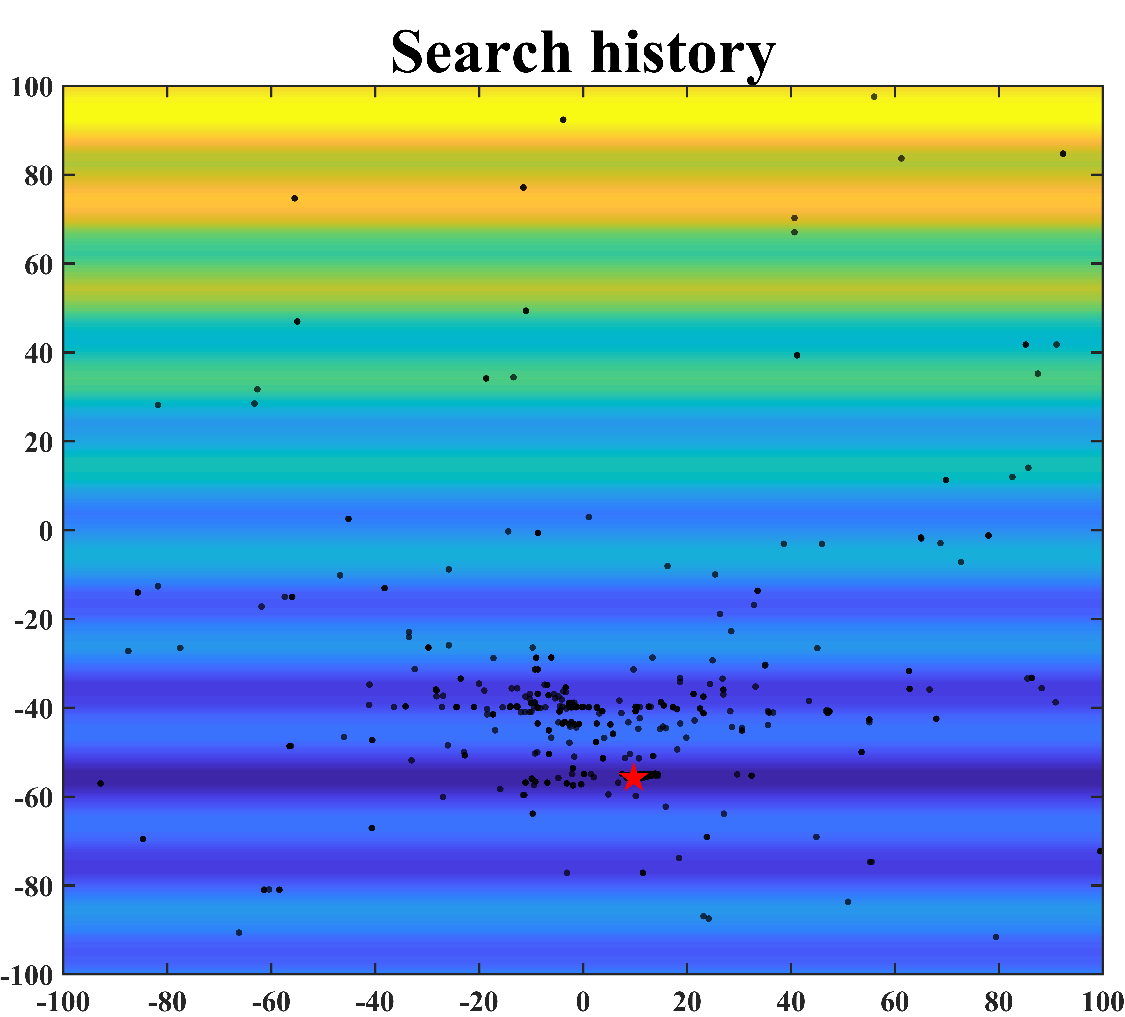}
					&
					\includegraphics[width=0.25\textwidth,height=0.12\textheight,keepaspectratio]{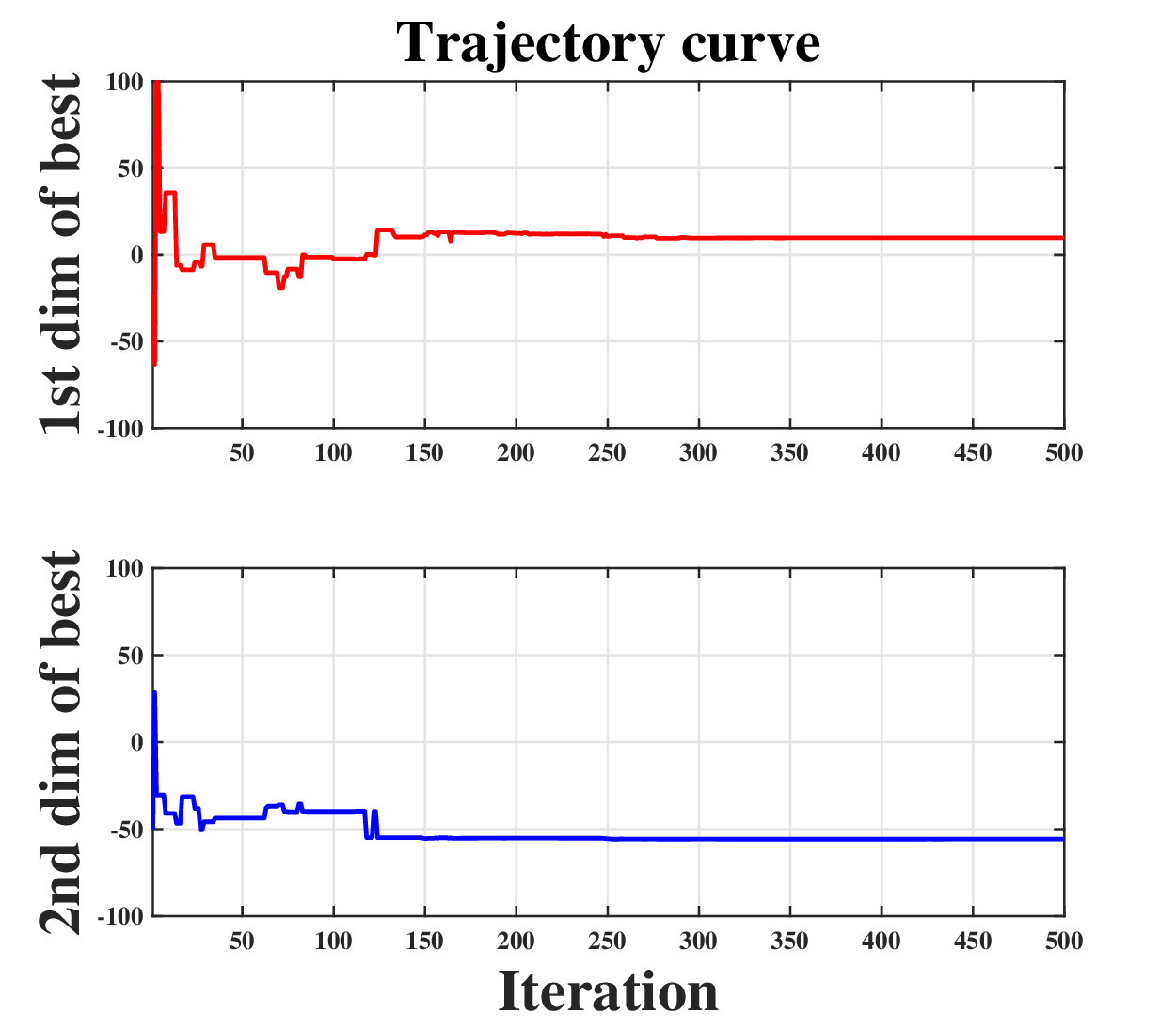}
					&
					\includegraphics[width=0.25\textwidth,height=0.12\textheight,keepaspectratio]{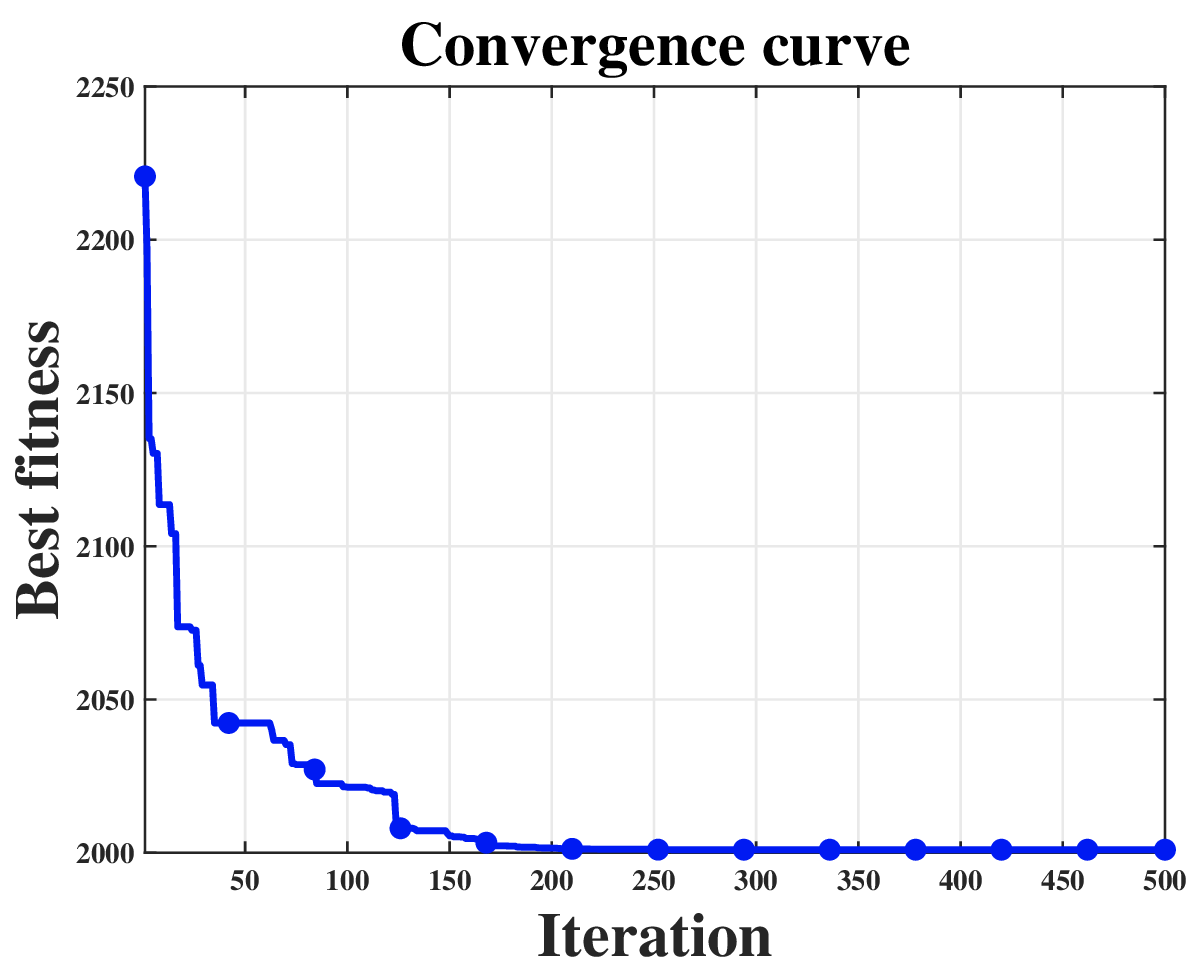}
					\\[0.5mm]
					
					\includegraphics[width=0.25\textwidth,height=0.12\textheight,keepaspectratio]{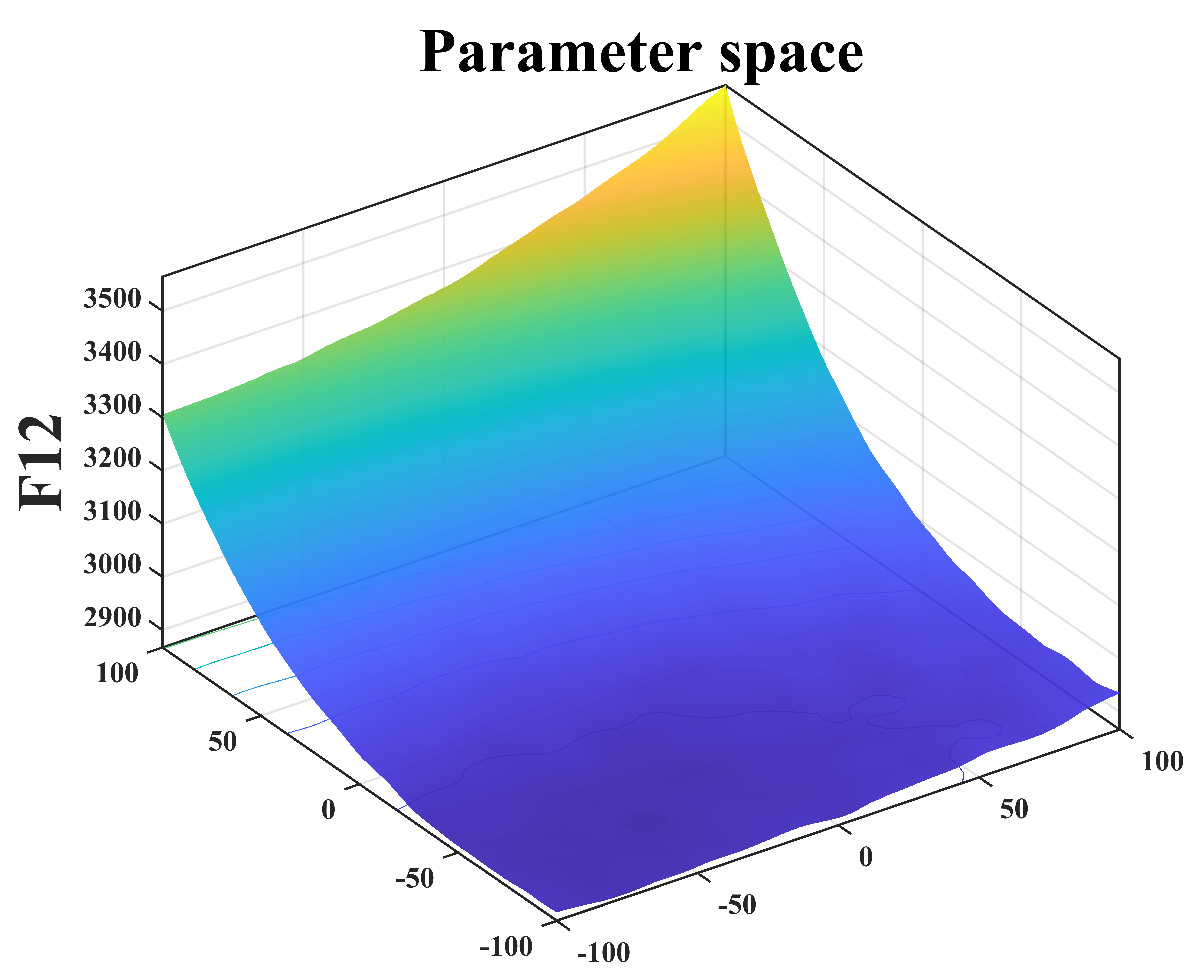}
					&
					\includegraphics[width=0.25\textwidth,height=0.12\textheight,keepaspectratio]{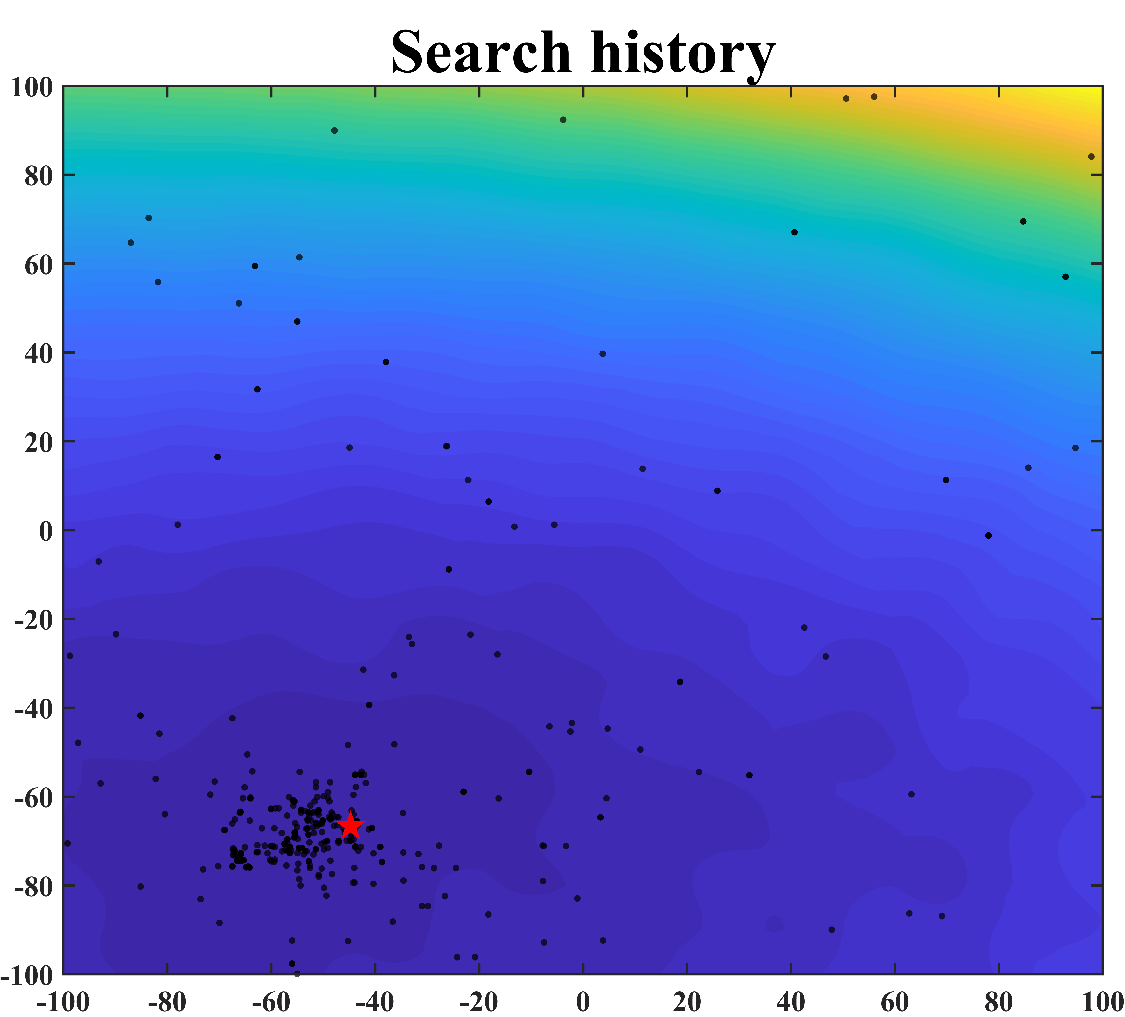}
					&
					\includegraphics[width=0.25\textwidth,height=0.12\textheight,keepaspectratio]{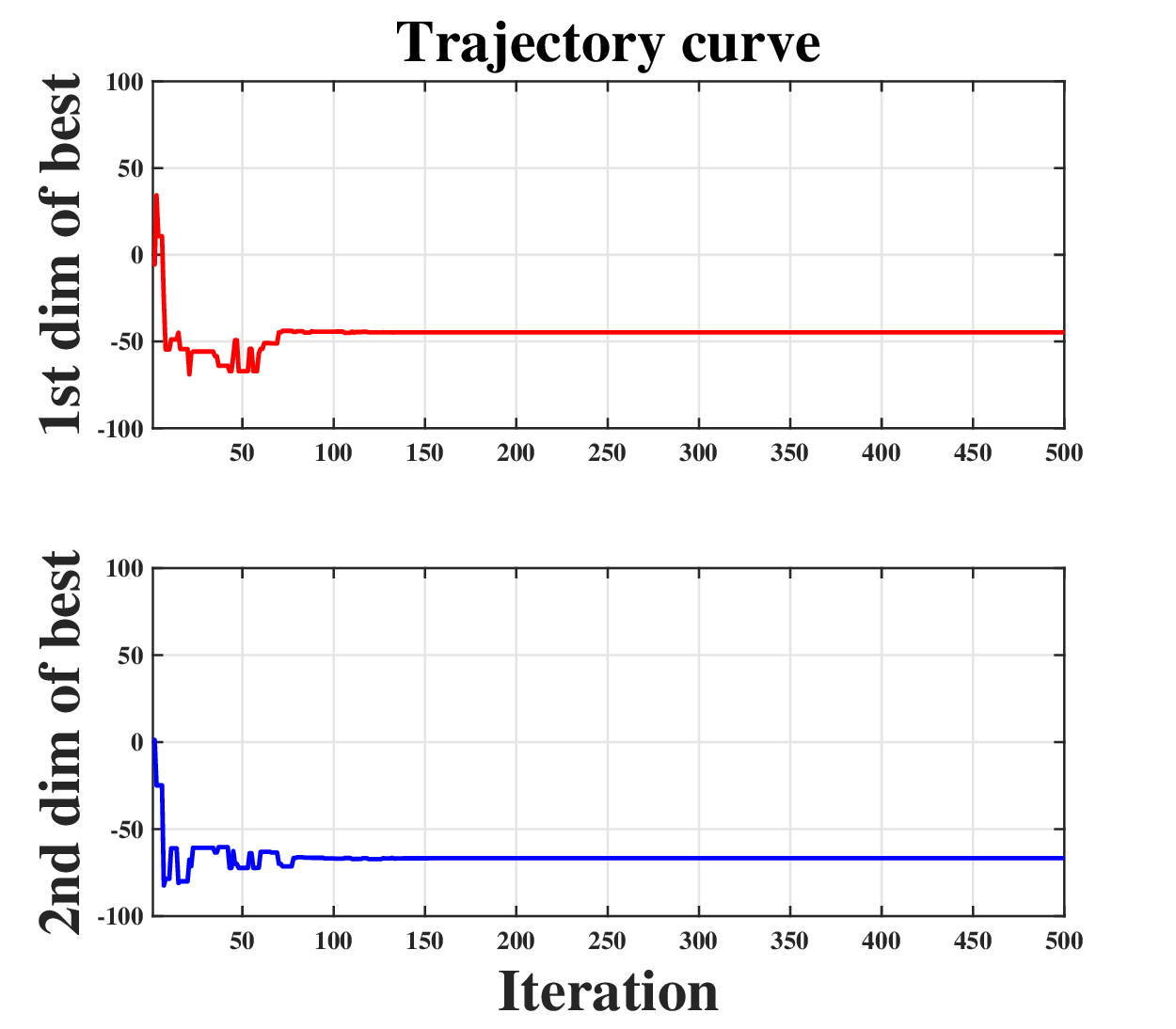}
					&
					\includegraphics[width=0.25\textwidth,height=0.12\textheight,keepaspectratio]{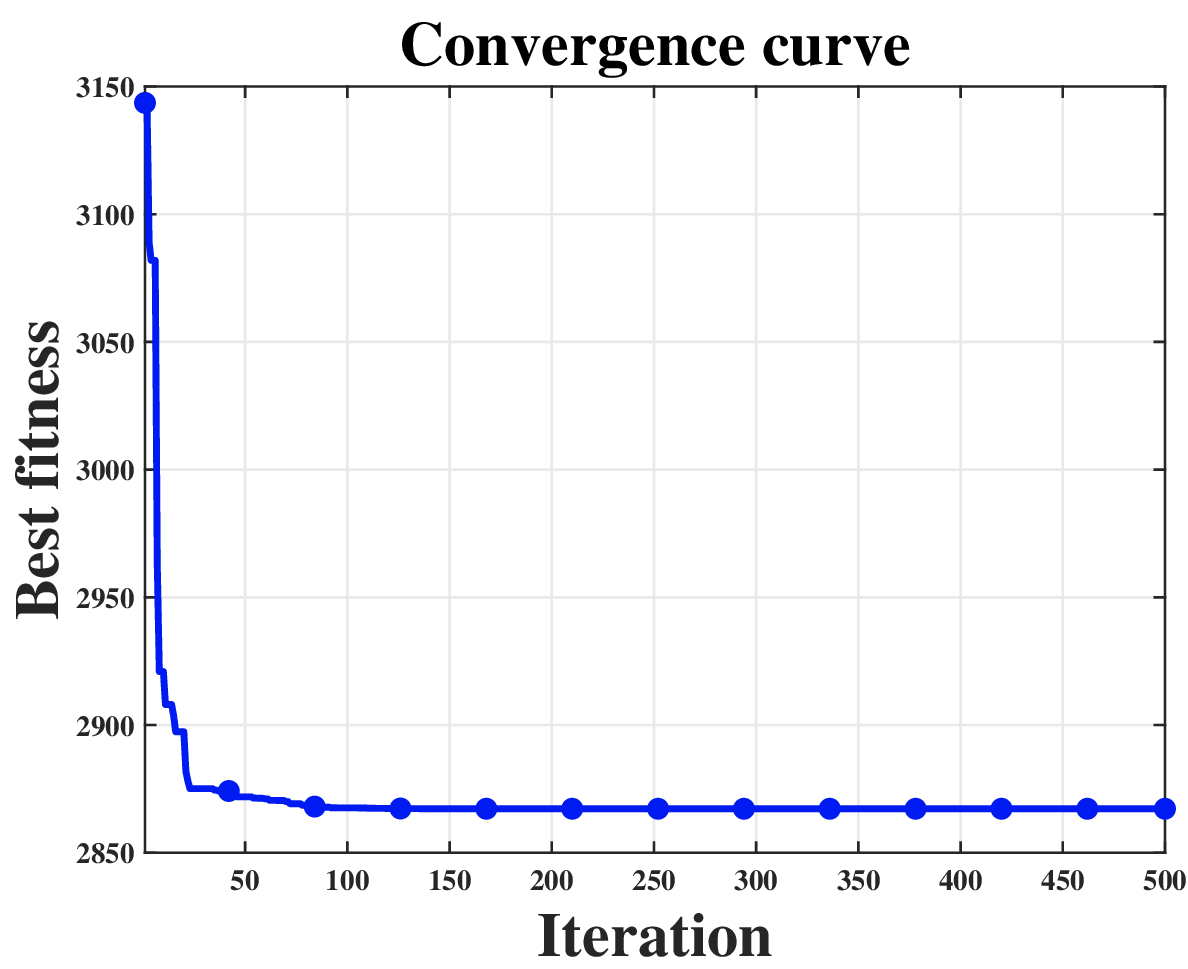}
					
				\end{tabular}%
			}
			
			\par\vspace{1mm}
			
			\captionof{figure}{Search behavior analysis of IWOO on representative CEC2022 functions}
			\label{fig:iwoo_search_behavior_cec2022}
			
		\end{minipage}
	\end{lrbox}
	
	\IWOOReservePageBottom

	In addition, the two-sided Wilcoxon signed-rank test is employed to evaluate the statistical significance of the paired differences between IWOO and each competitor.
	The comparison is performed using the run-wise paired results obtained from the same 30 run indices.
	When a statistically significant difference is detected, the comparison direction is determined according to the mean objective values.
	The symbols ``+'', ``='', and ``$-$'' indicate that IWOO performs significantly better than the corresponding algorithm, exhibits no statistically significant difference, and performs significantly worse than the corresponding algorithm, respectively~\cite{Derrac2011NonparametricTests}.
	
	Convergence curves are used to illustrate the convergence speed, late-stage exploitation capability, and stagnation behavior of the algorithms during the iterative process, whereas box plots are employed to visualize the median, interquartile range, dispersion, and outliers of the results obtained from 30 independent runs.
	Representative functions are selected for visualization in the main text.
	Specifically, F1 and F6 are selected for CEC2019; F1, F4, F7, and F12 are selected for CEC2022; and F1, F10, F20, and F30 are selected for CEC2017.

	\subsection{Algorithm analysis and verification}
	\label{subsec:algorithm_analysis}
	
	\subsubsection{Qualitative analysis of search behavior}
	\label{subsubsec:qualitative_search_behavior}

	To further investigate the search dynamics and convergence behavior of IWOO, a multi-perspective visualization analysis is conducted to qualitatively examine its optimization process~\cite{Xiao2024MSAO,Chen2026CZSCSO}.
	Four representative functions, namely F1, F4, F7, and F12, are selected from the CEC2022 benchmark suite to represent unimodal, multimodal, hybrid, and composition landscapes, respectively. These functions provide a diverse basis for examining the exploration, exploitation, population diversity, and convergence behavior of IWOO under different search complexities.
	
	The search behavior of IWOO on the four representative functions is presented in Fig.~\ref{fig:iwoo_search_behavior_cec2022}.
	Each row corresponds to one test function, and the four columns provide complementary views of the optimization process. From left to right, the columns show the function landscape, the historical population distribution, the evolution of the best individual's first two coordinates, and the convergence curve of the best fitness value. These visualizations jointly reveal how IWOO adjusts its search distribution, updates promising solutions, and approaches the optimum throughout the iterative process.
	
	A common stage-wise search pattern can be observed across the four functions. During the early iterations, IWOO maintains a relatively broad search distribution and explores multiple regions of the landscape. As the optimization proceeds, the population gradually concentrates around promising areas, the trajectory of the best individual becomes more stable, and the convergence process changes from rapid improvement to a relatively steady stage.
	
	For the relatively simple unimodal function F1, the population rapidly contracts from a broad search range toward the optimum and eventually forms a dense distribution around the promising region. The best-individual trajectory changes mainly during the early iterations and then stabilizes, while the convergence curve decreases sharply at the beginning and subsequently becomes smooth. This behavior indicates that IWOO can quickly identify a promising region and achieve stable local exploitation on a simple search landscape.
	
	For the more complex functions F4, F7, and F12, the search process exhibits more pronounced stage-wise adjustments. On the multimodal F4, the population remains distributed over a relatively broad region rather than prematurely concentrating in a single area, indicating that IWOO preserves population diversity and reduces the risk of premature convergence. On the hybrid F7, most individuals gradually move toward promising regions while some search points remain farther away, showing that global exploration is retained while local exploitation is strengthened. For the composition function F12, the population initially covers a broad region and then progressively converges toward promising areas, with the objective value continuing to improve before reaching a stable level.
	
	Overall, the search histories, best-individual trajectories, and convergence curves provide consistent visual evidence that IWOO can coordinate global exploration and local exploitation across different search landscapes. The algorithm maintains diversity during the early search stage, progressively focuses its search on promising regions, and achieves stable convergence in the later stage. These characteristics demonstrate the adaptability of IWOO to unimodal, multimodal, hybrid, and composition optimization problems.
	
	\FloatBarrier
	\subsubsection{Parameter sensitivity analysis}
	\label{subsubsec:parameter_sensitivity}
	
	The CGOL mechanism in IWOO partitions the decision variables into several groups through variable-correlation analysis and subsequently treats each variable group, rather than each individual decision variable, as the basic factor for orthogonal learning. The number of variable groups $G$ directly determines the grouping granularity and therefore affects both the search accuracy and function-evaluation overhead of CGOL. A smaller $G$ can reduce the scale of orthogonal learning, but may result in a loss of search accuracy due to coarser variable combinations. In contrast, a larger $G$ generally provides finer combination granularity and can improve search accuracy, although the associated evaluation overhead may increase. Therefore, it is necessary to analyze the influence of $G$ on CGOL and determine an appropriate setting.
	
	To investigate the sensitivity of CGOL to the number of variable groups, $G$ is selected from
	$G\in\{5,10,15,20,25,31,D\}$.
	Three test functions with different search characteristics, namely F1, F7, and F15 from the CEC2017 benchmark suite, are selected for the experiments. Figure~\ref{fig:cgol_group_convergence} presents the average convergence curves obtained under different values of $G$.
	
	As shown in Fig.~\ref{fig:cgol_group_convergence}, the average fitness generally improves as $G$ increases across the three test functions, indicating that finer variable grouping is beneficial for improving search accuracy. Conversely, when $G$ is reduced, the grouping becomes coarser, leading to lower combination resolution and, in general, higher average fitness values.
	
	However, increasing the number of variable groups does not imply that $G$ should be increased without limit, because the function-evaluation overhead of CGOL is constrained by the size of the orthogonal array.
	
\begin{figure*}[!t]
		\centering
		\captionsetup{skip=2pt}
		
		\begin{minipage}[t]{0.31\textwidth}
			\centering
			\includegraphics[
			width=\linewidth,
			height=0.17\textheight,
			keepaspectratio
			]{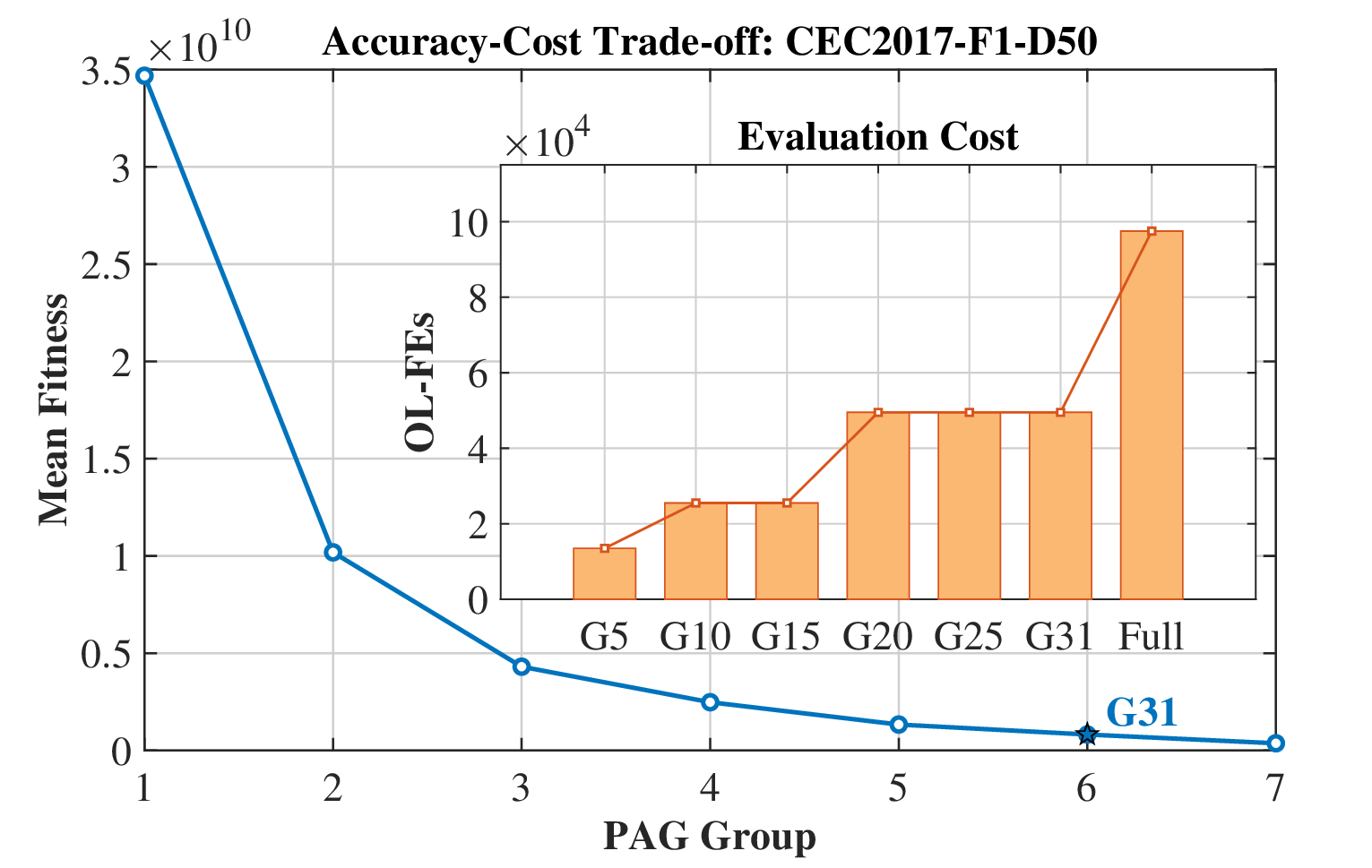}
			
			\par\vspace{1pt}
			{\footnotesize (a) F1}
		\end{minipage}
		\hfill
		\begin{minipage}[t]{0.31\textwidth}
			\centering
			\includegraphics[
			width=\linewidth,
			height=0.17\textheight,
			keepaspectratio
			]{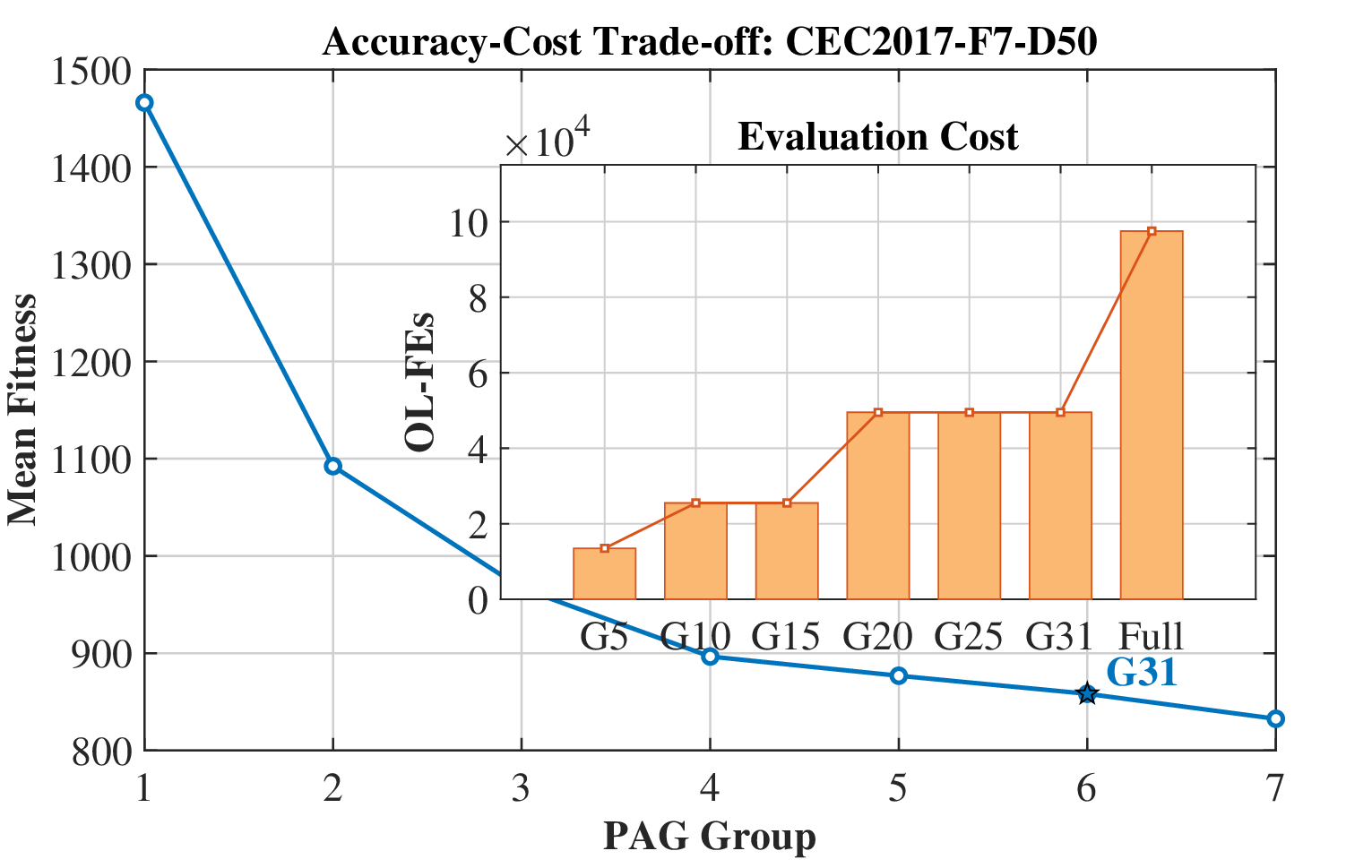}
			
			\par\vspace{1pt}
			{\footnotesize (b) F7}
		\end{minipage}
		\hfill
		\begin{minipage}[t]{0.31\textwidth}
			\centering
			\includegraphics[
			width=\linewidth,
			height=0.17\textheight,
			keepaspectratio
			]{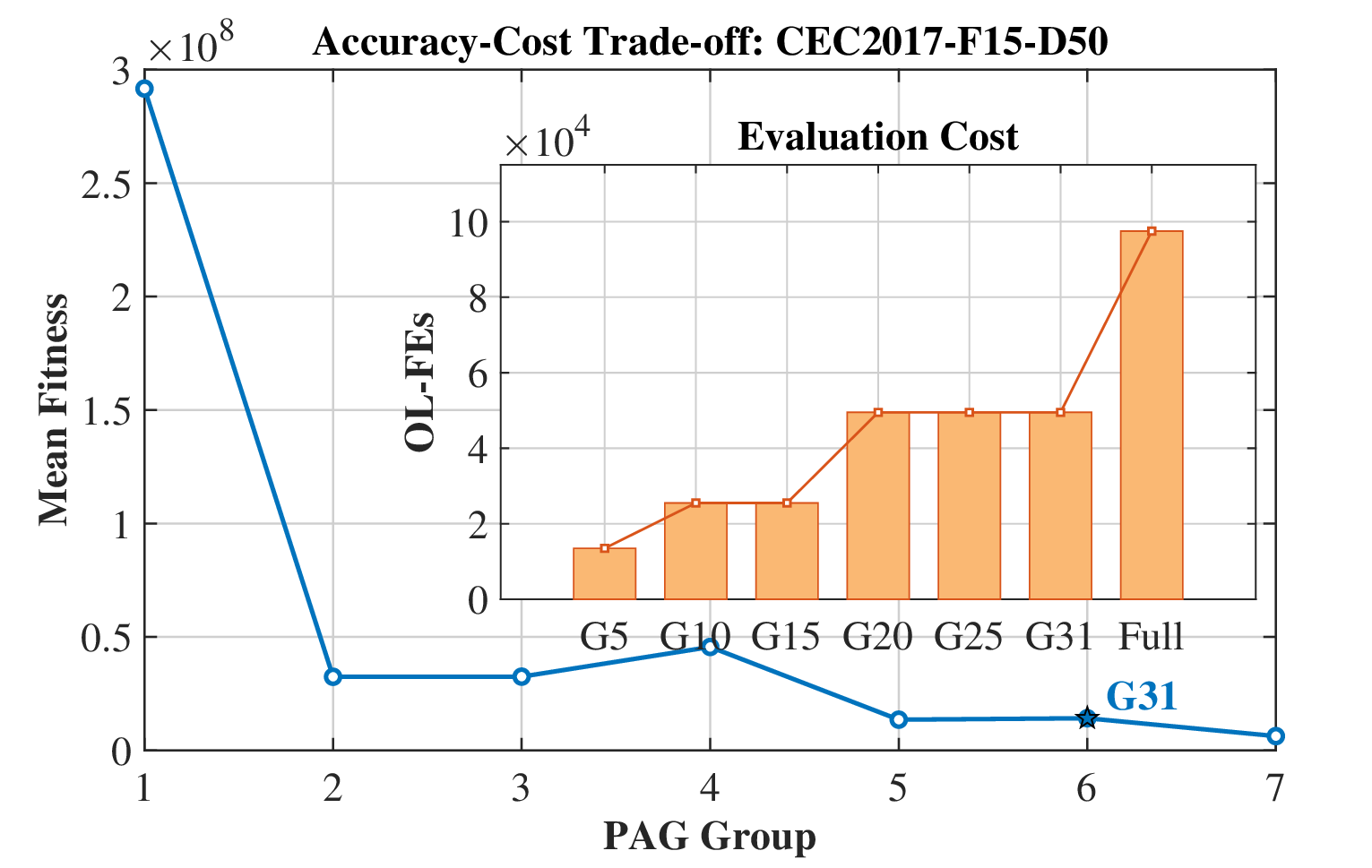}
			
			\par\vspace{1pt}
			{\footnotesize (c) F15}
		\end{minipage}
		
		\caption{Effect of group number on CGOL accuracy and evaluation cost}
		\label{fig:cgol_group_tradeoff}
	\end{figure*}

	As shown in Fig.~\ref{fig:cgol_group_tradeoff}, increasing $G$ generally improves the search accuracy of CGOL, while the function-evaluation overhead increases in a stepwise manner. According to Eq.~\eqref{eq:cgol_cost}, $G=20$, $25$, and $31$ correspond to the same orthogonal-array size. Therefore, within this range, increasing the number of variable groups does not introduce additional function evaluations for a single invocation of orthogonal learning.
	
	Considering the relationship between $G$ and search accuracy discussed above, $G=31$ achieves finer variable grouping and better search accuracy under the same function-evaluation overhead as $G=20$ and $G=25$.
	
	Therefore, the number of variable groups in CGOL is finally set to $G=31$, and this setting is kept unchanged in all subsequent experiments.
	
	\par\vspace{2pt}\noindent
	\begin{minipage}{\linewidth}
		\centering
		\captionsetup{font=footnotesize,skip=2pt}
		
		\includegraphics[
		width=\linewidth,
		height=0.17\textheight
		]{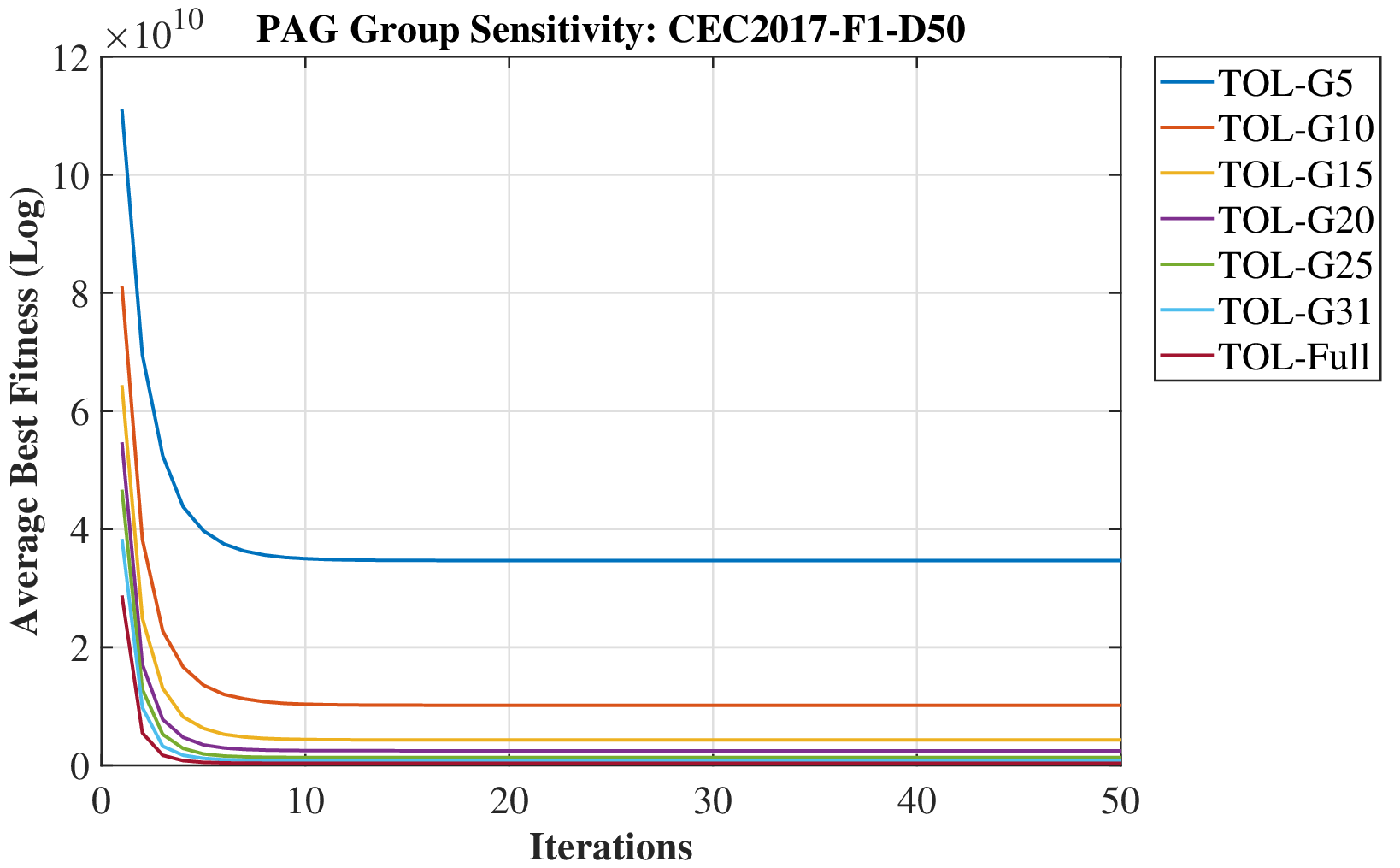}
		
		\par\vspace{1pt}
		{\footnotesize (a) F1}
		\par\vspace{2pt}
		
		\includegraphics[
		width=\linewidth,
		height=0.17\textheight
		]{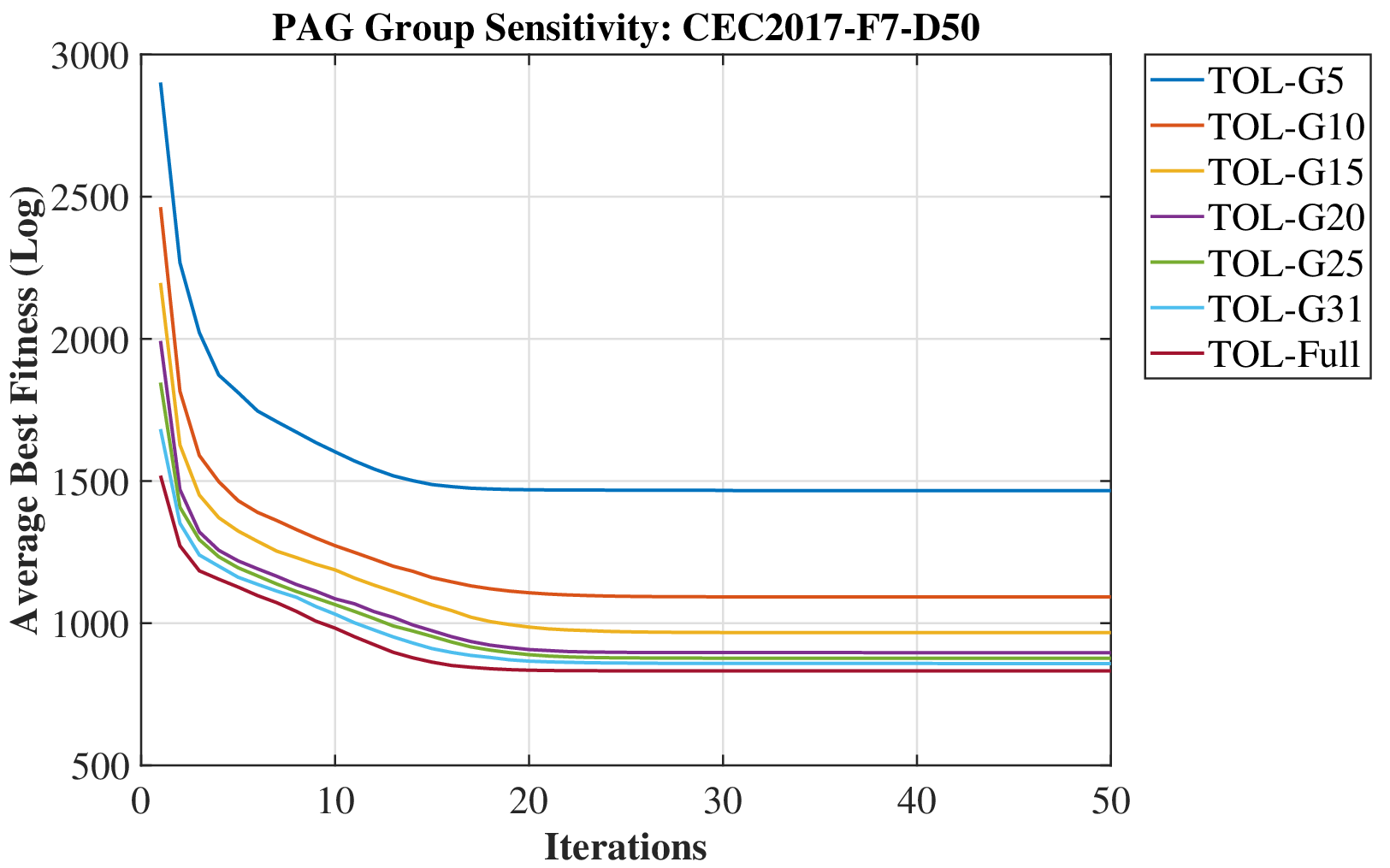}
		
		\par\vspace{1pt}
		{\footnotesize (b) F7}
		\par\vspace{2pt}
		
		\includegraphics[
		width=\linewidth,
		height=0.17\textheight
		]{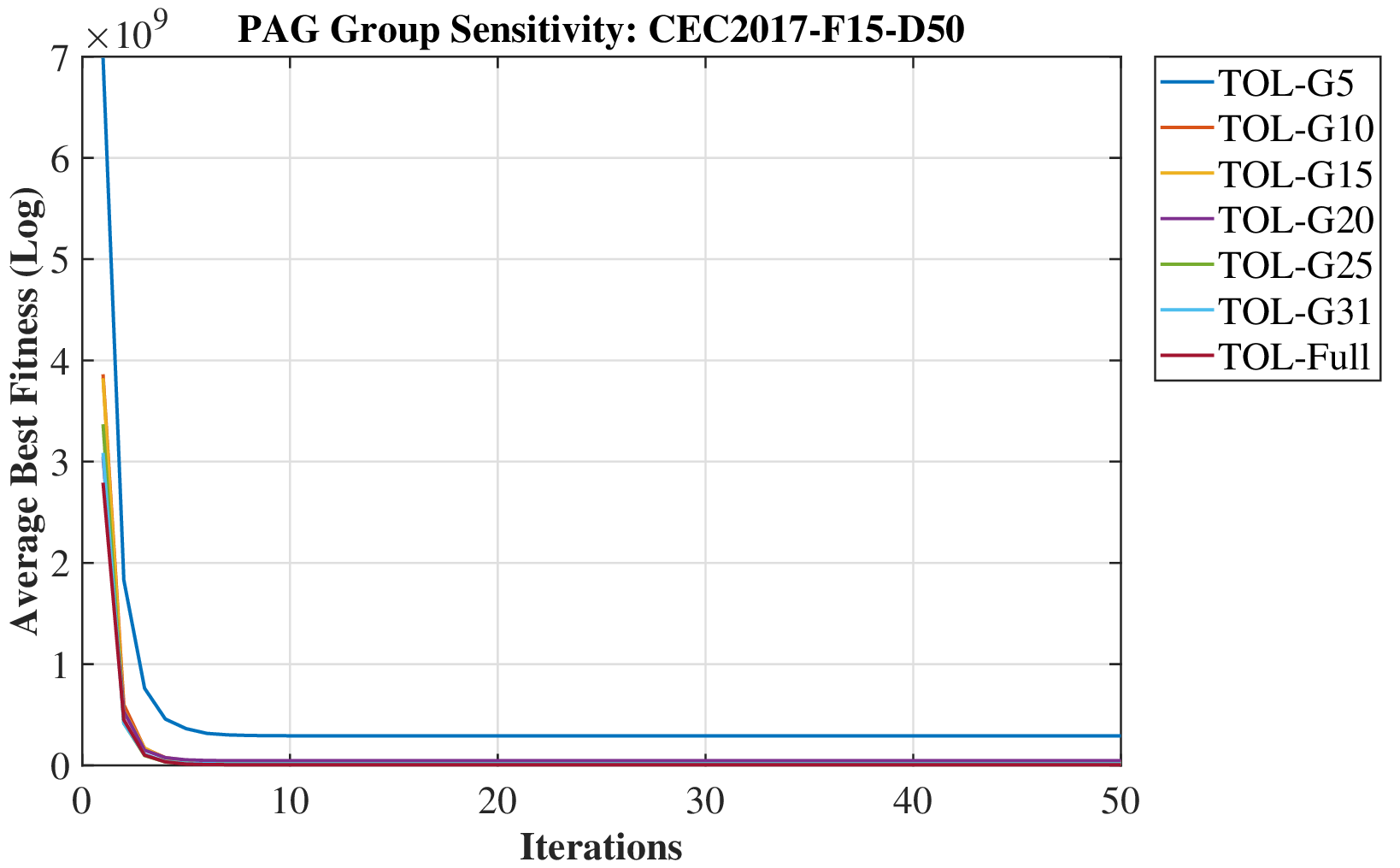}
		
		\par\vspace{1pt}
		{\footnotesize (c) F15}
		
		\captionof{figure}{CGOL convergence under different group numbers}
		\label{fig:cgol_group_convergence}
	\end{minipage}
	\par\vspace{2pt}
	
	\FloatBarrier
	\subsection{Experiments on IEEE CEC2019 benchmark functions}
	\label{subsec:cec2019_results}
	\suppressfloats[t]
	
	CEC2019 consists of 10 low-dimensional complex functions with different scales and search landscapes. In this study, F1 and F6 are selected for dynamic performance analysis. F1 is used to examine the early-stage convergence capability and differences in final solution accuracy in a large-scale search space, whereas F6 is employed to investigate sustained exploitation capability and run-to-run stability under complex nonlinear characteristics.
	
	\subsubsection{Comparative performance and statistical analysis}
	\label{subsubsec:cec2019_numerical}
	
	As shown in Table~\ref{tab:cec2019_results_basic}, IWOO achieves the lowest mean objective values on F2 and F4--F10, demonstrating strong solution accuracy across the CEC2019 benchmark suite. In terms of overall ranking, IWOO obtains a Friedman rank of 1.8000, ranking first among all compared algorithms and outperforming the original WOO, which obtains a rank of 3.4500. In addition, two-sided Wilcoxon signed-rank test yields a $9/0/1$ result for IWOO against WOO. Across the 120 pairwise comparisons, IWOO records 107 significant wins, 6 cases with no significant difference, and only 7 significant losses, indicating that its overall performance advantage is highly consistent across the benchmark functions.
	
	According to the No Free Lunch (NFL) theorem~\cite{Wolpert1997NFL}, no optimization algorithm can maintain absolute superiority over all possible optimization problems. On F1 and F3, IWOO does not achieve the lowest mean objective value; however, its weaker performance on a small number of functions does not alter its overall advantage across the benchmark suite. Furthermore, when compared with the high-performing algorithm group, IWOO achieves a Friedman rank of 2.3000 and ranks first overall, while obtaining the best results on half of the test functions. These results further demonstrate the cross-function adaptability and overall competitiveness of IWOO.
	
	\FloatBarrier
	\subsubsection{Convergence and robustness analysis}
	\label{subsubsec:cec2019_convergence}
	\suppressfloats[t]
	
	To further investigate the dynamic search behavior of IWOO, the average convergence curves and box plots for F1 and F6 are analyzed, as shown in Fig.~\ref{fig:cec2019_basic_plots}. Figures~\ref{fig:cec2019_basic_plots}(a)--(b) show that, on F1, IWOO rapidly enters a low-fitness region during the early search stage and exhibits a relatively concentrated distribution of final solutions, indicating high search efficiency and favorable run-to-run stability. For F6, Figs.~\ref{fig:cec2019_basic_plots}(c)--(d) show that IWOO gradually stabilizes after approximately 100--150 iterations, while the distribution of its final solutions is better than that of most compared algorithms. Overall, IWOO exhibits both fast convergence and stable optimization performance on the representative CEC2019 functions.
	
	\par\addvspace{5pt}
	\noindent
	\begin{minipage}{\columnwidth}
		\centering
		\captionsetup{skip=3pt}
		
		\begin{minipage}[b]{0.48\linewidth}
			\centering
			\includegraphics[width=\linewidth,height=0.15\textheight,keepaspectratio]{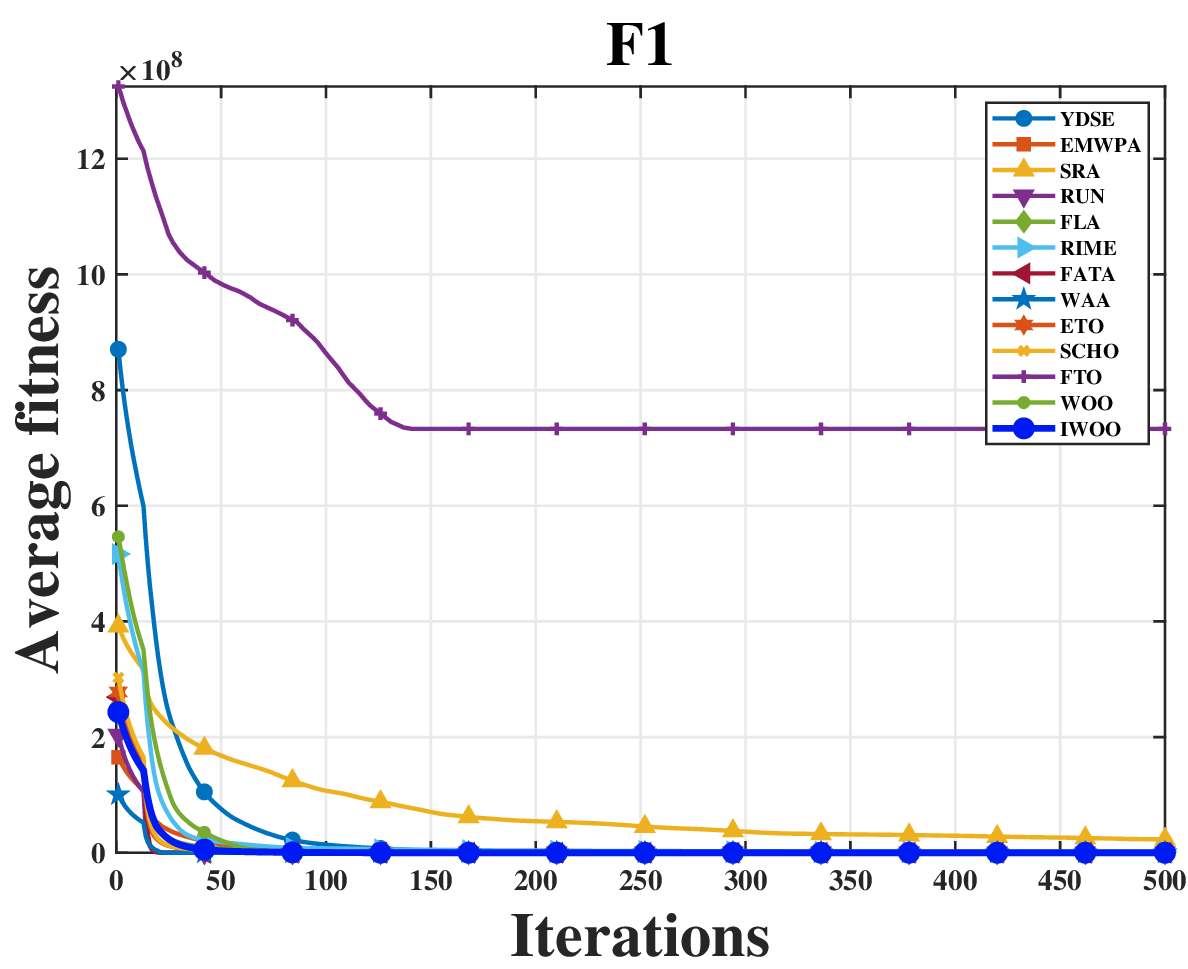}
			\par\vspace{-1mm}
			{\footnotesize\textbf{(a)}}
		\end{minipage}\hfill
		\begin{minipage}[b]{0.48\linewidth}
			\centering
			\includegraphics[width=\linewidth,height=0.15\textheight,keepaspectratio]{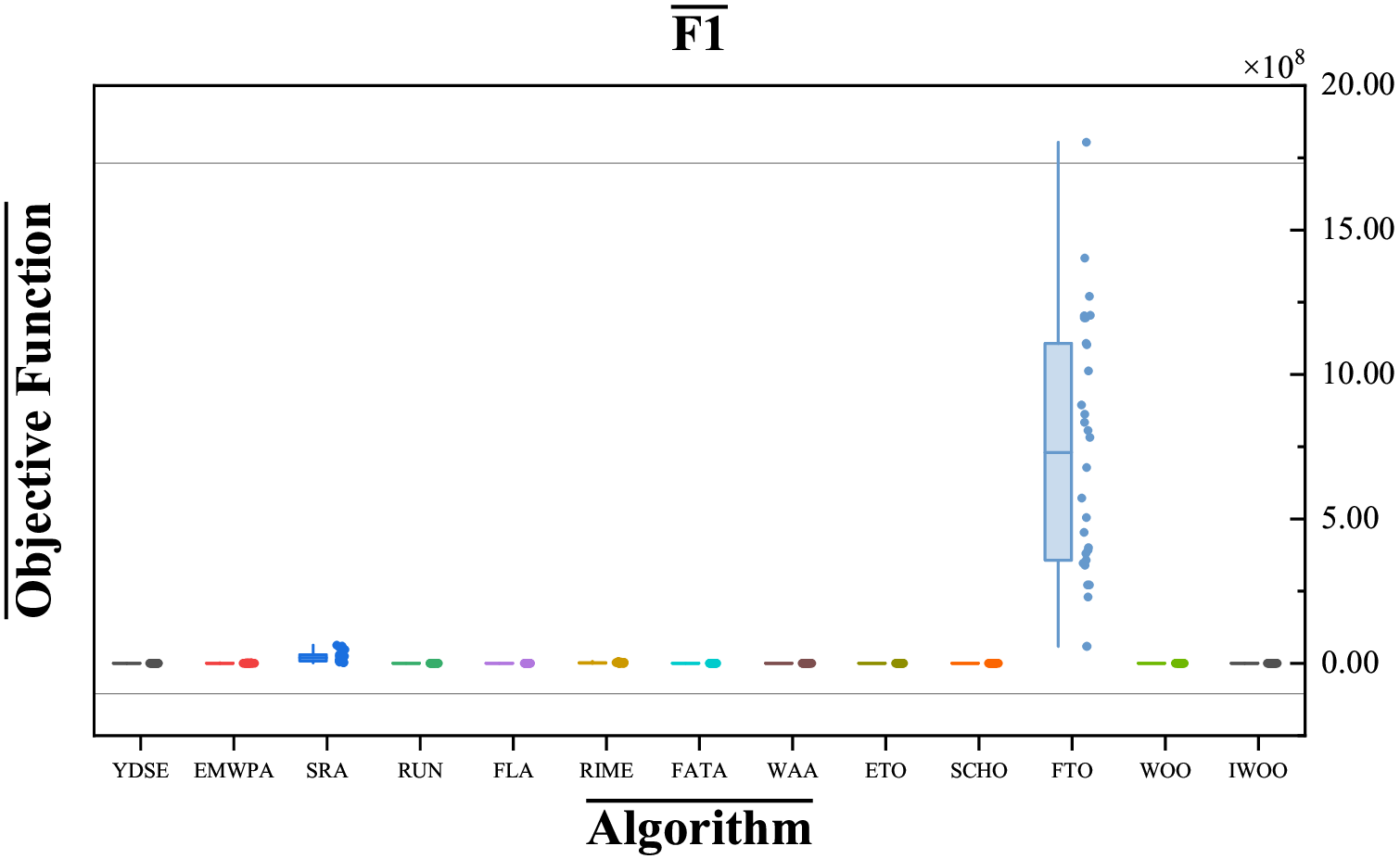}
			\par\vspace{-1mm}
			{\footnotesize\textbf{(b)}}
		\end{minipage}
		
		\par\vspace{2mm}
		
		\begin{minipage}[b]{0.48\linewidth}
			\centering
			\includegraphics[width=\linewidth,height=0.15\textheight,keepaspectratio]{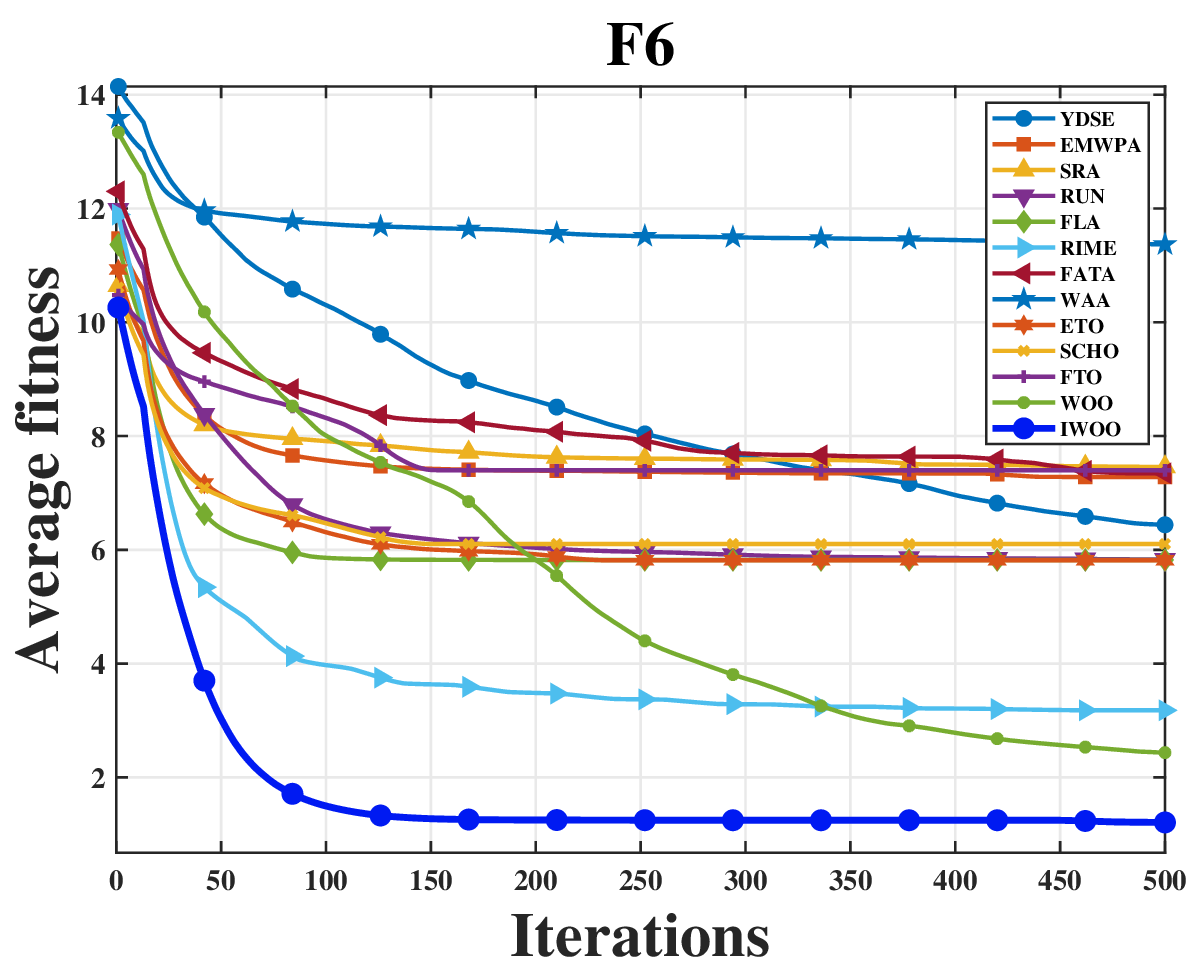}
			\par\vspace{-1mm}
			{\footnotesize\textbf{(c)}}
		\end{minipage}\hfill
		\begin{minipage}[b]{0.48\linewidth}
			\centering
			\includegraphics[width=\linewidth,height=0.15\textheight,keepaspectratio]{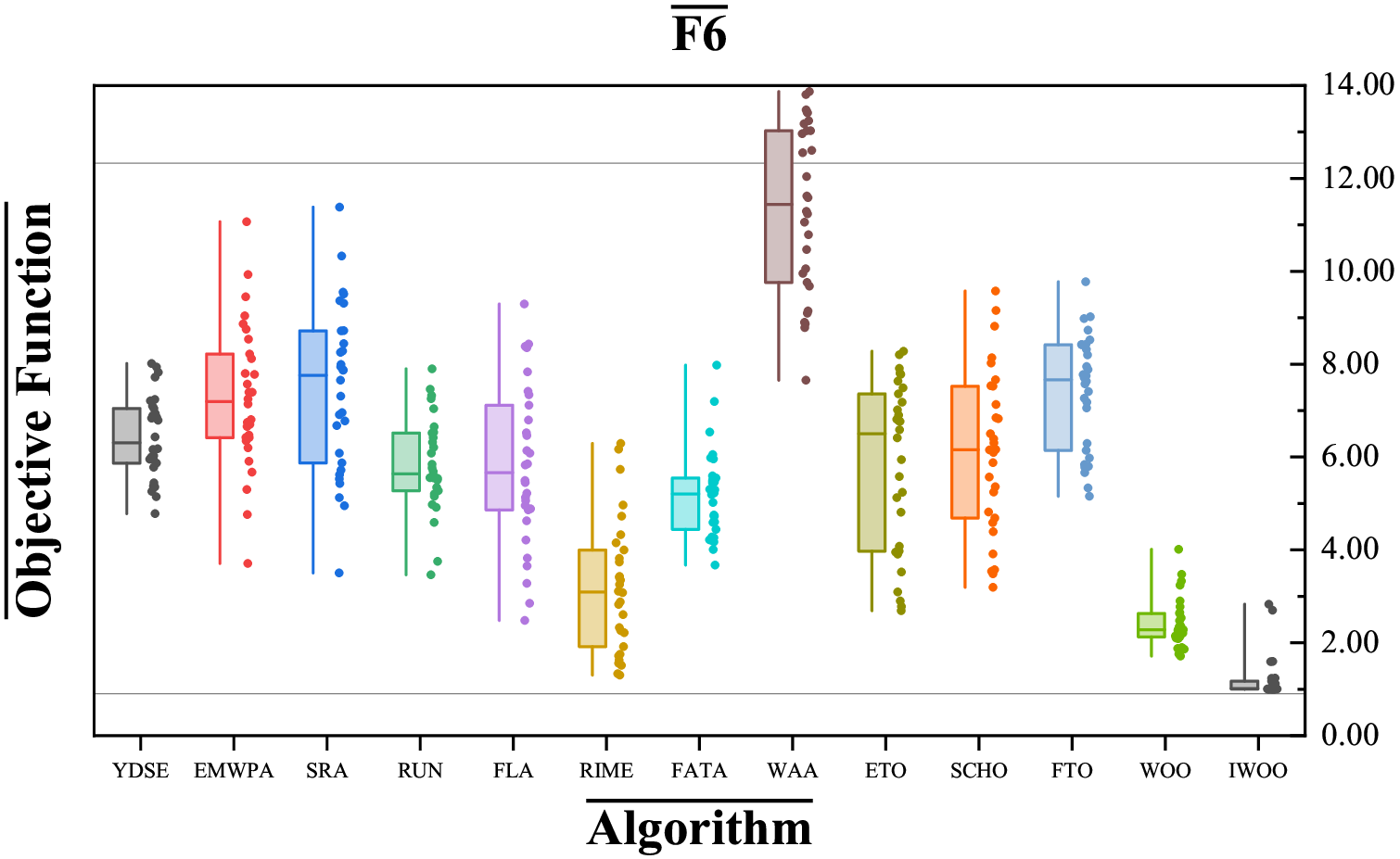}
			\par\vspace{-1mm}
			{\footnotesize\textbf{(d)}}
		\end{minipage}
		
		\par\vspace{1mm}
		\captionof{figure}{Convergence curves and box plots on representative CEC2019 benchmark functions}
		\label{fig:cec2019_basic_plots}
	\end{minipage}
	\par\addvspace{5pt}
	
	\subsection{Experiments on IEEE CEC2022 benchmark functions}
	\label{subsec:cec2022_results}
	
	CEC2022 consists of 12 complex functions, and the problem dimensionality is uniformly set to 20 to evaluate the performance of the algorithms under medium-dimensional conditions. To cover different function characteristics, F1, F4, F7, and F12 are selected as representative functions, corresponding to unimodal, multimodal, hybrid, and composition functions, respectively. These functions are used to examine the local exploitation capability, global exploration capability, exploration--exploitation balance, and adaptability of IWOO to complex search spaces.
	
	\par
	\makeatletter
	\global\@dbltopnum=-1
	\makeatother
	\begin{figure*}[!t]
		\centering
		\captionsetup{skip=3pt}
		\setlength{\tabcolsep}{0.8pt}
		\begin{tabular}{@{}cccc@{}}
			\includegraphics[width=0.242\textwidth,height=0.15\textheight,keepaspectratio]{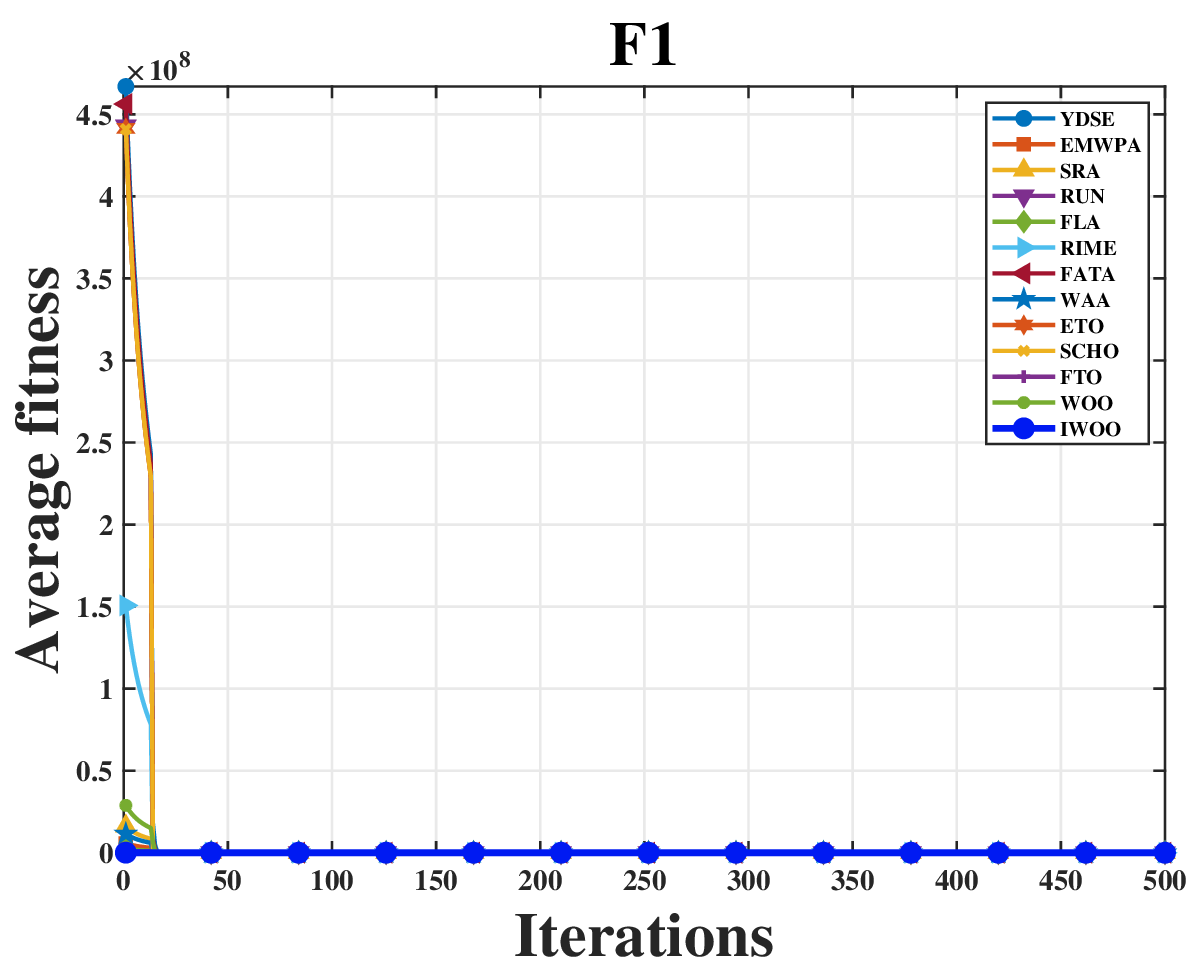}
			& \includegraphics[width=0.242\textwidth,height=0.15\textheight,keepaspectratio]{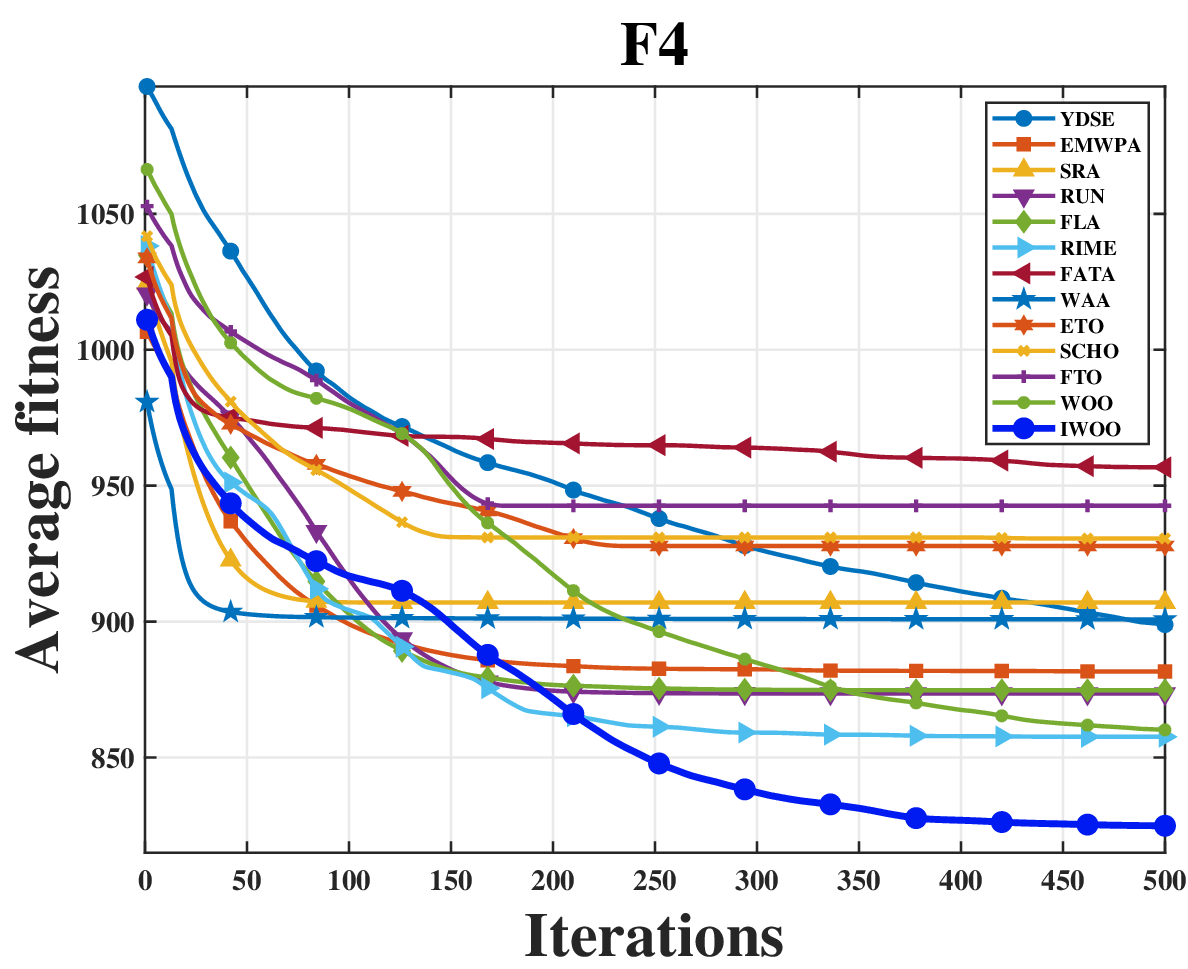}
			& \includegraphics[width=0.242\textwidth,height=0.15\textheight,keepaspectratio]{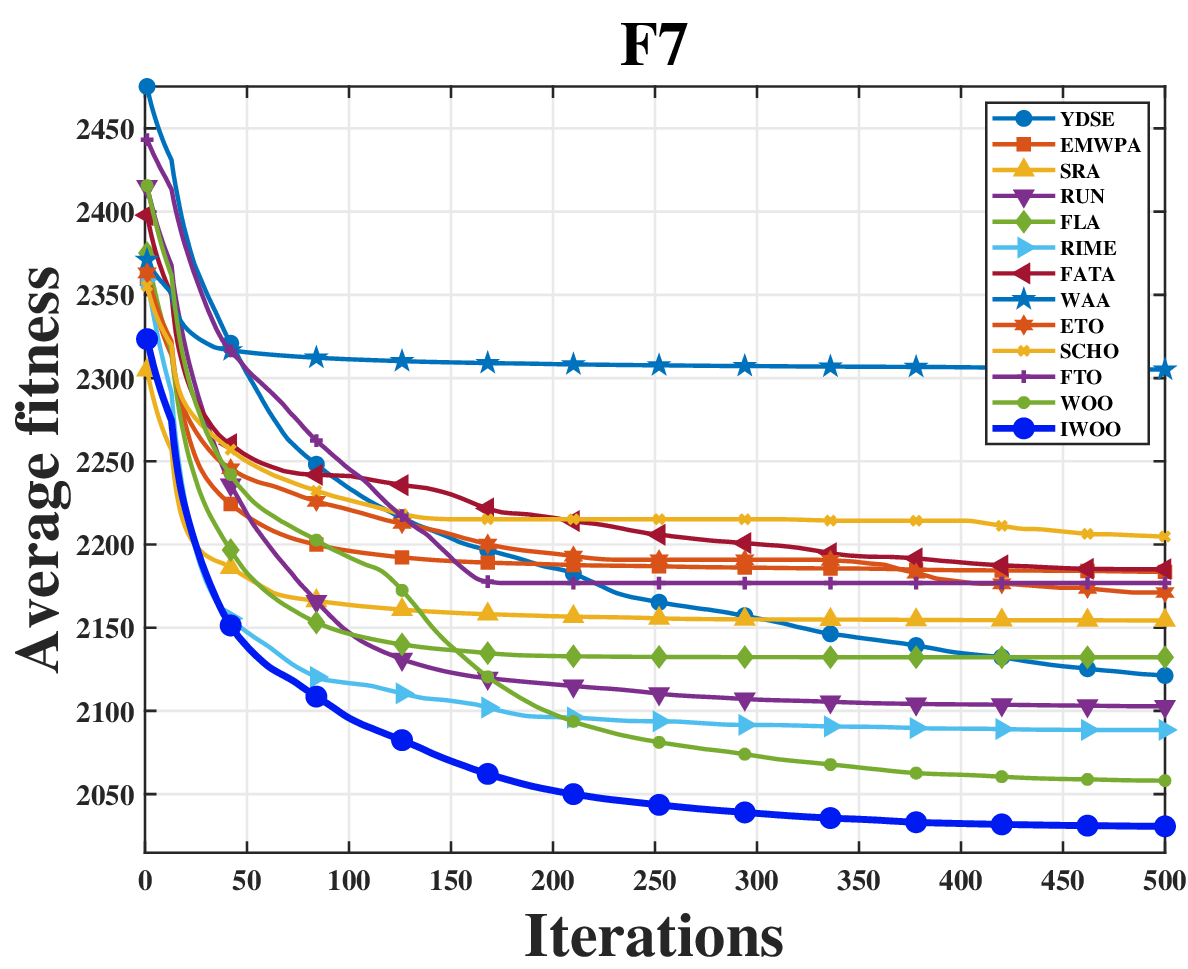}
			& \includegraphics[width=0.242\textwidth,height=0.15\textheight,keepaspectratio]{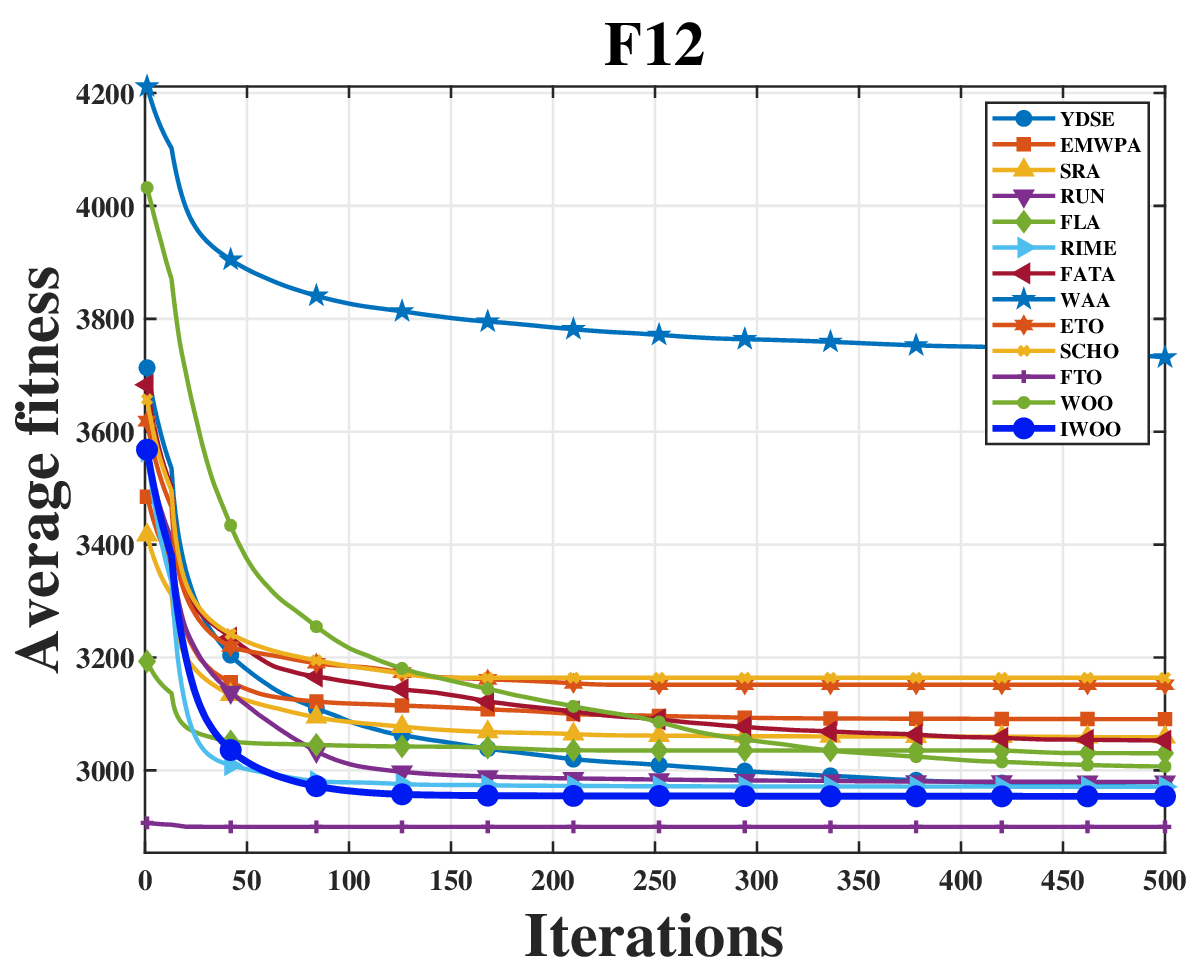}
			\\[0.3mm]
			{\footnotesize\textbf{(a)}} & {\footnotesize\textbf{(c)}} & {\footnotesize\textbf{(e)}} & {\footnotesize\textbf{(g)}}
			\\[2mm]
			\includegraphics[width=0.242\textwidth,height=0.15\textheight,keepaspectratio]{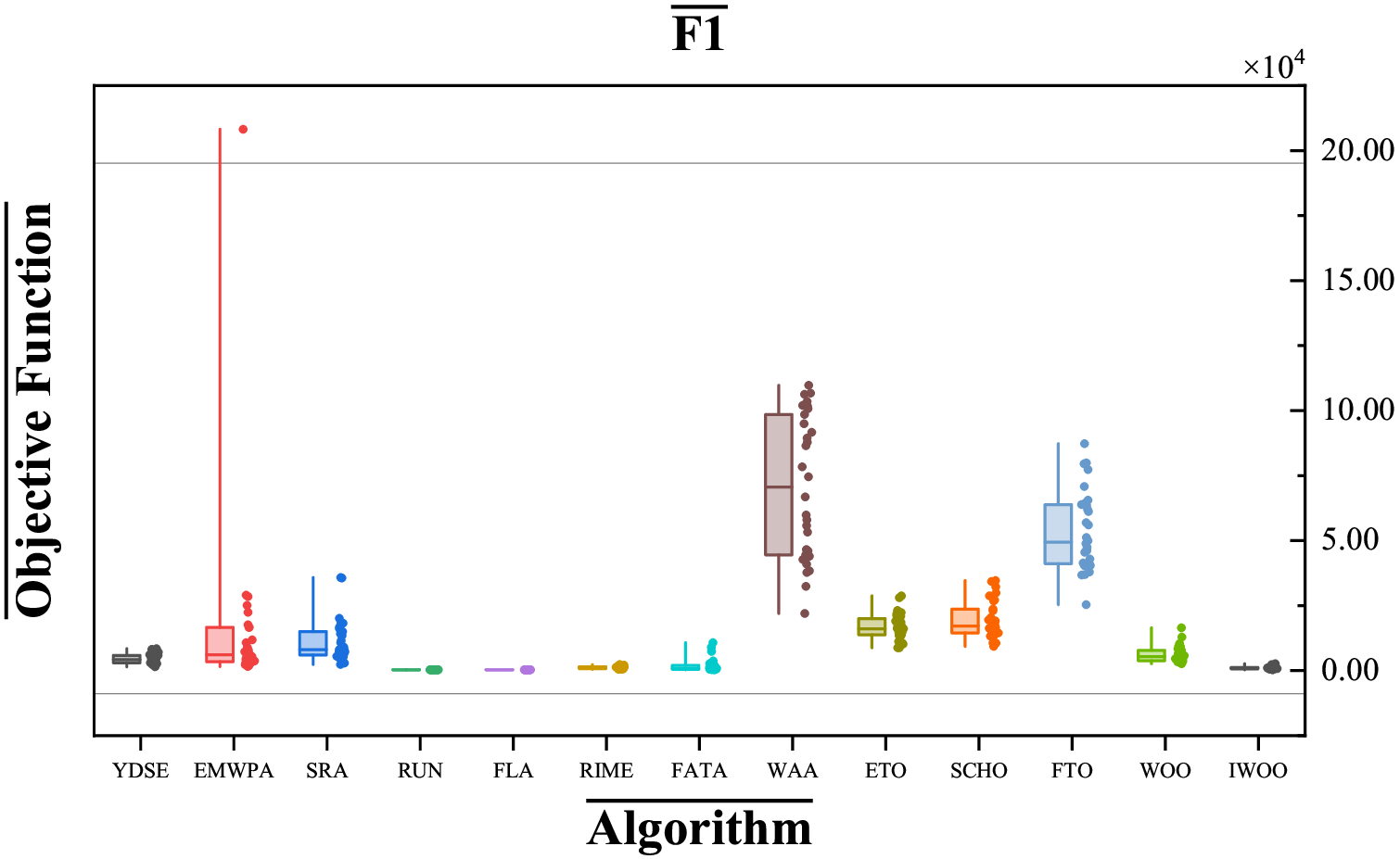}
			& \includegraphics[width=0.242\textwidth,height=0.15\textheight,keepaspectratio]{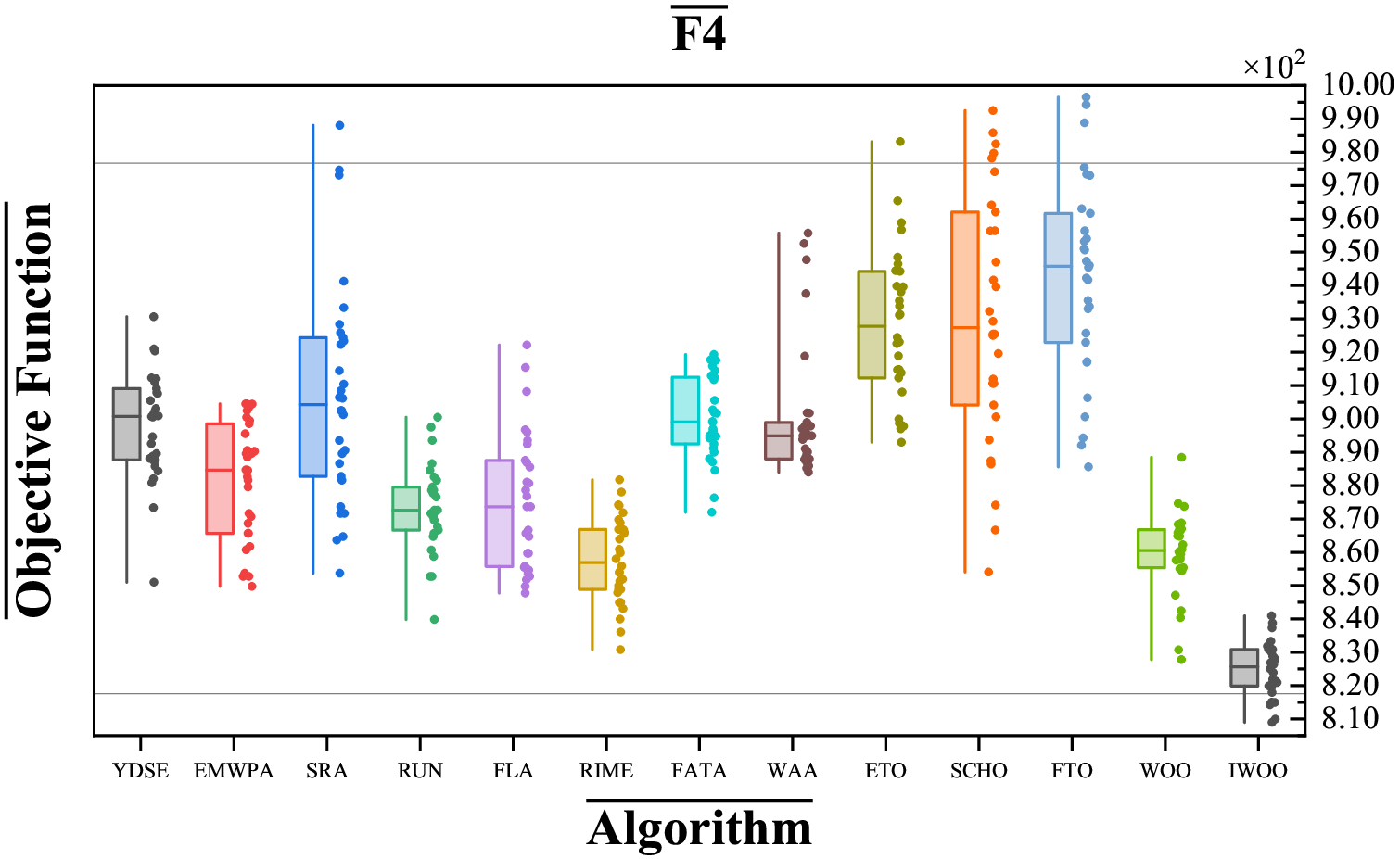}
			& \includegraphics[width=0.242\textwidth,height=0.15\textheight,keepaspectratio]{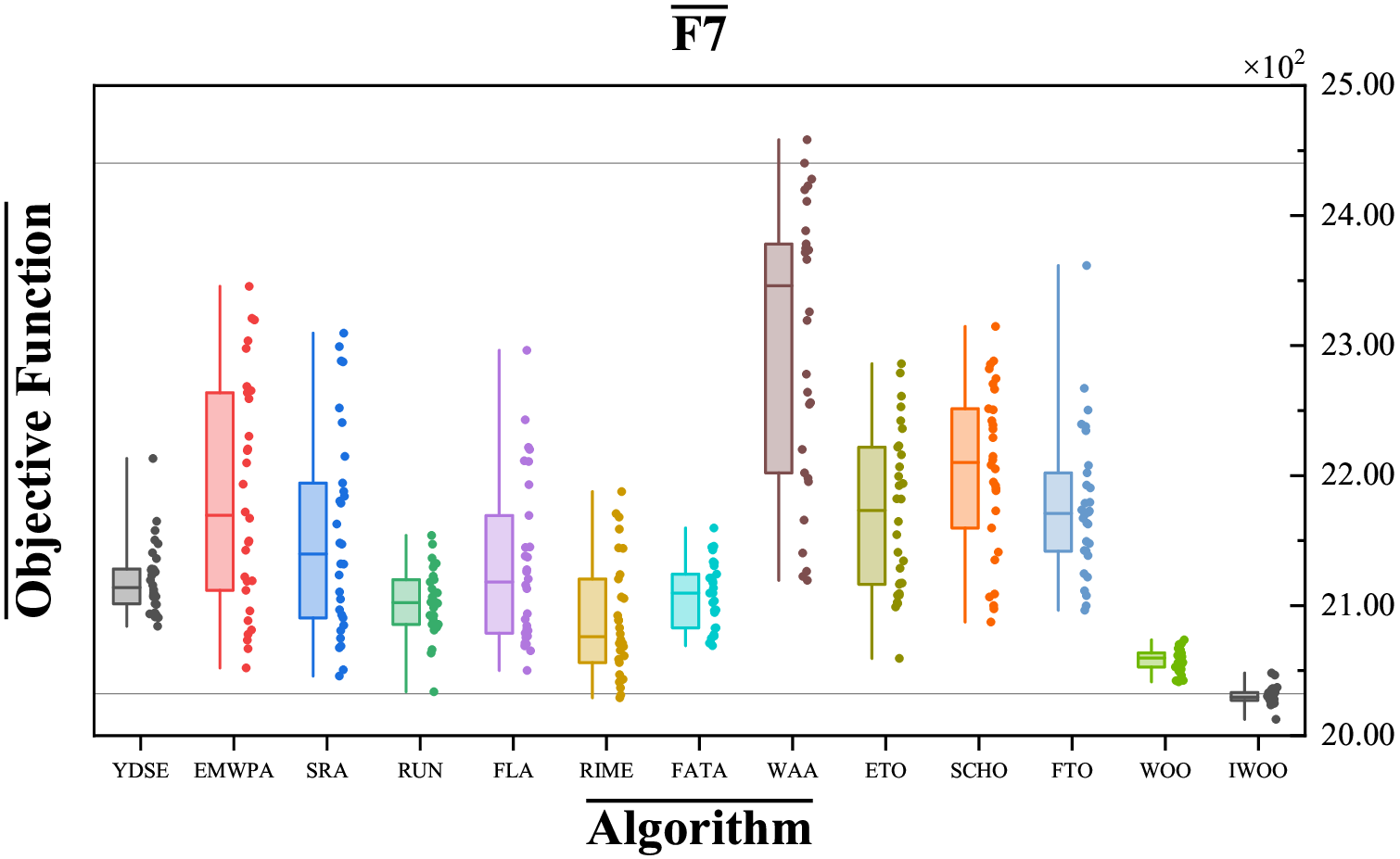}
			& \includegraphics[width=0.242\textwidth,height=0.15\textheight,keepaspectratio]{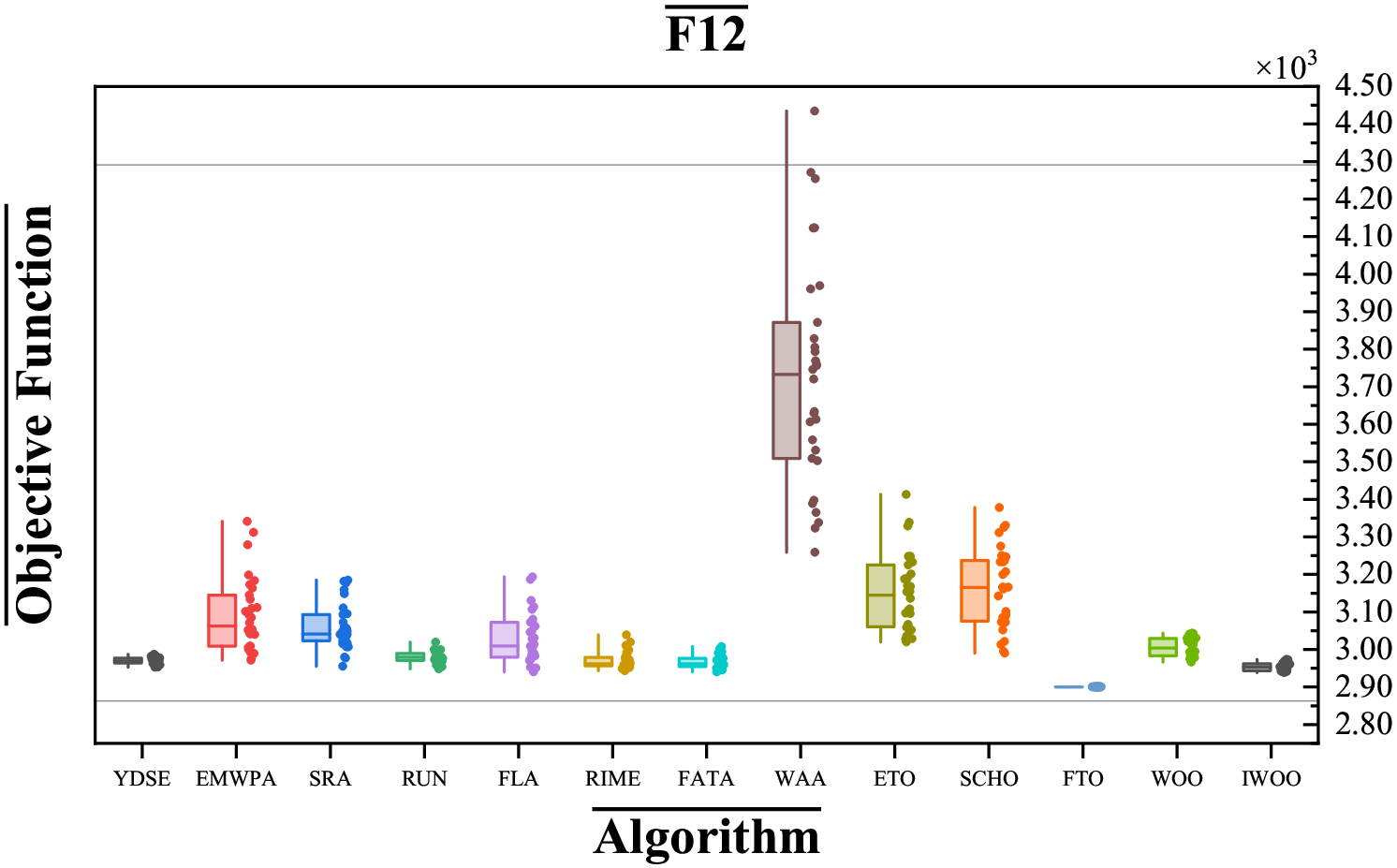}
			\\[0.3mm]
			{\footnotesize\textbf{(b)}} & {\footnotesize\textbf{(d)}} & {\footnotesize\textbf{(f)}} & {\footnotesize\textbf{(h)}}
		\end{tabular}
		\caption{Convergence curves and box plots on representative CEC2022 benchmark functions}
		\label{fig:cec2022_basic_all_plots}
	\end{figure*}
	
	\subsubsection{Comparative performance and statistical analysis}
	\label{subsubsec:cec2022_numerical}
	
	As shown in Table~\ref{tab:cec2022_results_basic}, IWOO achieves the lowest mean objective values on eight test functions, namely F3--F5 and F7--F11. It obtains a Friedman rank of 1.5833, ranking first among all compared algorithms. The two-sided Wilcoxon signed-rank test further shows that IWOO achieves a $12/0/0$ result against each of WOO, YDSE, SRA, WAA, ETO, and SCHO. Across the 144 pairwise comparisons, IWOO records 134 significant wins, 5 cases with no significant difference, and only 5 significant losses, demonstrating a consistent overall advantage across the CEC2022 benchmark suite.
	
	Nevertheless, IWOO does not achieve the best performance on F1, F2, F6, and F12, indicating that different algorithms retain certain complementary advantages on specific test functions. When compared with the high-performing algorithm group on CEC2022, IWOO obtains a Friedman rank of 3.0833 and ranks first overall, while achieving the lowest mean objective values on five test functions. These results further confirm the overall competitiveness of IWOO.
	
	\subsubsection{Convergence and robustness analysis}
	\label{subsubsec:cec2022_convergence}
	
	To further investigate the convergence characteristics of IWOO, F1, F4, F7, and F12 are selected as representative functions, and their average convergence curves and box plots are presented in Fig.~\ref{fig:cec2022_basic_all_plots}. As shown in Figs.~\ref{fig:cec2022_basic_all_plots}(c)--(f), on F4 and F7, IWOO continuously searches toward lower-fitness regions and produces relatively low and concentrated final-result distributions, demonstrating favorable solution accuracy and run-to-run stability. For F1 and F12, Figs.~\ref{fig:cec2022_basic_all_plots}(a)--(b) and (g)--(h) show that IWOO exhibits relatively fast early-stage convergence, although its final solution accuracy is slightly inferior to that achieved by FLA, RUN, and FTO on these functions. Overall, the results indicate that IWOO achieves a favorable balance among convergence speed, solution accuracy, and run-to-run stability on most representative CEC2022 functions.
	
	\subsection{Experiments on IEEE CEC2017 benchmark functions}
	\label{subsec:cec2017_50d_results}
	
	The CEC2017 benchmark suite contains a variety of complex functions, including unimodal, multimodal, hybrid, and composition functions. In this experiment, F1 and F3--F30, comprising 29 test functions in total, are employed for performance evaluation. Among them, F1, F10, F20, and F30 are selected as representative functions for visualization to cover different search characteristics.
	
	\par
	\makeatletter
	\global\@dbltopnum=-1
	\makeatother
	\begin{figure*}[!t]
		\centering
		\captionsetup{skip=3pt}
		\setlength{\tabcolsep}{0.8pt}
		\begin{tabular}{@{}cccc@{}}
			\includegraphics[width=0.242\textwidth,height=0.15\textheight,keepaspectratio]{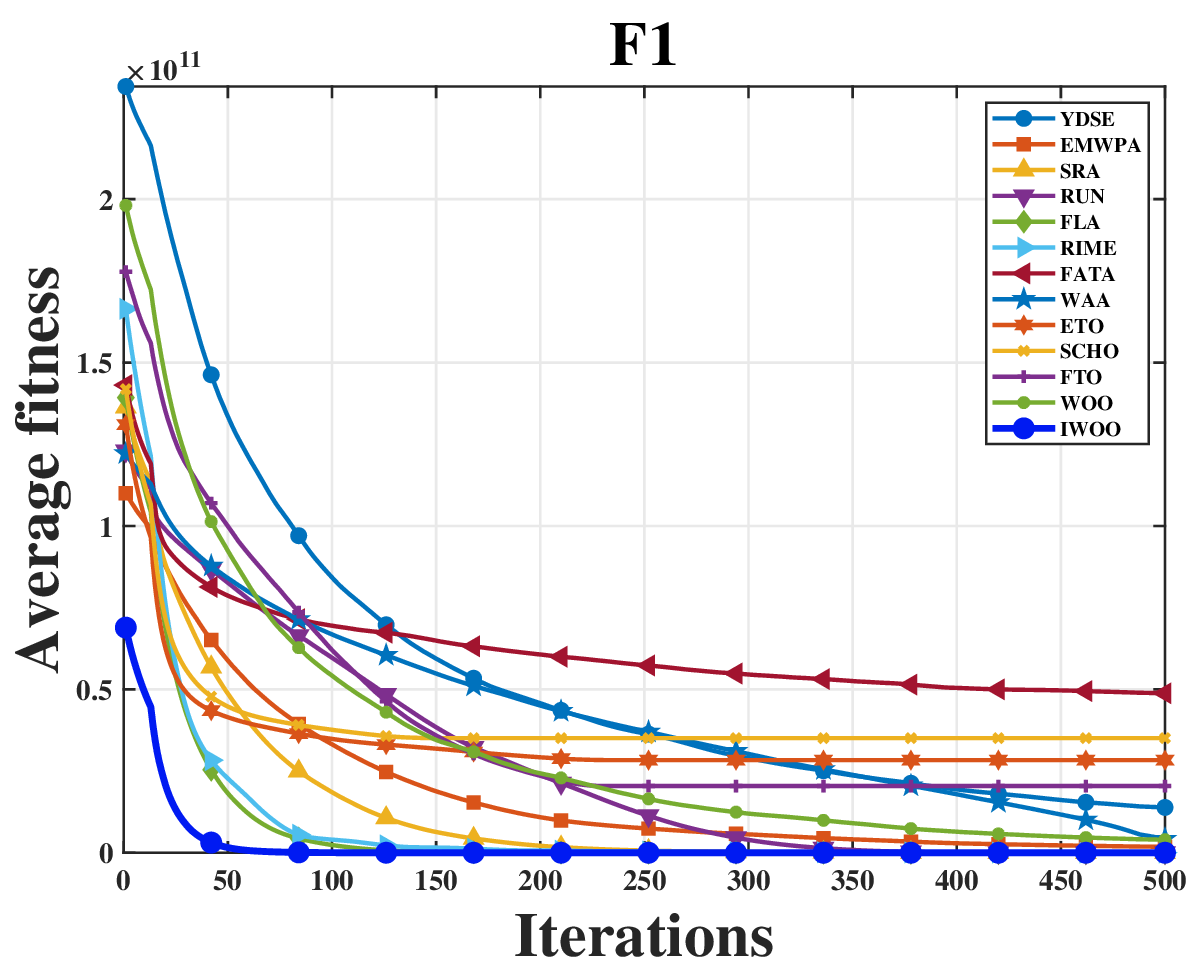}
			& \includegraphics[width=0.242\textwidth,height=0.15\textheight,keepaspectratio]{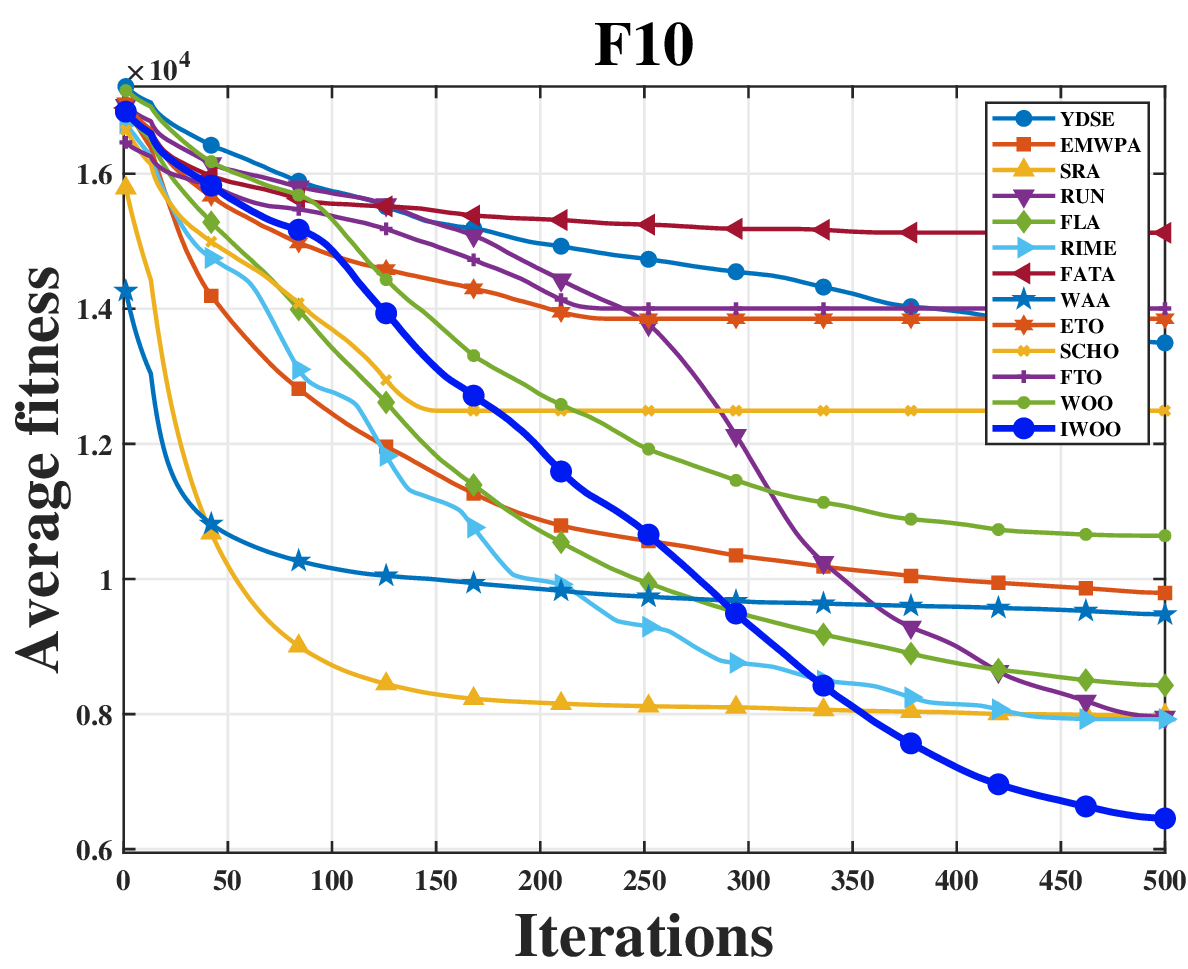}
			& \includegraphics[width=0.242\textwidth,height=0.15\textheight,keepaspectratio]{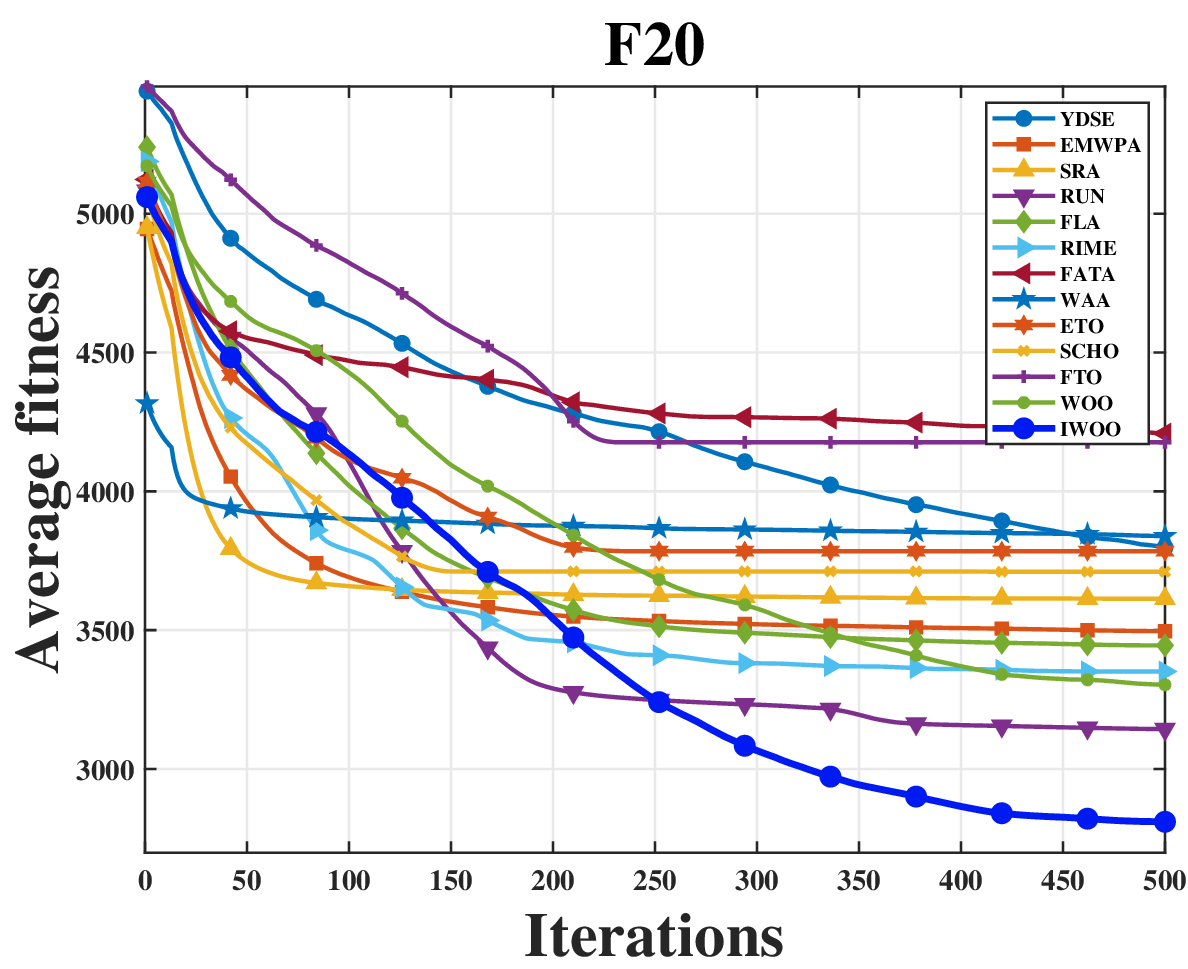}
			& \includegraphics[width=0.242\textwidth,height=0.15\textheight,keepaspectratio]{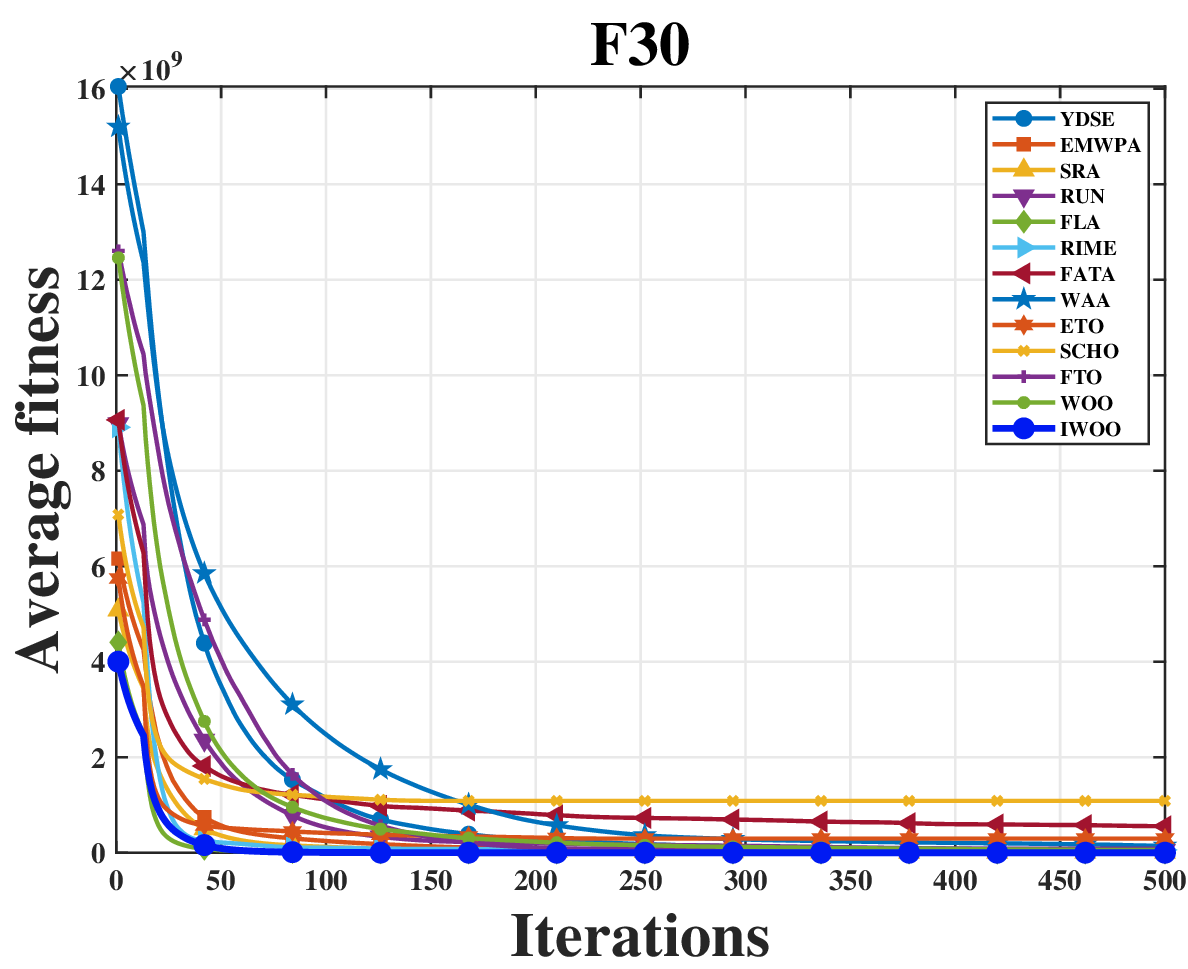}
			\\[0.3mm]
			{\footnotesize\textbf{(a)}} & {\footnotesize\textbf{(c)}} & {\footnotesize\textbf{(e)}} & {\footnotesize\textbf{(g)}}
			\\[2mm]
			\includegraphics[width=0.242\textwidth,height=0.15\textheight,keepaspectratio]{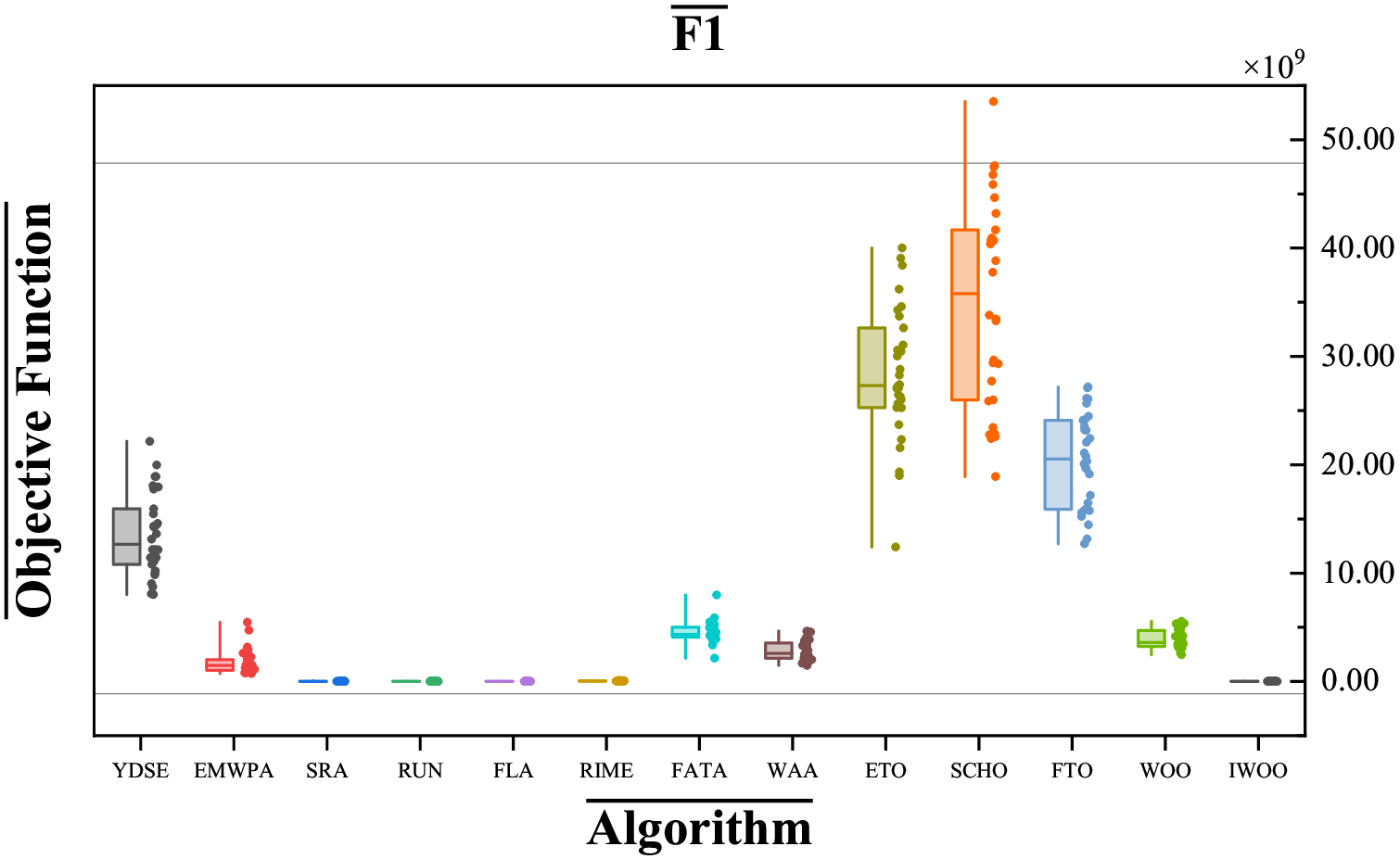}
			& \includegraphics[width=0.242\textwidth,height=0.15\textheight,keepaspectratio]{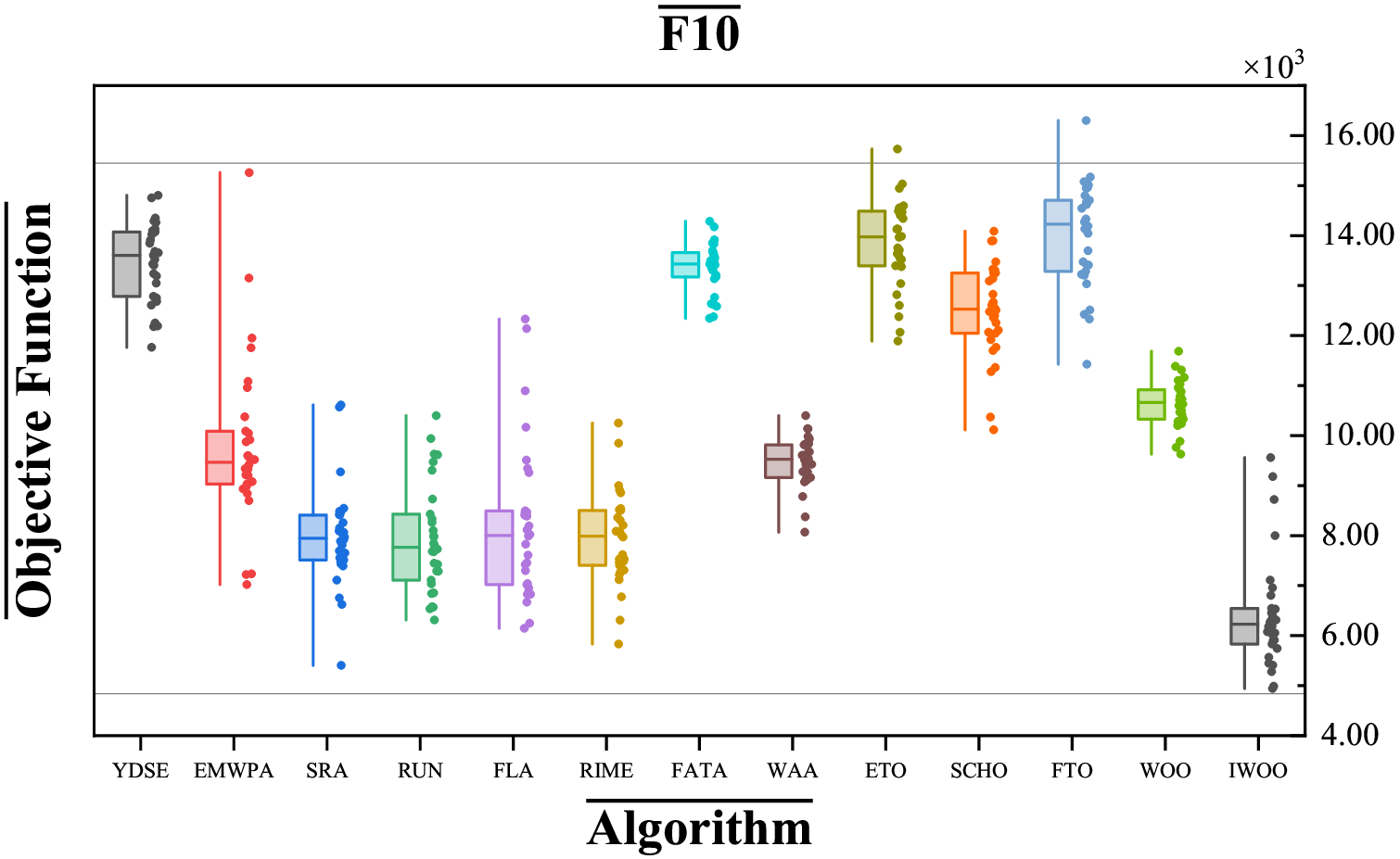}
			& \includegraphics[width=0.242\textwidth,height=0.15\textheight,keepaspectratio]{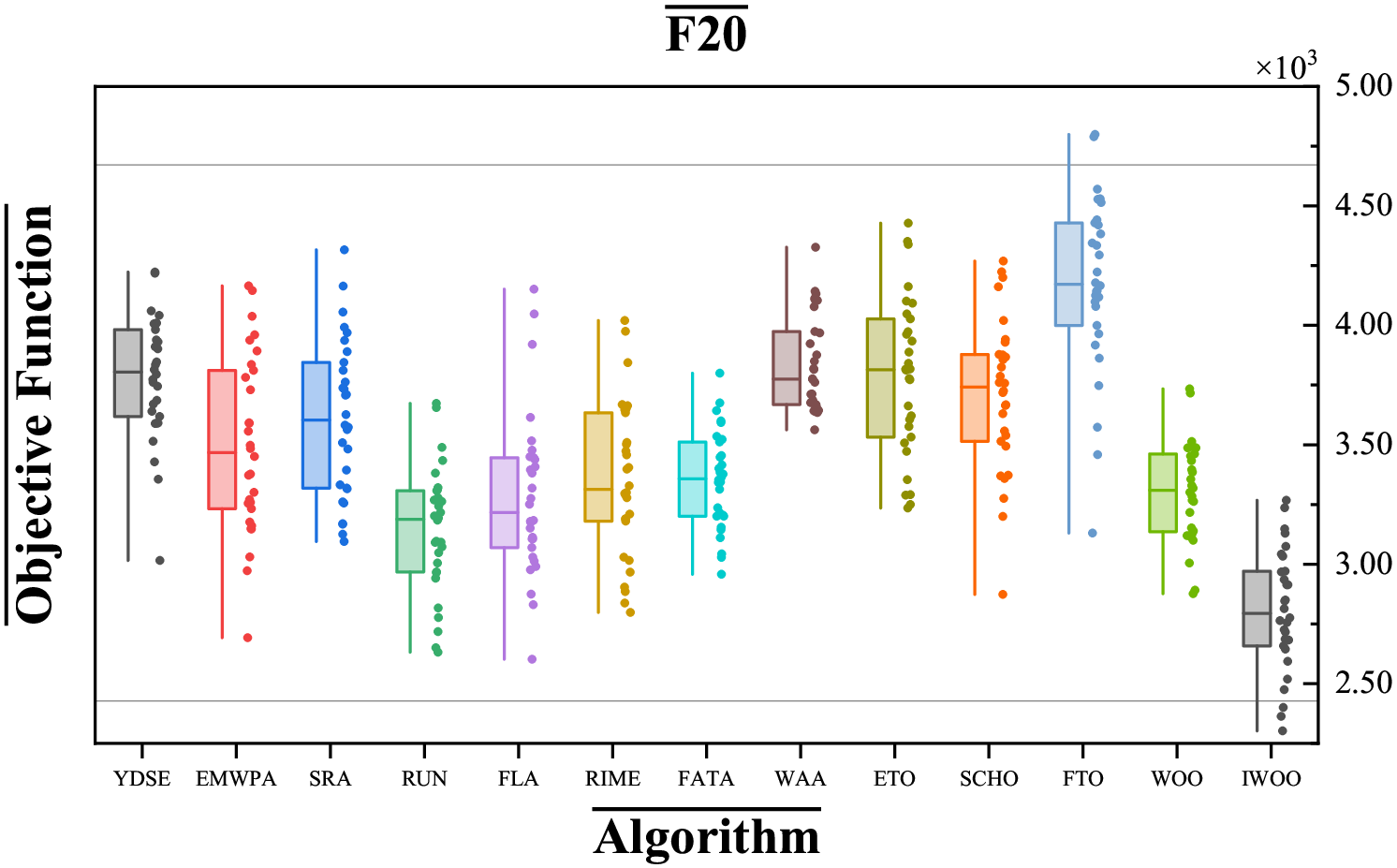}
			& \includegraphics[width=0.242\textwidth,height=0.15\textheight,keepaspectratio]{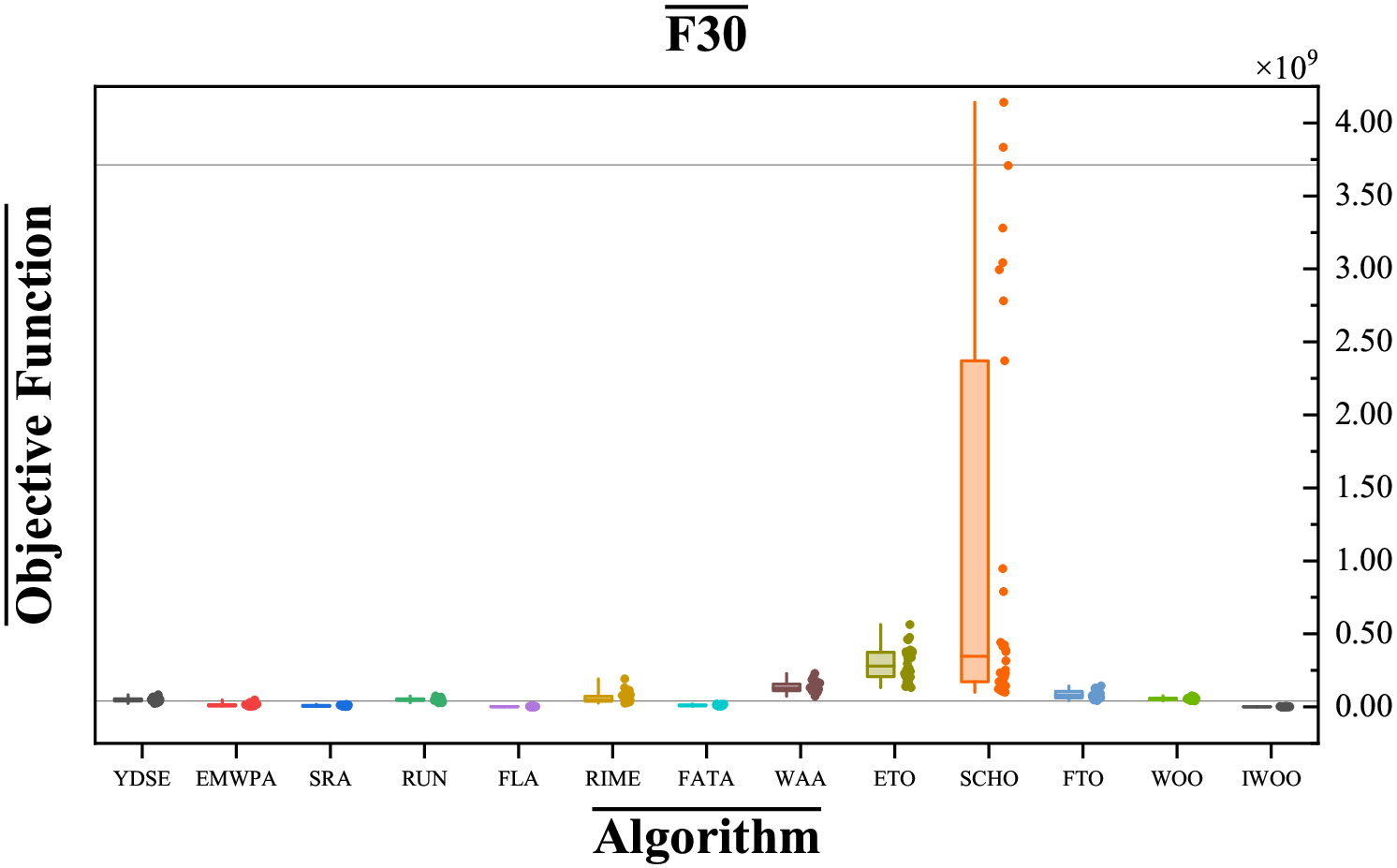}
			\\[0.3mm]
			{\footnotesize\textbf{(b)}} & {\footnotesize\textbf{(d)}} & {\footnotesize\textbf{(f)}} & {\footnotesize\textbf{(h)}}
		\end{tabular}
		\caption{Convergence curves and box plots on representative CEC2017 benchmark functions}
		\label{fig:cec2017_50d_basic_plots}
	\end{figure*}
	
	\subsubsection{Comparative performance and statistical analysis}
	\label{subsubsec:cec2017_50d_numerical}
	
	Table~\ref{tab:cec2017_50d_results_basic} presents the statistical results of the compared algorithms on CEC2017. IWOO achieves the lowest mean objective values on 19 of the 29 test functions, with its advantages extending across multiple categories of complex functions.
	
	IWOO obtains a Friedman rank of 1.4483 and ranks first among all compared algorithms, substantially outperforming FLA, RUN, RIME, and the original WOO. In the two-sided Wilcoxon signed-rank test, IWOO achieves a $28/1/0$ result against the original WOO. Across the 348 pairwise comparisons, IWOO records 334 significant wins, 1 case with no significant difference, and only 13 significant losses. These results indicate that the performance advantage of IWOO is consistently observed across the majority of the test functions rather than being driven by isolated cases. When compared with the high-performing algorithm group, IWOO further achieves the best Friedman rank of 1.8276, confirming its strong overall performance in high-dimensional search spaces.
	
	\subsubsection{Convergence and robustness analysis}
	\label{subsubsec:cec2017_50d_convergence}
	
	Figure~\ref{fig:cec2017_50d_basic_plots} presents the average convergence curves and box plots of IWOO on the representative CEC2017 functions F1, F10, F20, and F30. As shown in Figs.~\ref{fig:cec2017_50d_basic_plots}(c)--(h), on F10, F20, and F30, IWOO continues to converge toward lower-fitness regions even after several compared algorithms gradually enter stagnation. The corresponding final-result distributions are also relatively concentrated, indicating favorable sustained exploitation capability and run-to-run stability. For F1, Figs.~\ref{fig:cec2017_50d_basic_plots}(a)--(b) show that IWOO decreases rapidly during the early search stage, while the magnitude of improvement becomes limited in later iterations. Nevertheless, IWOO still achieves the best final solution accuracy among the compared algorithms. Overall, IWOO effectively coordinates global exploration and local exploitation on most representative functions, although further improvement remains possible on certain specific functions.
	
	\FloatBarrier

	\begin{sidewaystable*}[p]
		\centering
		
		\setlength{\abovecaptionskip}{3pt}
		\setlength{\belowcaptionskip}{3pt}
		
		\caption{Test result for CEC 2019 (Dim$<$20) - Basic Group}
		\label{tab:cec2019_results_basic}
		
		\scriptsize
		\setlength{\tabcolsep}{3.6pt}
		\renewcommand{\arraystretch}{0.78}
		
		\begin{adjustbox}{
				width=0.98\textheight,
				max totalheight=0.88\textwidth,
				keepaspectratio,
				center
			}
			\begin{tabular}{c c ccccccccccccc} 
				\toprule
				Function & & YDSE & EMWPA & SRA & RUN & FLA & RIME & FATA & WAA & ETO & SCHO & FTO & WOO & IWOO \\
				\midrule
				
				\multirow{5}{*}{F1} 
				& min       & 4.8282e+02 & 1.9748e+04 & 1.4377e+06 & 1.0000e+00 & 1.0042e+00 & 1.9489e+05 & 1.0000e+00 & 1.0000e+00 & 1.0000e+00 & 1.0000e+00 & 5.8331e+07 & 1.0000e+00 & 1.0000e+00 \\
				& mean      & 2.4219e+03 & 2.8281e+05 & 2.3223e+07 & \textbf{1.0000e+00} & 6.2633e+02 & 1.6679e+06 & \textbf{1.0000e+00} & \textbf{1.0000e+00} & \textbf{1.0000e+00} & \textbf{1.0000e+00} & 7.3289e+08 & \textbf{1.0000e+00} & 1.0001e+00 \\
				& std       & 1.8734e+03 & 3.3870e+05 & 1.8442e+07 & 9.1652e-15 & 1.1362e+03 & 1.2774e+06 & 3.8946e-10 & 0.0000e+00 & 0.0000e+00 & 0.0000e+00 & 4.4407e+08 & 0.0000e+00 & 4.5346e-04 \\
				& p-value   & 1.7344e-06 & 1.7344e-06 & 1.7344e-06 & 1.2279e-05 & 1.7344e-06 & 1.7344e-06 & 6.1261e-05 & 1.2279e-05 & 1.2279e-05 & 1.2279e-05 & 1.7344e-06 & 1.2279e-05 & - - \\
				& + / = / - & +          & +          & +          & -          & +          & +          & -          & -          & -          & -          & +          & -          & - - \\
				\midrule
				
				\multirow{5}{*}{F2} 
				& min       & 8.0831e+01 & 6.6775e+01 & 4.1457e+03 & 4.2208e+00 & 1.4673e+02 & 3.3792e+02 & 4.2168e+00 & 4.9577e+00 & 1.3906e+01 & 4.9941e+00 & 4.5590e+03 & 3.9092e+00 & 3.2496e+00 \\
				& mean      & 1.5671e+02 & 9.1151e+02 & 8.6104e+03 & 4.2775e+00 & 4.1487e+02 & 1.0081e+03 & 4.3451e+00 & 4.9979e+00 & 1.1698e+03 & 4.9997e+00 & 1.8926e+04 & 4.3823e+00 & \textbf{4.0308e+00} \\
				& std       & 3.4635e+01 & 9.2748e+02 & 2.4968e+03 & 4.4885e-02 & 1.2513e+02 & 4.4273e+02 & 2.2787e-01 & 8.3902e-03 & 7.4721e+02 & 1.2043e-03 & 7.2731e+03 & 2.4633e-01 & 3.3094e-01 \\
				& p-value   & 1.7344e-06 & 1.7344e-06 & 1.7344e-06 & 7.1570e-04 & 1.7344e-06 & 1.7344e-06 & 8.3071e-04 & 1.7344e-06 & 1.7344e-06 & 1.7344e-06 & 1.7344e-06 & 3.5888e-04 & - - \\
				& + / = / - & +          & +          & +          & +          & +          & +          & +          & +          & +          & +          & +          & +          & - - \\
				\midrule
				
				\multirow{5}{*}{F3} 
				& min       & 6.5963e+00 & 1.4100e+00 & 1.0002e+00 & 1.0000e+00 & 1.4091e+00 & 1.4162e+00 & 1.0000e+00 & 2.7786e+00 & 1.6766e+00 & 1.5474e+00 & 3.7991e+00 & 2.8686e+00 & 1.4091e+00 \\
				& mean      & 8.6758e+00 & 4.5628e+00 & 5.9138e+00 & \textbf{1.9852e+00} & 2.7798e+00 & 6.1866e+00 & 2.0717e+00 & 5.3377e+00 & 4.1093e+00 & 6.0344e+00 & 6.4832e+00 & 4.4821e+00 & 2.1769e+00 \\
				& std       & 8.5435e-01 & 2.5590e+00 & 2.6652e+00 & 1.5095e+00 & 2.5387e+00 & 2.3685e+00 & 1.2626e+00 & 1.7851e+00 & 1.1291e+00 & 3.0358e+00 & 1.2159e+00 & 8.6505e-01 & 6.9735e-01 \\
				& p-value   & 1.7344e-06 & 1.2506e-04 & 1.3601e-05 & 2.7029e-02 & 3.7094e-01 & 2.8786e-06 & 2.0589e-01 & 1.7344e-06 & 3.8822e-06 & 3.1817e-06 & 1.7344e-06 & 1.7344e-06 & - - \\
				& + / = / - & +          & +          & +          & -          & =          & +          & =          & +          & +          & +          & +          & +          & - - \\
				\midrule
				
				\multirow{5}{*}{F4} 
				& min       & 9.8380e+00 & 1.6919e+01 & 1.4929e+01 & 8.9597e+00 & 8.9597e+00 & 5.0059e+00 & 1.7510e+01 & 2.8904e+01 & 2.3215e+01 & 2.2269e+01 & 2.5321e+01 & 1.0736e+01 & 2.9899e+00 \\
				& mean      & 2.2941e+01 & 3.9140e+01 & 4.0851e+01 & 3.2781e+01 & 3.0771e+01 & 1.6503e+01 & 3.0864e+01 & 7.1675e+01 & 3.7062e+01 & 4.2710e+01 & 5.0760e+01 & 1.5684e+01 & \textbf{5.7986e+00} \\
				& std       & 5.7084e+00 & 1.5745e+01 & 1.3429e+01 & 1.0342e+01 & 1.1929e+01 & 4.6175e+00 & 1.0435e+01 & 2.3069e+01 & 8.8024e+00 & 1.3205e+01 & 1.4547e+01 & 3.0569e+00 & 1.7583e+00 \\
				& p-value   & 1.9209e-06 & 1.7344e-06 & 1.7344e-06 & 1.7344e-06 & 1.7333e-06 & 1.7344e-06 & 1.7344e-06 & 1.7344e-06 & 1.7344e-06 & 1.7344e-06 & 1.7344e-06 & 1.7344e-06 & - - \\
				& + / = / - & +          & +          & +          & +          & +          & +          & +          & +          & +          & +          & +          & +          & - - \\
				\midrule
				
				\multirow{5}{*}{F5} 
				& min       & 1.4032e+00 & 1.1057e+00 & 1.1525e+00 & 1.0552e+00 & 1.0295e+00 & 1.1180e+00 & 1.4403e+00 & 1.3540e+00 & 2.0545e+00 & 2.0103e+00 & 2.4069e+00 & 1.1977e+00 & 1.0000e+00 \\
				& mean      & 1.6460e+00 & 1.5792e+00 & 1.7739e+00 & 1.4983e+00 & 1.1497e+00 & 1.2640e+00 & 1.8482e+00 & 1.8689e+00 & 4.8661e+00 & 1.4066e+01 & 3.5844e+00 & 1.3585e+00 & \textbf{1.0365e+00} \\
				& std       & 9.2751e-02 & 4.4980e-01 & 5.5284e-01 & 4.4228e-01 & 8.0237e-02 & 1.2464e-01 & 1.6910e-01 & 4.1383e-01 & 5.3292e+00 & 1.5741e+01 & 7.9199e-01 & 6.0109e-02 & 2.6088e-02 \\
				& p-value   & 1.7344e-06 & 1.7344e-06 & 1.7344e-06 & 1.7344e-06 & 7.6909e-06 & 1.7344e-06 & 1.7344e-06 & 1.7344e-06 & 1.7344e-06 & 1.7344e-06 & 1.7344e-06 & 1.7344e-06 & - - \\
				& + / = / - & +          & +          & +          & +          & +          & +          & +          & +          & +          & +          & +          & +          & - - \\
				\midrule
				
				\multirow{5}{*}{F6} 
				& min       & 4.7782e+00 & 3.7084e+00 & 3.5018e+00 & 3.4622e+00 & 2.4816e+00 & 1.3035e+00 & 3.6735e+00 & 7.6528e+00 & 2.6906e+00 & 3.1944e+00 & 5.1522e+00 & 1.7105e+00 & 1.0000e+00 \\
				& mean      & 6.4352e+00 & 7.2800e+00 & 7.4545e+00 & 5.8314e+00 & 5.7935e+00 & 3.1783e+00 & 5.1803e+00 & 1.1352e+01 & 5.8161e+00 & 6.1021e+00 & 7.3982e+00 & 2.4178e+00 & \textbf{1.2043e+00} \\
				& std       & 8.7819e-01 & 1.5601e+00 & 1.7938e+00 & 1.0170e+00 & 1.7403e+00 & 1.4155e+00 & 9.5568e-01 & 1.8262e+00 & 1.8218e+00 & 1.7531e+00 & 1.2371e+00 & 5.2818e-01 & 4.5149e-01 \\
				& p-value   & 1.7344e-06 & 1.7344e-06 & 1.7344e-06 & 1.7344e-06 & 1.7344e-06 & 1.0246e-05 & 1.7344e-06 & 1.7344e-06 & 1.7344e-06 & 1.7344e-06 & 1.7344e-06 & 6.3391e-06 & - - \\
				& + / = / - & +          & +          & +          & +          & +          & +          & +          & +          & +          & +          & +          & +          & - - \\
				\midrule
				
				\multirow{5}{*}{F7} 
				& min       & 6.2756e+02 & 8.1829e+02 & 5.8622e+02 & 2.0747e+02 & 4.7079e+02 & 4.5537e+00 & 5.4535e+02 & 7.1183e+02 & 4.6776e+02 & 6.5061e+02 & 1.1657e+03 & 3.4624e+02 & 1.0625e+00 \\
				& mean      & 1.0406e+03 & 1.3242e+03 & 1.1358e+03 & 7.3457e+02 & 9.5141e+02 & 6.4183e+02 & 1.0449e+03 & 1.4013e+03 & 1.2522e+03 & 1.1935e+03 & 1.6457e+03 & 6.9658e+02 & \textbf{2.2845e+02} \\
				& std       & 1.9456e+02 & 3.2799e+02 & 2.7264e+02 & 2.5093e+02 & 2.8700e+02 & 2.9822e+02 & 1.8033e+02 & 3.7519e+02 & 3.3269e+02 & 3.1287e+02 & 2.6388e+02 & 1.5966e+02 & 1.6661e+02 \\
				& p-value   & 1.7344e-06 & 1.7344e-06 & 1.7344e-06 & 2.1266e-06 & 1.9209e-06 & 1.4936e-05 & 1.7344e-06 & 1.7344e-06 & 1.7344e-06 & 1.7344e-06 & 1.7344e-06 & 1.7344e-06 & - - \\
				& + / = / - & +          & +          & +          & +          & +          & +          & +          & +          & +          & +          & +          & +          & - - \\
				\midrule
				
				\multirow{5}{*}{F8} 
				& min       & 3.9347e+00 & 4.1216e+00 & 3.8094e+00 & 3.1100e+00 & 2.8345e+00 & 3.1271e+00 & 3.8067e+00 & 4.3668e+00 & 3.7293e+00 & 3.5949e+00 & 4.0215e+00 & 2.7922e+00 & 1.6707e+00 \\
				& mean      & 4.4634e+00 & 4.6026e+00 & 4.6170e+00 & 3.8524e+00 & 3.8046e+00 & 3.9320e+00 & 4.3594e+00 & 5.0183e+00 & 4.4421e+00 & 4.6315e+00 & 4.6224e+00 & 3.7013e+00 & \textbf{2.8184e+00} \\
				& std       & 2.2307e-01 & 2.7049e-01 & 3.6622e-01 & 4.2082e-01 & 4.3913e-01 & 4.4808e-01 & 2.2553e-01 & 2.0407e-01 & 3.5224e-01 & 3.7899e-01 & 2.2321e-01 & 2.9685e-01 & 5.1828e-01 \\
				& p-value   & 1.7344e-06 & 1.7344e-06 & 1.7344e-06 & 1.7344e-06 & 2.3534e-06 & 1.7344e-06 & 1.7344e-06 & 1.7344e-06 & 1.7344e-06 & 1.7344e-06 & 1.7344e-06 & 3.5152e-06 & - - \\
				& + / = / - & +          & +          & +          & +          & +          & +          & +          & +          & +          & +          & +          & +          & - - \\
				\midrule
				
				\multirow{5}{*}{F9} 
				& min       & 1.1786e+00 & 1.1569e+00 & 1.1330e+00 & 1.1832e+00 & 1.0348e+00 & 1.1129e+00 & 1.1495e+00 & 1.1831e+00 & 1.2106e+00 & 1.1863e+00 & 1.2729e+00 & 1.2440e+00 & 1.0364e+00 \\
				& mean      & 1.3136e+00 & 1.3547e+00 & 1.5722e+00 & 1.3201e+00 & 1.1662e+00 & 1.2671e+00 & 1.3072e+00 & 1.4528e+00 & 1.3678e+00 & 1.4570e+00 & 1.5225e+00 & 1.3321e+00 & \textbf{1.1168e+00} \\
				& std       & 6.5456e-02 & 1.5138e-01 & 3.2884e-01 & 8.3885e-02 & 1.0919e-01 & 9.6690e-02 & 1.0411e-01 & 1.7276e-01 & 1.2882e-01 & 3.4331e-01 & 1.2991e-01 & 4.5428e-02 & 5.6291e-02 \\
				& p-value   & 2.3534e-06 & 1.7344e-06 & 1.7344e-06 & 2.3534e-06 & 5.9836e-02 & 1.7344e-06 & 2.8786e-06 & 1.7344e-06 & 1.9209e-06 & 1.7344e-06 & 1.7344e-06 & 1.7344e-06 & - - \\
				& + / = / - & +          & +          & +          & +          & =          & +          & +          & +          & +          & +          & +          & +          & - - \\
				\midrule
				
				\multirow{5}{*}{F10} 
				& min       & 2.1236e+01 & 2.0991e+01 & 2.0970e+01 & 2.1000e+01 & 2.1000e+01 & 2.1007e+01 & 1.9113e+01 & 2.1065e+01 & 2.1213e+01 & 2.1226e+01 & 2.1051e+01 & 4.2901e+00 & 1.0000e+00 \\
				& mean      & 2.1421e+01 & 2.1092e+01 & 2.1015e+01 & 2.1331e+01 & 2.1019e+01 & 2.1057e+01 & 2.1038e+01 & 2.1253e+01 & 2.1421e+01 & 2.1458e+01 & 2.1218e+01 & 2.0003e+01 & \textbf{1.7810e+01} \\
				& std       & 1.0273e-01 & 1.7091e-01 & 4.9009e-02 & 1.5485e-01 & 6.2337e-02 & 4.2139e-02 & 6.9812e-01 & 8.2814e-02 & 1.0865e-01 & 9.3296e-02 & 8.8505e-02 & 3.9287e+00 & 7.3119e+00 \\
				& p-value   & 1.7344e-06 & 2.6230e-01 & 8.4508e-01 & 2.3534e-06 & 8.6121e-01 & 3.5888e-04 & 4.3896e-03 & 1.7344e-06 & 1.7344e-06 & 1.7344e-06 & 1.7344e-06 & 1.8910e-04 & - - \\
				& + / = / - & +          & =          & =          & +          & =          & +          & +          & +          & +          & +          & +          & +          & - - \\
				\midrule
				
				Friedman & Mean rank & 7.9000 & 8.6000 & 9.7000 & 5.1000 & 4.6000 & 5.7000 & 5.2000 & 9.5500 & 8.2500 & 9.5500 & 11.6000 & 3.4500 & \textbf{1.8000} \\
				\midrule
				ToL Wilcoxon & + / = / - & 10/0/0 & 9/1/0 & 9/1/0 & 8/0/2 & 7/3/0 & 10/0/0 & 8/1/1 & 9/0/1 & 9/0/1 & 9/0/1 & 10/0/0 & 9/0/1 & - - \\
				\bottomrule
			\end{tabular}
		\end{adjustbox}
	\end{sidewaystable*}

	\begin{sidewaystable*}[p]
		\centering
		
		\setlength{\abovecaptionskip}{3pt}
		\setlength{\belowcaptionskip}{3pt}
		
		\caption{Test results for CEC 2022 ($D=20$) -- Basic Group}
		\label{tab:cec2022_results_basic}
		
		\scriptsize
		\setlength{\tabcolsep}{3.4pt}
		\renewcommand{\arraystretch}{0.74}
		
		\begin{adjustbox}{
				width=0.98\textheight,
				max totalheight=0.88\textwidth,
				keepaspectratio,
				center
			}
			\begin{tabular}{c c ccccccccccccc}
				\toprule
				Function & & YDSE & EMWPA & SRA & RUN & FLA & RIME & FATA & WAA & ETO & SCHO & FTO & WOO & IWOO \\
				\midrule
				
				\multirow{5}{*}{F1}
				& min       & 1.4639e+03 & 1.5636e+03 & 2.3582e+03 & 3.0034e+02 & 3.0000e+02 & 4.6409e+02 & 3.3414e+02 & 2.1997e+04 & 8.7254e+03 & 9.3096e+03 & 2.5304e+04 & 2.6163e+03 & 3.2994e+02 \\
				& mean      & 4.4733e+03 & 1.5954e+04 & 1.1301e+04 & 3.0423e+02 & \textbf{3.0000e+02} & 1.2062e+03 & 1.8520e+03 & 7.0689e+04 & 1.6699e+04 & 1.9407e+04 & 5.3567e+04 & 6.2188e+03 & 9.8987e+02 \\
				& std       & 1.8459e+03 & 3.7182e+04 & 8.2099e+03 & 4.9466e+00 & 0.0000e+00 & 4.9683e+02 & 2.5819e+03 & 2.7426e+04 & 5.0329e+03 & 7.1803e+03 & 1.5273e+04 & 3.1658e+03 & 5.0559e+02 \\
				& p-value   & 1.7344e-06 & 1.9209e-06 & 1.7344e-06 & 1.7344e-06 & 1.7344e-06 & 5.7096e-02 & 4.6528e-01 & 1.7344e-06 & 1.7344e-06 & 1.7344e-06 & 1.7344e-06 & 1.7344e-06 & -- \\
				& $+ / = / -$ & + & + & + & - & - & = & = & + & + & + & + & + & -- \\
				\midrule
				
				\multirow{5}{*}{F2}
				& min       & 4.5524e+02 & 4.4959e+02 & 4.4491e+02 & 4.3175e+02 & 4.0000e+02 & 4.4494e+02 & 4.4924e+02 & 4.5042e+02 & 4.8992e+02 & 4.7912e+02 & 5.0373e+02 & 4.5504e+02 & 4.0001e+02 \\
				& mean      & 4.6967e+02 & 4.7139e+02 & 4.7611e+02 & 4.5620e+02 & \textbf{4.3630e+02} & 4.5673e+02 & 4.7276e+02 & 5.0837e+02 & 5.8676e+02 & 6.3733e+02 & 6.0012e+02 & 4.7207e+02 & 4.4950e+02 \\
				& std       & 1.4384e+01 & 1.6187e+01 & 3.3907e+01 & 1.1503e+01 & 2.3476e+01 & 1.7076e+01 & 2.3754e+01 & 2.8867e+01 & 8.1560e+01 & 1.1898e+02 & 5.8050e+01 & 1.0281e+01 & 1.1262e+01 \\
				& p-value   & 1.1265e-05 & 3.1817e-06 & 6.8923e-05 & 1.5927e-03 & 4.8603e-05 & 2.0671e-02 & 1.2381e-05 & 1.7344e-06 & 1.7344e-06 & 1.7344e-06 & 1.7344e-06 & 2.1266e-06 & -- \\
				& $+ / = / -$ & + & + & + & + & - & + & + & + & + & + & + & + & -- \\
				\midrule
				
				\multirow{5}{*}{F3}
				& min       & 6.1602e+02 & 6.3297e+02 & 6.1814e+02 & 6.1712e+02 & 6.0956e+02 & 6.0065e+02 & 6.2293e+02 & 6.5523e+02 & 6.2690e+02 & 6.1799e+02 & 6.1689e+02 & 6.0258e+02 & 6.0000e+02 \\
				& mean      & 6.2702e+02 & 6.4984e+02 & 6.3388e+02 & 6.3419e+02 & 6.2751e+02 & 6.0549e+02 & 6.4167e+02 & 6.7171e+02 & 6.4508e+02 & 6.4912e+02 & 6.3746e+02 & 6.0388e+02 & \textbf{6.0000e+02} \\
				& std       & 7.2666e+00 & 1.0355e+01 & 1.1034e+01 & 1.1640e+01 & 9.7502e+00 & 4.4204e+00 & 7.4931e+00 & 7.0993e+00 & 1.1362e+01 & 1.4995e+01 & 1.4126e+01 & 1.1253e+00 & 1.6905e-05 \\
				& p-value   & 1.7344e-06 & 1.7344e-06 & 1.7344e-06 & 1.7344e-06 & 1.7344e-06 & 1.7344e-06 & 1.7344e-06 & 1.7344e-06 & 1.7344e-06 & 1.7344e-06 & 1.7344e-06 & 1.7344e-06 & -- \\
				& $+ / = / -$ & + & + & + & + & + & + & + & + & + & + & + & + & -- \\
				\midrule
				
				\multirow{5}{*}{F4}
				& min       & 8.5099e+02 & 8.4976e+02 & 8.5373e+02 & 8.3980e+02 & 8.4776e+02 & 8.3075e+02 & 8.7198e+02 & 8.8402e+02 & 8.9298e+02 & 8.5406e+02 & 8.8561e+02 & 8.2776e+02 & 8.0895e+02 \\
				& mean      & 8.9813e+02 & 8.8163e+02 & 9.0701e+02 & 8.7349e+02 & 8.7475e+02 & 8.5761e+02 & 9.0020e+02 & 9.0079e+02 & 9.2781e+02 & 9.3058e+02 & 9.4260e+02 & 8.5971e+02 & \textbf{8.2486e+02} \\
				& std       & 1.6074e+01 & 1.7791e+01 & 3.3049e+01 & 1.3066e+01 & 2.0472e+01 & 1.2889e+01 & 1.2507e+01 & 2.0340e+01 & 2.2742e+01 & 3.7750e+01 & 2.9638e+01 & 1.2501e+01 & 8.0409e+00 \\
				& p-value   & 1.7344e-06 & 1.7344e-06 & 1.7344e-06 & 1.7344e-06 & 1.7344e-06 & 1.7344e-06 & 1.7344e-06 & 1.7344e-06 & 1.7344e-06 & 1.7344e-06 & 1.7344e-06 & 1.7344e-06 & -- \\
				& $+ / = / -$ & + & + & + & + & + & + & + & + & + & + & + & + & -- \\
				\midrule
				
				\multirow{5}{*}{F5}
				& min       & 1.0148e+03 & 1.4991e+03 & 1.7085e+03 & 1.3206e+03 & 1.0680e+03 & 9.0545e+02 & 1.3883e+03 & 2.6135e+03 & 1.4956e+03 & 2.0898e+03 & 1.6259e+03 & 9.6695e+02 & 9.0000e+02 \\
				& mean      & 1.6787e+03 & 2.2703e+03 & 3.4115e+03 & 1.9981e+03 & 1.5250e+03 & 1.1694e+03 & 2.2690e+03 & 2.9311e+03 & 3.0074e+03 & 3.6138e+03 & 2.7874e+03 & 1.0545e+03 & \textbf{9.0103e+02} \\
				& std       & 3.7571e+02 & 5.7114e+02 & 1.0993e+03 & 2.9208e+02 & 4.1119e+02 & 4.3552e+02 & 3.5086e+02 & 2.9288e+02 & 8.5481e+02 & 7.0665e+02 & 7.2729e+02 & 7.1918e+01 & 1.3175e+00 \\
				& p-value   & 1.7344e-06 & 1.7344e-06 & 1.7344e-06 & 1.7344e-06 & 1.7344e-06 & 1.7344e-06 & 1.7344e-06 & 1.7344e-06 & 1.7344e-06 & 1.7344e-06 & 1.7344e-06 & 1.7344e-06 & -- \\
				& $+ / = / -$ & + & + & + & + & + & + & + & + & + & + & + & + & -- \\
				\midrule
				
				\multirow{5}{*}{F6}
				& min       & 5.1608e+03 & 1.8987e+03 & 1.9640e+03 & 1.9662e+03 & 1.8728e+03 & 3.2280e+03 & 1.5359e+04 & 1.2401e+04 & 5.7064e+05 & 2.3104e+05 & 1.6389e+06 & 6.7296e+04 & 1.8741e+03 \\
				& mean      & 1.5791e+04 & 5.1694e+03 & 7.8645e+03 & \textbf{3.5122e+03} & 1.2060e+04 & 1.2784e+04 & 1.1015e+05 & 2.8244e+04 & 1.0280e+07 & 1.5302e+07 & 1.7033e+07 & 2.4874e+05 & 3.7889e+03 \\
				& std       & 1.1286e+04 & 3.9886e+03 & 4.8243e+03 & 9.4946e+02 & 8.1629e+03 & 8.8479e+03 & 9.8867e+04 & 1.1621e+04 & 1.2827e+07 & 3.1660e+07 & 1.4523e+07 & 1.2546e+05 & 2.7767e+03 \\
				& p-value   & 3.1817e-06 & 1.5286e-01 & 9.6266e-04 & 4.5281e-01 & 1.2506e-04 & 5.7517e-06 & 1.7344e-06 & 1.7344e-06 & 1.7344e-06 & 1.7344e-06 & 1.7344e-06 & 1.7344e-06 & -- \\
				& $+ / = / -$ & + & = & + & = & + & + & + & + & + & + & + & + & -- \\
				\midrule
				
				\multirow{5}{*}{F7}
				& min       & 2.0840e+03 & 2.0520e+03 & 2.0459e+03 & 2.0337e+03 & 2.0501e+03 & 2.0290e+03 & 2.0692e+03 & 2.1195e+03 & 2.0593e+03 & 2.0873e+03 & 2.0965e+03 & 2.0414e+03 & 2.0125e+03 \\
				& mean      & 2.1205e+03 & 2.1835e+03 & 2.1544e+03 & 2.1027e+03 & 2.1307e+03 & 2.0886e+03 & 2.1080e+03 & 2.3048e+03 & 2.1712e+03 & 2.2048e+03 & 2.1768e+03 & 2.0581e+03 & \textbf{2.0306e+03} \\
				& std       & 2.7217e+01 & 8.8252e+01 & 7.8304e+01 & 2.5731e+01 & 6.3325e+01 & 4.5174e+01 & 2.5301e+01 & 1.0748e+02 & 6.2052e+01 & 6.4578e+01 & 5.6934e+01 & 8.4688e+00 & 6.6860e+00 \\
				& p-value   & 1.7344e-06 & 1.7344e-06 & 1.7344e-06 & 1.7344e-06 & 1.7344e-06 & 2.3534e-06 & 1.7344e-06 & 1.7344e-06 & 1.7344e-06 & 1.7344e-06 & 1.7344e-06 & 1.7344e-06 & -- \\
				& $+ / = / -$ & + & + & + & + & + & + & + & + & + & + & + & + & -- \\
				\midrule
				
				\multirow{5}{*}{F8}
				& min       & 2.2319e+03 & 2.2249e+03 & 2.2238e+03 & 2.2241e+03 & 2.2212e+03 & 2.2252e+03 & 2.2234e+03 & 2.2297e+03 & 2.2337e+03 & 2.2288e+03 & 2.2431e+03 & 2.2264e+03 & 2.2226e+03 \\
				& mean      & 2.2386e+03 & 2.3393e+03 & 2.3503e+03 & 2.2322e+03 & 2.2808e+03 & 2.2456e+03 & 2.2532e+03 & 2.4378e+03 & 2.2941e+03 & 2.3191e+03 & 2.2660e+03 & 2.2301e+03 & \textbf{2.2249e+03} \\
				& std       & 4.3625e+00 & 1.1766e+02 & 1.1277e+02 & 2.1603e+01 & 6.9836e+01 & 3.8449e+01 & 4.8159e+01 & 1.9568e+02 & 6.1259e+01 & 8.5924e+01 & 1.8989e+01 & 2.1480e+00 & 1.2439e+00 \\
				& p-value   & 1.7344e-06 & 2.3534e-06 & 1.7344e-06 & 1.3601e-05 & 2.8308e-04 & 2.6033e-06 & 9.3157e-06 & 1.7344e-06 & 1.7344e-06 & 1.7344e-06 & 1.7344e-06 & 1.7344e-06 & -- \\
				& $+ / = / -$ & + & + & + & + & + & + & + & + & + & + & + & + & -- \\
				\midrule
				
				\multirow{5}{*}{F9}
				& min       & 2.4830e+03 & 2.4808e+03 & 2.4808e+03 & 2.4808e+03 & 2.4807e+03 & 2.4808e+03 & 2.4809e+03 & 2.4963e+03 & 2.5045e+03 & 2.5206e+03 & 2.4924e+03 & 2.4839e+03 & 2.4800e+03 \\
				& mean      & 2.4866e+03 & 2.4825e+03 & 2.4811e+03 & 2.4813e+03 & 2.4808e+03 & 2.4816e+03 & 2.4819e+03 & 2.5673e+03 & 2.5529e+03 & 2.5852e+03 & 2.5493e+03 & 2.4860e+03 & \textbf{2.4800e+03} \\
				& std       & 1.9913e+00 & 7.3476e+00 & 3.3020e-01 & 3.3772e-01 & 7.9776e-03 & 5.5361e-01 & 2.2102e+00 & 5.2628e+01 & 3.6843e+01 & 4.6729e+01 & 3.9505e+01 & 1.0833e+00 & 2.0562e-05 \\
				& p-value   & 1.7344e-06 & 1.7344e-06 & 1.7344e-06 & 1.7344e-06 & 1.7344e-06 & 1.7344e-06 & 1.7344e-06 & 1.7344e-06 & 1.7344e-06 & 1.7344e-06 & 1.7344e-06 & 1.7344e-06 & -- \\
				& $+ / = / -$ & + & + & + & + & - & + & + & + & + & + & + & + & -- \\
				\midrule
				
				\multirow{5}{*}{F10}
				& min       & 2.5013e+03 & 2.5009e+03 & 2.5005e+03 & 2.5010e+03 & 2.5003e+03 & 2.4325e+03 & 2.5020e+03 & 2.5013e+03 & 2.5016e+03 & 2.5092e+03 & 2.5016e+03 & 2.5005e+03 & 2.4021e+03 \\
				& mean      & 3.4482e+03 & 4.0515e+03 & 3.4109e+03 & 2.7647e+03 & 4.1008e+03 & 2.7045e+03 & 2.5120e+03 & 4.7037e+03 & 4.1085e+03 & 3.9569e+03 & 2.9828e+03 & 2.5241e+03 & \textbf{2.4973e+03} \\
				& std       & 1.3126e+03 & 9.1886e+02 & 5.8420e+02 & 3.7424e+02 & 1.0414e+03 & 2.4887e+02 & 8.2112e+00 & 9.8590e+02 & 1.1752e+03 & 1.0240e+03 & 9.7948e+02 & 7.5085e+01 & 1.7987e+01 \\
				& p-value   & 1.7344e-06 & 1.7344e-06 & 1.7344e-06 & 1.7344e-06 & 6.3391e-06 & 2.3704e-05 & 1.7344e-06 & 1.7344e-06 & 1.7344e-06 & 1.7344e-06 & 1.7344e-06 & 1.6394e-05 & -- \\
				& $+ / = / -$ & + & + & + & + & + & + & + & + & + & + & + & + & -- \\
				\midrule
				
				\multirow{5}{*}{F11}
				& min       & 4.3565e+03 & 2.9342e+03 & 2.9000e+03 & 2.9003e+03 & 2.9000e+03 & 2.9699e+03 & 2.6108e+03 & 2.9417e+03 & 3.5819e+03 & 3.4213e+03 & 5.2919e+03 & 3.2887e+03 & 2.9000e+03 \\
				& mean      & 4.9571e+03 & 3.1540e+03 & 2.9724e+03 & 2.9262e+03 & \textbf{2.9000e+03} & 3.0188e+03 & 2.9247e+03 & 3.6560e+03 & 4.5862e+03 & 5.2342e+03 & 5.5223e+03 & 3.5388e+03 & \textbf{2.9000e+03} \\
				& std       & 2.9887e+02 & 1.9578e+02 & 1.0471e+02 & 7.2682e+01 & 4.2222e-13 & 2.4960e+01 & 1.3201e+02 & 8.1591e+02 & 8.1097e+02 & 9.3427e+02 & 1.1255e+03 & 1.3231e+02 & 4.7017e-13 \\
				& p-value   & 1.7344e-06 & 1.7344e-06 & 1.7344e-06 & 1.7344e-06 & 1.0000e+00 & 1.7344e-06 & 2.2551e-03 & 1.7344e-06 & 1.7344e-06 & 1.7344e-06 & 1.7344e-06 & 1.7344e-06 & -- \\
				& $+ / = / -$ & + & + & + & + & = & + & + & + & + & + & + & + & -- \\
				\midrule
				
				\multirow{5}{*}{F12}
				& min       & 2.9527e+03 & 2.9712e+03 & 2.9555e+03 & 2.9481e+03 & 2.9401e+03 & 2.9431e+03 & 2.9401e+03 & 3.2588e+03 & 3.0198e+03 & 2.9901e+03 & 2.9000e+03 & 2.9666e+03 & 2.9384e+03 \\
				& mean      & 2.9697e+03 & 3.0908e+03 & 3.0587e+03 & 2.9797e+03 & 3.0284e+03 & 2.9711e+03 & 2.9664e+03 & 3.7280e+03 & 3.1517e+03 & 3.1641e+03 & \textbf{2.9000e+03} & 3.0065e+03 & 2.9541e+03 \\
				& std       & 9.6491e+00 & 9.8725e+01 & 6.0526e+01 & 1.5264e+01 & 6.6747e+01 & 2.2738e+01 & 1.8188e+01 & 3.0203e+02 & 1.0140e+02 & 1.0699e+02 & 9.3127e-06 & 2.4985e+01 & 1.0150e+01 \\
				& p-value   & 1.7988e-05 & 1.7344e-06 & 1.7344e-06 & 1.6394e-05 & 5.7517e-06 & 1.4839e-03 & 5.6672e-03 & 1.7344e-06 & 1.7344e-06 & 1.7344e-06 & 1.7344e-06 & 1.9209e-06 & -- \\
				& $+ / = / -$ & + & + & + & + & + & + & + & + & + & + & - & + & -- \\
				\midrule
				Friedman & Mean rank & 6.3333 & 8.3333 & 7.7500 & 4.0833 & 4.7500 & 4.1667 & 5.9167 & 11.3333 & 10.5000 & 11.6667 & 9.5000 & 5.0833 & \textbf{1.5833} \\
				\midrule
				ToL Wilcoxon & $+ / = / -$ & 12/0/0 & 11/1/0 & 12/0/0 & 10/1/1 & 8/1/3 & 11/1/0 & 11/1/0 & 12/0/0 & 12/0/0 & 12/0/0 & 11/0/1 & 12/0/0 & -- \\
				\bottomrule
			\end{tabular}
		\end{adjustbox}
	\end{sidewaystable*}

	\begin{sidewaystable*}[p]
		\centering
		
		\setlength{\abovecaptionskip}{3pt}
		\setlength{\belowcaptionskip}{3pt}
		
		\caption{Test results for CEC 2017 ($D=50$) -- Basic Group}
		\label{tab:cec2017_50d_results_basic}
		
		\scriptsize
		\setlength{\tabcolsep}{2.8pt}
		\renewcommand{\arraystretch}{0.72}
		
		\begin{adjustbox}{
				width=0.98\textheight,
				max totalheight=0.88\textwidth,
				keepaspectratio,
				center
			}
			\begin{tabular}{c c ccccccccccccc}
				\toprule
				Function & & YDSE & EMWPA & SRA & RUN & FLA & RIME & FATA & WAA & ETO & SCHO & FTO & WOO & IWOO \\
				\midrule
				
				\multirow{5}{*}{F1}
				& min       & 8.0258e+09 & 7.3917e+08 & 5.6304e+03 & 1.9224e+05 & 4.9150e+02 & 1.7735e+07 & 2.1503e+09 & 1.4711e+09 & 1.2412e+10 & 1.8905e+10 & 1.2715e+10 & 2.4631e+09 & 1.0002e+02 \\
				& mean      & 1.3539e+10 & 1.7702e+09 & 3.3876e+06 & 1.1712e+06 & 1.0285e+04 & 4.0995e+07 & 4.5647e+09 & 2.8397e+09 & 2.8336e+10 & 3.5074e+10 & 2.0410e+10 & 3.9074e+09 & \textbf{1.7274e+03} \\
				& std       & 3.7882e+09 & 1.1157e+09 & 9.0802e+06 & 9.4095e+05 & 1.0396e+04 & 1.6295e+07 & 9.7490e+08 & 9.3271e+08 & 6.2334e+09 & 9.5101e+09 & 4.3482e+09 & 8.4623e+08 & 2.6363e+03 \\
				& p-value   & 1.7344e-06 & 1.7344e-06 & 1.7344e-06 & 1.7344e-06 & 1.2506e-04 & 1.7344e-06 & 1.7344e-06 & 1.7344e-06 & 1.7344e-06 & 1.7344e-06 & 1.7344e-06 & 1.7344e-06 & -- \\
				& $+ / = / -$ & + & + & + & + & + & + & + & + & + & + & + & + & -- \\
				\midrule
				
				\multirow{5}{*}{F3}
				& min       & 7.8280e+04 & 1.1346e+05 & 2.0043e+05 & 4.6212e+04 & 3.0000e+02 & 1.4584e+05 & 3.4281e+04 & 2.2962e+05 & 1.2834e+05 & 1.5948e+05 & 1.9006e+05 & 6.9179e+04 & 6.4634e+04 \\
				& mean      & 9.9806e+04 & 2.7090e+05 & 3.3529e+05 & 6.5188e+04 & \textbf{3.0000e+02} & 1.9879e+05 & 7.5198e+04 & 2.6241e+05 & 1.7008e+05 & 2.4145e+05 & 3.3663e+05 & 1.1281e+05 & 8.7236e+04 \\
				& std       & 1.5970e+04 & 1.2909e+05 & 8.3433e+04 & 9.2061e+03 & 1.8283e-14 & 3.6715e+04 & 1.7496e+04 & 3.9580e+04 & 1.7659e+04 & 5.2767e+04 & 9.9534e+04 & 1.8884e+04 & 1.2750e+04 \\
				& p-value   & 2.4147e-03 & 1.7344e-06 & 1.7344e-06 & 3.5152e-06 & 1.7344e-06 & 1.7344e-06 & 9.2710e-03 & 1.7344e-06 & 1.7344e-06 & 1.7344e-06 & 1.7344e-06 & 5.3070e-05 & -- \\
				& $+ / = / -$ & + & + & + & - & - & + & - & + & + & + & + & + & -- \\
				\midrule
				
				\multirow{5}{*}{F4}
				& min       & 1.0899e+03 & 7.3790e+02 & 6.0884e+02 & 5.0798e+02 & 4.0000e+02 & 5.4214e+02 & 5.8023e+02 & 9.2487e+02 & 2.1417e+03 & 2.3714e+03 & 1.2092e+03 & 1.0726e+03 & 4.2934e+02 \\
				& mean      & 1.6978e+03 & 9.4334e+02 & 7.1297e+02 & 6.3078e+02 & \textbf{4.4116e+02} & 6.7522e+02 & 6.8516e+02 & 1.2017e+03 & 4.8853e+03 & 6.0671e+03 & 2.2735e+03 & 1.2662e+03 & 5.2829e+02 \\
				& std       & 3.0643e+02 & 1.7025e+02 & 7.8994e+01 & 5.9163e+01 & 4.5352e+01 & 7.2945e+01 & 5.2728e+01 & 1.8443e+02 & 1.5277e+03 & 2.2922e+03 & 5.8922e+02 & 1.3205e+02 & 5.9031e+01 \\
				& p-value   & 1.7344e-06 & 1.7344e-06 & 1.7344e-06 & 1.6394e-05 & 4.8603e-05 & 6.9838e-06 & 1.7344e-06 & 1.7344e-06 & 1.7344e-06 & 1.7344e-06 & 1.7344e-06 & 1.7344e-06 & -- \\
				& $+ / = / -$ & + & + & + & + & - & + & + & + & + & + & + & + & -- \\
				\midrule
				
				\multirow{5}{*}{F5}
				& min       & 8.7349e+02 & 7.7833e+02 & 8.0854e+02 & 7.5783e+02 & 6.7909e+02 & 6.9099e+02 & 9.3663e+02 & 8.6305e+02 & 9.5167e+02 & 8.7270e+02 & 8.5409e+02 & 7.7035e+02 & 5.6169e+02 \\
				& mean      & 9.7322e+02 & 9.0061e+02 & 9.0479e+02 & 8.4446e+02 & 8.3004e+02 & 7.6951e+02 & 9.8336e+02 & 9.1327e+02 & 1.0329e+03 & 1.0244e+03 & 9.5130e+02 & 8.0656e+02 & \textbf{5.8928e+02} \\
				& std       & 3.1974e+01 & 5.5230e+01 & 6.0531e+01 & 3.2164e+01 & 7.2774e+01 & 4.4398e+01 & 2.7500e+01 & 2.5558e+01 & 5.1906e+01 & 7.5261e+01 & 5.8955e+01 & 1.8782e+01 & 1.7086e+01 \\
				& p-value   & 1.7344e-06 & 1.7344e-06 & 1.7344e-06 & 1.7344e-06 & 1.7344e-06 & 1.7344e-06 & 1.7344e-06 & 1.7344e-06 & 1.7344e-06 & 1.7344e-06 & 1.7344e-06 & 1.7344e-06 & -- \\
				& $+ / = / -$ & + & + & + & + & + & + & + & + & + & + & + & + & -- \\
				\midrule
				
				\multirow{5}{*}{F6}
				& min       & 6.5283e+02 & 6.5576e+02 & 6.3311e+02 & 6.4405e+02 & 6.2801e+02 & 6.1742e+02 & 6.6971e+02 & 6.7172e+02 & 6.6055e+02 & 6.5929e+02 & 6.4304e+02 & 6.1063e+02 & 6.0002e+02 \\
				& mean      & 6.6535e+02 & 6.6818e+02 & 6.4590e+02 & 6.5894e+02 & 6.4998e+02 & 6.3197e+02 & 6.7517e+02 & 6.7511e+02 & 6.7885e+02 & 6.7955e+02 & 6.7437e+02 & 6.1606e+02 & \textbf{6.0019e+02} \\
				& std       & 6.4382e+00 & 8.4173e+00 & 7.1028e+00 & 6.1432e+00 & 9.8974e+00 & 1.0155e+01 & 3.3166e+00 & 2.3775e+00 & 9.5702e+00 & 1.0261e+01 & 1.6241e+01 & 2.0950e+00 & 1.5788e-01 \\
				& p-value   & 1.7344e-06 & 1.7344e-06 & 1.7344e-06 & 1.7344e-06 & 1.7344e-06 & 1.7344e-06 & 1.7344e-06 & 1.7344e-06 & 1.7344e-06 & 1.7344e-06 & 1.7344e-06 & 1.7344e-06 & -- \\
				& $+ / = / -$ & + & + & + & + & + & + & + & + & + & + & + & + & -- \\
				\midrule
				
				\multirow{5}{*}{F7}
				& min       & 1.3865e+03 & 1.5287e+03 & 1.8504e+03 & 1.2841e+03 & 1.1704e+03 & 9.9971e+02 & 1.2636e+03 & 1.8253e+03 & 1.4040e+03 & 1.3696e+03 & 1.3034e+03 & 1.2025e+03 & 8.0091e+02 \\
				& mean      & 1.4734e+03 & 1.7517e+03 & 2.3086e+03 & 1.4706e+03 & 1.4838e+03 & 1.0930e+03 & 1.4662e+03 & 1.8989e+03 & 1.5956e+03 & 1.6551e+03 & 1.4422e+03 & 1.2832e+03 & \textbf{8.3478e+02} \\
				& std       & 6.5757e+01 & 1.3359e+02 & 3.0440e+02 & 1.0085e+02 & 1.6281e+02 & 5.4512e+01 & 1.2913e+02 & 2.6717e+01 & 9.6541e+01 & 1.1583e+02 & 8.1245e+01 & 3.7781e+01 & 2.0289e+01 \\
				& p-value   & 1.7344e-06 & 1.7344e-06 & 1.7344e-06 & 1.7344e-06 & 1.7344e-06 & 1.7344e-06 & 1.7344e-06 & 1.7344e-06 & 1.7344e-06 & 1.7344e-06 & 1.7344e-06 & 1.7344e-06 & -- \\
				& $+ / = / -$ & + & + & + & + & + & + & + & + & + & + & + & + & -- \\
				\midrule
				
				\multirow{5}{*}{F8}
				& min       & 1.1771e+03 & 1.0569e+03 & 1.0827e+03 & 1.0956e+03 & 1.0089e+03 & 9.5541e+02 & 1.2618e+03 & 1.1555e+03 & 1.2634e+03 & 1.1834e+03 & 1.1829e+03 & 1.0561e+03 & 8.5373e+02 \\
				& mean      & 1.2642e+03 & 1.2083e+03 & 1.2267e+03 & 1.1455e+03 & 1.1277e+03 & 1.0641e+03 & 1.3063e+03 & 1.2232e+03 & 1.3398e+03 & 1.2972e+03 & 1.2662e+03 & 1.1067e+03 & \textbf{8.8282e+02} \\
				& std       & 3.8513e+01 & 5.4382e+01 & 6.9261e+01 & 2.6311e+01 & 5.6588e+01 & 6.6800e+01 & 2.9522e+01 & 3.4587e+01 & 4.3540e+01 & 7.5690e+01 & 5.7757e+01 & 2.6073e+01 & 1.5138e+01 \\
				& p-value   & 1.7344e-06 & 1.7344e-06 & 1.7344e-06 & 1.7344e-06 & 1.7344e-06 & 1.7344e-06 & 1.7344e-06 & 1.7344e-06 & 1.7344e-06 & 1.7344e-06 & 1.7344e-06 & 1.7344e-06 & -- \\
				& $+ / = / -$ & + & + & + & + & + & + & + & + & + & + & + & + & -- \\
				\midrule
				
				\multirow{5}{*}{F9}
				& min       & 1.2760e+04 & 1.2178e+04 & 1.0328e+04 & 8.3804e+03 & 6.3492e+03 & 3.5508e+03 & 1.8431e+04 & 2.1466e+04 & 2.4263e+04 & 2.8855e+04 & 1.0507e+04 & 3.0861e+03 & 9.4813e+02 \\
				& mean      & 2.0994e+04 & 1.5683e+04 & 1.8286e+04 & 1.1809e+04 & 1.0218e+04 & 1.0579e+04 & 2.6240e+04 & 2.6872e+04 & 3.4958e+04 & 4.3847e+04 & 2.4573e+04 & 4.9761e+03 & \textbf{1.0884e+03} \\
				& std       & 3.9156e+03 & 2.8636e+03 & 5.0333e+03 & 1.4869e+03 & 2.4405e+03 & 4.5983e+03 & 3.1285e+03 & 2.7496e+03 & 4.8953e+03 & 9.4654e+03 & 8.2941e+03 & 1.3169e+03 & 1.3626e+02 \\
				& p-value   & 1.7344e-06 & 1.7344e-06 & 1.7344e-06 & 1.7344e-06 & 1.7344e-06 & 1.7344e-06 & 1.7344e-06 & 1.7344e-06 & 1.7344e-06 & 1.7344e-06 & 1.7344e-06 & 1.7344e-06 & -- \\
				& $+ / = / -$ & + & + & + & + & + & + & + & + & + & + & + & + & -- \\
				\midrule
				
				\multirow{5}{*}{F10}
				& min       & 1.1763e+04 & 7.0212e+03 & 5.4029e+03 & 6.3095e+03 & 6.1467e+03 & 5.8307e+03 & 1.2346e+04 & 8.0661e+03 & 1.1890e+04 & 1.0114e+04 & 1.1426e+04 & 9.6302e+03 & 4.9426e+03 \\
				& mean      & 1.3473e+04 & 9.7824e+03 & 7.9837e+03 & 7.9396e+03 & 8.2117e+03 & 7.9259e+03 & 1.3358e+04 & 9.4610e+03 & 1.3852e+04 & 1.2492e+04 & 1.4003e+04 & 1.0641e+04 & \textbf{6.4364e+03} \\
				& std       & 7.9142e+02 & 1.6651e+03 & 9.9820e+02 & 1.0975e+03 & 1.5641e+03 & 9.3347e+02 & 4.8875e+02 & 5.0045e+02 & 8.9438e+02 & 9.4537e+02 & 1.0422e+03 & 4.6358e+02 & 1.1194e+03 \\
				& p-value   & 1.7344e-06 & 3.1817e-06 & 3.7243e-05 & 6.3198e-05 & 1.3601e-05 & 1.7423e-04 & 1.7344e-06 & 1.9209e-06 & 1.7344e-06 & 1.7344e-06 & 1.7344e-06 & 1.7344e-06 & -- \\
				& $+ / = / -$ & + & + & + & + & + & + & + & + & + & + & + & + & -- \\
				\midrule
				
				\multirow{5}{*}{F11}
				& min       & 2.2266e+03 & 1.7381e+03 & 1.4245e+03 & 1.2883e+03 & 1.3074e+03 & 1.5876e+03 & 1.2730e+03 & 2.3698e+03 & 3.6392e+03 & 3.9463e+03 & 3.4881e+03 & 1.5805e+03 & 1.2116e+03 \\
				& mean      & 3.1394e+03 & 2.2795e+03 & 1.5904e+03 & 1.3719e+03 & 1.4771e+03 & 1.7651e+03 & 1.4023e+03 & 3.5935e+03 & 9.0511e+03 & 1.1523e+04 & 5.5395e+03 & 1.7529e+03 & \textbf{1.2646e+03} \\
				& std       & 6.2449e+02 & 4.8769e+02 & 1.2061e+02 & 6.4944e+01 & 9.9522e+01 & 1.0437e+02 & 1.0545e+02 & 4.9853e+02 & 3.1759e+03 & 3.5530e+03 & 2.2262e+03 & 1.0331e+02 & 3.2788e+01 \\
				& p-value   & 1.7344e-06 & 1.7344e-06 & 1.7344e-06 & 3.1817e-06 & 1.7344e-06 & 1.7344e-06 & 2.1266e-06 & 1.7344e-06 & 1.7344e-06 & 1.7344e-06 & 1.7344e-06 & 1.7344e-06 & -- \\
				& $+ / = / -$ & + & + & + & + & + & + & + & + & + & + & + & + & -- \\
				\bottomrule
			\end{tabular}
		\end{adjustbox}
	\end{sidewaystable*}

	\begin{sidewaystable*}[p]
		\centering
		
		\setlength{\abovecaptionskip}{3pt}
		\setlength{\belowcaptionskip}{3pt}
		
		\addtocounter{table}{-1}
		\caption[]{Test results for CEC 2017 ($D=50$) -- Basic Group (continued)}
		
		\scriptsize
		\setlength{\tabcolsep}{2.8pt}
		\renewcommand{\arraystretch}{0.72}
		
		\begin{adjustbox}{
				width=0.98\textheight,
				max totalheight=0.88\textwidth,
				keepaspectratio,
				center
			}
			\begin{tabular}{c c ccccccccccccc}
				\toprule
				Function & & YDSE & EMWPA & SRA & RUN & FLA & RIME & FATA & WAA & ETO & SCHO & FTO & WOO & IWOO \\
				\midrule
				
				\multirow{5}{*}{F12}
				& min       & 1.8341e+08 & 5.2134e+06 & 5.0178e+06 & 1.2975e+07 & 1.2182e+04 & 3.1151e+07 & 2.4313e+08 & 6.4005e+07 & 1.5618e+09 & 2.1989e+09 & 1.0231e+09 & 1.5120e+08 & 2.8973e+05 \\
				& mean      & 6.1785e+08 & 1.2761e+08 & 2.9228e+07 & 4.4684e+07 & \textbf{6.4233e+04} & 1.4466e+08 & 4.6928e+08 & 2.6371e+08 & 8.3065e+09 & 1.4195e+10 & 2.8176e+09 & 3.2558e+08 & 9.7720e+05 \\
				& std       & 2.6099e+08 & 1.3372e+08 & 1.8656e+07 & 2.6596e+07 & 2.8748e+04 & 9.4906e+07 & 1.6291e+08 & 1.4812e+08 & 4.3025e+09 & 7.6963e+09 & 1.6105e+09 & 9.6369e+07 & 5.8517e+05 \\
				& p-value   & 1.7344e-06 & 1.7344e-06 & 1.7344e-06 & 1.7344e-06 & 1.7344e-06 & 1.7344e-06 & 1.7344e-06 & 1.7344e-06 & 1.7344e-06 & 1.7344e-06 & 1.7344e-06 & 1.7344e-06 & -- \\
				& $+ / = / -$ & + & + & + & + & - & + & + & + & + & + & + & + & -- \\
				\midrule
				
				\multirow{5}{*}{F13}
				& min       & 3.1199e+06 & 3.7771e+04 & 3.6689e+04 & 1.9433e+04 & 3.2491e+03 & 2.4818e+05 & 3.5640e+06 & 5.9315e+05 & 2.1121e+08 & 3.5195e+07 & 7.0394e+07 & 2.7247e+06 & 1.4551e+03 \\
				& mean      & 1.4775e+07 & 2.7061e+05 & 1.1036e+05 & 3.6548e+04 & 2.0543e+04 & 4.8456e+05 & 1.4657e+07 & 1.5606e+06 & 1.1272e+09 & 5.4655e+09 & 2.7890e+08 & 9.9122e+06 & \textbf{3.2793e+03} \\
				& std       & 7.9751e+06 & 4.2627e+05 & 6.0561e+04 & 1.4482e+04 & 1.3331e+04 & 1.8052e+05 & 8.3429e+06 & 1.3041e+06 & 1.9325e+09 & 6.7531e+09 & 2.2295e+08 & 4.9383e+06 & 2.5057e+03 \\
				& p-value   & 1.7344e-06 & 1.7344e-06 & 1.7344e-06 & 1.7344e-06 & 2.6033e-06 & 1.7344e-06 & 1.7344e-06 & 1.7344e-06 & 1.7344e-06 & 1.7344e-06 & 1.7344e-06 & 1.7344e-06 & -- \\
				& $+ / = / -$ & + & + & + & + & + & + & + & + & + & + & + & + & -- \\
				\midrule
				
				\multirow{5}{*}{F14}
				& min       & 3.0000e+03 & 2.7016e+04 & 1.7070e+04 & 1.4096e+04 & 1.8273e+03 & 3.0616e+04 & 4.0282e+04 & 2.4620e+05 & 3.6733e+05 & 2.9756e+05 & 4.1763e+05 & 4.3799e+03 & 3.3159e+03 \\
				& mean      & 5.9402e+03 & 9.9906e+05 & 6.5609e+05 & 9.3047e+04 & \textbf{3.0841e+03} & 7.3046e+05 & 2.1523e+05 & 2.9462e+06 & 3.0277e+06 & 4.9699e+06 & 1.9738e+06 & 5.6266e+04 & 3.2732e+04 \\
				& std       & 2.6031e+03 & 3.0890e+06 & 6.0876e+05 & 6.7914e+04 & 8.2795e+02 & 5.4197e+05 & 1.3765e+05 & 2.2585e+06 & 2.0699e+06 & 4.5864e+06 & 1.9318e+06 & 3.7193e+04 & 3.0660e+04 \\
				& p-value   & 8.4661e-06 & 4.7292e-06 & 1.7344e-06 & 2.2248e-04 & 1.7344e-06 & 1.9209e-06 & 4.2857e-06 & 1.7344e-06 & 1.7344e-06 & 1.7344e-06 & 1.7344e-06 & 2.4308e-02 & -- \\
				& $+ / = / -$ & - & + & + & + & - & + & + & + & + & + & + & + & -- \\
				\midrule
				
				\multirow{5}{*}{F15}
				& min       & 1.1900e+05 & 4.1069e+03 & 1.9025e+04 & 1.1991e+04 & 2.4079e+03 & 4.1300e+04 & 1.0360e+05 & 5.8359e+04 & 1.1449e+07 & 1.4668e+06 & 8.3413e+05 & 1.6764e+05 & 1.8875e+03 \\
				& mean      & 3.2333e+05 & 1.6889e+04 & 6.4270e+04 & 2.3258e+04 & 1.4586e+04 & 1.1745e+05 & 3.4242e+05 & 1.8623e+05 & 1.2493e+08 & 6.9222e+08 & 1.9912e+07 & 4.9154e+05 & \textbf{6.6601e+03} \\
				& std       & 1.5933e+05 & 9.7986e+03 & 3.4328e+04 & 4.2676e+03 & 9.3994e+03 & 5.1190e+04 & 2.5681e+05 & 8.9514e+04 & 2.4489e+08 & 1.0829e+09 & 3.5705e+07 & 2.1511e+05 & 3.3723e+03 \\
				& p-value   & 1.7344e-06 & 4.8603e-05 & 1.7344e-06 & 1.7344e-06 & 4.1955e-04 & 1.7344e-06 & 1.7344e-06 & 1.7344e-06 & 1.7344e-06 & 1.7344e-06 & 1.7344e-06 & 1.7344e-06 & -- \\
				& $+ / = / -$ & + & + & + & + & + & + & + & + & + & + & + & + & -- \\
				\midrule
				
				\multirow{5}{*}{F16}
				& min       & 4.0403e+03 & 2.3847e+03 & 2.1748e+03 & 2.5361e+03 & 2.6015e+03 & 2.9473e+03 & 2.5706e+03 & 2.7631e+03 & 3.5476e+03 & 3.6224e+03 & 3.4423e+03 & 3.5818e+03 & 1.9757e+03 \\
				& mean      & 4.6318e+03 & 3.6522e+03 & 3.9216e+03 & 3.4425e+03 & 3.5925e+03 & 3.7937e+03 & 3.2140e+03 & 5.2462e+03 & 4.9934e+03 & 4.6688e+03 & 5.2516e+03 & 4.0388e+03 & \textbf{2.8025e+03} \\
				& std       & 3.2535e+02 & 4.6671e+02 & 6.8300e+02 & 4.5612e+02 & 5.3753e+02 & 4.4423e+02 & 3.7083e+02 & 7.6652e+02 & 6.7535e+02 & 6.7613e+02 & 6.8979e+02 & 2.1440e+02 & 3.5950e+02 \\
				& p-value   & 1.7344e-06 & 1.1265e-05 & 6.3391e-06 & 1.0246e-05 & 3.4053e-05 & 2.8786e-06 & 1.7423e-04 & 1.9209e-06 & 1.7344e-06 & 1.7344e-06 & 1.7344e-06 & 1.7344e-06 & -- \\
				& $+ / = / -$ & + & + & + & + & + & + & + & + & + & + & + & + & -- \\
				\midrule
				
				\multirow{5}{*}{F17}
				& min       & 3.0132e+03 & 2.5706e+03 & 3.0415e+03 & 2.6788e+03 & 2.8921e+03 & 2.6203e+03 & 2.2225e+03 & 3.0470e+03 & 3.3508e+03 & 3.0074e+03 & 3.0311e+03 & 3.1331e+03 & 2.1745e+03 \\
				& mean      & 3.6262e+03 & 3.6767e+03 & 3.8646e+03 & 3.2701e+03 & 3.6986e+03 & 3.4012e+03 & 3.1159e+03 & 3.7608e+03 & 4.0107e+03 & 3.7492e+03 & 4.0797e+03 & 3.7448e+03 & \textbf{2.6682e+03} \\
				& std       & 2.9534e+02 & 4.4009e+02 & 4.1270e+02 & 3.3842e+02 & 5.6900e+02 & 3.7748e+02 & 4.2619e+02 & 3.6243e+02 & 4.6789e+02 & 4.2750e+02 & 4.6720e+02 & 3.2141e+02 & 2.4282e+02 \\
				& p-value   & 1.7344e-06 & 2.8786e-06 & 1.7344e-06 & 4.2857e-06 & 2.1266e-06 & 2.3534e-06 & 1.7988e-05 & 1.7344e-06 & 1.7344e-06 & 1.7344e-06 & 1.7344e-06 & 1.7344e-06 & -- \\
				& $+ / = / -$ & + & + & + & + & + & + & + & + & + & + & + & + & -- \\
				\midrule
				
				\multirow{5}{*}{F18}
				& min       & 5.6019e+04 & 2.1811e+05 & 2.4672e+05 & 1.2349e+05 & 3.5747e+03 & 7.7770e+05 & 5.7198e+05 & 7.3578e+05 & 2.8969e+06 & 1.0510e+06 & 2.7506e+06 & 1.3469e+05 & 7.3185e+04 \\
				& mean      & 1.5631e+05 & 1.9060e+06 & 2.2594e+06 & 7.3007e+05 & \textbf{1.2837e+04} & 5.3723e+06 & 1.9705e+06 & 9.5538e+06 & 1.2076e+07 & 3.0796e+07 & 1.4838e+07 & 4.4874e+05 & 3.6090e+05 \\
				& std       & 9.3548e+04 & 2.0256e+06 & 1.8070e+06 & 3.4628e+05 & 6.6165e+03 & 3.6949e+06 & 1.0213e+06 & 5.8639e+06 & 1.1330e+07 & 3.2017e+07 & 1.2682e+07 & 3.8079e+05 & 4.8685e+05 \\
				& p-value   & 1.9646e-03 & 2.1266e-06 & 1.7344e-06 & 2.2248e-04 & 1.7344e-06 & 1.9209e-06 & 2.6033e-06 & 1.7344e-06 & 1.7344e-06 & 1.7344e-06 & 1.7344e-06 & 6.2683e-02 & -- \\
				& $+ / = / -$ & - & + & + & + & - & + & + & + & + & + & + & = & -- \\
				\midrule
				
				\multirow{5}{*}{F19}
				& min       & 1.1221e+05 & 8.9111e+03 & 5.4713e+03 & 6.7220e+04 & 2.2380e+03 & 3.1593e+04 & 9.0842e+04 & 6.8656e+04 & 1.1719e+07 & 2.6122e+06 & 5.7141e+05 & 1.5095e+05 & 7.5298e+03 \\
				& mean      & 3.4342e+05 & 2.7306e+04 & 3.4405e+04 & 2.1201e+05 & 2.4180e+04 & 4.0440e+05 & 4.6709e+05 & 2.6375e+05 & 1.6018e+08 & 3.2666e+08 & 4.7362e+06 & 3.5408e+05 & \textbf{1.5851e+04} \\
				& std       & 1.8174e+05 & 2.9053e+04 & 2.1394e+04 & 9.2051e+04 & 1.4666e+04 & 3.3083e+05 & 2.5509e+05 & 3.5296e+05 & 4.2289e+08 & 5.2147e+08 & 5.1893e+06 & 1.3850e+05 & 5.4240e+03 \\
				& p-value   & 1.7344e-06 & 2.5637e-02 & 1.8910e-04 & 1.7344e-06 & 9.2710e-03 & 1.7344e-06 & 1.7344e-06 & 1.7344e-06 & 1.7344e-06 & 1.7344e-06 & 1.7344e-06 & 1.7344e-06 & -- \\
				& $+ / = / -$ & + & + & + & + & + & + & + & + & + & + & + & + & -- \\
				\midrule
				
				\multirow{5}{*}{F20}
				& min       & 3.0150e+03 & 2.6915e+03 & 3.0945e+03 & 2.6312e+03 & 2.6018e+03 & 2.7974e+03 & 2.9574e+03 & 3.5626e+03 & 3.2353e+03 & 2.8730e+03 & 3.1299e+03 & 2.8765e+03 & 2.3017e+03 \\
				& mean      & 3.7830e+03 & 3.4952e+03 & 3.6130e+03 & 3.1428e+03 & 3.2827e+03 & 3.3513e+03 & 3.3521e+03 & 3.8369e+03 & 3.7845e+03 & 3.7101e+03 & 4.1767e+03 & 3.3012e+03 & \textbf{2.8064e+03} \\
				& std       & 2.5978e+02 & 3.7153e+02 & 3.2682e+02 & 2.6482e+02 & 3.4260e+02 & 3.2600e+02 & 2.0531e+02 & 2.0223e+02 & 3.3383e+02 & 3.2144e+02 & 3.7179e+02 & 2.1127e+02 & 2.5226e+02 \\
				& p-value   & 1.9209e-06 & 2.6033e-06 & 1.7344e-06 & 2.4118e-04 & 9.3157e-06 & 3.1817e-06 & 2.6033e-06 & 1.7344e-06 & 1.7344e-06 & 1.9209e-06 & 1.7344e-06 & 7.6909e-06 & -- \\
				& $+ / = / -$ & + & + & + & + & + & + & + & + & + & + & + & + & -- \\
				\midrule
				
				\multirow{5}{*}{F21}
				& min       & 2.6896e+03 & 2.6324e+03 & 2.5908e+03 & 2.5235e+03 & 2.4909e+03 & 2.4788e+03 & 2.5322e+03 & 2.7889e+03 & 2.7597e+03 & 2.7090e+03 & 2.6436e+03 & 2.5655e+03 & 2.3448e+03 \\
				& mean      & 2.7625e+03 & 2.7821e+03 & 2.7133e+03 & 2.6343e+03 & 2.6693e+03 & 2.5556e+03 & 2.6281e+03 & 3.0341e+03 & 2.8894e+03 & 2.8727e+03 & 2.7737e+03 & 2.6069e+03 & \textbf{2.3789e+03} \\
				& std       & 3.2234e+01 & 6.7722e+01 & 7.2202e+01 & 5.5779e+01 & 7.3464e+01 & 4.5567e+01 & 4.5931e+01 & 1.3507e+02 & 6.7407e+01 & 6.8013e+01 & 7.5174e+01 & 2.2686e+01 & 1.7626e+01 \\
				& p-value   & 1.7344e-06 & 1.7344e-06 & 1.7344e-06 & 1.7344e-06 & 1.7344e-06 & 1.7344e-06 & 1.7344e-06 & 1.7344e-06 & 1.7344e-06 & 1.7344e-06 & 1.7344e-06 & 1.7344e-06 & -- \\
				& $+ / = / -$ & + & + & + & + & + & + & + & + & + & + & + & + & -- \\
				\bottomrule
			\end{tabular}
		\end{adjustbox}
	\end{sidewaystable*}

	\begin{sidewaystable*}[p]
		\centering
		
		\setlength{\abovecaptionskip}{3pt}
		\setlength{\belowcaptionskip}{3pt}
		
		\addtocounter{table}{-1}
		\caption[]{Test results for CEC 2017 ($D=50$) -- Basic Group (continued)}
		
		\scriptsize
		\setlength{\tabcolsep}{2.8pt}
		\renewcommand{\arraystretch}{0.72}
		
		\begin{adjustbox}{
				width=0.98\textheight,
				max totalheight=0.88\textwidth,
				keepaspectratio,
				center
			}
			\begin{tabular}{c c ccccccccccccc}
				\toprule
				Function & & YDSE & EMWPA & SRA & RUN & FLA & RIME & FATA & WAA & ETO & SCHO & FTO & WOO & IWOO \\
				\midrule
				
				\multirow{5}{*}{F22}
				& min       & 1.3823e+04 & 9.7669e+03 & 7.5613e+03 & 7.9554e+03 & 7.4753e+03 & 7.9478e+03 & 1.2090e+04 & 9.2840e+03 & 1.4234e+04 & 1.2432e+04 & 1.1991e+04 & 4.6563e+03 & 2.3000e+03 \\
				& mean      & 1.5365e+04 & 1.1973e+04 & 9.7827e+03 & 1.0152e+04 & 9.7153e+03 & 9.9114e+03 & 1.5440e+04 & 1.1358e+04 & 1.5896e+04 & 1.4560e+04 & 1.4629e+04 & 1.1390e+04 & \textbf{8.6461e+03} \\
				& std       & 6.0021e+02 & 1.5911e+03 & 9.2594e+02 & 2.0625e+03 & 1.1075e+03 & 9.8821e+02 & 1.1351e+03 & 9.6056e+02 & 9.1211e+02 & 1.0255e+03 & 9.0297e+02 & 2.6345e+03 & 2.5337e+03 \\
				& p-value   & 1.7344e-06 & 1.9729e-05 & 2.3038e-02 & 1.6566e-02 & 3.8723e-02 & 3.0010e-02 & 1.7344e-06 & 1.1265e-05 & 1.7344e-06 & 1.7344e-06 & 1.7344e-06 & 3.5888e-04 & -- \\
				& $+ / = / -$ & + & + & + & + & + & + & + & + & + & + & + & + & -- \\
				\midrule
				
				\multirow{5}{*}{F23}
				& min       & 3.1951e+03 & 3.4302e+03 & 3.0892e+03 & 3.0114e+03 & 3.0810e+03 & 2.9550e+03 & 3.0035e+03 & 3.8831e+03 & 3.3143e+03 & 3.4259e+03 & 3.1511e+03 & 3.0840e+03 & 2.7990e+03 \\
				& mean      & 3.2969e+03 & 3.8107e+03 & 3.3356e+03 & 3.1599e+03 & 3.3488e+03 & 3.0639e+03 & 3.1413e+03 & 4.4686e+03 & 3.5796e+03 & 3.6391e+03 & 3.2998e+03 & 3.1835e+03 & \textbf{2.8548e+03} \\
				& std       & 3.8860e+01 & 2.1048e+02 & 1.0871e+02 & 9.4218e+01 & 1.2585e+02 & 6.2586e+01 & 5.9506e+01 & 2.7595e+02 & 1.2876e+02 & 1.2045e+02 & 6.8796e+01 & 8.3843e+01 & 3.1245e+01 \\
				& p-value   & 1.7344e-06 & 1.7344e-06 & 1.7344e-06 & 1.7344e-06 & 1.7344e-06 & 1.7344e-06 & 1.7344e-06 & 1.7344e-06 & 1.7344e-06 & 1.7344e-06 & 1.7344e-06 & 1.7344e-06 & -- \\
				& $+ / = / -$ & + & + & + & + & + & + & + & + & + & + & + & + & -- \\
				\midrule
				
				\multirow{5}{*}{F24}
				& min       & 3.3964e+03 & 3.4798e+03 & 3.2842e+03 & 3.0508e+03 & 3.2973e+03 & 3.0838e+03 & 3.3752e+03 & 4.0600e+03 & 3.6435e+03 & 3.6541e+03 & 3.4001e+03 & 3.2972e+03 & 2.9780e+03 \\
				& mean      & 3.4718e+03 & 4.1691e+03 & 3.6199e+03 & 3.2326e+03 & 3.5906e+03 & 3.1924e+03 & 3.5074e+03 & 4.5482e+03 & 3.8286e+03 & 3.8621e+03 & 3.4957e+03 & 3.4505e+03 & \textbf{3.0138e+03} \\
				& std       & 4.7078e+01 & 3.8826e+02 & 2.1172e+02 & 9.0159e+01 & 1.8019e+02 & 7.0135e+01 & 8.7957e+01 & 1.9256e+02 & 1.1139e+02 & 1.1940e+02 & 5.6481e+01 & 4.4279e+01 & 2.4703e+01 \\
				& p-value   & 1.7344e-06 & 1.7344e-06 & 1.7344e-06 & 1.7344e-06 & 1.7344e-06 & 1.7344e-06 & 1.7344e-06 & 1.7344e-06 & 1.7344e-06 & 1.7344e-06 & 1.7344e-06 & 1.7344e-06 & -- \\
				& $+ / = / -$ & + & + & + & + & + & + & + & + & + & + & + & + & -- \\
				\midrule
				
				\multirow{5}{*}{F25}
				& min       & 3.7013e+03 & 3.2158e+03 & 3.0943e+03 & 3.1231e+03 & 2.9610e+03 & 3.0695e+03 & 3.1068e+03 & 3.3332e+03 & 4.0406e+03 & 4.0313e+03 & 3.7568e+03 & 3.4013e+03 & 3.0121e+03 \\
				& mean      & 4.1378e+03 & 3.4221e+03 & 3.1832e+03 & 3.1702e+03 & \textbf{3.0359e+03} & 3.1469e+03 & 3.1969e+03 & 3.5048e+03 & 5.7208e+03 & 5.9855e+03 & 4.2552e+03 & 3.6173e+03 & 3.0747e+03 \\
				& std       & 3.6796e+02 & 1.5997e+02 & 6.3370e+01 & 3.2505e+01 & 4.4170e+01 & 3.9943e+01 & 4.6569e+01 & 9.5831e+01 & 8.3221e+02 & 8.3474e+02 & 3.5737e+02 & 1.0846e+02 & 2.7085e+01 \\
				& p-value   & 1.7344e-06 & 1.7344e-06 & 1.7344e-06 & 1.7344e-06 & 7.7122e-04 & 1.9209e-06 & 1.7344e-06 & 1.7344e-06 & 1.7344e-06 & 1.7344e-06 & 1.7344e-06 & 1.7344e-06 & -- \\
				& $+ / = / -$ & + & + & + & + & - & + & + & + & + & + & + & + & -- \\
				\midrule
				
				\multirow{5}{*}{F26}
				& min       & 8.4413e+03 & 5.2842e+03 & 8.4272e+03 & 6.6051e+03 & 8.0498e+03 & 5.7860e+03 & 3.8248e+03 & 1.1778e+04 & 8.4077e+03 & 8.7475e+03 & 8.3873e+03 & 6.9567e+03 & 4.3978e+03 \\
				& mean      & 9.4463e+03 & 1.2389e+04 & 1.0727e+04 & 1.1368e+04 & 1.0150e+04 & 7.1544e+03 & \textbf{4.9406e+03} & 1.2725e+04 & 1.1518e+04 & 1.2169e+04 & 1.0204e+04 & 8.7869e+03 & 5.2014e+03 \\
				& std       & 4.3258e+02 & 2.1645e+03 & 1.4111e+03 & 1.4134e+03 & 1.4430e+03 & 7.9326e+02 & 1.9850e+03 & 7.6897e+02 & 8.9033e+02 & 1.2042e+03 & 6.9514e+02 & 7.9814e+02 & 3.7250e+02 \\
				& p-value   & 1.7344e-06 & 1.7344e-06 & 1.7344e-06 & 1.7344e-06 & 1.7344e-06 & 1.7344e-06 & 2.7653e-03 & 1.7344e-06 & 1.7344e-06 & 1.7344e-06 & 1.7344e-06 & 1.7344e-06 & -- \\
				& $+ / = / -$ & + & + & + & + & + & + & - & + & + & + & + & + & -- \\
				\midrule
				
				\multirow{5}{*}{F27}
				& min       & 3.6850e+03 & 3.7340e+03 & 3.6123e+03 & 3.7026e+03 & 3.4715e+03 & 3.3749e+03 & 3.4728e+03 & 4.6814e+03 & 4.0357e+03 & 4.2281e+03 & 3.2000e+03 & 3.5832e+03 & 3.3626e+03 \\
				& mean      & 3.8364e+03 & 4.1638e+03 & 4.0813e+03 & 3.8738e+03 & 3.7568e+03 & 3.5847e+03 & 3.6464e+03 & 6.3419e+03 & 4.6131e+03 & 4.7205e+03 & \textbf{3.2000e+03} & 3.6928e+03 & 3.5101e+03 \\
				& std       & 9.3944e+01 & 3.1325e+02 & 2.5008e+02 & 1.1339e+02 & 2.5440e+02 & 1.1926e+02 & 9.6242e+01 & 1.0034e+03 & 3.0834e+02 & 2.9629e+02 & 5.3415e-05 & 6.9165e+01 & 7.4959e+01 \\
				& p-value   & 1.7344e-06 & 1.7344e-06 & 1.7344e-06 & 1.7344e-06 & 5.7517e-06 & 1.6566e-02 & 5.2165e-06 & 1.7344e-06 & 1.7344e-06 & 1.7344e-06 & 1.7344e-06 & 2.3534e-06 & -- \\
				& $+ / = / -$ & + & + & + & + & + & + & + & + & + & + & - & + & -- \\
				\midrule
				
				\multirow{5}{*}{F28}
				& min       & 3.9832e+03 & 3.4441e+03 & 3.4664e+03 & 3.4413e+03 & 3.2588e+03 & 3.3704e+03 & 3.4015e+03 & 4.2261e+03 & 4.7301e+03 & 4.5733e+03 & 3.3000e+03 & 3.9649e+03 & 3.2795e+03 \\
				& mean      & 5.0297e+03 & 3.7879e+03 & 3.6790e+03 & 3.5501e+03 & 3.3213e+03 & 3.4524e+03 & 3.5460e+03 & 4.6079e+03 & 5.8267e+03 & 6.2981e+03 & \textbf{3.3000e+03} & 4.2719e+03 & 3.3162e+03 \\
				& std       & 4.9152e+02 & 1.9384e+02 & 2.0429e+02 & 6.8005e+01 & 1.3783e+02 & 4.1050e+01 & 7.5074e+01 & 2.5485e+02 & 7.3611e+02 & 9.2133e+02 & 6.4567e-05 & 2.3259e+02 & 2.0925e+01 \\
				& p-value   & 1.7344e-06 & 1.7344e-06 & 1.7344e-06 & 1.7344e-06 & 2.8308e-04 & 1.7344e-06 & 1.7344e-06 & 1.7344e-06 & 1.7344e-06 & 1.7344e-06 & 1.6394e-05 & 1.7344e-06 & -- \\
				& $+ / = / -$ & + & + & + & + & + & + & + & + & + & + & - & + & -- \\
				\midrule
				
				\multirow{5}{*}{F29}
				& min       & 5.2008e+03 & 4.9434e+03 & 4.6317e+03 & 4.6449e+03 & 4.4496e+03 & 4.2455e+03 & 3.9275e+03 & 5.7778e+03 & 5.2779e+03 & 5.8373e+03 & 4.8715e+03 & 4.8360e+03 & 3.3180e+03 \\
				& mean      & 6.0576e+03 & 5.7803e+03 & 5.3544e+03 & 5.5932e+03 & 5.2723e+03 & 5.1685e+03 & 4.4263e+03 & 7.6299e+03 & 7.1450e+03 & 7.0785e+03 & 6.1770e+03 & 5.9007e+03 & \textbf{3.7209e+03} \\
				& std       & 4.0223e+02 & 4.6923e+02 & 4.9310e+02 & 5.5005e+02 & 5.4133e+02 & 4.3392e+02 & 2.7643e+02 & 9.8279e+02 & 7.8838e+02 & 1.3324e+03 & 6.3170e+02 & 3.5822e+02 & 2.6358e+02 \\
				& p-value   & 1.7344e-06 & 1.7344e-06 & 1.7344e-06 & 1.7344e-06 & 1.7344e-06 & 1.7344e-06 & 1.7344e-06 & 1.7344e-06 & 1.7344e-06 & 1.7344e-06 & 1.7344e-06 & 1.7344e-06 & -- \\
				& $+ / = / -$ & + & + & + & + & + & + & + & + & + & + & + & + & -- \\
				\midrule
				
				\multirow{5}{*}{F30}
				& min       & 2.2153e+07 & 4.7280e+06 & 1.7902e+06 & 2.6809e+07 & 7.3482e+05 & 2.4882e+07 & 3.1509e+06 & 7.1824e+07 & 1.3093e+08 & 1.0023e+08 & 4.2659e+07 & 3.9616e+07 & 7.8251e+05 \\
				& mean      & 4.8862e+07 & 1.1490e+07 & 6.7241e+06 & 4.8485e+07 & 1.1283e+06 & 5.9860e+07 & 1.0013e+07 & 1.3592e+08 & 2.9531e+08 & 1.0862e+09 & 8.4328e+07 & 5.3810e+07 & \textbf{9.1707e+05} \\
				& std       & 1.2874e+07 & 8.3165e+06 & 3.5287e+06 & 1.0540e+07 & 2.4586e+05 & 3.4282e+07 & 3.7867e+06 & 3.3964e+07 & 1.1211e+08 & 1.3822e+09 & 2.6304e+07 & 9.2319e+06 & 7.8295e+04 \\
				& p-value   & 1.7344e-06 & 1.7344e-06 & 1.7344e-06 & 1.7344e-06 & 8.9187e-05 & 1.7344e-06 & 1.7344e-06 & 1.7344e-06 & 1.7344e-06 & 1.7344e-06 & 1.7344e-06 & 1.7344e-06 & -- \\
				& $+ / = / -$ & + & + & + & + & + & + & + & + & + & + & + & + & -- \\
				\midrule
				Friedman & Mean rank & 7.8621 & 7.4483 & 6.5862 & 4.4828 & 3.6897 & 4.5862 & 6.4483 & 9.7931 & 11.4828 & 11.6207 & 9.5517 & 6.0000 & \textbf{1.4483} \\
				\midrule
				ToL Wilcoxon & $+ / = / -$ & 27/0/2 & 29/0/0 & 29/0/0 & 28/0/1 & 23/0/6 & 29/0/0 & 27/0/2 & 29/0/0 & 29/0/0 & 29/0/0 & 27/0/2 & 28/1/0 & -- \\
				\bottomrule
			\end{tabular}
		\end{adjustbox}
	\end{sidewaystable*}

	\makeatletter
	\newsavebox{\IWOORankBottomFigure}
	\newdimen\IWOORankBottomSpace
	\newcommand{\IWOOReserveRankBottom}{%
		\par
		\if@firstcolumn\else
		\PackageError{IWOO layout}{Insert the cross-benchmark section in the left column}{The page bottom must be reserved before the right column is built.}%
		\fi
		\global\setbox\IWOORankBottomFigure=\vbox{%
			\hbox{\copy\IWOORankBottomFigure}\kern0pt
		}%
		\global\IWOORankBottomSpace=\dimexpr
		\ht\IWOORankBottomFigure+\dp\IWOORankBottomFigure+10pt\relax
		\global\let\IWOORankOriginalCombineFloats\@combinedblfloats
		\gdef\@combinedblfloats{%
			\IWOORankOriginalCombineFloats
			\global\let\@combinedblfloats\IWOORankOriginalCombineFloats
			\setbox\@outputbox=\vbox to\textheight{%
				\unvbox\@outputbox
				\vfil
				\vskip10pt
				\nointerlineskip
				\copy\IWOORankBottomFigure
				\kern0pt
			}%
		}%
		\global\advance\@colht-\IWOORankBottomSpace
		\global\@colroom=\@colht
		\global\vsize=\@colht
		\global\pagegoal=\@colht
	}
	\makeatother
	
	\begingroup
	\flushbottom
	\setlength{\parskip}{0pt plus 20pt}
	\baselineskip=\dimexpr\baselineskip\relax plus 1pt
	\begin{lrbox}{\IWOORankBottomFigure}
		\begin{minipage}{\textwidth}
			\centering
			\captionsetup{skip=3pt}
			\includegraphics[width=0.95\textwidth,height=0.27\textheight,keepaspectratio]{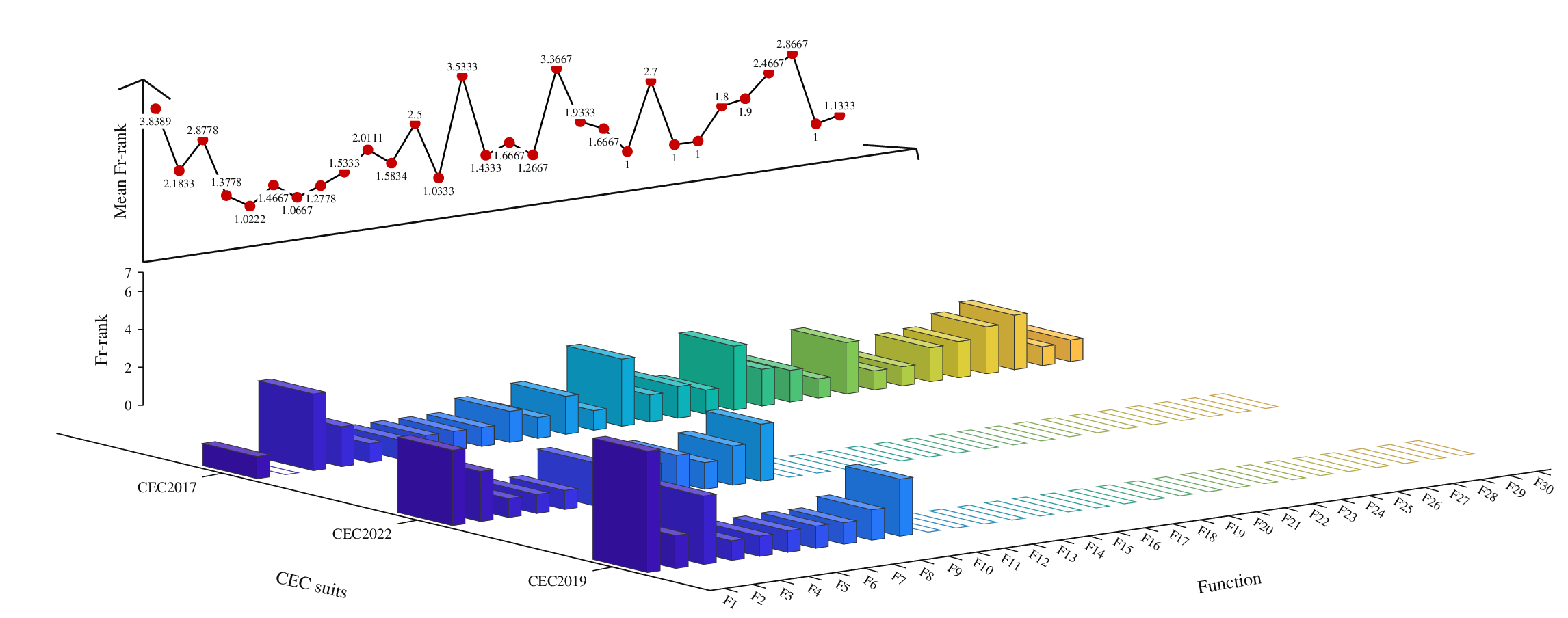}
			\captionof{figure}{Function-wise Fr-ranks of IWOO across benchmark suites}
			\label{fig:functionwise_rank_summary}
		\end{minipage}
	\end{lrbox}
	\IWOOReserveRankBottom
	\makeatletter
	\def\@BAlancecol{%
		\if@twocolumn
		\setbox0=\vbox{\unvbox\@outputbox\unskip}%
		\@tempdima=\ht0
		\advance\@tempdima by\topskip
		\advance\@tempdima by-\baselineskip
		\divide\@tempdima by2
		\splittopskip=\topskip
		{\vbadness=\@M
			\loop
			\global\setbox3=\copy0
			\global\setbox1=\vsplit3 to\@tempdima
			\ifdim\ht3>\@tempdima
			\global\advance\@tempdima by1pt
			\repeat}%
		\setbox1=\vbox{\unvbox1\unskip}%
		\setbox3=\vbox{\unvbox3\unskip}%
		\@tempdima=\ht1
		\ifdim\ht3>\@tempdima \@tempdima=\ht3\fi
		\setbox\@leftcolumn=\vbox to\@tempdima{\unvbox1}%
		\setbox\@outputbox=\vbox to\@tempdima{\unvbox3}%
		\fi
	}
	\makeatother
	\makeatletter
	\let\IWOORankBalanceOutput\@BAdblcol
	\def\@BAdblcol{%
		\if@firstcolumn
		\setbox\@outputbox=\vbox{\unvbox\@outputbox\unskip}%
		\fi
		\IWOORankBalanceOutput
	}
	\makeatother
	\balance
	
	\subsection{Extended performance analysis}
	\label{subsec:cross_suite_summary}
	\baselineskip=\dimexpr\baselineskip\relax plus 1pt
	
	To further evaluate the overall performance of IWOO across different benchmark suites, the Friedman ranking results obtained on CEC2019, CEC2022, and CEC2017 are jointly analyzed. Figure~\ref{fig:functionwise_rank_summary} summarizes the function-wise Friedman ranks of IWOO on the three benchmark suites together with their ranking variation trends. The three-dimensional bar chart in the lower part of the figure presents the Friedman rank of IWOO on each test function, while the polyline in the upper part summarizes the available ranking results associated with identical function indices to provide an intuitive view of ranking variations under different benchmark settings.
	
	As shown in Fig.~\ref{fig:functionwise_rank_summary}, IWOO achieves relatively low Friedman ranks on most test functions across the three benchmark suites, with the overall ranking results predominantly concentrated in the favorable low-rank region. Although some variations are observed among individual functions, IWOO maintains strong competitive performance across different benchmark suites, indicating that its performance advantage is not restricted to a specific benchmark suite or a small subset of test functions. Overall, IWOO exhibits good adaptability to different function characteristics and benchmark settings, together with relatively stable overall optimization performance.
	
	\FloatBarrier
	\clearpage
	\nobalance
	\endgroup

	\FloatBarrier
	
	\begingroup
	\captionsetup{font=footnotesize,skip=2pt}
	
	\subsection{Ablation study}
	\label{subsec:ablation_study}
	
	To evaluate the individual contribution of each proposed improvement strategy and the combined effect of the complete algorithm, four ablation variants are constructed using the original WOO as the baseline. Specifically, WOO denotes the original Wave Optics Optimizer and serves as the baseline algorithm; WOO1 incorporates only the TriEDM multi-path search strategy; WOO2 incorporates only the CGOL strategy; WOO3 incorporates only the ASRK correction strategy; and WOO4 incorporates only the GPS--OBL initialization strategy. IWOO denotes the complete improved algorithm integrating all four strategies.
	
	The ablation experiments are conducted on three benchmark suites, namely CEC2019, CEC2022, and CEC2017. The Friedman rank is employed to evaluate the overall performance of the different variants.
	
	\par\smallskip\noindent
	\begin{minipage}{\linewidth}
		\centering
		\captionof{table}{Average Fr-ranks of the ablation variants across different benchmark suites}
		\label{tab:ablation_frank}
		
		\small
		\setlength{\tabcolsep}{3pt}
		\renewcommand{\arraystretch}{1.12}
		
		\resizebox{\linewidth}{!}{%
			\begin{tabular}{lccc}
				\toprule
				Algorithm & CEC2019 & CEC2022 & CEC2017 \\
				\midrule
				WOO  & 5.1500 & 5.5000 & 5.5517 \\
				WOO1 & 4.0000 & 3.4167 & 3.8276 \\
				WOO2 & 2.1500 & 1.9583 & 1.6897 \\
				WOO3 & 3.3500 & 3.2500 & 3.4483 \\
				WOO4 & 4.5500 & 5.5000 & 4.8966 \\
				IWOO & \textbf{1.8000}
				& \textbf{1.3750}
				& \textbf{1.5862} \\
				\midrule
				Friedman $p$-value
				& $9.2822\times10^{-5}$
				& $7.0938\times10^{-10}$
				& $5.3462\times10^{-22}$ \\
				\bottomrule
			\end{tabular}%
		}
	\end{minipage}
	\par\smallskip
	
	As shown in Table~\ref{tab:ablation_frank}, the Friedman test yields $p$-values below 0.05 across all three benchmark suites, indicating statistically significant performance differences among the ablation variants. IWOO achieves the lowest Fr-rank across all three benchmark suites, demonstrating that the integration of the four improvement strategies provides more consistent overall performance than both the original WOO and the single-strategy variants.
	
	Considering the individual strategies, WOO2 consistently achieves relatively low average Fr-ranks across all three benchmark suites, indicating that CGOL provides the most consistent individual improvement in overall optimization performance. WOO1 and WOO3 also consistently outperform the original WOO, demonstrating the effectiveness of TriEDM in multi-directional search and ASRK in stagnation correction, respectively. WOO4 outperforms the original WOO on most benchmark suites and achieves comparable performance on CEC2022, indicating that GPS--OBL can effectively improve the quality of the initial population and provide favorable initial conditions for the subsequent search process.
	
	The function-wise Fr-rank comparisons and mean Fr-rank radar plots on the CEC2019, CEC2022, and CEC2017 benchmark suites are presented in Figs.~\ref{fig:ablation_cec2019}, \ref{fig:ablation_cec2022}, and \ref{fig:ablation_cec2017}, respectively. For each benchmark suite, the function-wise comparison shows the rankings of the ablation variants on individual test functions, whereas the radar plot summarizes their mean Fr-ranks. Together, these visualizations provide complementary views of the performance differences among the variants at the individual-function and overall benchmark-suite levels.
	
	\par
	\newpage
	
	\noindent
	\begin{minipage}{\linewidth}
		\centering
		
		\begin{minipage}[c]{0.64\linewidth}
			\centering
			\includegraphics[
			width=\linewidth,
			height=0.175\textheight,
			keepaspectratio
			]{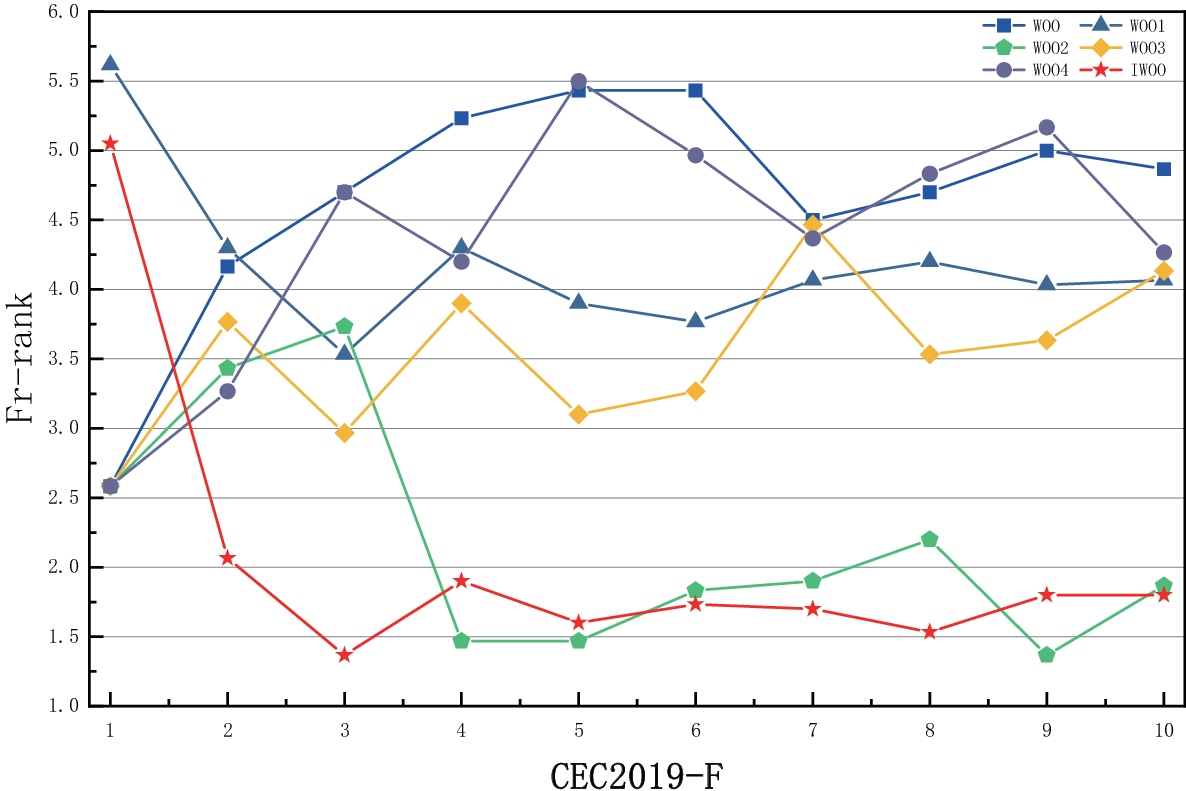}
			
			\par\vspace{-1mm}
			{\small (a)}
		\end{minipage}
		\hfill
		\begin{minipage}[c]{0.30\linewidth}
			\centering
			\includegraphics[
			width=\linewidth,
			height=0.175\textheight,
			keepaspectratio
			]{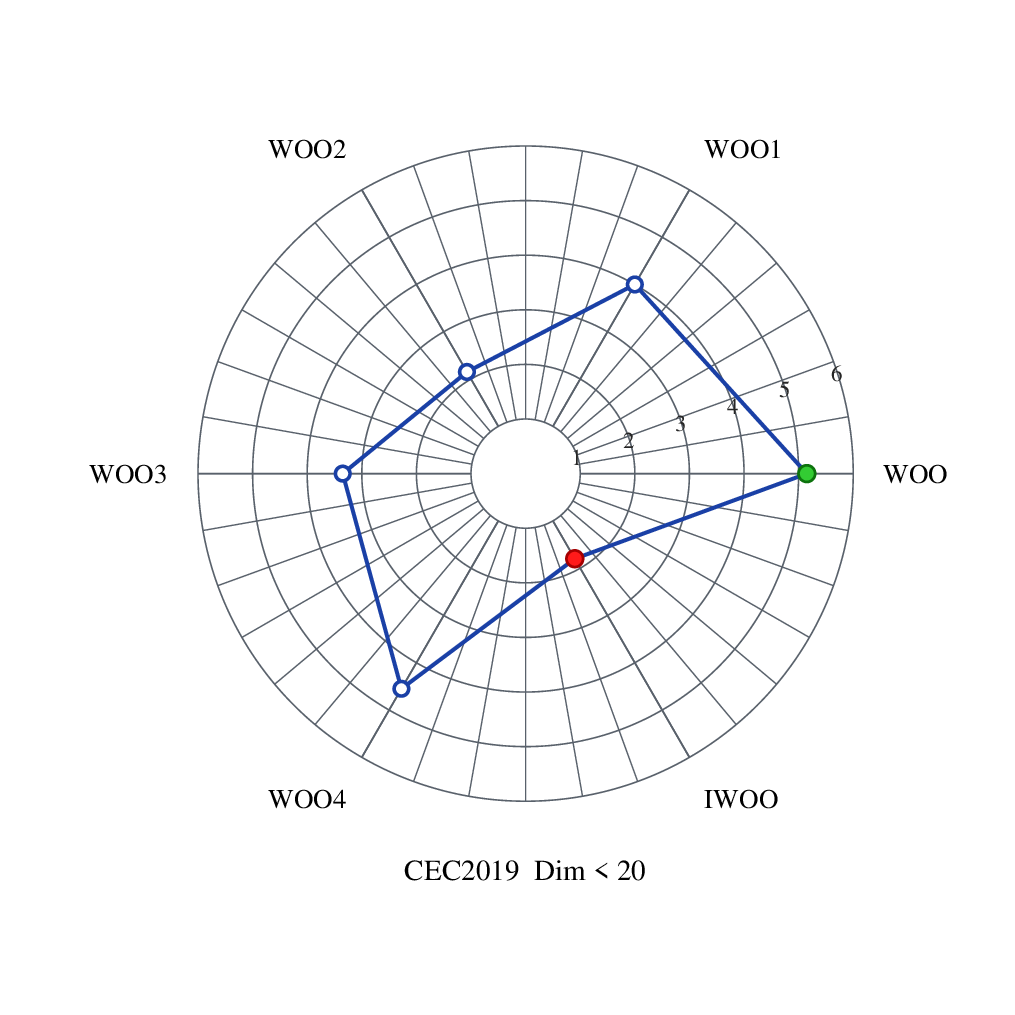}
			
			\par\vspace{-1mm}
			{\small (b)}
		\end{minipage}
		
		\par\vspace{0.5mm}
		\captionof{figure}{Ablation study results on the CEC2019 benchmark suite}
		\label{fig:ablation_cec2019}
	\end{minipage}
	\par\smallskip
	
	\noindent
	\begin{minipage}{\linewidth}
		\centering
		
		\begin{minipage}[c]{0.64\linewidth}
			\centering
			\includegraphics[
			width=\linewidth,
			height=0.175\textheight,
			keepaspectratio
			]{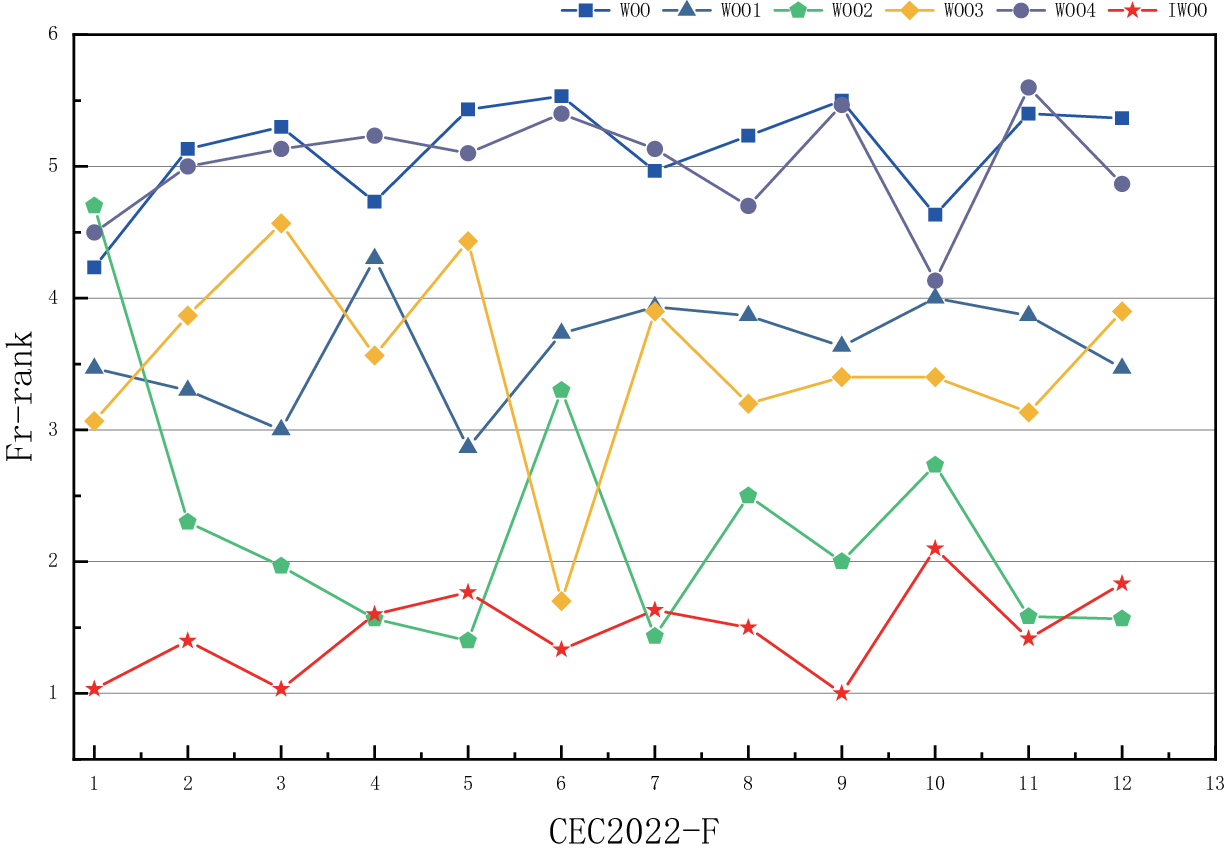}
			
			\par\vspace{-1mm}
			{\small (a)}
		\end{minipage}
		\hfill
		\begin{minipage}[c]{0.30\linewidth}
			\centering
			\includegraphics[
			width=\linewidth,
			height=0.175\textheight,
			keepaspectratio
			]{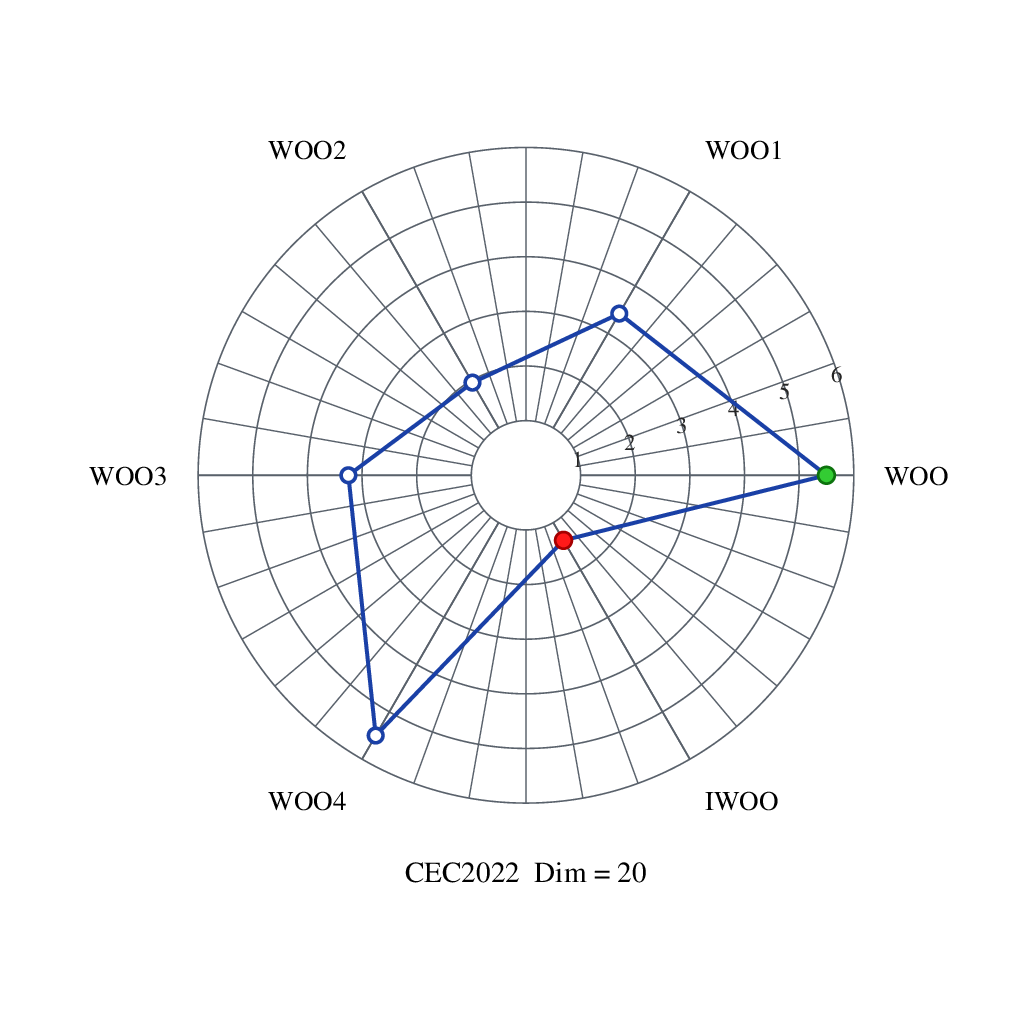}
			
			\par\vspace{-1mm}
			{\small (b)}
		\end{minipage}
		
		\par\vspace{0.5mm}
		\captionof{figure}{Ablation study results on the CEC2022 benchmark suite}
		\label{fig:ablation_cec2022}
	\end{minipage}
	\par\smallskip
	
	\noindent
	\begin{minipage}{\linewidth}
		\centering
		
		\begin{minipage}[c]{0.64\linewidth}
			\centering
			\includegraphics[
			width=\linewidth,
			height=0.175\textheight,
			keepaspectratio
			]{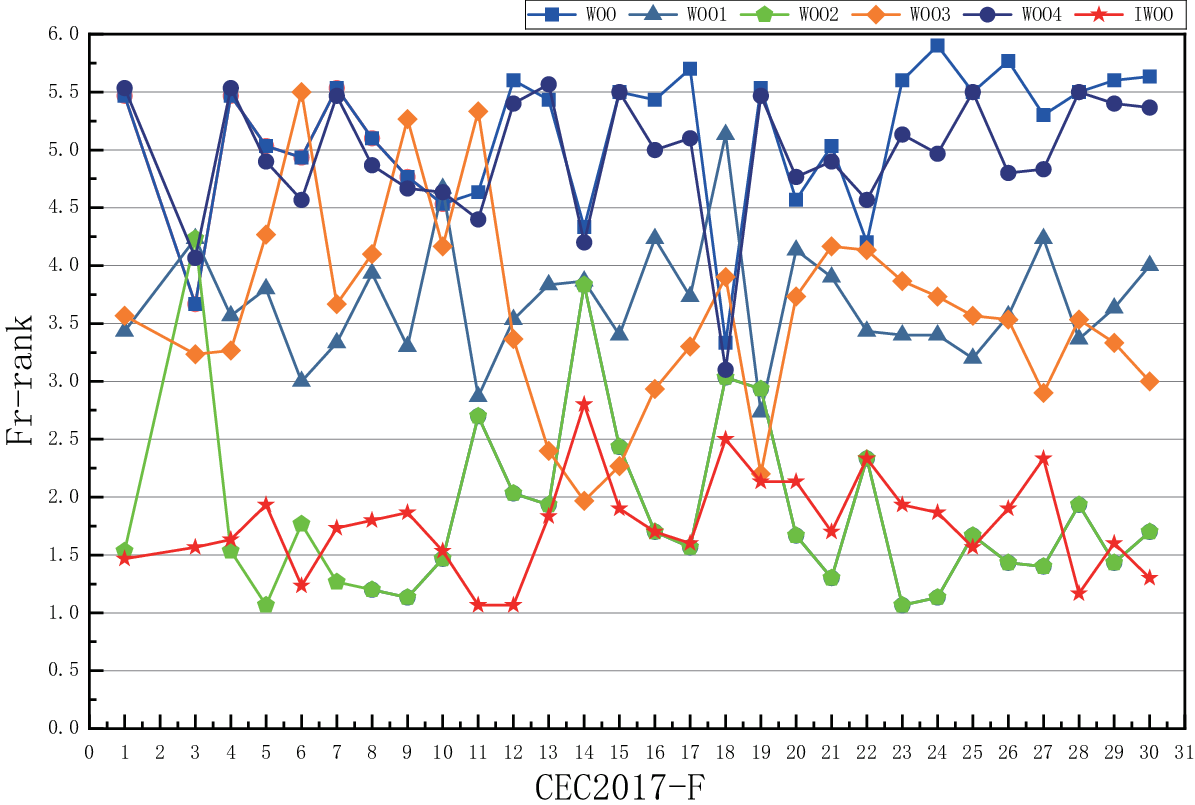}
			
			\par\vspace{-1mm}
			{\small (a)}
		\end{minipage}
		\hfill
		\begin{minipage}[c]{0.30\linewidth}
			\centering
			\includegraphics[
			width=\linewidth,
			height=0.175\textheight,
			keepaspectratio
			]{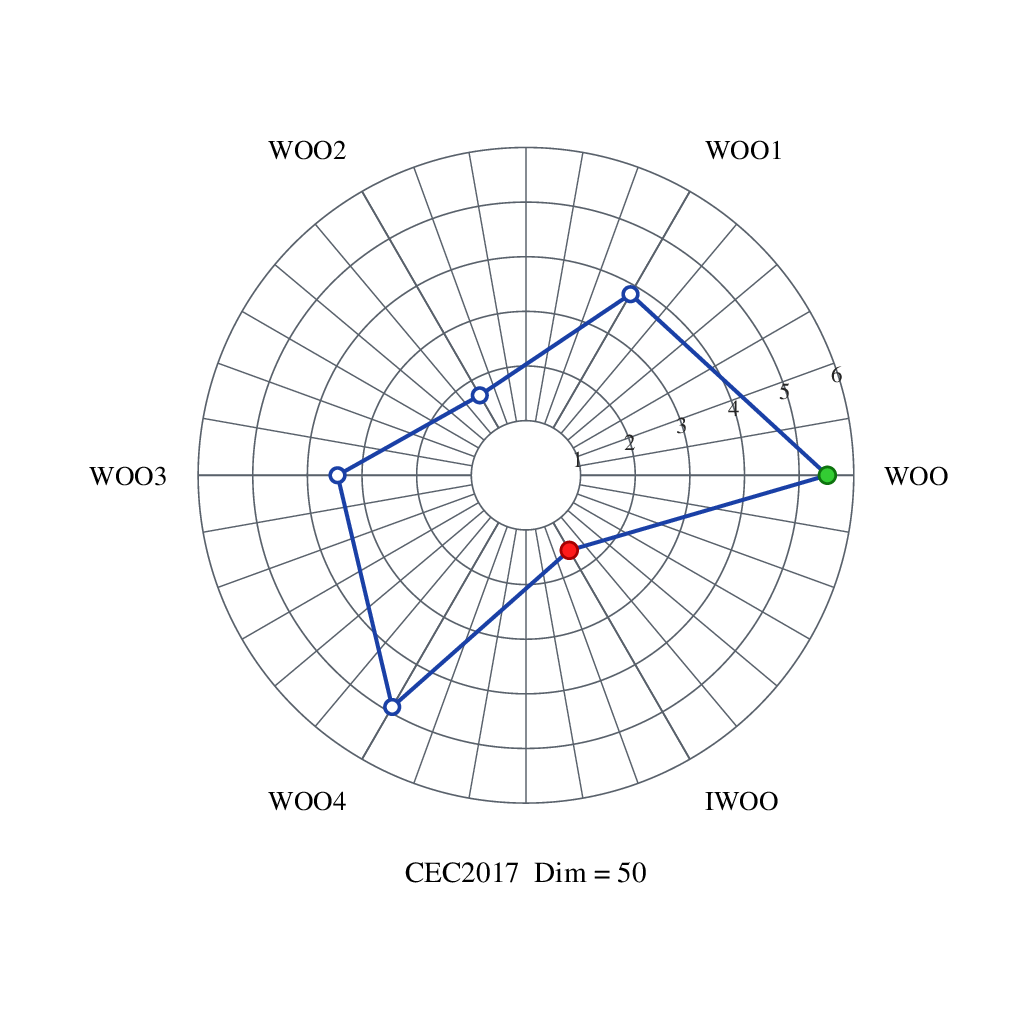}
			
			\par\vspace{-1mm}
			{\small (b)}
		\end{minipage}
		
		\par\vspace{0.5mm}
		\captionof{figure}{Ablation study results on the CEC2017 benchmark suite}
		\label{fig:ablation_cec2017}
	\end{minipage}
	\par\smallskip
	
	Figures~\ref{fig:ablation_cec2019}--\ref{fig:ablation_cec2017} further present the function-wise ranking behavior of the different variants across the benchmark functions. Overall, the Fr-ranks of IWOO are predominantly concentrated in the lower-rank region, followed by WOO2, whereas the original WOO generally exhibits higher ranks. This observation is consistent with the statistical results reported in Table~\ref{tab:ablation_frank}. Taken together, the ablation results demonstrate that each proposed strategy provides a different degree of individual improvement, while the complete IWOO exhibits a more consistent combined effect. These results verify both the effectiveness and the complementary roles of the four proposed improvement strategies.
	
	\par
	\endgroup
	\FloatBarrier

	\section{Engineering design and practical application cases}
	\label{sec:engineering_problems}
	
	To evaluate the applicability of IWOO to practical engineering optimization problems, this section considers several constrained engineering design problems, together with representative real-world applications involving economic load dispatch, wireless sensor network deployment, and rigid point cloud registration.
	
	\subsection{Welded beam design problem}
	\label{subsec:welded_beam_design}
	
	The welded beam design problem is a classical constrained optimization problem in mechanical engineering, aiming to minimize the structural manufacturing cost while satisfying stress, deflection, buckling load, and geometric constraints~\cite{Ragsdell1976WeldedStructures,Cagnina2008ConstrainedPSO}. The problem involves four design variables, namely the weld thickness, weld length, beam height, and beam thickness, which are defined as
	
	\begin{equation}
		\boldsymbol{x}=[x_1,x_2,x_3,x_4]=[h,l,t,b]
		\label{eq:welded_variables}
	\end{equation}
	
\begin{figure}[t]
		\centering
		\captionsetup{skip=3pt}
		
		\begin{minipage}[b]{0.49\linewidth}
			\centering
			\includegraphics[
			width=\linewidth
			]{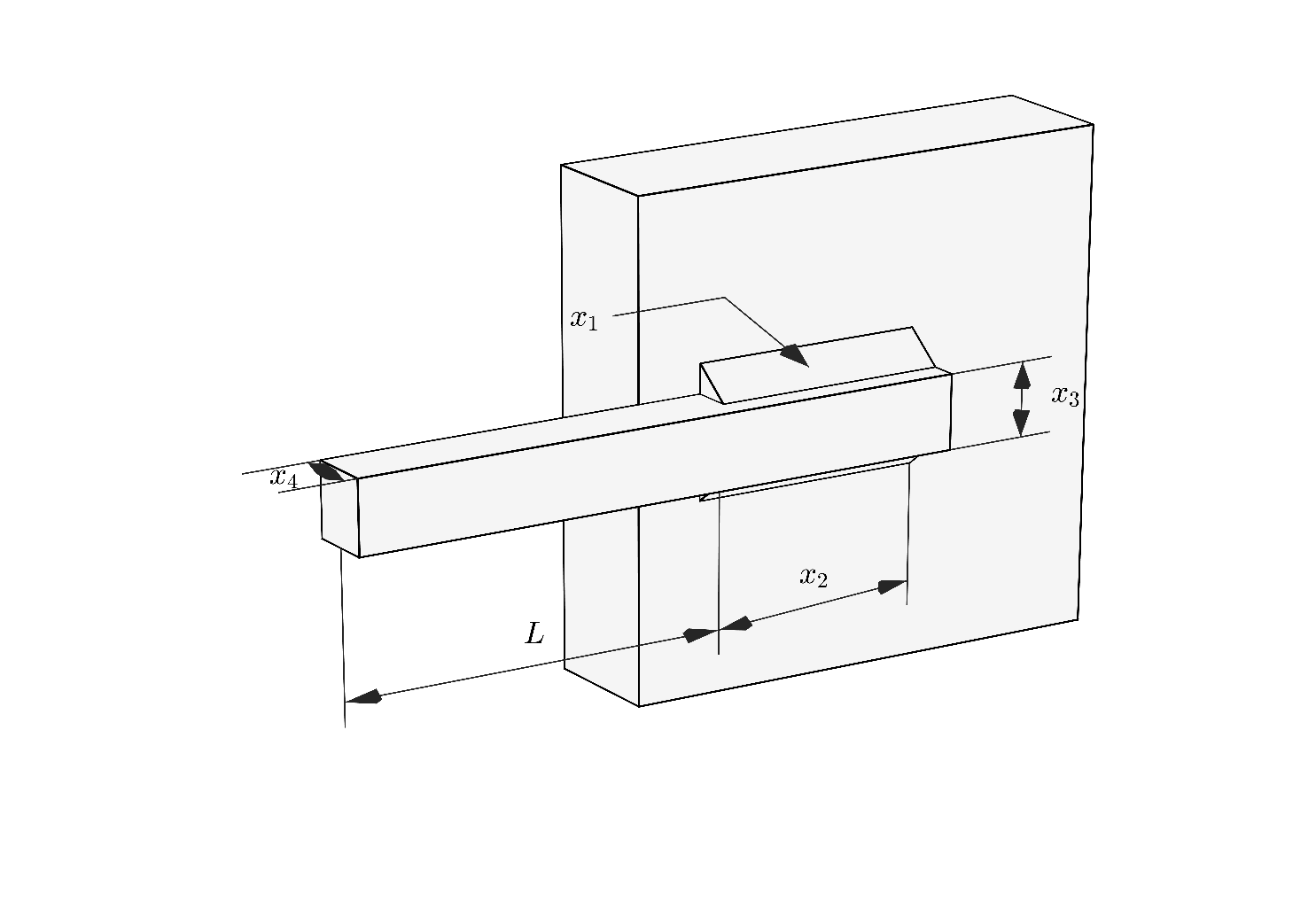}
			
			\par\vspace{2pt}
			{\small (a) Structural parameters}
		\end{minipage}%
		\hfill
		\begin{minipage}[b]{0.49\linewidth}
			\centering
			\includegraphics[
			width=\linewidth
			]{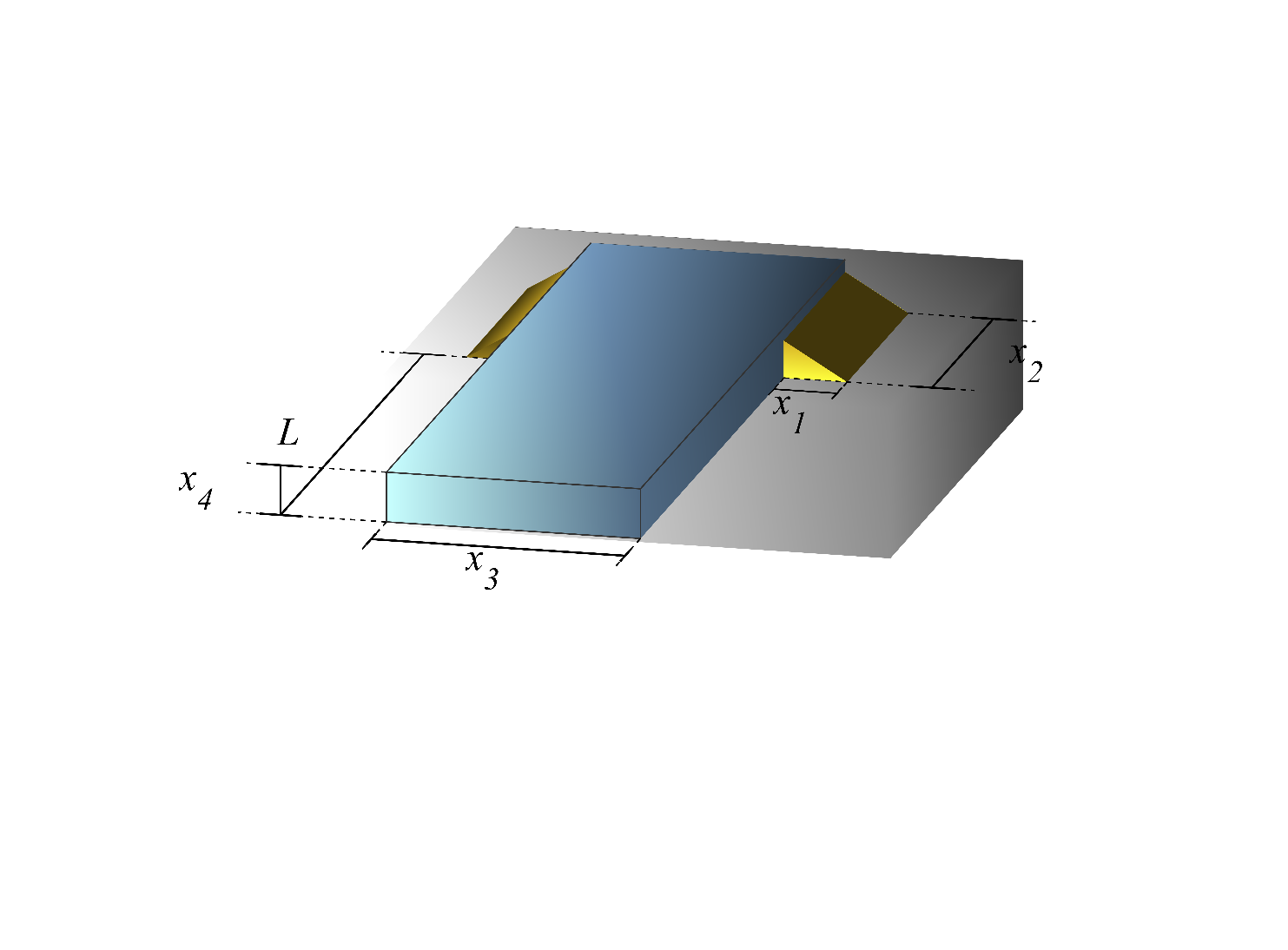}
			
			\par\vspace{2pt}
			{\small (b) 3D geometric model}
		\end{minipage}
		
		\caption{Welded beam model}
		\label{fig:welded_beam_model}
	\end{figure}

	The structural configuration and corresponding design variables are illustrated in Fig.~\ref{fig:welded_beam_model}
	
	The mathematical formulation of the welded beam design problem is given by
	\begin{equation}
		\min f(\boldsymbol{x})
		=
		1.10471x_1^2x_2
		+
		0.04811x_3x_4(14+x_2)
		\label{eq:welded_objective}
	\end{equation}
	\begin{equation}
		\left\{
		\begin{aligned}
			g_1(\boldsymbol{x})&=\tau(\boldsymbol{x})-13600\leq 0\\
			g_2(\boldsymbol{x})&=\sigma(\boldsymbol{x})-30000\leq 0\\
			g_3(\boldsymbol{x})&=x_1-x_4\leq 0\\
			g_4(\boldsymbol{x})&=0.10471x_1^2+0.04811x_3x_4(14+x_2)-5.0\leq 0\\
			g_5(\boldsymbol{x})&=0.125-x_1\leq 0\\
			g_6(\boldsymbol{x})&=\delta(\boldsymbol{x})-0.25\leq 0\\
			g_7(\boldsymbol{x})&=6000-P_c(\boldsymbol{x})\leq 0
		\end{aligned}
		\right.
		\label{eq:welded_constraints}
	\end{equation}
	where the shear stress and its auxiliary quantities are defined as
	\begin{equation}
		\left\{
		\begin{alignedat}{4}
			\tau(\boldsymbol{x})
			&=
			\sqrt{
				(\tau')^2
				+
				\frac{\tau'\tau''x_2}{R}
				+
				(\tau'')^2
			}
			&\qquad
			&\\
			\tau'
			&=
			\frac{6000}{\sqrt{2}x_1x_2},
			&\qquad
			\tau''
			&=
			\frac{MR}{J}
			\\
			M
			&=
			6000\left(14+\frac{x_2}{2}\right),
			&\qquad
			R
			&=
			\sqrt{
				\frac{x_2^2}{4}
				+
				\left(
				\frac{x_1+x_3}{2}
				\right)^2
			}
			\\
			J
			&=
			2\sqrt{2}x_1x_2
			\left[
			\frac{x_2^2}{12}
			+
			\left(
			\frac{x_1+x_3}{2}
			\right)^2
			\right]
			&\qquad
			&
		\end{alignedat}
		\right.
		\label{eq:welded_auxiliary}
	\end{equation}
	
	The bending stress, tip deflection, and critical buckling load are respectively given by
	\begingroup
	\normalsize
	\setlength{\jot}{1pt}
	
	\begin{equation}
		\left\{
		\begin{aligned}
			&\sigma(\boldsymbol{x})
			=\frac{504000}{x_4x_3^2},
			\quad
			\delta(\boldsymbol{x})
			=\frac{65856000}{30\times10^6x_4x_3^3}
			\\
			&P_c(\boldsymbol{x})
			=
			\frac{
				4.013(30\times10^6)
				\sqrt{x_3^2x_4^6/36}
			}{196}
			\\
			&\phantom{P_c(\boldsymbol{x})={}}
			\times
			\left(
			1-\frac{x_3}{28}
			\sqrt{
				\frac{30\times10^6}
				{4(12\times10^6)}
			}
			\right)
		\end{aligned}
		\right.
		\label{eq:welded_response}
	\end{equation}
	\endgroup
	
	The design variables are bounded by
	\begin{equation}
		0.1\leq x_1,x_4\leq 2.0,
		\qquad
		0.1\leq x_2,x_3\leq 10.0
		\label{eq:welded_bounds}
	\end{equation}
	
	The reference optimum for this problem is
	$\boldsymbol{x}^{*}=(0.205730,\,3.470489,\,9.036624,\,0.205729)$, with the corresponding objective value $f(\boldsymbol{x}^{*})=1.724852$.
	
	To evaluate the performance of IWOO on the welded beam design problem, all algorithms were independently executed 30 times under the same experimental conditions, and a unified penalty-function strategy was adopted to handle the constraints~\cite{CoelloCoello2002ConstraintHandling}.The statistical results are reported in Table~\ref{tab:welded_results}
	
	\begin{table}[!htbp]
		\centering
		\caption{Experimental results of the welded beam design optimization problem}
		\label{tab:welded_results}
		\scriptsize
		\setlength{\tabcolsep}{3.2pt}
		\renewcommand{\arraystretch}{1.05}
		\resizebox{\linewidth}{!}{
			\begin{tabular}{lcccccccc}
				\toprule
				\multirow{2}{*}{Algorithm}
				& \multicolumn{4}{c}{Optimal Variables}
				& \multirow{2}{*}{Min Cost}
				& \multirow{2}{*}{Mean Cost}
				& \multirow{2}{*}{Std}
				& \multirow{2}{*}{Fr-rank} \\
				\cmidrule(lr){2-5}
				& $x_1$ & $x_2$ & $x_3$ & $x_4$ & & & & \\
				\midrule
				FLA
				& 0.2057296363 & 3.4704887786 & 9.0366238412 & 0.2057296429
				& \textbf{1.7248523305} & 1.7278451584 & $1.6327\times10^{-2}$ & 2.1333 \\
				
				IWOO
				& 0.2057296078 & 3.4704893651 & 9.0366238924 & 0.2057296406
				& 1.7248523567 & \textbf{1.7248528003} & \textbf{$4.7273\times10^{-7}$} & \textbf{1.1333} \\
				
				EMWPA
				& 0.2057275111 & 3.4705344806 & 9.0366239104 & 0.2057296398
				& 1.7248551905 & 1.8806640130 & $4.4077\times10^{-1}$ & 5.4000 \\
				
				RUN
				& 0.2056573548 & 3.4720744202 & 9.0366054315 & 0.2057304868
				& 1.7249574624 & 1.7639633031 & $3.7304\times10^{-2}$ & 7.1000 \\
				
				FATA
				& 0.2057322246 & 3.4715029202 & 9.0363332158 & 0.2057458385
				& 1.7250672953 & 1.7267541033 & $1.3822\times10^{-3}$ & 3.5667 \\
				
				WOO
				& 0.2056237963 & 3.4815504860 & 9.0340927157 & 0.2058630936
				& 1.7267674692 & 1.7320650135 & $3.3279\times10^{-3}$ & 5.2333 \\
				
				YDSE
				& 0.2052191019 & 3.4814713938 & 9.0551344734 & 0.2056474616
				& 1.7281181778 & 1.7345910623 & $4.4331\times10^{-3}$ & 5.7000 \\
				
				SCHO
				& 0.2037373206 & 3.5116086731 & 9.0485595810 & 0.2056724867
				& 1.7289211120 & 1.9052360092 & $2.4506\times10^{-1}$ & 8.6000 \\
				
				SRA
				& 0.2041414900 & 3.5251532088 & 9.0366722880 & 0.2057307961
				& 1.7297797796 & 2.0360085425 & $2.9058\times10^{-1}$ & 10.1000 \\
				
				ETO
				& 0.2058822637 & 3.5145851083 & 9.0219384218 & 0.2067258399
				& 1.7361295232 & 1.7747506512 & $4.1174\times10^{-2}$ & 8.2000 \\
				
				RIME
				& 0.2169548158 & 3.3428866125 & 8.7692293962 & 0.2184897115
				& 1.7724579483 & 2.6189563647 & $6.9154\times10^{-1}$ & 12.2333 \\
				
				WAA
				& 0.1986132260 & 3.7364047488 & 8.8733989262 & 0.2164054731
				& 1.8013721426 & 2.1554396348 & $2.7942\times10^{-1}$ & 11.3667 \\
				
				FTO
				& 0.2078669102 & 3.6556633560 & 8.9596864257 & 0.2137728980
				& 1.8014108450 & 1.9863395894 & $1.9387\times10^{-1}$ & 10.2333 \\
				\bottomrule
			\end{tabular}
		}
	\end{table}
	
	As shown in Table~\ref{tab:welded_results}, FLA obtains the lowest single-run cost of 1.7248523305. However, IWOO achieves the best mean cost of 1.7248528003, the smallest standard deviation of $4.7273\times10^{-7}$, and the best Friedman rank of 1.1333 among all compared algorithms. These results indicate that IWOO provides both high solution accuracy and strong robustness on the welded beam design problem.
	
\begin{figure}[t]
		\centering
		\begin{tabular}[b]{@{}p{0.46\linewidth}@{}}
			\centering
			\includegraphics[width=\linewidth]{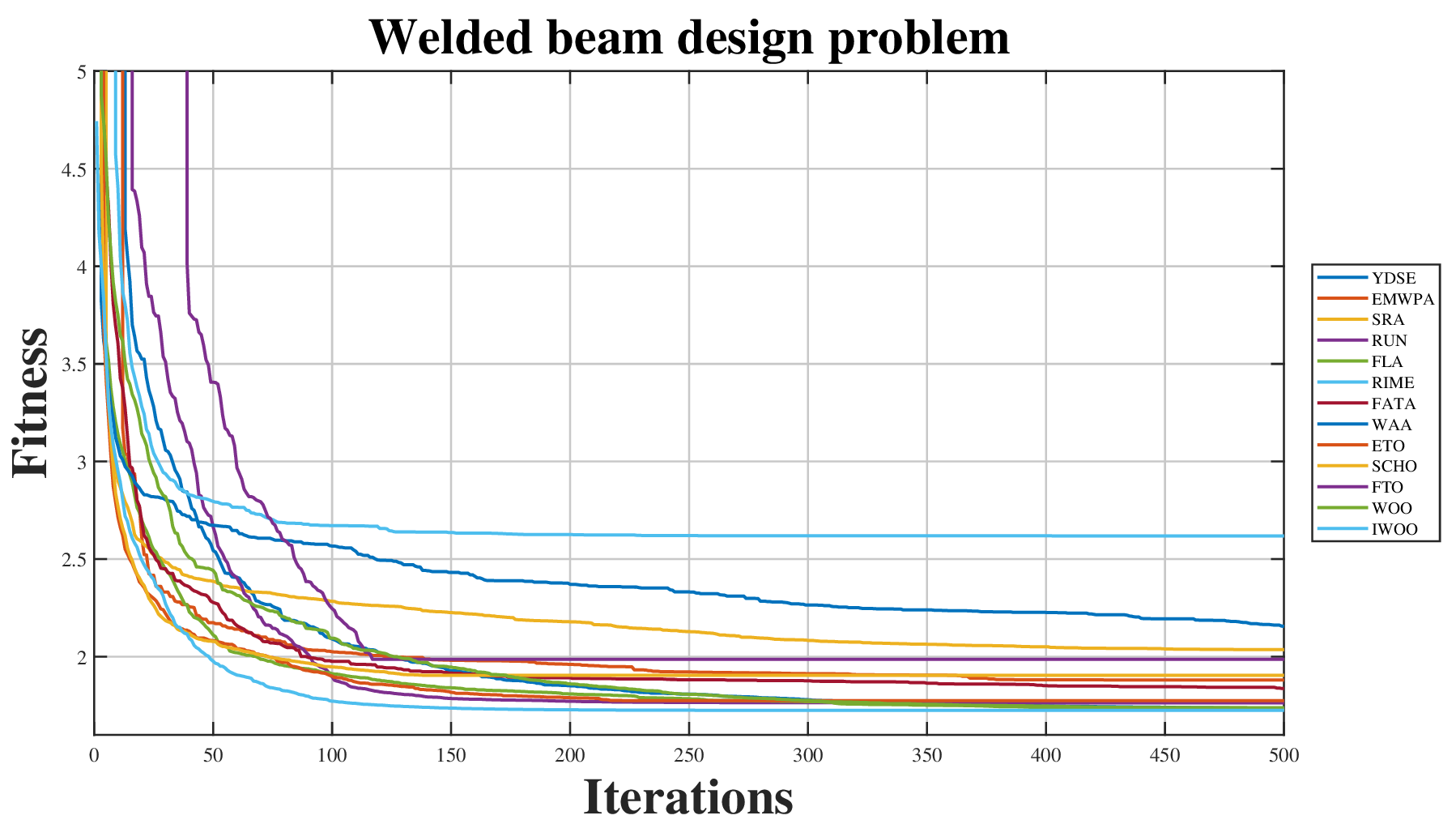}
			\\
			\small (a) Convergence curves
		\end{tabular}
		\hfill
		\begin{tabular}[b]{@{}p{0.46\linewidth}@{}}
			\centering
			\includegraphics[width=\linewidth]{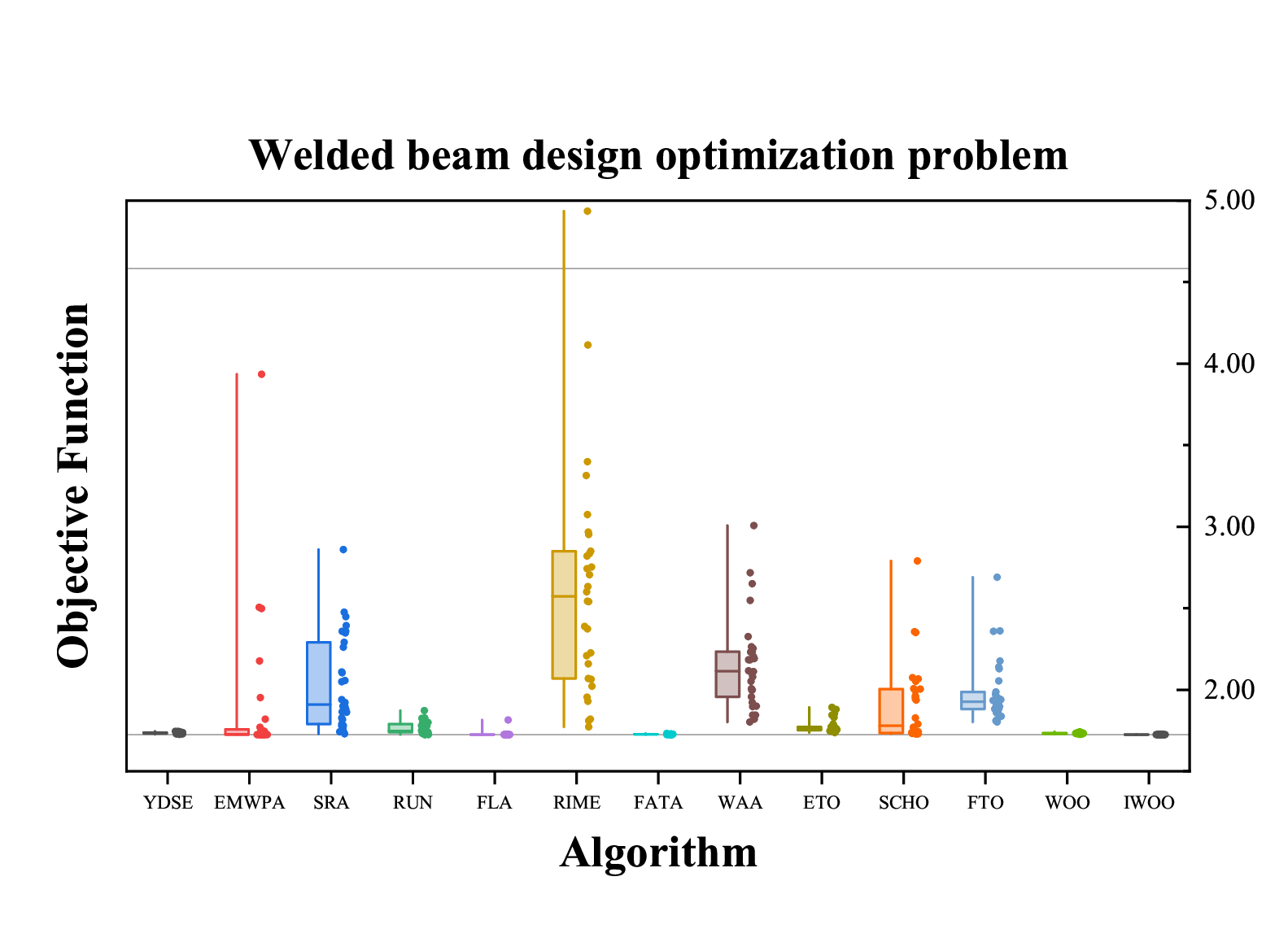}
			\\
			\small (b) Box plots
		\end{tabular}
		\caption{Performance comparison for the welded beam problem}
		\label{fig:welded_results}
	\end{figure}

	The average convergence curves and box plots of all algorithms are shown in Fig.~\ref{fig:welded_results}. IWOO rapidly converges to the low-cost region and exhibits a compact distribution across independent runs, further demonstrating its stable optimization performance.
	
	\subsection{Speed reducer design problem}
	\label{subsec:speed_reducer}
	
	\begingroup
	\allowdisplaybreaks[4]
	
	The speed reducer design problem is a complex constrained optimization problem in mechanical engineering that aims to minimize the total weight of the reducer while satisfying gear strength, shaft deflection, and dimensional constraints~\cite{Golinski1973AdaptiveOptimization}. The problem involves seven key design variables, which are defined as
	
	\begin{equation}
		\boldsymbol{x}
		=
		[x_1,x_2,x_3,x_4,x_5,x_6,x_7]
		=
		[b,m,z,l_1,l_2,d_1,d_2]
		\label{eq:speed_reducer_variables}
	\end{equation}
	where $b$, $m$, and $z$ denote the gear face width, module, and number of teeth, respectively; $l_1$ and $l_2$ represent the bearing spans of the two shafts; and $d_1$ and $d_2$ denote the corresponding shaft diameters. The three-dimensional geometric model of the speed reducer is illustrated in Fig.~\ref{fig:speed_reducer_model}.
	
	The objective function of the speed reducer design problem is given by
	\begingroup
	\small
	\begin{equation}
		\begin{aligned}
			\min f(\boldsymbol{x})={}&
			0.7854x_1x_2^2\bigl(3.3333x_3^2\\
			&\quad+14.9334x_3-43.0934\bigr)
			-1.508x_1(x_6^2+x_7^2)\\
			&+7.4777(x_6^3+x_7^3)
			+0.7854(x_4x_6^2+x_5x_7^2).
		\end{aligned}
		\label{eq:speed_reducer_objective}
	\end{equation}
	\endgroup
	
	\par\vspace{4pt}
	\noindent
	\begin{minipage}{\linewidth}
		\centering
		\captionsetup{font=footnotesize,skip=2pt}
		
		\includegraphics[
		width=0.72\linewidth,
		height=0.22\textheight,
		keepaspectratio
		]{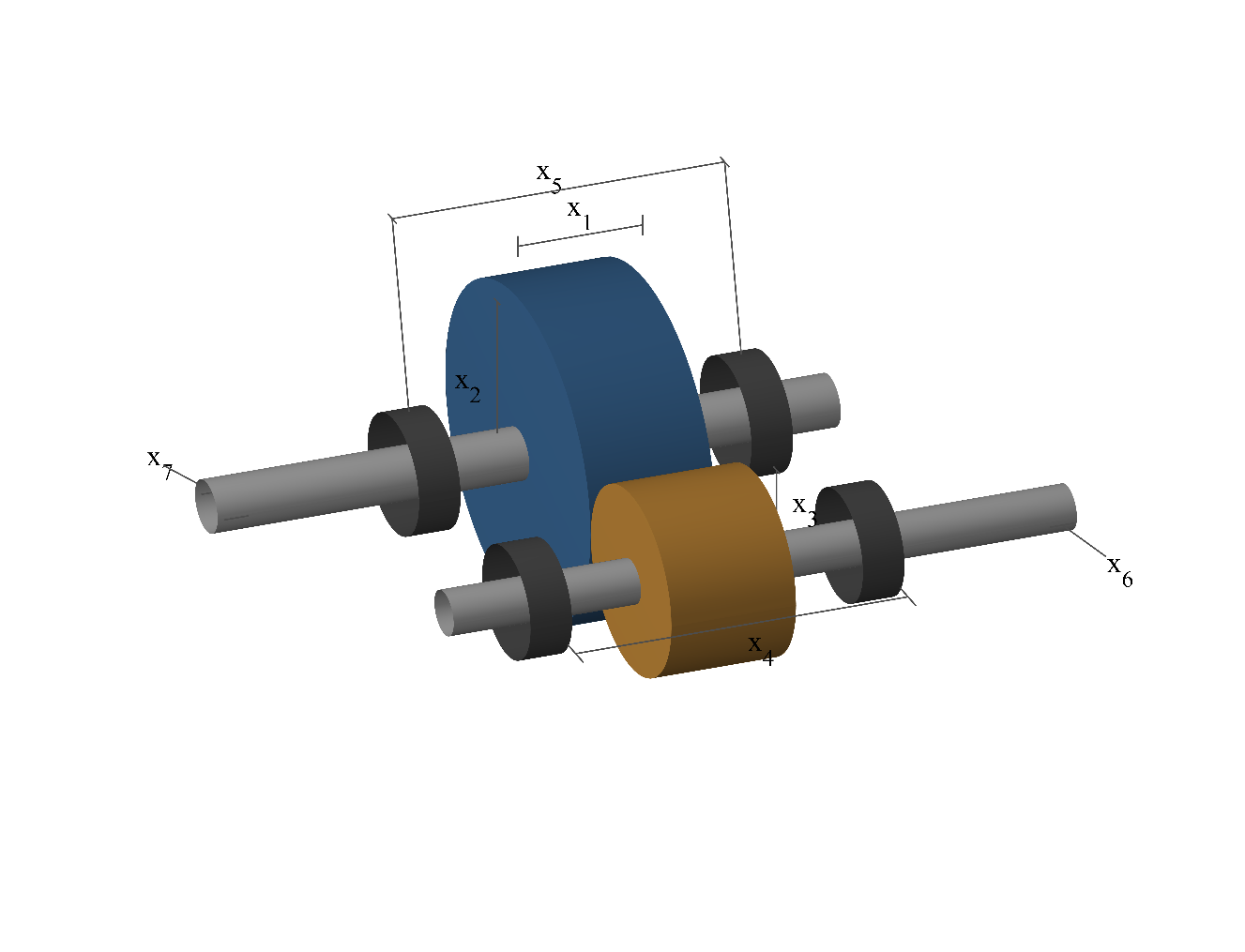}
		
		\captionof{figure}{3D geometric model of the speed reducer}
		\label{fig:speed_reducer_model}
	\end{minipage}
	\par\vspace{4pt}
	
	The corresponding constraints are expressed as
	\begin{align}
		&\left\{
		\begin{aligned}
			g_1(\boldsymbol{x})
			&=
			\frac{27}{x_1x_2^2x_3}-1
			\leq 0\\
			g_2(\boldsymbol{x})
			&=
			\frac{397.5}{x_1x_2^2x_3^2}-1
			\leq 0\\
			g_3(\boldsymbol{x})
			&=
			\frac{1.93x_4^3}{x_2x_3x_6^4}-1
			\leq 0\\
			g_4(\boldsymbol{x})
			&=
			\frac{1.93x_5^3}{x_2x_3x_7^4}-1
			\leq 0\\
			g_5(\boldsymbol{x})
			&=
			\frac{1}{110x_6^3}
			\sqrt{
				\left(
				\frac{745x_4}{x_2x_3}
				\right)^2
				+16.9\times10^6
			}
			-1
			\leq 0\\
			g_6(\boldsymbol{x})
			&=
			\frac{1}{85x_7^3}
			\sqrt{
				\left(
				\frac{745x_5}{x_2x_3}
				\right)^2
				+157.5\times10^6
			}
			-1
			\leq 0
		\end{aligned}
		\right.
		\label{eq:speed_reducer_constraints_a}
		\\
		&\left\{
		\begin{aligned}
			g_7(\boldsymbol{x})
			&=
			\frac{x_2x_3}{40}-1
			\leq 0\\
			g_8(\boldsymbol{x})
			&=
			\frac{5x_2}{x_1}-1
			\leq 0\\
			g_9(\boldsymbol{x})
			&=
			\frac{x_1}{12x_2}-1
			\leq 0\\
			g_{10}(\boldsymbol{x})
			&=
			\frac{1.5x_6+1.9}{x_4}-1
			\leq 0\\
			g_{11}(\boldsymbol{x})
			&=
			\frac{1.1x_7+1.9}{x_5}-1
			\leq 0
		\end{aligned}
		\right.
		\label{eq:speed_reducer_constraints_b}
	\end{align}
	
	The design variables are bounded by
	\begin{equation}
		\begin{alignedat}{3}
			2.6 &\leq x_1\leq 3.6,
			&\qquad
			0.7 &\leq x_2\leq 0.8,
			&\qquad
			x_3 &\in\{17,18,\ldots,28\},\\
			7.3 &\leq x_4\leq 8.3,
			&
			7.8 &\leq x_5\leq 8.3,
			&
			2.9 &\leq x_6\leq 3.9,\\
			5.0 &\leq x_7\leq 5.5
		\end{alignedat}
		\label{eq:speed_reducer_bounds}
	\end{equation}
	
	The reference optimum objective value for this problem is $2996.348165$.
	
	\begin{tabular}{@{}p{\linewidth}@{}}
		\centering
		\captionof{table}{Experimental results of the speed reducer design problem}
		\label{tab:speed_reducer_results}
		
		\scriptsize
		\setlength{\tabcolsep}{3.7pt}
		\renewcommand{\arraystretch}{0.92}
		\resizebox{\linewidth}{!}{%
			\begin{tabular}{lcccc}
				\toprule
				Algorithm & Min Cost & Mean Cost & Std & Fr-rank \\
				\midrule
				IWOO  & \textbf{2996.3481650} & \textbf{2996.3481650} & \textbf{$7.3126\times10^{-12}$} & \textbf{1.0000} \\
				FLA   & 2996.3481651 & 3009.7932696 & $4.5698\times10^{1}$  & 3.4333 \\
				RUN   & 2996.3528184 & 2996.3981962 & $1.5234\times10^{-1}$ & 3.6000 \\
				WOO   & 2996.5657798 & 2996.9495451 & $2.0715\times10^{-1}$ & 5.1667 \\
				YDSE  & 2996.6092269 & 2997.7008759 & $7.5765\times10^{-1}$ & 6.2000 \\
				EMWPA & 2996.3481650 & 3124.8822159 & $4.5130\times10^{2}$  & 6.3667 \\
				SRA   & 2996.3529135 & 3077.6628088 & $1.8211\times10^{2}$  & 6.8333 \\
				RIME  & 2996.6613755 & 2999.7828663 & $2.3555\times10^{0}$  & 7.1333 \\
				FATA  & 2996.7428509 & 3001.4432478 & $3.2569\times10^{0}$  & 7.7333 \\
				SCHO  & 3003.8523387 & 3013.5391219 & $5.3105\times10^{0}$  & 9.7333 \\
				WAA   & 3005.6887646 & 3075.5136959 & $1.6434\times10^{2}$  & 10.6000 \\
				ETO   & 3012.1817753 & 3023.5566178 & $7.2883\times10^{0}$  & 10.9000 \\
				FTO   & 3029.1649028 & 3111.8807501 & $5.1209\times10^{1}$  & 12.3000 \\
				\bottomrule
			\end{tabular}
		}
	\end{tabular}

	As shown in Table~\ref{tab:speed_reducer_results}, IWOO achieves both the minimum cost and the mean cost of $2996.3481650$, reaching the reference optimum value. Moreover, IWOO obtains an extremely small standard deviation of $7.3126\times10^{-12}$ and ranks first with an Fr-rank of $1.0000$. These results demonstrate that IWOO provides high solution accuracy and excellent stability on the speed reducer design problem.
	
	The average convergence curves and box plots of all algorithms are presented in Fig.~\ref{fig:speed_reducer_curves}. IWOO converges rapidly toward the optimum region and exhibits a highly concentrated distribution across independent runs, further demonstrating its strong convergence performance and robustness.
	
	\begin{figure}[t]
		\centering
		\begin{tabular}[b]{@{}p{0.46\linewidth}@{}}
			\centering
			\includegraphics[width=\linewidth]{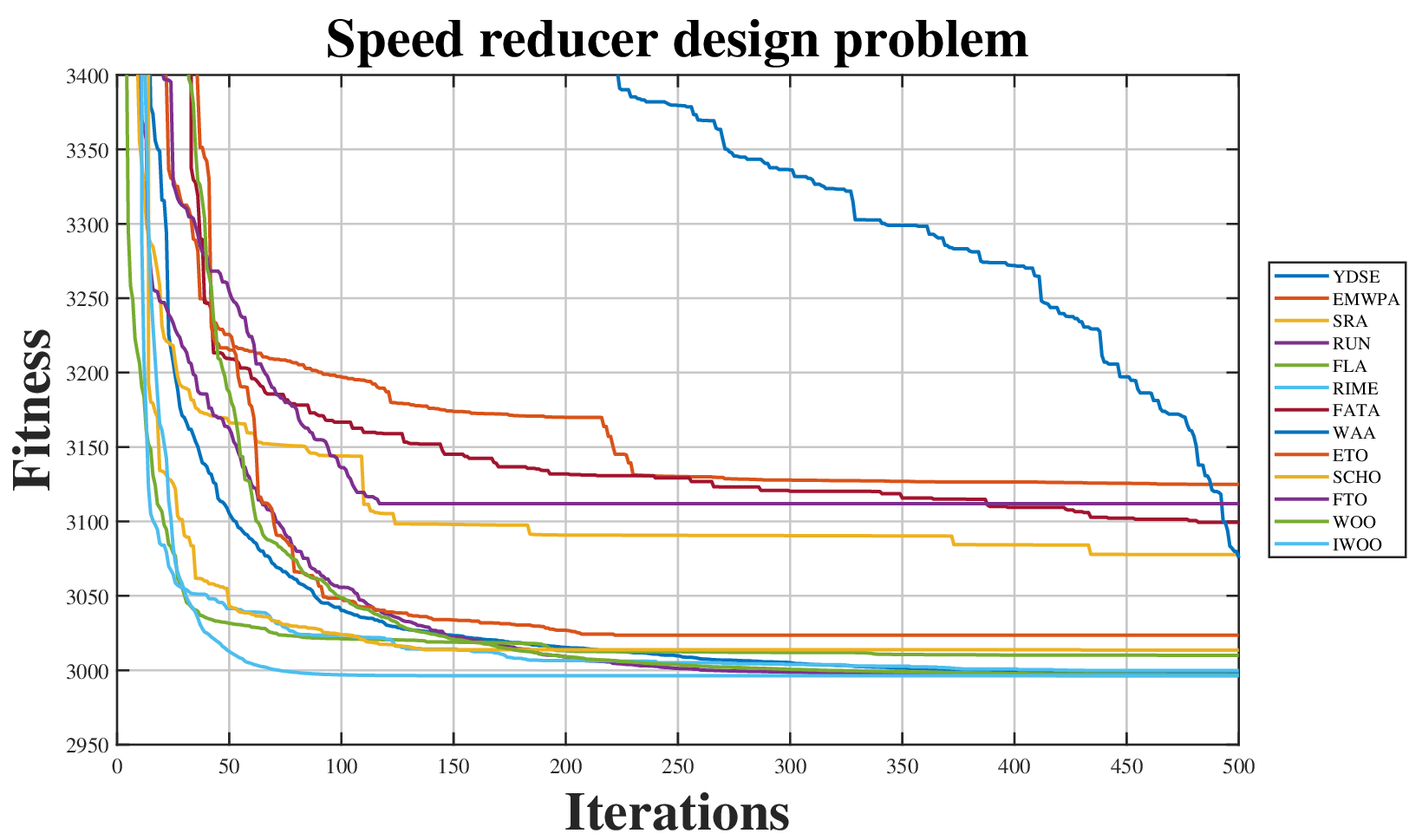}
			\\
			\small (a) Convergence curves
		\end{tabular}
		\hfill
		\begin{tabular}[b]{@{}p{0.46\linewidth}@{}}
			\centering
			\includegraphics[width=\linewidth]{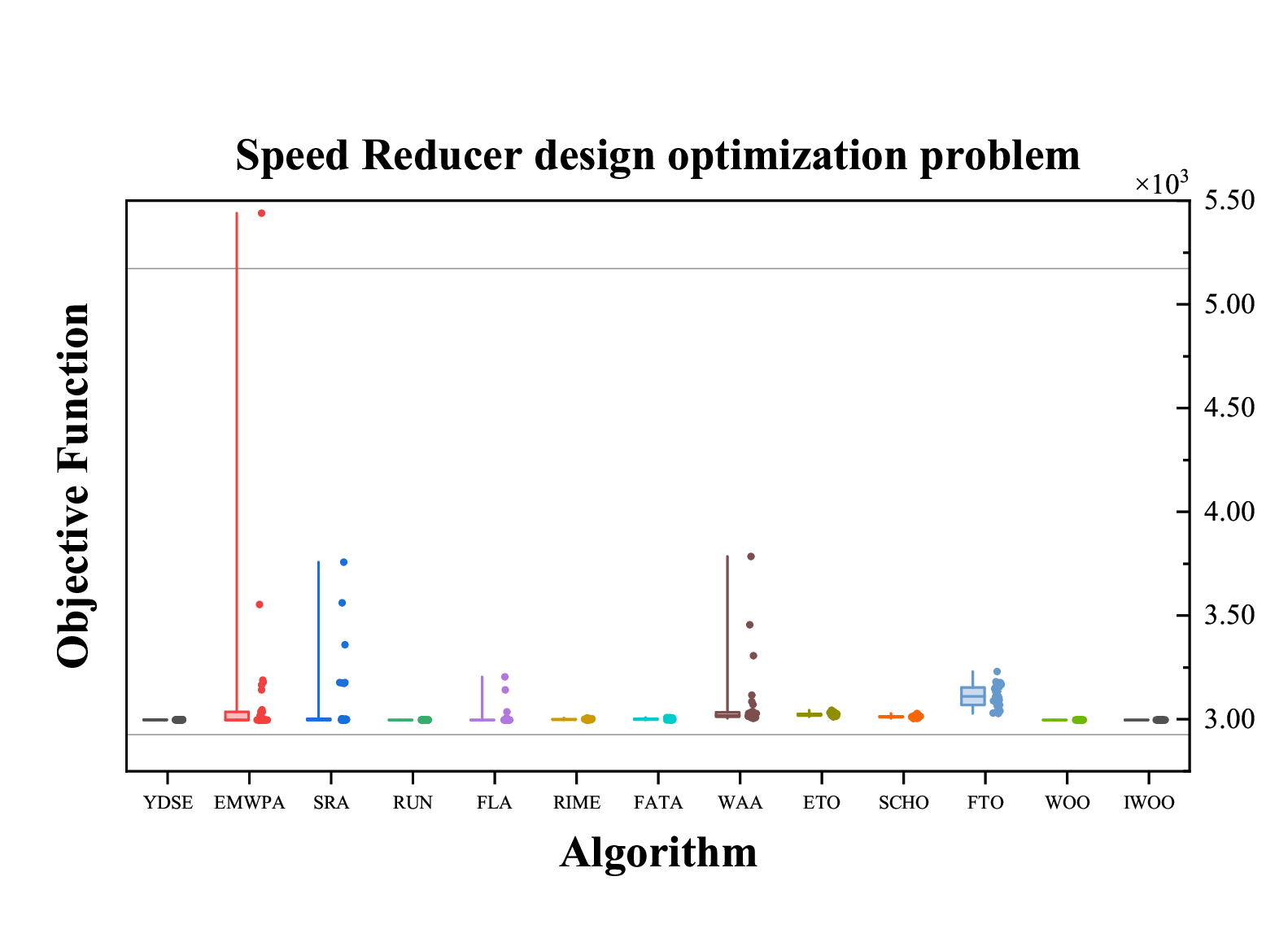}
			\\
			\small (b) Box plots
		\end{tabular}
		\captionof{figure}{Performance comparison on the speed reducer design problem}
		\label{fig:speed_reducer_curves}
	\end{figure}
	\endgroup

	\subsection{Gear train design problem}
	\label{subsec:gear_train}
	
	The gear train design problem is a classical integer engineering optimization problem, originally formulated by Sandgren~\cite{Sandgren1990DiscreteProgramming} and subsequently adopted by Gandomi et al.~\cite{Gandomi2013CuckooSearch} for structural optimization studies. The objective is to determine the numbers of teeth of four gears such that the actual gear ratio is as close as possible to a prescribed target value.
	
	The four design variables are the numbers of teeth $T_d$, $T_b$, $T_a$, and $T_f$, which are defined within the following integer range
	
	\begin{equation}
		\boldsymbol{x}
		=
		[T_d,T_b,T_a,T_f],
		\qquad
		T_d,T_b,T_a,T_f\in\{12,13,\ldots,60\}
		\label{eq:gear_train_vars}
	\end{equation}
	
	The actual gear ratio is given by $T_dT_b/(T_aT_f)$, whereas the target gear ratio is $1/6.931$, Accordingly, the mathematical formulation of the problem is expressed as
	
	\begin{equation}
		\min f(\boldsymbol{x})
		=
		\left(
		\frac{1}{6.931}
		-
		\frac{T_dT_b}{T_aT_f}
		\right)^2
		\label{eq:gear_train_obj}
	\end{equation}
	
\begin{figure}[t]
		\centering
		\captionsetup{skip=3pt}
		
		\begin{minipage}[b]{0.49\linewidth}
			\centering
			\includegraphics[
			width=\linewidth
			]{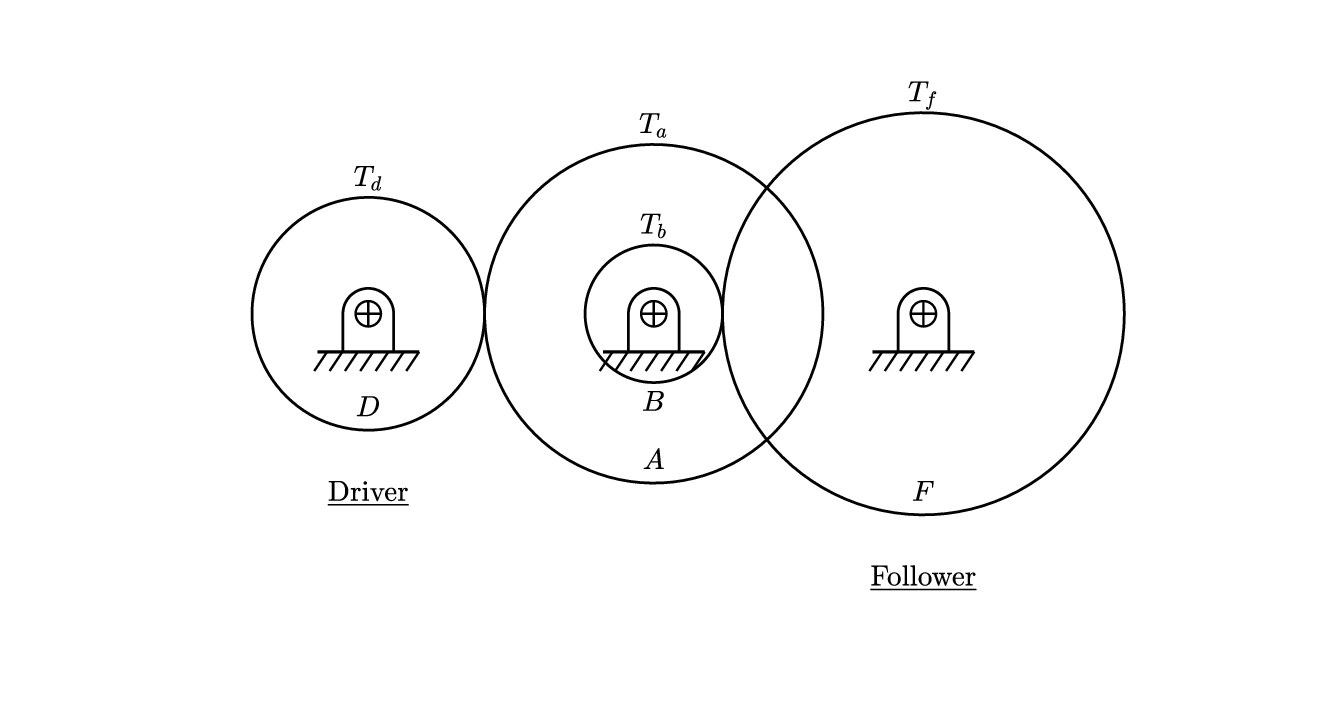}
			
			\par\vspace{2pt}
			{\small (a) 2D structural model}
		\end{minipage}%
		\hfill
		\begin{minipage}[b]{0.49\linewidth}
			\centering
			\includegraphics[
			width=\linewidth
			]{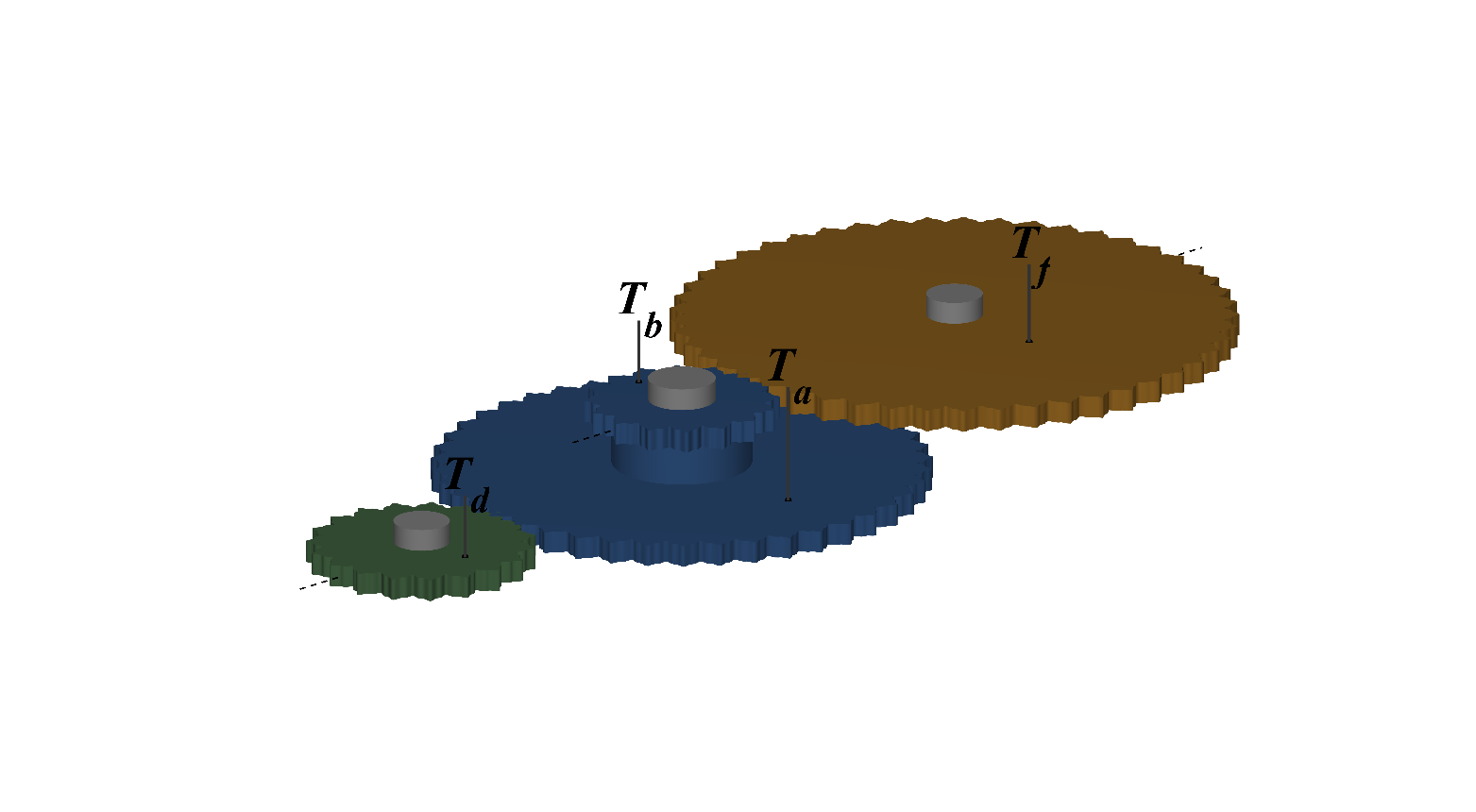}
			
			\par\vspace{2pt}
			{\small (b) 3D geometric model}
		\end{minipage}
		
		\caption{Gear train models}
		\label{fig:gear_train_model}
	\end{figure}

	The two-dimensional structural configuration and three-dimensional geometric model of the gear train system are illustrated in Fig.~\ref{fig:gear_train_model}
	
	Table ~\ref{tab:gear_train_results} summarizes the best gear combinations obtained by the compared algorithms together with their statistical results.
	
	%
	
	\begin{table}[H]
		\centering
		\caption{Experimental results of the gear train design optimization problem}
		\label{tab:gear_train_results}

		\scriptsize
		\setlength{\tabcolsep}{3.4pt}
		\renewcommand{\arraystretch}{0.96}
		
		\resizebox{0.94\linewidth}{!}{
			\begin{tabular}{lcccccccc}
				\toprule
				Algorithm & $T_d$ & $T_b$ & $T_a$ & $T_f$ & Best Obj. & Mean Obj. & Std & Fr-rank \\
				\midrule
				IWOO  & 16 & 19 & 49 & 43 & $\mathbf{2.7009\times10^{-12}}$ & $\mathbf{1.7767\times10^{-11}}$ & $\mathbf{2.8593\times10^{-11}}$ & $\mathbf{1.8833}$ \\
				WOO   & 19 & 16 & 49 & 43 & $2.7009\times10^{-12}$ & $2.3487\times10^{-10}$ & $4.1338\times10^{-10}$ & $3.2333$ \\
				FATA  & 19 & 16 & 43 & 49 & $2.7009\times10^{-12}$ & $4.0517\times10^{-10}$ & $4.5432\times10^{-10}$ & $4.3333$ \\
				YDSE  & 13 & 30 & 51 & 53 & $2.3078\times10^{-11}$ & $3.3179\times10^{-10}$ & $3.7982\times10^{-10}$ & $4.3833$ \\
				RUN   & 16 & 19 & 49 & 43 & $2.7009\times10^{-12}$ & $5.6524\times10^{-10}$ & $4.7893\times10^{-10}$ & $5.0333$ \\
				ETO   & 13 & 30 & 51 & 53 & $2.3078\times10^{-11}$ & $1.3321\times10^{-9}$  & $8.6336\times10^{-10}$ & $7.4000$ \\
				SCHO  & 13 & 20 & 34 & 53 & $2.3078\times10^{-11}$ & $3.1769\times10^{-9}$  & $5.6969\times10^{-9}$  & $7.4167$ \\
				RIME  & 13 & 20 & 53 & 34 & $2.3078\times10^{-11}$ & $3.4090\times10^{-9}$  & $6.6453\times10^{-9}$  & $7.6833$ \\
				FLA   & 19 & 16 & 43 & 49 & $2.7009\times10^{-12}$ & $4.4827\times10^{-9}$  & $6.6993\times10^{-9}$  & $8.4333$ \\
				FTO   & 13 & 20 & 53 & 34 & $2.3078\times10^{-11}$ & $1.0184\times10^{-8}$  & $1.2645\times10^{-8}$  & $9.5333$ \\
				EMWPA & 19 & 16 & 43 & 49 & $2.7009\times10^{-12}$ & $1.8773\times10^{-7}$  & $3.5074\times10^{-7}$  & $10.2667$ \\
				SRA   & 16 & 19 & 43 & 49 & $2.7009\times10^{-12}$ & $7.0679\times10^{-8}$  & $1.0440\times10^{-7}$  & $10.6500$ \\
				WAA   & 12 & 37 & 57 & 54 & $8.8876\times10^{-10}$ & $7.6341\times10^{-8}$  & $2.1618\times10^{-7}$  & $10.7500$ \\
				\bottomrule
			\end{tabular}
		}
	\end{table}

	
	As shown in Table~\ref{tab:gear_train_results}, IWOO obtains a best objective value of $2.7009\times10^{-12}$, which is tied with the best values achieved by several compared algorithms. More importantly, IWOO achieves the lowest mean objective value of $1.7767\times10^{-11}$ and the smallest standard deviation of $2.8593\times10^{-11}$, together with the best Fr-rank of $1.8833$ among all compared algorithms. These results demonstrate that IWOO provides both high solution accuracy and strong stability on the gear train design problem.
	
\begin{figure}[H]
		\centering
		
		\begin{tabular}[c]{@{}p{0.47\linewidth}@{}}
			\centering
			\includegraphics[
			width=0.94\linewidth,
			keepaspectratio
			]{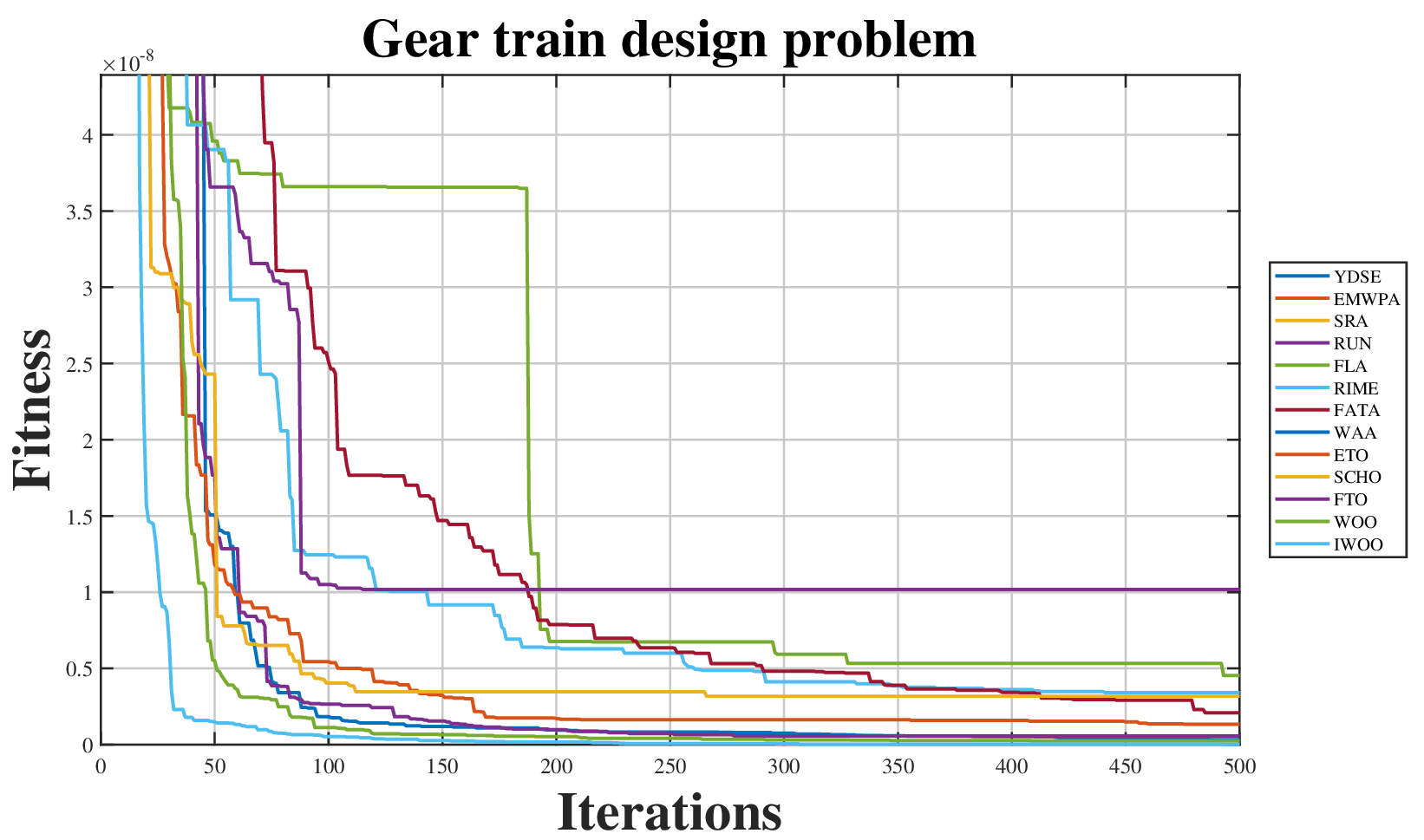}
			\\
			\small (a)
		\end{tabular}
		\hfill
		\begin{tabular}[c]{@{}p{0.47\linewidth}@{}}
			\centering
			\includegraphics[
			width=0.94\linewidth,
			keepaspectratio
			]{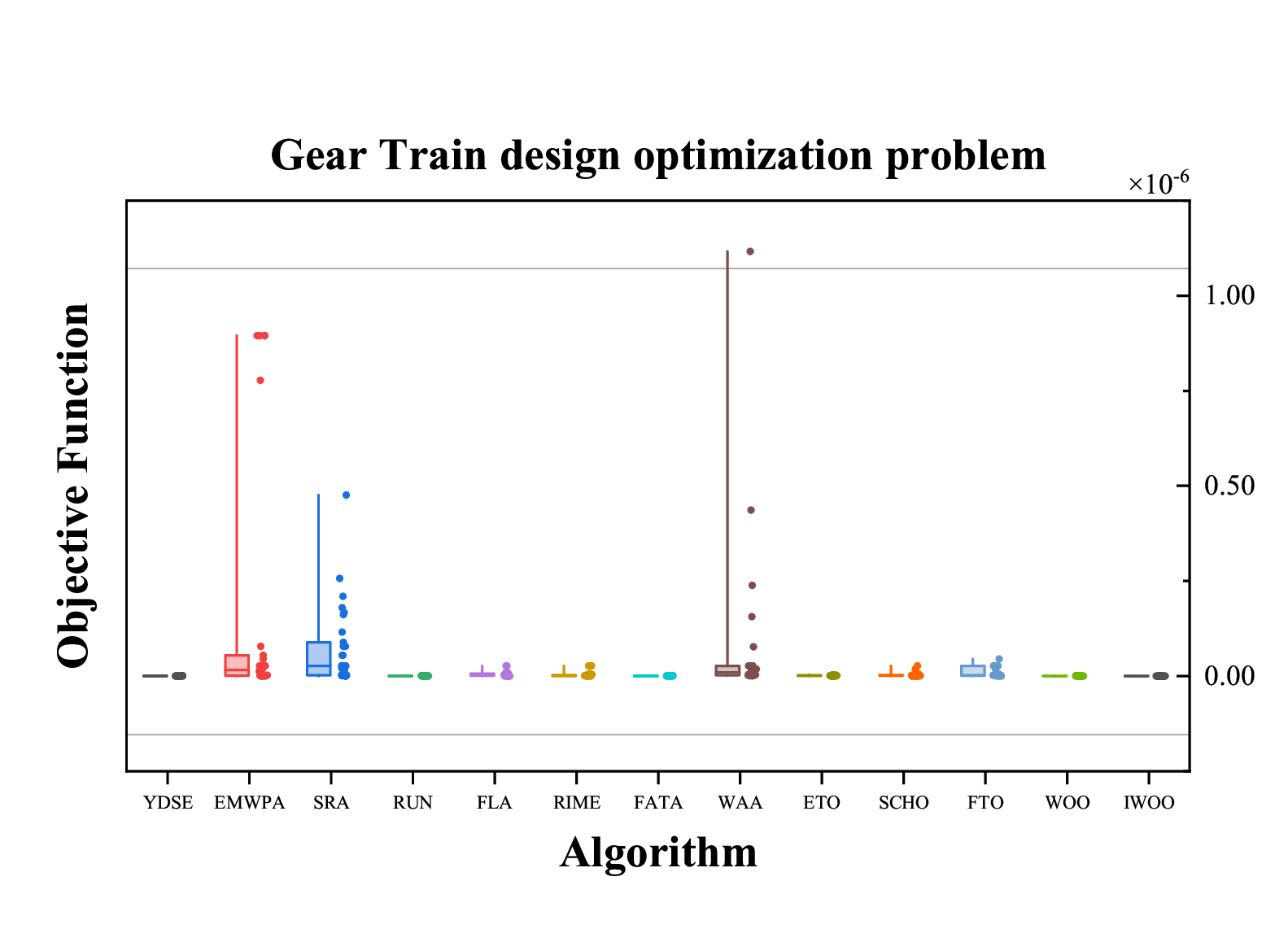}
			\\
			\small (b)
		\end{tabular}
		
		\caption{Performance comparison on the gear train design optimization problem:
			(a) convergence curves; (b) box plots}
		\label{fig:gear_train_curves}
	\end{figure}

	The average convergence curves and box plots shown in Fig.~\ref{fig:gear_train_curves} further indicate that IWOO converges rapidly toward the optimal region and exhibits a relatively concentrated distribution across independent runs, confirming its favorable convergence behavior and stability.
	
	The optimal design obtained by IWOO corresponds to
	$T_d=16$, $T_b=19$, $T_a=49$, and $T_f=43$,
	with an objective value of $2.7009\times10^{-12}$. This result indicates that the obtained gear configuration yields an actual gear ratio extremely close to the prescribed target ratio.
	
	\subsection{Economic load dispatch problem}
	\label{subsec:economic_load_dispatch}
	
	\begingroup
	
	The economic load dispatch (ELD) problem is a classical optimization problem in power system operation that aims to minimize the total fuel cost by optimally allocating the active power outputs of the generating units while satisfying the system load demand, power balance, and generating capacity constraints. In practical generating units, the opening of steam turbine valves produces valve-point effects, resulting in nonsmooth and nonconvex fuel-cost curves with multiple local minima and thereby increasing the difficulty of conventional gradient-based optimization methods. In this study, the 13-unit ELD test system with valve-point effects reported by Sinha et al.~\cite{Sinha2003EconomicLoadDispatch} is adopted. The system load demand is further set to $2520$~MW to construct a higher-load ELD scenario.
	
	The design variables are the active power outputs of the 13 generating units:
	\begin{equation}
		\boldsymbol{P}
		=
		[P_1,P_2,\ldots,P_{13}],
		\qquad
		P_D=2520\ \mathrm{MW}
		\label{eq:eld_variables_demand}
	\end{equation}
	where $P_i$ denotes the active power output of the $i$th generating unit in MW. Considering the valve-point effects, the fuel-cost function of the $i$th generating unit is given by
	\begin{equation}
		F_i(P_i)
		=
		a_iP_i^2+b_iP_i+c_i
		+
		\left|
		e_i\sin\left(f_i(P_i^{\min}-P_i)\right)
		\right|
		\label{eq:eld_unit_cost}
	\end{equation}
	where $a_i$, $b_i$, and $c_i$ are the conventional quadratic fuel-cost coefficients, $e_i$ and $f_i$ are the valve-point effect coefficients, and $P_i^{\min}$ denotes the minimum power output of the $i$th generating unit. The objective function of the ELD problem is formulated as
	\begin{equation}
		\min F(\boldsymbol{P})
		=
		\sum_{i=1}^{13}F_i(P_i)
		\label{eq:eld_objective}
	\end{equation}
	
	Transmission losses are neglected in this study, i.e., $P_L=0$. Therefore, the power balance and generating capacity constraints are expressed as
	\begin{equation}
		\begin{cases}
			\displaystyle
			\sum_{i=1}^{13}P_i=P_D\\
			P_i^{\min}\leq P_i\leq P_i^{\max},
			\quad i=1,2,\ldots,13
		\end{cases}
		\label{eq:eld_constraints}
	\end{equation}
	
	To satisfy the power balance constraint exactly, a dependent-unit strategy is adopted, with Unit 1 selected as the dependent generating unit. Its power output is determined by the remaining generating units as
	\begin{equation}
		P_1
		=
		P_D-\sum_{i=2}^{13}P_i
		\label{eq:eld_dependent_unit}
	\end{equation}
	
	A penalty function is employed to handle violations of the generating capacity constraints. The corresponding fitness function is defined as
	\begin{equation}
		\left\{
		\begin{aligned}
			\mathrm{Fitness}
			&=
			F(\boldsymbol{P})
			+
			\lambda
			\left[
			\left(\sum_{i=1}^{13}P_i-P_D\right)^2
			+
			\sum_{i=1}^{13}V_i^2
			\right],\\
			V_i
			&=
			\max(0,P_i^{\min}-P_i)
			+
			\max(0,P_i-P_i^{\max})
		\end{aligned}
		\right.
		\label{eq:eld_fitness_violation}
	\end{equation}
	where $\lambda$ is the penalty factor, which is set to $\lambda=1000$ in this study, and $V_i$ measures the degree of violation of the lower and upper generation limits of the $i$th unit.
	
	The statistical results of all compared algorithms are reported in Table~\ref{tab:eld_results}.
	
	\begin{tabular}{@{}p{\linewidth}@{}}
		\centering
		\captionof{table}{Experimental results of the economic load dispatch problem}
		\label{tab:eld_results}
		
		\scriptsize
		\setlength{\tabcolsep}{3.3pt}
		\renewcommand{\arraystretch}{0.92}
		\resizebox{\linewidth}{!}{%
			\begin{tabular}{lcccccc}
				\toprule
				Algorithm & Min Cost & Mean Cost & Std & Feasible Rate(\%) & MaxVio & Fr-rank \\
				\midrule
				IWOO  & \textbf{24170.260956} & \textbf{24196.960330} & \textbf{$2.3303\times10^{1}$} & 100.00 & \textbf{0.0000} & \textbf{1.0333} \\
				WAA   & 24276.973399 & 24289.633520 & $2.4855\times10^{1}$ & 100.00 & 0.0000 & 2.2333 \\
				RIME  & 24214.349423 & 24385.274607 & $6.8755\times10^{1}$ & 100.00 & 0.0000 & 4.1000 \\
				WOO   & 24290.341422 & 24397.293201 & $4.6913\times10^{1}$ & 100.00 & 0.0000 & 4.3667 \\
				FATA  & 24253.464392 & 24460.136197 & $1.1913\times10^{2}$ & 100.00 & 0.0000 & 5.6333 \\
				RUN   & 24329.025670 & 24470.246052 & $7.5482\times10^{1}$ & 100.00 & 0.0000 & 6.0667 \\
				YDSE  & 24473.729383 & 24566.480798 & $3.2411\times10^{1}$ & 100.00 & 0.0000 & 8.0667 \\
				SRA   & 24231.754371 & 24626.962103 & $1.8147\times10^{2}$ & 100.00 & 0.0000 & 8.3000 \\
				SCHO  & 24522.395264 & 24617.175496 & $1.5847\times10^{2}$ & 100.00 & 0.0000 & 8.6667 \\
				FLA   & 24329.627923 & 24689.818229 & $2.2651\times10^{2}$ & 100.00 & 0.0000 & 9.1667 \\
				EMWPA & 24395.017521 & 24795.786049 & $2.8245\times10^{2}$ & 93.33  & $6.6667\times10^{-1}$ & 10.1333 \\
				ETO   & 24542.299582 & 24776.310628 & $1.0931\times10^{2}$ & 100.00 & 0.0000 & 10.6000 \\
				FTO   & 24822.031277 & 25038.717603 & $6.8391\times10^{1}$ & 100.00 & 0.0000 & 12.6333 \\
				\bottomrule
			\end{tabular}%
		}
	\end{tabular}
	\par\vspace{10pt}

	As shown in Table~\ref{tab:eld_results}, IWOO achieves the lowest minimum cost of $24170.260956$ and the lowest mean cost of $24196.960330$ among all compared algorithms. It also obtains a standard deviation of $2.3303\times10^{1}$ and the best Fr-rank of $1.0333$. Moreover, IWOO achieves a feasibility rate of $100\%$ with a maximum constraint violation of zero, indicating that it can consistently obtain low-cost dispatch solutions while fully satisfying the imposed constraints.
	
	\par\smallskip
	\noindent
	\begin{minipage}{\linewidth}
		\centering
		\captionsetup{skip=2pt}
		
		\includegraphics[
		width=\linewidth,
		keepaspectratio
		]{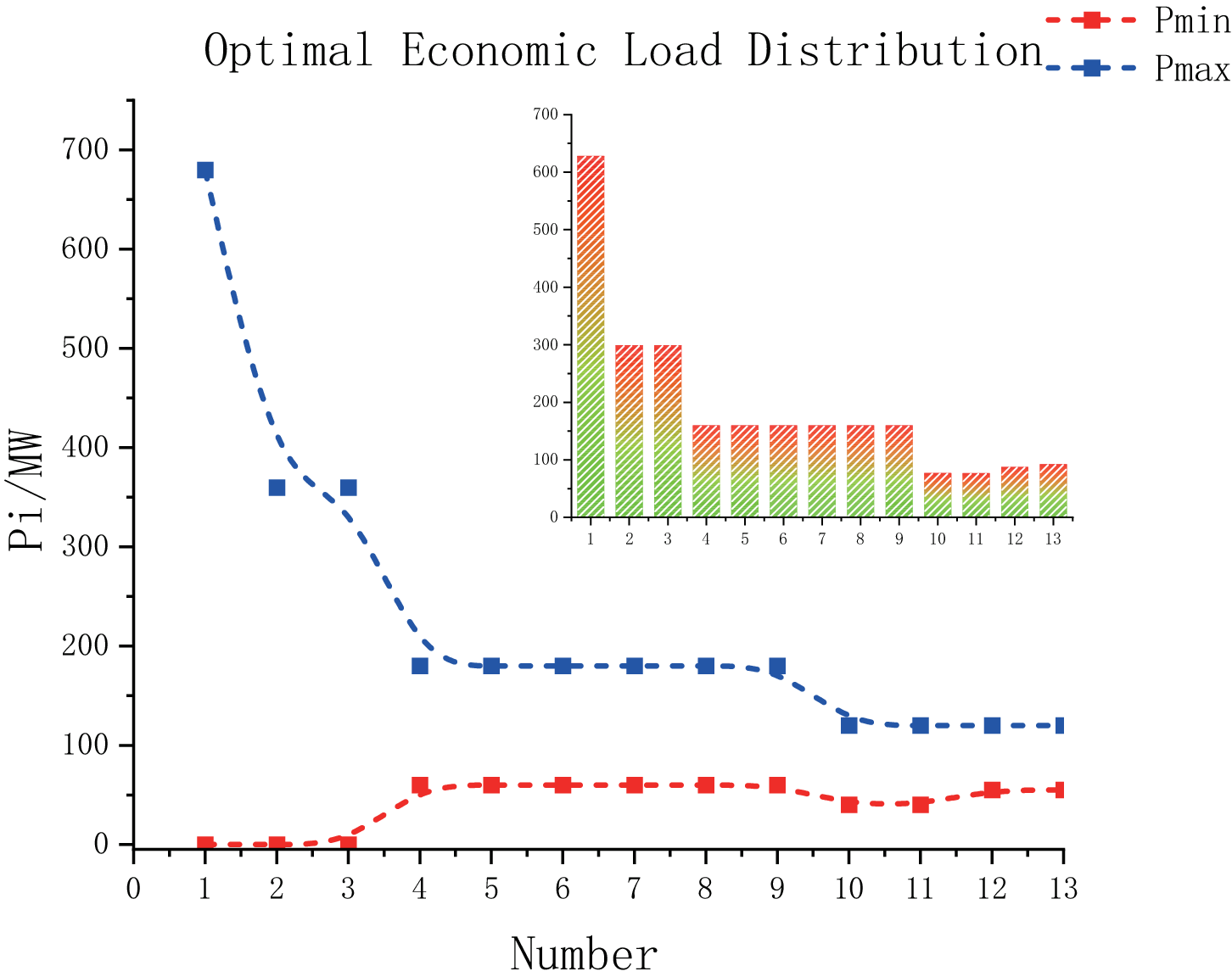}
		
		\captionof{figure}{Optimal economic load dispatch obtained by IWOO}
		\label{fig:eld_dispatch}
	\end{minipage}
	\par\medskip
	
	
	\noindent
	\begin{minipage}{\linewidth}
		\centering
		\captionsetup{skip=2pt}
		
		\captionof{table}{Optimal power schedule obtained by IWOO}
		\label{tab:eld_best_power}
		
		\footnotesize
		\setlength{\tabcolsep}{3pt}
		\renewcommand{\arraystretch}{1.05}
		
		\begin{tabular*}{\linewidth}{
				@{\extracolsep{\fill}}cccc@{}
			}
			\toprule
			Unit & $P_i^{\min}$ & $P_i$ & $P_i^{\max}$ \\
			\midrule
			1  & 0  & 628.318163 & 680 \\
			2  & 0  & 299.193155 & 360 \\
			3  & 0  & 299.198720 & 360 \\
			4  & 60 & 159.733054 & 180 \\
			5  & 60 & 159.728070 & 180 \\
			6  & 60 & 159.730281 & 180 \\
			7  & 60 & 159.733090 & 180 \\
			8  & 60 & 159.731844 & 180 \\
			9  & 60 & 159.718071 & 180 \\
			10 & 40 & 77.390855  & 120 \\
			11 & 40 & 76.967285  & 120 \\
			12 & 55 & 88.159085  & 120 \\
			13 & 55 & 92.397962  & 120 \\
			\midrule
			Total & -- & \textbf{2520.000000} & -- \\
			\bottomrule
		\end{tabular*}
	\end{minipage}
	\par\smallskip
	
	As shown in Fig.~\ref{fig:eld_dispatch} and Table~\ref{tab:eld_best_power}, the power outputs of all 13 generating units obtained by IWOO remain within their respective generation limits, while the total power output is exactly $2520$~MW, which satisfies the specified system load demand.
	
	The average convergence curves and box plots shown in Fig.~\ref{fig:eld_curves} indicate that IWOO converges rapidly toward the low-cost region and exhibits a relatively concentrated distribution across independent runs, further demonstrating its high solution accuracy and strong stability on the economic load dispatch problem.
	
	\par\medskip\noindent
	\begin{minipage}{\linewidth}
		\centering
		\begin{tabular}[b]{@{}p{0.46\linewidth}@{}}
			\centering
			\includegraphics[width=\linewidth]{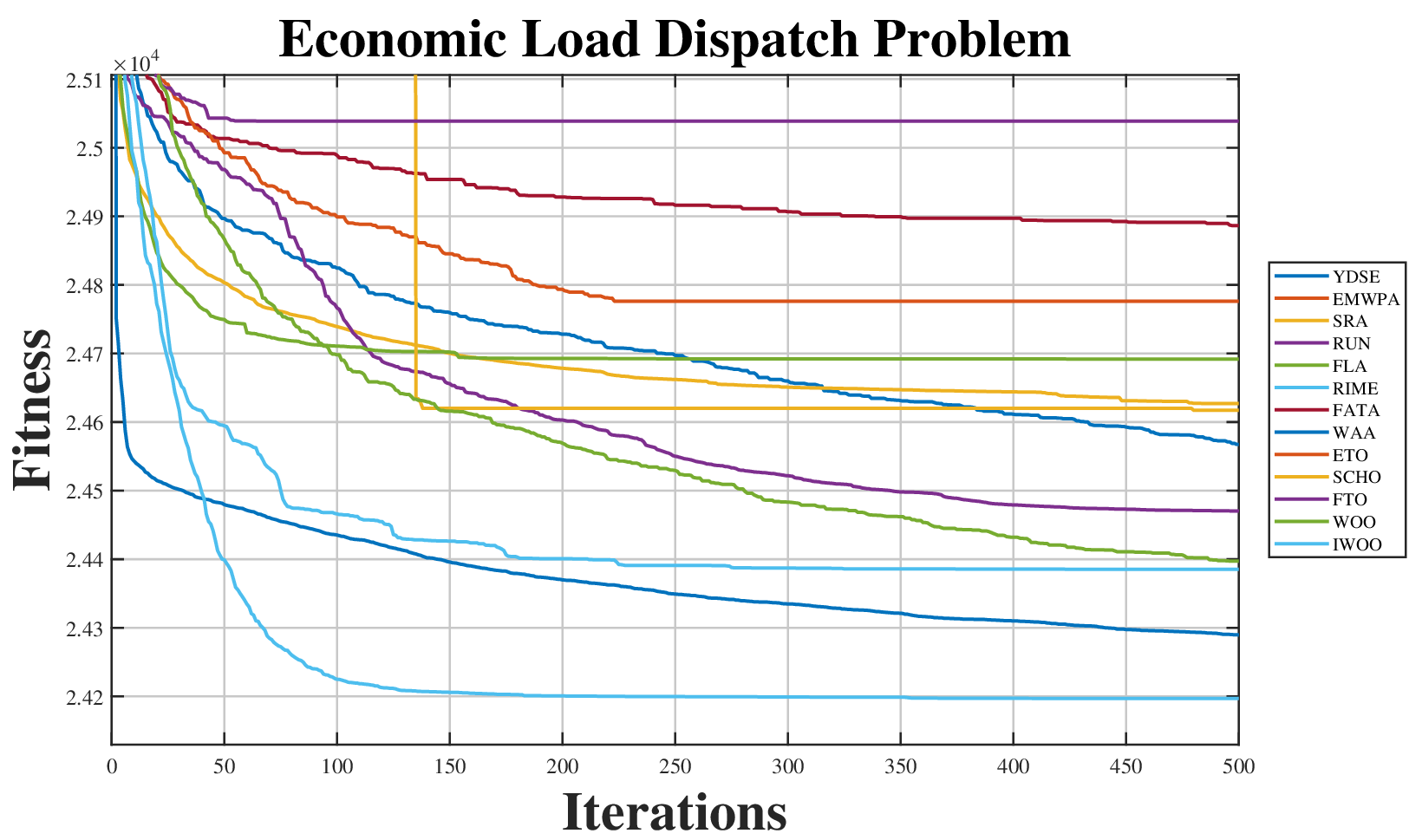}
			\\
			\small (a) Convergence curves
		\end{tabular}
		\hfill
		\begin{tabular}[b]{@{}p{0.46\linewidth}@{}}
			\centering
			\includegraphics[width=\linewidth]{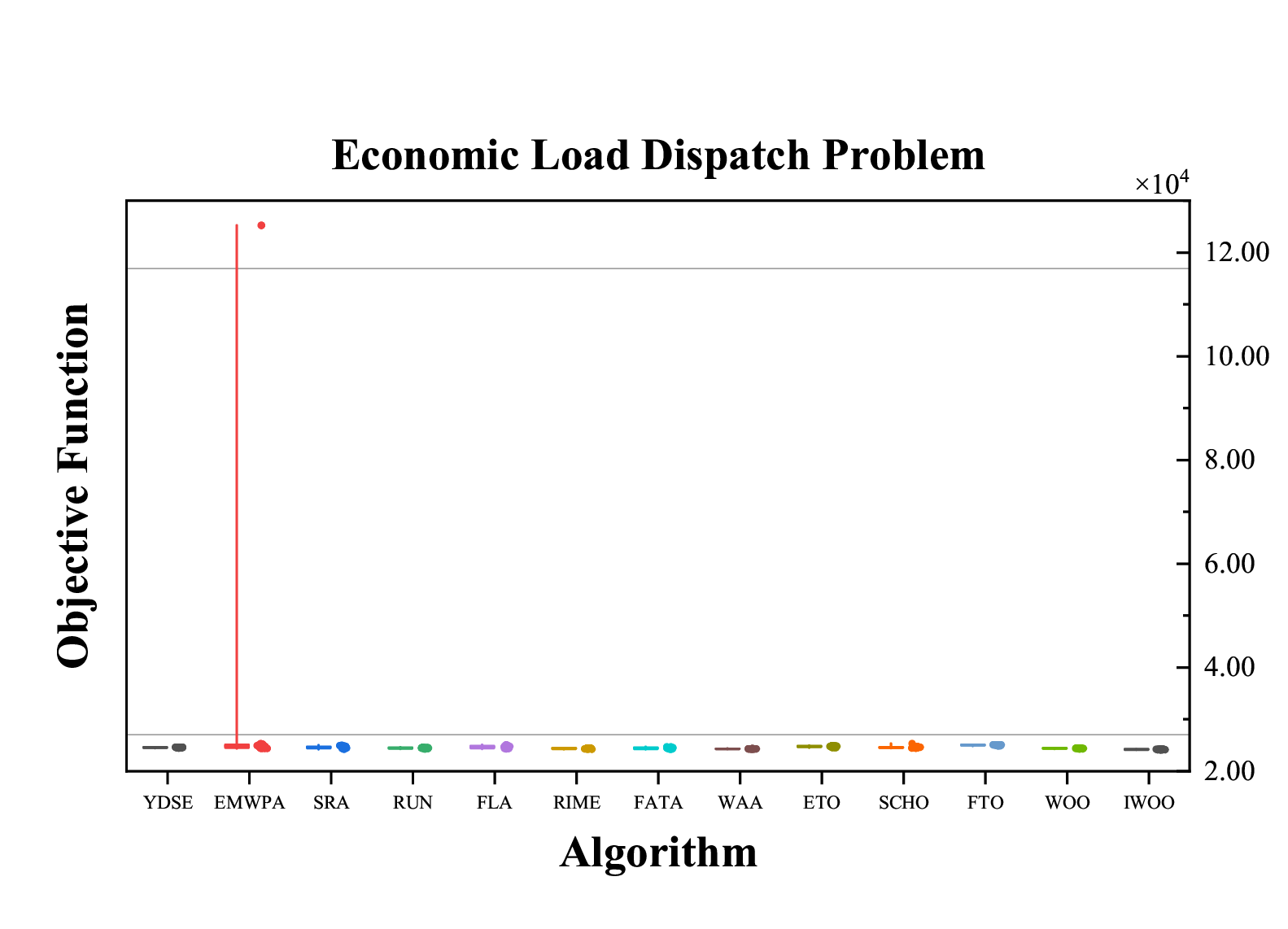}
			\\
			\small (b) Box plots
		\end{tabular}
		\captionof{figure}{Performance comparison on the economic load dispatch problem}
		\label{fig:eld_curves}
	\end{minipage}
	\par\medskip
	\endgroup
	
	\FloatBarrier
	\begingroup
	\raggedbottom
	\par\onecolumn
	\newsavebox{\IWOOWSNSafeBottom}
	\newsavebox{\IWOOWSNSafeText}
	\newsavebox{\IWOOWSNSafeTrial}
	\newsavebox{\IWOOWSNSafeLeft}
	\newsavebox{\IWOOWSNSafeRight}
	\newsavebox{\IWOOWSNSafeTop}
	\newsavebox{\IWOOWSNSafePerformance}
	\newlength{\IWOOWSNSafeWidth}
	\newlength{\IWOOWSNSafeRoom}
	\newlength{\IWOOWSNSafeGoal}
	\newlength{\IWOOWSNSafeLimit}
	\newlength{\IWOOWSNSafeRightRoom}
	\setlength{\IWOOWSNSafeWidth}{\dimexpr(\textwidth-\columnsep)/2\relax}
	
	\begin{lrbox}{\IWOOWSNSafeBottom}
		\begin{minipage}{\textwidth}
			\centering
			\begin{minipage}[b]{0.245\textwidth}
				\centering
				\includegraphics[width=\linewidth,height=0.15\textheight,keepaspectratio]{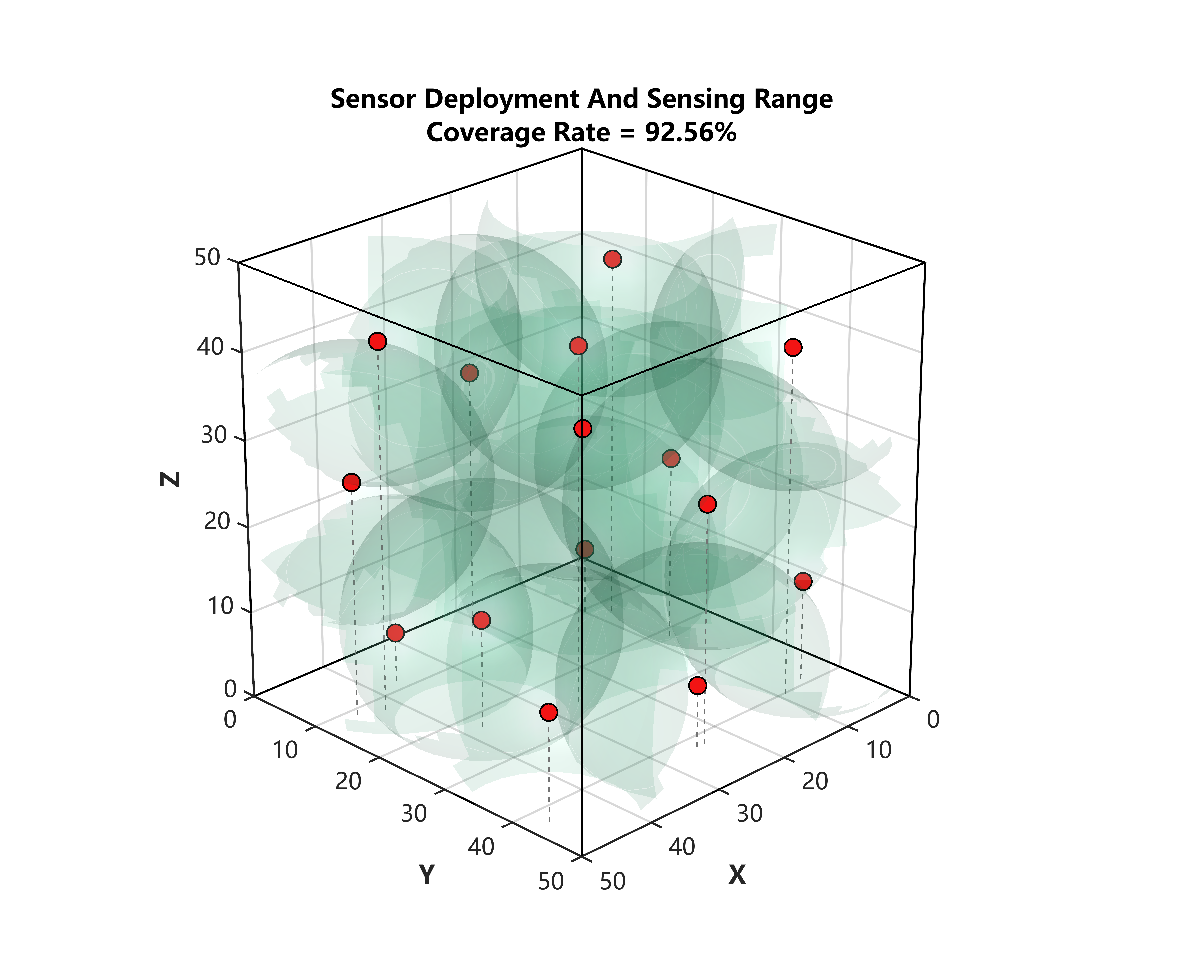}
			\end{minipage}%
			\hfill
			\begin{minipage}[b]{0.245\textwidth}
				\centering
				\includegraphics[width=\linewidth,height=0.15\textheight,keepaspectratio]{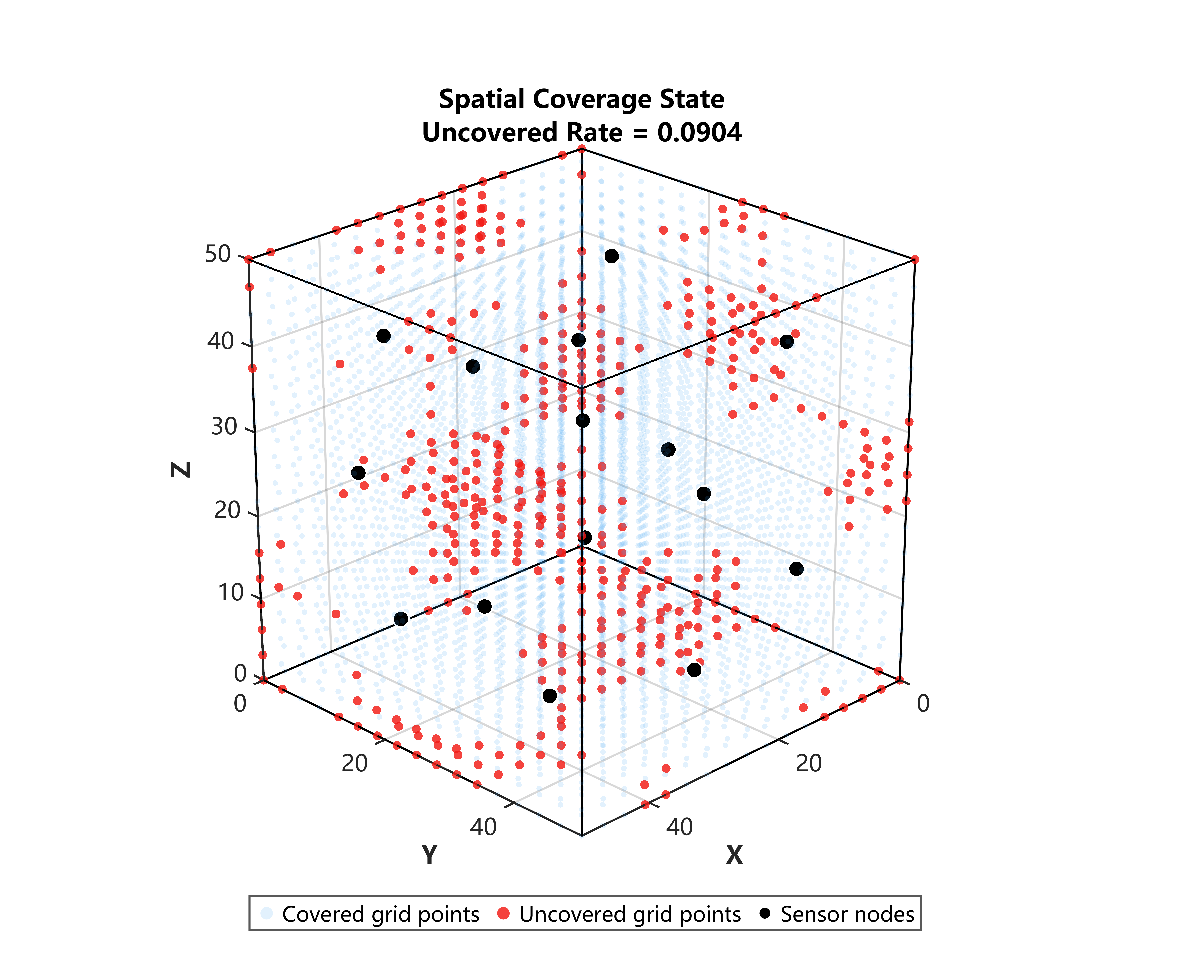}
			\end{minipage}%
			\hfill
			\begin{minipage}[b]{0.245\textwidth}
				\centering
				\includegraphics[width=\linewidth,height=0.15\textheight,keepaspectratio]{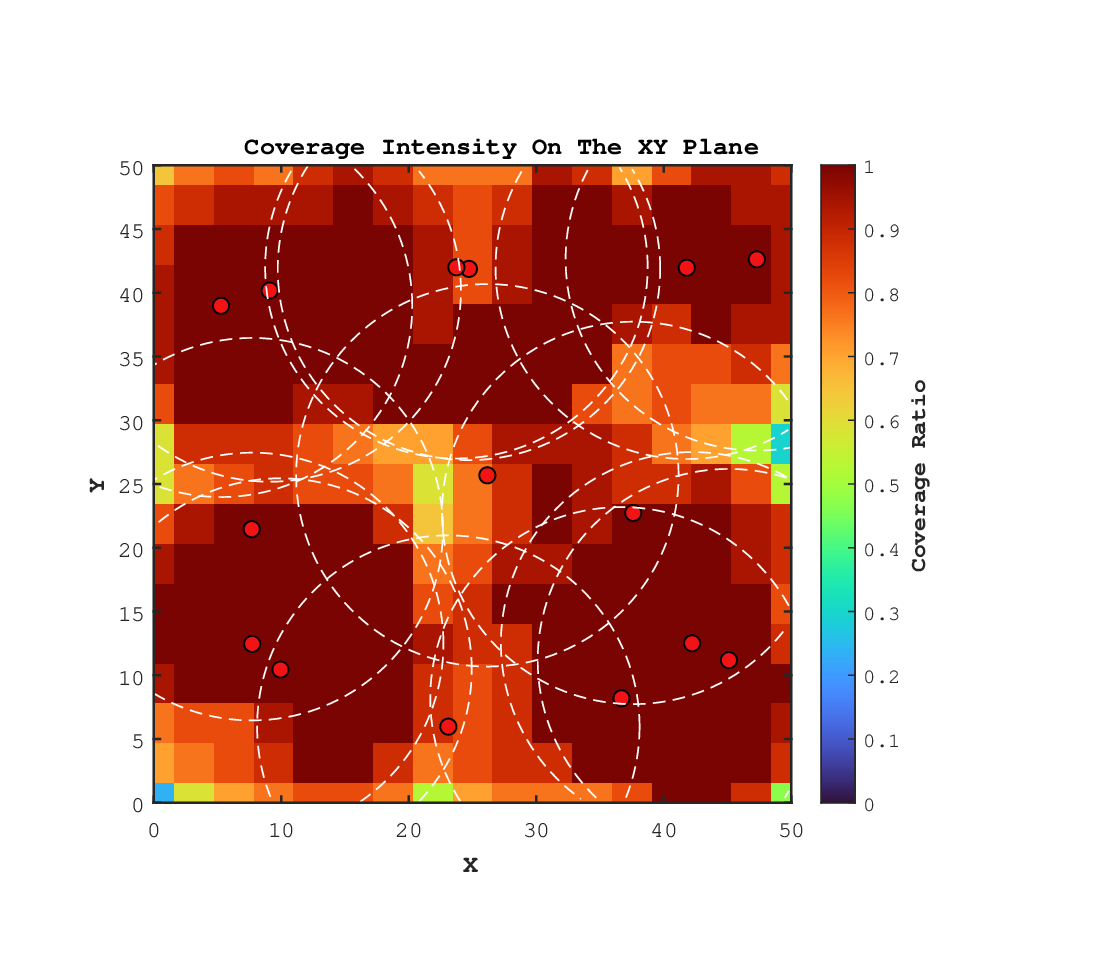}
			\end{minipage}%
			\hfill
			\begin{minipage}[b]{0.245\textwidth}
				\centering
				\includegraphics[width=\linewidth,height=0.15\textheight,keepaspectratio]{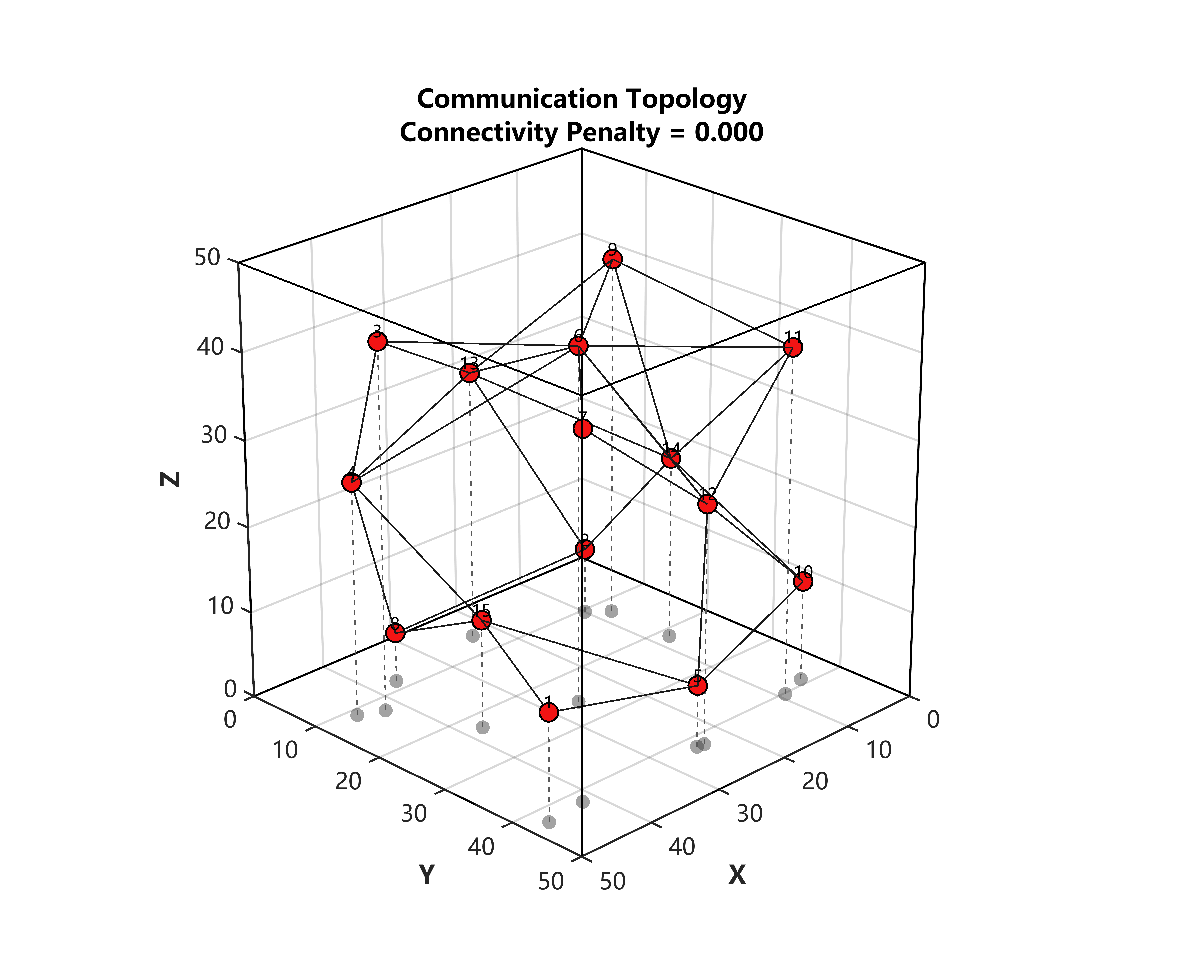}
			\end{minipage}%
			
			\centering
			\captionsetup{font=footnotesize,skip=2pt}
			\begin{minipage}{0.245\textwidth}\centering{\footnotesize (a)}\end{minipage}%
			\hfill
			\begin{minipage}{0.245\textwidth}\centering{\footnotesize (b)}\end{minipage}%
			\hfill
			\begin{minipage}{0.245\textwidth}\centering{\footnotesize (c)}\end{minipage}%
			\hfill
			\begin{minipage}{0.245\textwidth}\centering{\footnotesize (d)}\end{minipage}%
			\par
			\captionof{figure}{Optimal 3D WSN deployment obtained by IWOO
				(a) sensor deployment and sensing range
				(b) spatial coverage
				(c) $XY$-plane coverage intensity
				(d) communication topology}
			\label{fig:wsn_visual}
		\end{minipage}
	\end{lrbox}
	\setbox\IWOOWSNSafeBottom=\vbox{\hbox{\usebox{\IWOOWSNSafeBottom}}\kern0pt}
	\setlength{\IWOOWSNSafeRoom}{\dimexpr\textheight
		-\ht\IWOOWSNSafeBottom-\dp\IWOOWSNSafeBottom-12pt\relax}
	
	\begin{lrbox}{\IWOOWSNSafePerformance}
		\begin{minipage}{\IWOOWSNSafeWidth}
			\par\smallskip\noindent
			\begin{minipage}{\linewidth}
				\centering
				
				\par\vspace{5pt}
				\noindent
				\begin{tabular}{
						@{}c@{\hspace{0.02\linewidth}}c@{}
					}
					\includegraphics[
					width=0.49\linewidth
					]{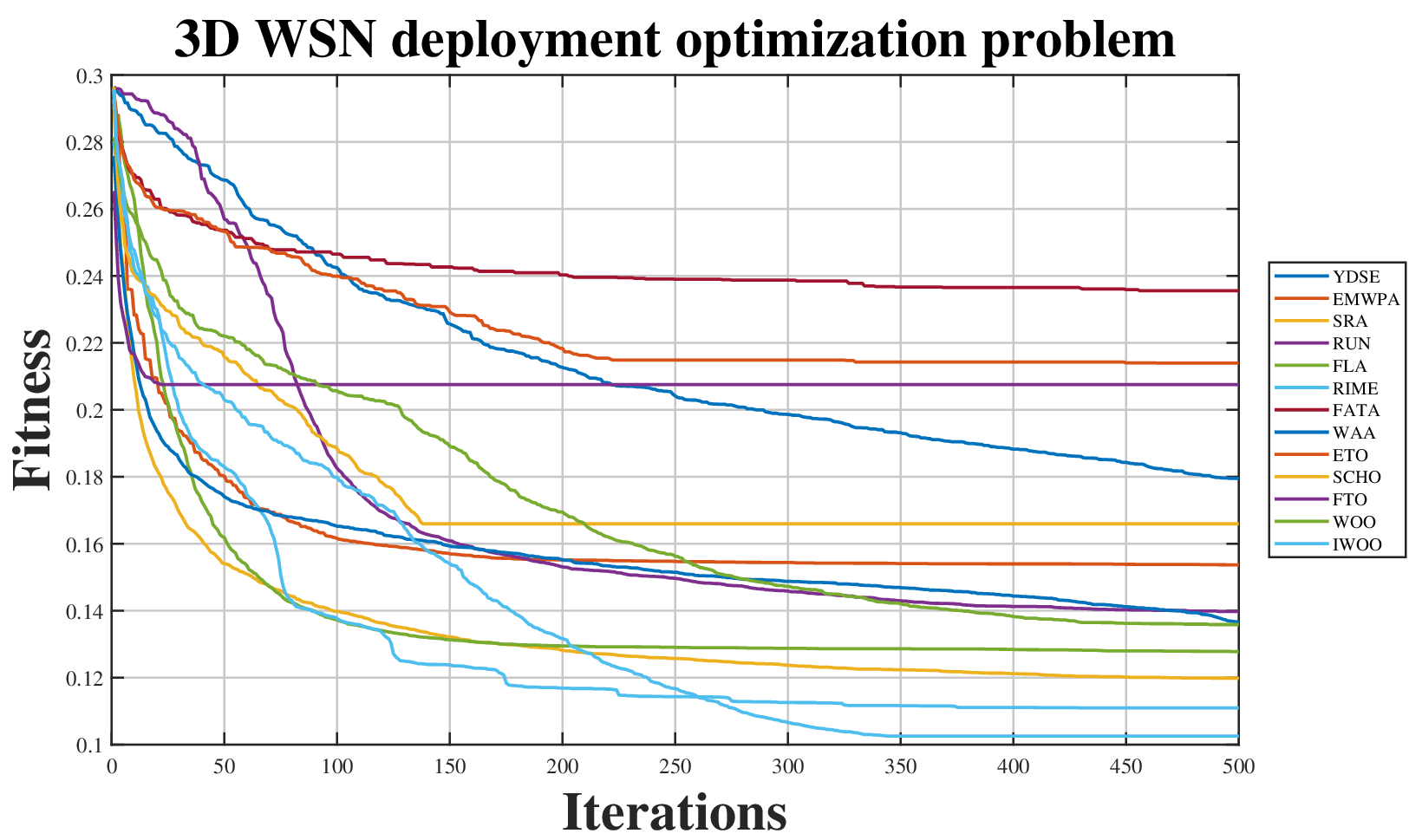}
					&
					\includegraphics[
					width=0.49\linewidth
					]{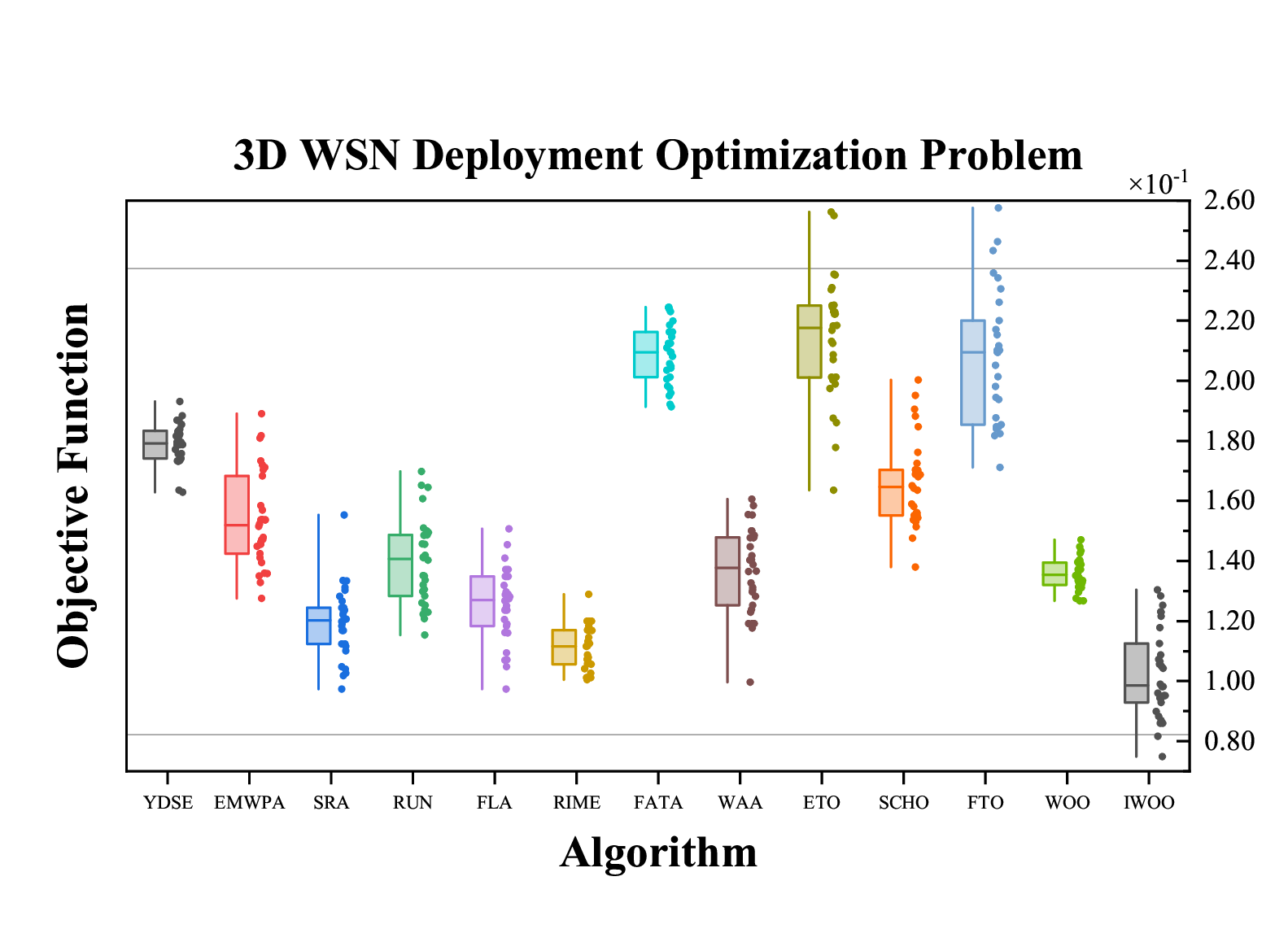}
					\\[2pt]
					{\small (a)}
					&
					{\small (b)}
				\end{tabular}
				\par

				\captionof{figure}{Performance comparison for the 3D WSN deployment optimization problem
					(a) convergence curves
					(b) box plots}
				\label{fig:wsn_curve_box}
			\end{minipage}\par\smallskip
		\end{minipage}
	\end{lrbox}
	\setbox\IWOOWSNSafePerformance=\vbox{\hbox{\usebox{\IWOOWSNSafePerformance}}\kern0pt}
	\setlength{\IWOOWSNSafeRightRoom}{\dimexpr\IWOOWSNSafeRoom
		-\ht\IWOOWSNSafePerformance-\dp\IWOOWSNSafePerformance-6pt\relax}
	
	\setbox\IWOOWSNSafeText=\vbox{%
		\hsize=\IWOOWSNSafeWidth
		\linewidth=\hsize
		\columnwidth=\hsize
		\subsection{3D WSN deployment optimization problem}
		\label{subsec:wsn_deployment}

		\begingroup
		
		The three-dimensional wireless sensor network (3D WSN) deployment optimization problem aims to determine the optimal positions of sensor nodes within a bounded three-dimensional region of interest (ROI), so as to improve spatial coverage while maintaining communication connectivity and reducing blind areas and redundant coverage. Compared with two-dimensional deployment, the additional spatial degree of freedom in three-dimensional environments substantially enlarges the search space and increases the difficulty of the deployment optimization problem. Metaheuristic algorithms have been widely applied to sensor node deployment and coverage optimization in WSNs~\cite{Houssein2024MetaheuristicWSN}. Related studies have further investigated three-dimensional deployment and connectivity optimization for heterogeneous sensor nodes~\cite{Lu2025WSN3DDeployment}, as well as coverage enhancement using particle swarm optimization~\cite{Zhang2020WSNCoverage}.
		\par\endgroup

		In this study, a 3D WSN deployment optimization model is formulated by jointly considering the spatial uncovered ratio, communication connectivity, coverage redundancy, and inter-node spacing. A cubic monitoring region with a side length of $50$ is considered, containing $N=15$ sensor nodes. Since each sensor node is represented by three spatial coordinates, the dimensionality of the decision space is $D=45$. Let the position of the $i$th sensor node be $(x_i,y_i,z_i)$. The decision vector is then defined as
		\begin{equation}
			\boldsymbol{X}
			=
			[x_1,y_1,z_1,x_2,y_2,z_2,\ldots,x_{15},y_{15},z_{15}]
			\label{eq:wsn_variables}
		\end{equation}
		subject to
		\begin{equation}
			\begin{aligned}
				0\leq x_i\leq 50,\qquad 0\leq y_i\leq 50,\\
				0\leq z_i\leq 50,\qquad i=1,2,\ldots,15
			\end{aligned}
			\label{eq:wsn_bounds}
		\end{equation}
		
		In the experiments, the sensing radius is set to $R_s=15$, the communication radius to $R_c=28$, and the minimum inter-node distance to $d_{\min}=8$. The objective function is formulated as
		\begin{equation}
			\begin{aligned}
				\min F(\boldsymbol{X})
				={}&
				w_{\mathrm{cov}}U(\boldsymbol{X})
				+w_{\mathrm{conn}}P_{\mathrm{conn}}(\boldsymbol{X})\\
				&+w_{\mathrm{red}}P_{\mathrm{red}}(\boldsymbol{X})
				+w_{\mathrm{spa}}P_{\mathrm{spa}}(\boldsymbol{X})
			\end{aligned}
			\label{eq:wsn_obj}
		\end{equation}
		where, $U(\boldsymbol{X})$ denotes the spatial uncovered ratio, while $P_{\mathrm{conn}}(\boldsymbol{X})$, $P_{\mathrm{red}}(\boldsymbol{X})$ and $P_{\mathrm{spa}}(\boldsymbol{X})$ denote the penalty terms associated with communication connectivity, coverage redundancy, and inter-node spacing, respectively. The corresponding weights are set as
		\begin{equation}
			\begin{aligned}
				w_{\mathrm{cov}}&=1.00, & w_{\mathrm{conn}}&=0.25,\\
				w_{\mathrm{red}}&=0.03, & w_{\mathrm{spa}}&=0.03
			\end{aligned}
			\label{eq:wsn_weights}
		\end{equation}
		
		As shown in Table~\ref{tab:wsn_results}, IWOO achieves the lowest best objective value of $7.4903\times10^{-2}$ and the best Fr-rank of $1.7667$, ranking first among all compared algorithms. Its mean objective value is $1.0254\times10^{-1}$, with an average coverage rate of $89.7971\%$, indicating highly competitive overall performance. By comparison, RIME achieves the second-lowest mean
		objective value of $1.1099\times10^{-1}$ and the second-highest
		average coverage rate of $88.9482\%$, while WOO obtains the smallest standard deviation of $5.2656\times10^{-3}$. Considering the overall statistical results, IWOO provides a favorable balance between solution quality, coverage performance, and stability on the 3D WSN deployment optimization problem.
		
		The optimal deployment obtained by IWOO is illustrated in Fig.~\ref{fig:wsn_visual}. The sensor nodes are distributed relatively uniformly throughout the monitoring region and form a fully connected communication topology, indicating that the resulting deployment achieves a favorable balance among spatial coverage, network connectivity, and inter-node spacing.
		
		As shown in Fig.~\ref{fig:wsn_curve_box}, IWOO converges rapidly toward the low-objective-value region, while the results obtained over multiple independent runs remain at a relatively favorable level, further demonstrating its competitive convergence performance on the 3D WSN deployment optimization problem.

		\par
		\par
	}
	
	\newcommand{\IWOOWSNSafeTake}[1]{%
		\setbox\IWOOWSNSafeTrial=\copy\IWOOWSNSafeText
		\setbox#1=\vsplit\IWOOWSNSafeTrial to\IWOOWSNSafeGoal
		\setbox#1=\vbox{\unvbox#1}%
		\ifdim\dimexpr\ht#1+\dp#1\relax>\IWOOWSNSafeLimit
		\ifdim\IWOOWSNSafeGoal>\baselineskip
		\advance\IWOOWSNSafeGoal-\baselineskip
		\IWOOWSNSafeTake{#1}%
		\else
		\setbox#1=\vbox{}%
		\setbox\IWOOWSNSafeTrial=\copy\IWOOWSNSafeText
		\fi
		\fi
	}
	\splittopskip=\topskip
	\splitmaxdepth=\maxdepth
	\setlength{\IWOOWSNSafeGoal}{\dimexpr
		(\ht\IWOOWSNSafeText+\dp\IWOOWSNSafeText+\ht\IWOOWSNSafePerformance+\dp\IWOOWSNSafePerformance+6pt)/2+4pt\relax}
	\ifdim\IWOOWSNSafeGoal>\IWOOWSNSafeRoom
	\setlength{\IWOOWSNSafeGoal}{\IWOOWSNSafeRoom}
	\fi
	\setlength{\IWOOWSNSafeLimit}{\IWOOWSNSafeRoom}
	\IWOOWSNSafeTake{\IWOOWSNSafeLeft}
	\setbox\IWOOWSNSafeText=\box\IWOOWSNSafeTrial
	\setlength{\IWOOWSNSafeGoal}{\IWOOWSNSafeRightRoom}
	\setlength{\IWOOWSNSafeLimit}{\IWOOWSNSafeRightRoom}
	\IWOOWSNSafeTake{\IWOOWSNSafeRight}
	\setbox\IWOOWSNSafeText=\box\IWOOWSNSafeTrial
	
	\setbox\IWOOWSNSafeTop=\vbox{%
		\hbox to\textwidth{%
			\vtop{\hsize=\IWOOWSNSafeWidth\kern0pt\unvbox\IWOOWSNSafeLeft}%
			\hfil
			\vtop{\hsize=\IWOOWSNSafeWidth\kern0pt
				\unvbox\IWOOWSNSafeRight
				\vskip6pt\nointerlineskip
				\hbox{\usebox{\IWOOWSNSafePerformance}}\kern0pt}%
		}\kern0pt
	}
	\noindent
	\begin{minipage}[b][\textheight][t]{\textwidth}
		\noindent\usebox{\IWOOWSNSafeTop}\par
		\vfill
		\vskip12pt
		\nointerlineskip
		\noindent\usebox{\IWOOWSNSafeBottom}\par
	\end{minipage}\par
	
	\twocolumn[{

		\begin{minipage}{\textwidth}
			\centering
			\captionsetup{font=footnotesize,skip=2pt}
			
			\centering
			
			\centering
			\begin{minipage}{0.98\linewidth}
				\centering
				\captionof{table}{Experimental results of the 3D WSN deployment optimization problem}
				\label{tab:wsn_results}

				\renewcommand{\arraystretch}{1.00}
				\setlength{\tabcolsep}{3.8pt}
				\scriptsize
				
				\resizebox{\linewidth}{!}{
					\begin{tabular}{lccccccc}
						\toprule
						Algorithm
						& Mean
						& Std
						& Best
						& Worst
						& Median
						& Avg. Coverage (\%)
						& Fr-rank \\
						\midrule
						
						IWOO
						& $\mathbf{1.0254\times10^{-1}}$
						& $1.4815\times10^{-2}$
						& $\mathbf{7.4903\times10^{-2}}$
						& $1.3040\times10^{-1}$
						& $9.8589\times10^{-2}$
						& $\mathbf{89.7971}$
						& $\mathbf{1.7667}$ \\
						
						RIME
						& $1.1099\times10^{-1}$
						& $7.3167\times10^{-3}$
						& $1.0042\times10^{-1}$
						& $1.2894\times10^{-1}$
						& $1.1164\times10^{-1}$
						& $88.9482$
						& $2.2333$ \\
						
						SRA
						& $1.1977\times10^{-1}$
						& $1.1723\times10^{-2}$
						& $9.7326\times10^{-2}$
						& $1.5526\times10^{-1}$
						& $1.2029\times10^{-1}$
						& $88.0666$
						& $3.3000$ \\
						
						FLA
						& $1.2529\times10^{-1}$
						& $1.2273\times10^{-2}$
						& $9.7357\times10^{-2}$
						& $1.5067\times10^{-1}$
						& $1.2703\times10^{-1}$
						& $87.5106$
						& $4.1333$ \\
						
						WAA
						& $1.3666\times10^{-1}$
						& $1.4475\times10^{-2}$
						& $9.9637\times10^{-2}$
						& $1.6057\times10^{-1}$
						& $1.3771\times10^{-1}$
						& $86.3862$
						& $5.5333$ \\
						
						WOO
						& $1.3586\times10^{-1}$
						& $\mathbf{5.2656\times10^{-3}}$
						& $1.2676\times10^{-1}$
						& $1.4702\times10^{-1}$
						& $1.3541\times10^{-1}$
						& $86.4688$
						& $5.8333$ \\
						
						RUN
						& $1.3977\times10^{-1}$
						& $1.4249\times10^{-2}$
						& $1.1534\times10^{-1}$
						& $1.6976\times10^{-1}$
						& $1.4071\times10^{-1}$
						& $86.0831$
						& $6.2333$ \\
						
						EMWPA
						& $1.5367\times10^{-1}$
						& $1.5814\times10^{-2}$
						& $1.2753\times10^{-1}$
						& $1.8907\times10^{-1}$
						& $1.5192\times10^{-1}$
						& $84.6907$
						& $7.4667$ \\
						
						SCHO
						& $1.6595\times10^{-1}$
						& $1.4567\times10^{-2}$
						& $1.3804\times10^{-1}$
						& $2.0029\times10^{-1}$
						& $1.6462\times10^{-1}$
						& $83.4510$
						& $8.8000$ \\
						
						YDSE
						& $1.7917\times10^{-1}$
						& $6.6893\times10^{-3}$
						& $1.6285\times10^{-1}$
						& $1.9307\times10^{-1}$
						& $1.7911\times10^{-1}$
						& $82.1412$
						& $9.9000$ \\
						
						FTO
						& $2.0755\times10^{-1}$
						& $2.2095\times10^{-2}$
						& $1.7121\times10^{-1}$
						& $2.5761\times10^{-1}$
						& $2.0947\times10^{-1}$
						& $79.3088$
						& $11.7333$ \\
						
						FATA
						& $2.0909\times10^{-1}$
						& $9.7507\times10^{-3}$
						& $1.9138\times10^{-1}$
						& $2.2458\times10^{-1}$
						& $2.0949\times10^{-1}$
						& $79.1510$
						& $11.9333$ \\
						
						ETO
						& $2.1394\times10^{-1}$
						& $2.0539\times10^{-2}$
						& $1.6356\times10^{-1}$
						& $2.5620\times10^{-1}$
						& $2.1756\times10^{-1}$
						& $78.6802$
						& $12.1333$ \\
						
						\bottomrule
					\end{tabular}
				}
			\end{minipage}
			
		\end{minipage}
		
		\vspace{5pt}
	}]
	
	\par\unvbox\IWOOWSNSafeText\par

	\subsection{3D rigid point cloud registration problem for Stanford Bunny}
	\label{subsec:point_cloud_registration}

	The Stanford Bunny 3D rigid point cloud registration problem aims to estimate the rotation and translation parameters that transform a source point cloud into the coordinate system of a target point cloud, thereby achieving accurate geometric alignment between the two point clouds in three-dimensional space. Since the registration error generally exhibits nonlinear characteristics with respect to the rigid transformation parameters, appropriate transformation parameters must be searched for in a continuous parameter space. Metaheuristic global optimization methods have been applied to three-dimensional point cloud registration~\cite{Chu2025PointCloudRegistration,Wu2023PointCloudRegistration}. The Stanford Bunny is a classical three-dimensional scanned dataset whose model reconstruction can be traced back to the work of Turk and Levoy~\cite{Turk1994ZipperedPolygonMeshes}. In this study, the Stanford Bunny point cloud is adopted, and a registration optimization model is formulated using six-degree-of-freedom rigid transformation parameters, with the registration error taken as the optimization objective.\par

	\begingroup
	
	Let the source and target point clouds be denoted by
	\begin{equation}
		\mathcal{P}
		=
		\left\{
		\boldsymbol{p}_i
		\right\}_{i=1}^{M},
		\qquad
		\mathcal{Q}
		=
		\left\{
		\boldsymbol{q}_i
		\right\}_{i=1}^{M}
		\label{eq:pc_reg_clouds}
	\end{equation}
	
	The six-degree-of-freedom rigid transformation parameters are used as the decision variables
	\begin{equation}
		\boldsymbol{x}
		=
		[t_x,t_y,t_z,r_x,r_y,r_z]
		\label{eq:pc_reg_variables}
	\end{equation}
	where $t_x$, $t_y$, and $t_z$ denote the translation components along the $x$-, $y$-, and $z$-axes, respectively, while $r_x$, $r_y$, and $r_z$ denote the corresponding rotation angles. The search ranges of the translation variables are set to $[-20,20]$, whereas those of the rotation variables are set to $[-\pi,\pi]$.
	
	Given the rotation matrix $\boldsymbol{R}(r_x,r_y,r_z)$ and the translation vector
	$\boldsymbol{t}=[t_x,t_y,t_z]^{\mathrm T}$, the $i$th point in the source point cloud is transformed as
	\begin{equation}
		\hat{\boldsymbol{p}}_i
		=
		\boldsymbol{R}(r_x,r_y,r_z)\boldsymbol{p}_i
		+
		\boldsymbol{t}
		\label{eq:pc_reg_transform}
	\end{equation}
	
	The root mean square error (RMSE) between the transformed source point cloud and the target point cloud is employed as the objective function
	\begin{equation}
		\min f(\boldsymbol{x})
		=
		\sqrt{
			\frac{1}{M}
			\sum_{i=1}^{M}
			\left\|
			\hat{\boldsymbol{p}}_i-\boldsymbol{q}_i
			\right\|_2^2
		}
		\label{eq:pc_reg_objective}
	\end{equation}
	
	A smaller RMSE indicates higher point cloud registration accuracy. Considering the periodicity and equivalent representations of Euler angles, the registration performance is quantitatively evaluated using RMSE, while the spatial alignment between the registered point clouds is further examined through visual comparison.
	
	\par\smallskip\noindent
	\begin{minipage}{\linewidth}
		\centering
		\captionof{table}{Experimental results of the Stanford Bunny 3D rigid point cloud registration problem}
		\label{tab:pc_bunny_results}
		
		\scriptsize
		\setlength{\tabcolsep}{6.2pt}
		\renewcommand{\arraystretch}{0.92}
		\resizebox{\linewidth}{!}{%
			\begin{tabular}{lcccc}
				\toprule
				Algorithm & Mean RMSE & Std & Best RMSE & Fr-rank \\
				\midrule
				IWOO  & $\mathbf{8.6254\times10^{-2}}$ & $\mathbf{2.7153\times10^{-3}}$ & $\mathbf{8.0916\times10^{-2}}$ & $\mathbf{1.0000}$ \\
				RUN   & $\mathbf{8.6254\times10^{-2}}$ & $\mathbf{2.7153\times10^{-3}}$ & $\mathbf{8.0916\times10^{-2}}$ & 3.8333 \\
				EMWPA & $7.7915\times10^{-1}$ & $1.3233\times10^{0}$ & $\mathbf{8.0916\times10^{-2}}$ & 4.6333 \\
				SRA   & $6.0800\times10^{-1}$ & $1.0785\times10^{0}$ & $\mathbf{8.0916\times10^{-2}}$ & 5.1000 \\
				WOO   & $8.6551\times10^{-2}$ & $2.6172\times10^{-3}$ & $8.1269\times10^{-2}$ & 5.3667 \\
				RIME  & $3.1265\times10^{-1}$ & $7.0132\times10^{-1}$ & $8.1105\times10^{-2}$ & 6.3667 \\
				FLA   & $2.0658\times10^{0}$  & $2.3105\times10^{0}$  & $8.1930\times10^{-2}$ & 7.2000 \\
				YDSE  & $1.1783\times10^{-1}$ & $2.0911\times10^{-2}$ & $8.6803\times10^{-2}$ & 7.4667 \\
				FATA  & $1.6240\times10^{-1}$ & $4.1608\times10^{-2}$ & $1.0130\times10^{-1}$ & 8.7333 \\
				SCHO  & $1.1015\times10^{0}$  & $1.4294\times10^{0}$  & $8.9214\times10^{-2}$ & 9.2667 \\
				WAA   & $8.1067\times10^{-1}$ & $1.3549\times10^{0}$  & $1.4131\times10^{-1}$ & 9.9333 \\
				ETO   & $7.3711\times10^{-1}$ & $1.0659\times10^{0}$  & $1.2613\times10^{-1}$ & 10.2667 \\
				FTO   & $2.4255\times10^{0}$  & $2.1435\times10^{0}$  & $2.0485\times10^{-1}$ & 11.8333 \\
				\bottomrule
			\end{tabular}
		}
	\end{minipage}\par\smallskip
	\par\vspace{10pt}

	As shown in Table~\ref{tab:pc_bunny_results}, IWOO and RUN achieve the lowest mean RMSE of $8.6254\times10^{-2}$. IWOO also shares the lowest best RMSE of $8.0916\times10^{-2}$ with RUN, EMWPA, and SRA. Although WOO obtains the smallest standard deviation of $2.6172\times10^{-3}$, IWOO achieves the best Fr-rank of $1.0000$ among all compared algorithms. These results indicate that IWOO provides high registration accuracy and strong overall optimization performance on the Stanford Bunny 3D rigid point cloud registration problem.
	
	\par\smallskip\noindent
	\begin{minipage}{\linewidth}
		\centering
		\includegraphics[
		width=\linewidth
		]{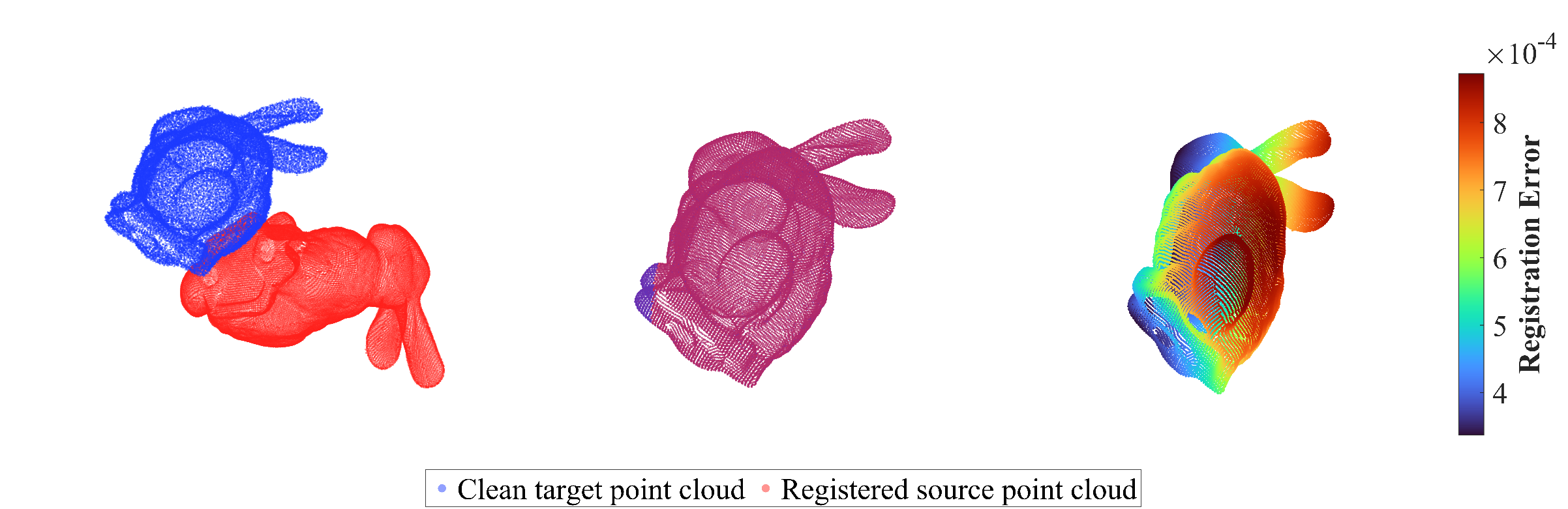}
		
		\captionof{figure}{Optimal Stanford Bunny 3D rigid point cloud registration obtained by IWOO}
		\label{fig:pc_bunny_visual}
	\end{minipage}
	\par\smallskip

	Figure~\ref{fig:pc_bunny_visual} illustrates the registration result obtained by IWOO. After applying the optimized rigid transformation, the transformed source point cloud exhibits close spatial alignment with the target point cloud in both the overall contour and local geometric structures.
	
	\par\smallskip\noindent
	\begin{minipage}{\linewidth}
		\centering
		\begin{tabular}[b]{@{}p{0.46\linewidth}@{}}
			\centering
			\includegraphics[width=\linewidth]{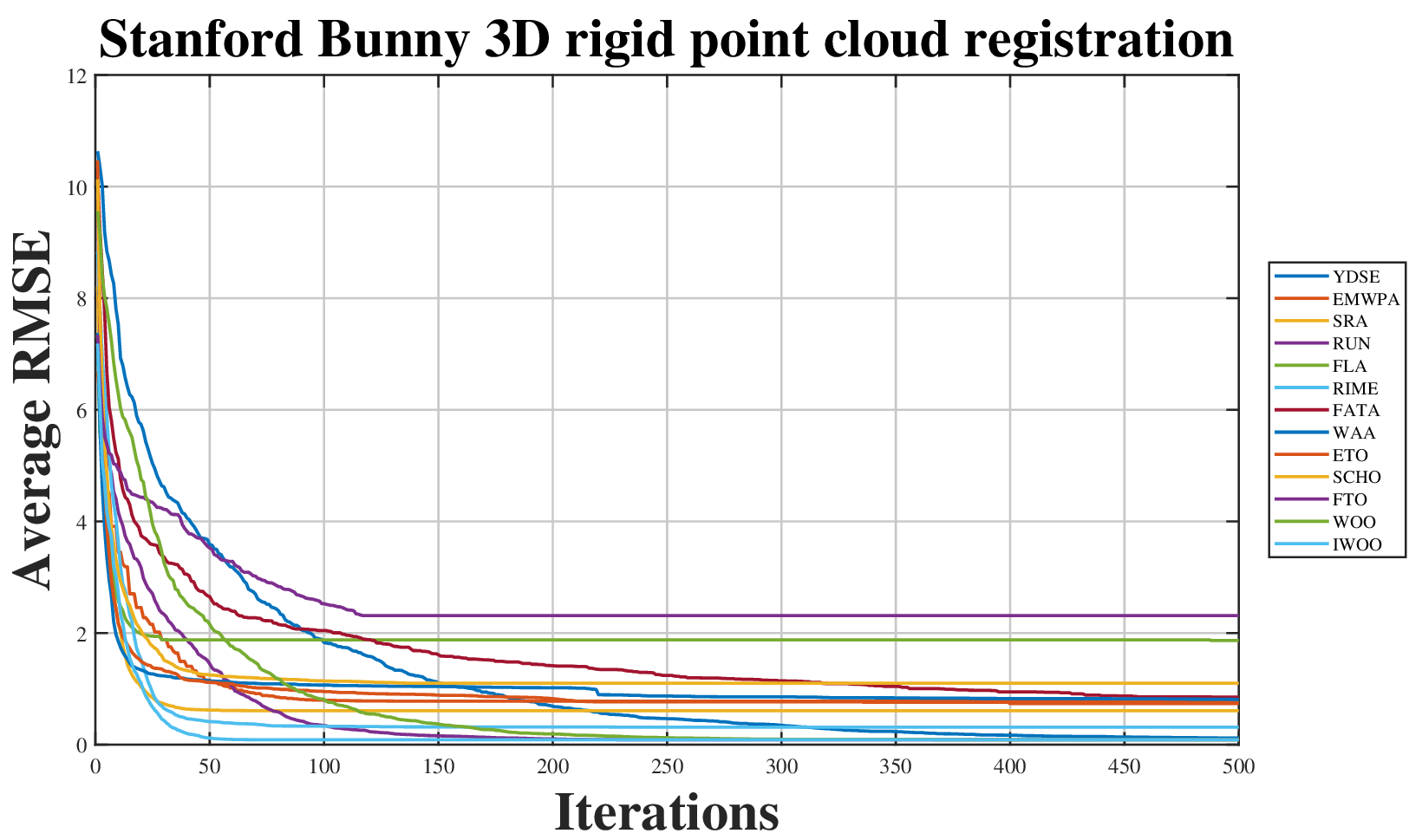}
			\\
			\small (a) Convergence curves
		\end{tabular}
		\hfill
		\begin{tabular}[b]{@{}p{0.46\linewidth}@{}}
			\centering
			\includegraphics[width=\linewidth]{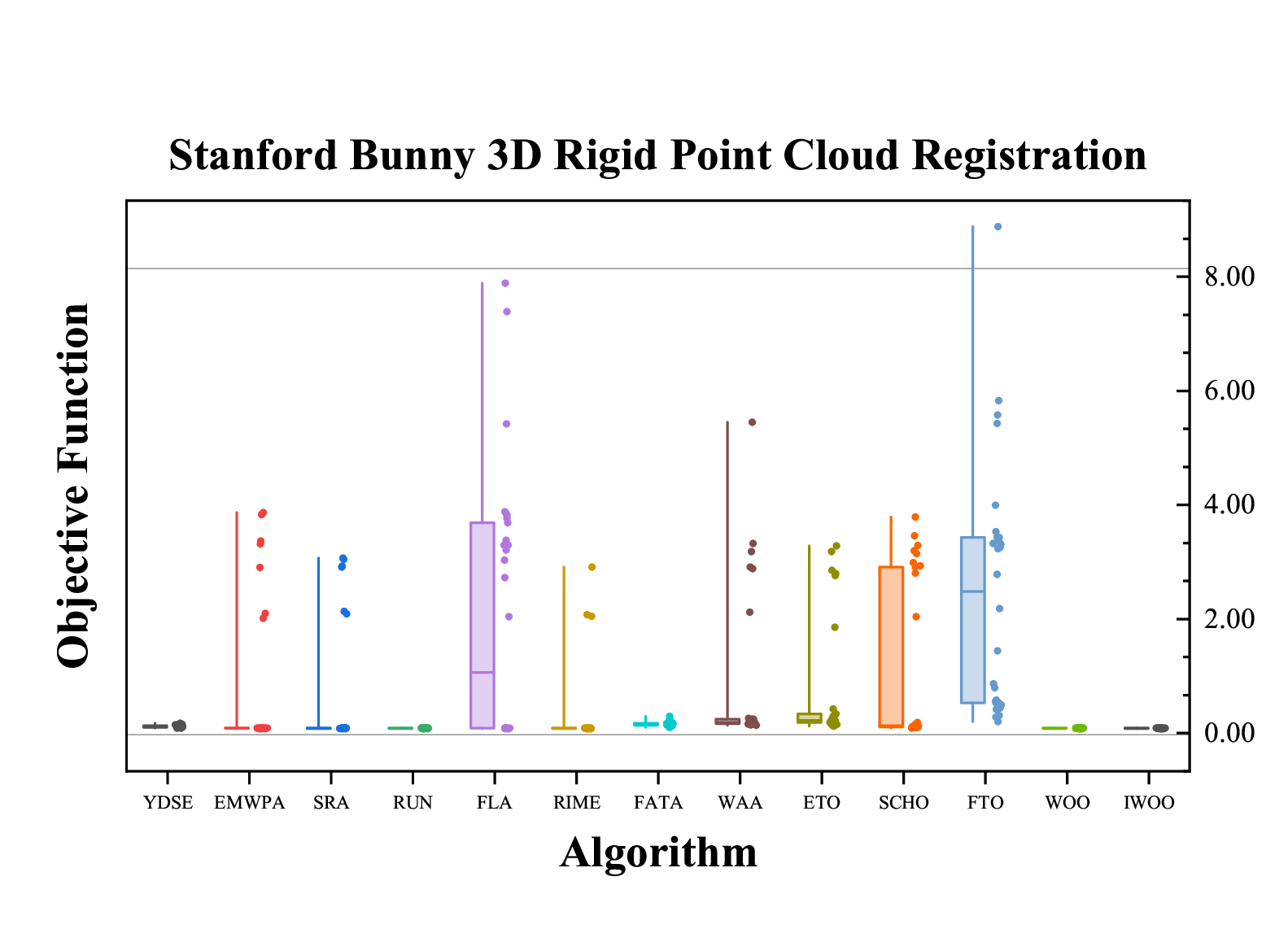}
			\\
			\small (b) Box plots
		\end{tabular}
		\captionof{figure}{Performance comparison for the Stanford Bunny 3D rigid point cloud registration problem}
		\label{fig:pc_bunny_curve_box}
	\end{minipage}\par\smallskip

	As shown in Fig.~\ref{fig:pc_bunny_curve_box}, IWOO converges rapidly toward the low-RMSE region, while the results obtained over multiple independent runs remain at a relatively favorable level, further demonstrating its competitive registration performance on the Stanford Bunny 3D rigid point cloud registration problem.
	\endgroup
	\FloatBarrier

	
	\section{Conclusion} 
	
	This study developed IWOO, an improved Wave Optics Optimizer, to address insufficient population diversity, limited local exploitation, and late-stage stagnation in complex optimization problems. Based on the original WOO framework, the proposed algorithm introduces two novel core mechanisms, namely the TriEDM multi-path search strategy and CGOL collaborative grouped orthogonal learning, supplemented by two novel auxiliary mechanisms, namely GPS–-OBL initialization and ASRK correction. These mechanisms jointly form a unified framework for improving the optimization performance of WOO. Extensive experiments on the CEC2019, CEC2022, and CEC2017 benchmark suites demonstrate that IWOO achieves competitive solution accuracy, convergence performance, and run-to-run stability across different benchmark settings. Comparisons with representative optimizers and a high-performing algorithm group further support its overall competitiveness. The ablation study supports the individual contributions of the proposed strategies and shows that the complete algorithm achieves more consistent overall performance than the baseline and single-strategy variants. Furthermore, experiments on several classical constrained engineering design problems and selected practical engineering applications demonstrate the applicability of IWOO to diverse practical optimization tasks. Future work will investigate more efficient implementations, adaptive strategy activation, and parameter adjustment mechanisms, while extending IWOO to higher-dimensional, multi-objective, dynamic, and noisy optimization problems.
	
	\section*{CRediT authorship contribution statement} 
	
	\textbf{Rui Du:} Writing -- original draft, Methodology, Software, Formal analysis, Validation, Visualization, Data curation, Conceptualization. 
	\textbf{Yaming Zhang:} Supervision, Conceptualization, Methodology, Project administration, Writing -- review \& editing. 
	\textbf{Yongpeng Wen:} Software, Validation, Formal analysis, Data curation. 
	\textbf{Linghong Cai:} Methodology, Software, Validation, Writing -- review \& editing. 
	\textbf{Nanqing Liu:} Validation, Data curation, Formal analysis. 
	
	\section*{Declaration of competing interest} 
	
	The authors declare that there are no conflicts of interest regarding  
	the publication of this paper. No financial or personal relationships  
	have influenced the outcomes of this research. 
	
	\section*{Data availability} 
	
	Data will be made available on request.

	\bibliographystyle{elsarticle-num}
	\bibliography{references}
	
	\clearpage
	
	\addcontentsline{toc}{section}{Appendix A}
	
	\setcounter{table}{0}
	\renewcommand{\thetable}{A\arabic{table}}
	
	\begin{sidewaystable*}[p]
		\centering
		
		\begin{minipage}{0.96\textheight}
			\raggedright
			\large\bfseries Appendix A
		\end{minipage}
		
		\vspace{3mm}
		
		\captionsetup{
			font=small,
			skip=3pt
		}
		
		\caption{Test results for CEC2019 (Dim$<20$) -- Advanced Group}
		\label{tab:cec2019_results_advanced}
		
		\scriptsize
		\setlength{\tabcolsep}{3.6pt}
		\renewcommand{\arraystretch}{0.78}
		
		\begin{adjustbox}{
				width=0.98\textheight,
				max totalheight=0.88\textwidth,
				keepaspectratio,
				center
			}
			
			\begin{tabular}{c c cccccccccc} 
				\toprule
				Function & & WOO & CGO & RWGWO & SaDE & SHADE & LSHADE & APSM\_jSO & NL\_SHADE\_RSP & LSRTDE & IWOO \\
				\midrule
				
				\multirow{5}{*}{F1} 
				& min       & 1.0000e+00 & 1.0000e+00 & 6.4380e+01 & 1.0121e+00 & 1.0000e+00 & 1.0001e+00 & 1.0000e+00 & 1.0000e+00 & 1.0000e+00 & 1.0000e+00 \\
				& mean      & \textbf{1.0000e+00} & \textbf{1.0000e+00} & 1.4375e+05 & 1.8504e+03 & 1.0001e+00 & 1.0116e+00 & 3.1647e+01 & 4.3922e+00 & \textbf{1.0000e+00} & 1.0001e+00 \\
				& std       & 0.0000e+00 & 0.0000e+00 & 2.4936e+05 & 4.2473e+03 & 5.9755e-04 & 1.4814e-02 & 1.6459e+02 & 1.5488e+01 & 1.0364e-06 & 4.5346e-04 \\
				& p-value   & 1.2279e-05 & 1.2279e-05 & 1.7344e-06 & 1.7344e-06 & 2.2888e-01 & 6.3391e-06 & 1.2506e-04 & 2.3534e-06 & 7.3433e-01 & - - \\
				& + / = / - & -          & -          & +          & +          & =          & +          & +          & +          & =          & - - \\
				\midrule
				
				\multirow{5}{*}{F2} 
				& min       & 3.9092e+00 & 3.9236e+00 & 1.7338e+02 & 6.3722e+01 & 1.3136e+01 & 6.4788e+00 & 3.6684e+01 & 2.2498e+01 & 2.8716e+00 & 3.2496e+00 \\
				& mean      & 4.3823e+00 & 4.2693e+00 & 8.2510e+02 & 1.3669e+02 & 2.3041e+02 & 2.0274e+02 & 1.3219e+02 & 7.7262e+01 & 6.3373e+00 & \textbf{4.0308e+00} \\
				& std       & 2.4633e-01 & 1.2272e-01 & 3.7772e+02 & 5.4487e+01 & 3.6659e+02 & 4.2057e+02 & 5.9143e+01 & 3.7061e+01 & 2.1679e+00 & 3.3094e-01 \\
				& p-value   & 3.5888e-04 & 1.2866e-03 & 1.7344e-06 & 1.7344e-06 & 1.7344e-06 & 1.7344e-06 & 1.7344e-06 & 1.7344e-06 & 5.7517e-06 & - - \\
				& + / = / - & +          & +          & +          & +          & +          & +          & +          & +          & +          & - - \\
				\midrule
				
				\multirow{5}{*}{F3} 
				& min       & 2.8686e+00 & 1.0000e+00 & 1.0194e+00 & 3.4383e+00 & 2.3874e+00 & 3.7517e+00 & 1.0000e+00 & 1.4641e+00 & 5.3460e+00 & 1.4091e+00 \\
				& mean      & 4.4821e+00 & \textbf{1.3824e+00} & 3.9939e+00 & 4.6360e+00 & 3.3832e+00 & 4.7746e+00 & 1.4819e+00 & 2.8048e+00 & 6.3694e+00 & 2.1769e+00 \\
				& std       & 8.6505e-01 & 1.0400e-01 & 2.7248e+00 & 6.3815e-01 & 6.3846e-01 & 6.5220e-01 & 2.8144e-01 & 7.6472e-01 & 6.1405e-01 & 6.9735e-01 \\
				& p-value   & 1.7344e-06 & 3.1817e-06 & 8.2167e-03 & 1.7344e-06 & 2.5967e-05 & 1.7344e-06 & 1.6046e-04 & 1.8326e-03 & 1.7344e-06 & - - \\
				& + / = / - & +          & -          & +          & +          & +          & +          & -          & +          & +          & - - \\
				\midrule
				
				\multirow{5}{*}{F4} 
				& min       & 1.0736e+01 & 5.9748e+00 & 5.2218e+00 & 7.1909e+00 & 4.7577e+00 & 4.7637e+00 & 4.9798e+00 & 3.9943e+00 & 2.0143e+00 & 2.9899e+00 \\
				& mean      & 1.5684e+01 & 1.4564e+01 & 1.3344e+01 & 1.2320e+01 & 7.7157e+00 & 6.8321e+00 & 7.3346e+00 & 7.3134e+00 & 1.4336e+01 & \textbf{5.7986e+00} \\
				& std       & 3.0569e+00 & 4.6684e+00 & 5.2554e+00 & 2.6726e+00 & 1.7132e+00 & 1.2105e+00 & 2.1622e+00 & 2.3267e+00 & 9.2069e+00 & 1.7583e+00 \\
				& p-value   & 1.7344e-06 & 1.7300e-06 & 5.2165e-06 & 2.3534e-06 & 2.5846e-03 & 1.3194e-02 & 6.2241e-03 & 2.9575e-03 & 1.7423e-04 & - - \\
				& + / = / - & +          & +          & +          & +          & +          & +          & +          & +          & +          & - - \\
				\midrule
				
				\multirow{5}{*}{F5} 
				& min       & 1.1977e+00 & 1.0148e+00 & 1.2885e+00 & 1.0186e+00 & 1.0008e+00 & 1.0065e+00 & 1.0074e+00 & 1.0004e+00 & 1.0000e+00 & 1.0000e+00 \\
				& mean      & 1.3585e+00 & 1.1325e+00 & 1.5922e+00 & 1.1116e+00 & 1.0495e+00 & 1.0475e+00 & 1.0441e+00 & 1.0279e+00 & \textbf{1.0106e+00} & 1.0365e+00 \\
				& std       & 6.0109e-02 & 9.4873e-02 & 1.1393e-01 & 6.2064e-02 & 2.8975e-02 & 3.7065e-02 & 2.5913e-02 & 2.6251e-02 & 1.0906e-02 & 2.6088e-02 \\
				& p-value   & 1.7344e-06 & 2.6033e-06 & 1.7344e-06 & 1.3601e-05 & 5.7096e-02 & 4.6528e-01 & 1.6817e-01 & 1.4704e-01 & 1.2506e-04 & - - \\
				& + / = / - & +          & +          & +          & +          & =          & =          & =          & =          & -          & - - \\
				\midrule
				
				\multirow{5}{*}{F6} 
				& min       & 1.7105e+00 & 1.7918e+00 & 1.2533e+00 & 1.0002e+00 & 1.0003e+00 & 1.0548e+00 & 1.0000e+00 & 1.0026e+00 & 1.0002e+00 & 1.0000e+00 \\
				& mean      & 2.4178e+00 & 3.6837e+00 & 2.3501e+00 & 1.0219e+00 & 1.2944e+00 & 1.3901e+00 & 1.1815e+00 & 1.1058e+00 & \textbf{1.0038e+00} & 1.2043e+00 \\
				& std       & 5.2818e-01 & 1.6434e+00 & 1.1164e+00 & 3.8613e-02 & 3.1663e-01 & 1.6814e-01 & 4.8295e-01 & 1.2875e-01 & 3.8249e-03 & 4.5149e-01 \\
				& p-value   & 6.3391e-06 & 1.7344e-06 & 4.8603e-05 & 1.2453e-02 & 8.2206e-02 & 2.7653e-03 & 7.0356e-01 & 8.9364e-01 & 1.0357e-03 & - - \\
				& + / = / - & +          & +          & +          & -          & =          & +          & =          & =          & -          & - - \\
				\midrule
				
				\multirow{5}{*}{F7} 
				& min       & 3.4624e+02 & 2.1477e+02 & 1.5420e+02 & 3.7849e+02 & 4.1189e+01 & 8.9129e+01 & 1.2498e+00 & 1.6280e+01 & 9.8059e+00 & 1.0625e+00 \\
				& mean      & 6.9658e+02 & 6.0591e+02 & 7.1670e+02 & 6.9908e+02 & 3.9001e+02 & 3.1175e+02 & 5.3582e+02 & 3.1001e+02 & 7.4440e+02 & \textbf{2.2845e+02} \\
				& std       & 1.5966e+02 & 3.1138e+02 & 2.0804e+02 & 1.7268e+02 & 1.7203e+02 & 1.0652e+02 & 2.4347e+02 & 1.9338e+02 & 4.4503e+02 & 1.6661e+02 \\
				& p-value   & 1.7344e-06 & 7.6909e-06 & 5.2165e-06 & 1.7344e-06 & 2.1053e-03 & 4.2767e-02 & 4.8603e-05 & 1.0639e-01 & 4.4493e-05 & - - \\
				& + / = / - & +          & +          & +          & +          & +          & +          & +          & =          & +          & - - \\
				\midrule
				
				\multirow{5}{*}{F8} 
				& min       & 2.7922e+00 & 2.3116e+00 & 2.4718e+00 & 2.9220e+00 & 3.0313e+00 & 2.7059e+00 & 2.2324e+00 & 2.7189e+00 & 2.3457e+00 & 1.6707e+00 \\
				& mean      & 3.7013e+00 & 3.3795e+00 & 3.6342e+00 & 3.7291e+00 & 3.5635e+00 & 3.4934e+00 & 3.2165e+00 & 3.4232e+00 & 3.2377e+00 & \textbf{2.8184e+00} \\
				& std       & 2.9685e-01 & 5.4404e-01 & 5.0178e-01 & 3.6959e-01 & 2.6664e-01 & 3.2041e-01 & 5.3316e-01 & 3.0598e-01 & 3.5622e-01 & 5.1828e-01 \\
				& p-value   & 3.5152e-06 & 1.4839e-03 & 7.6909e-06 & 6.9838e-06 & 5.2165e-06 & 9.3157e-06 & 6.8359e-03 & 5.3070e-05 & 8.3071e-04 & - - \\
				& + / = / - & +          & +          & +          & +          & +          & +          & +          & +          & +          & - - \\
				\midrule
				
				\multirow{5}{*}{F9} 
				& min       & 1.2440e+00 & 1.0181e+00 & 1.0974e+00 & 1.1345e+00 & 1.0966e+00 & 1.0698e+00 & 1.0327e+00 & 1.0435e+00 & 1.0581e+00 & 1.0364e+00 \\
				& mean      & 1.3321e+00 & 1.1565e+00 & 1.1923e+00 & 1.2134e+00 & 1.1979e+00 & 1.1499e+00 & \textbf{1.1116e+00} & 1.1395e+00 & 1.1281e+00 & 1.1168e+00 \\
				& std       & 4.5428e-02 & 8.8229e-02 & 6.1984e-02 & 4.7016e-02 & 5.5961e-02 & 3.1818e-02 & 4.7704e-02 & 6.0013e-02 & 3.7864e-02 & 5.6291e-02 \\
				& p-value   & 1.7344e-06 & 9.7772e-02 & 2.4118e-04 & 2.5967e-05 & 5.3070e-05 & 1.5927e-03 & 9.4261e-01 & 7.5213e-02 & 2.0589e-01 & - - \\
				& + / = / - & +          & =          & +          & +          & +          & +          & =          & =          & =          & - - \\
				\midrule
				
				\multirow{5}{*}{F10} 
				& min       & 4.2901e+00 & 2.0995e+01 & 2.1147e+01 & 1.6373e+01 & 1.7477e+01 & 1.3376e+01 & 3.3168e+00 & 7.9040e+00 & 1.3594e+00 & 1.0000e+00 \\
				& mean      & 2.0003e+01 & 2.1117e+01 & 2.1351e+01 & 2.0430e+01 & 2.0982e+01 & 2.0802e+01 & 2.0250e+01 & 2.0128e+01 & \textbf{1.7724e+01} & 1.7810e+01 \\
				& std       & 3.9287e+00 & 1.0156e-01 & 1.0326e-01 & 1.5596e+00 & 6.6295e-01 & 1.4145e+00 & 4.5690e+00 & 3.4256e+00 & 6.9556e+00 & 7.3119e+00 \\
				& p-value   & 1.8910e-04 & 1.6394e-05 & 1.7344e-06 & 1.5658e-02 & 1.9729e-05 & 1.6046e-04 & 1.6046e-04 & 1.1138e-03 & 1.6503e-01 & - - \\
				& + / = / - & +          & +          & +          & +          & +          & +          & +          & +          & =          & - - \\
				\midrule
				
				Friedman & Mean rank & 6.8500 & 5.6500 & 8.5000 & 7.2000 & 6.3000 & 5.8000 & 4.1000 & 3.9000 & 4.4000 & 2.3000 \\
				\midrule
				ToL Wilcoxon & + / = / - & 9/0/1 & 7/1/2 & 10/0/0 & 9/0/1 & 7/3/0 & 9/1/0 & 6/3/1 & 6/4/0 & 5/3/2 & - - \\
				\bottomrule
			\end{tabular}
		\end{adjustbox}
	\end{sidewaystable*}
	\clearpage
	
	\begin{sidewaystable*}[p]
		\centering
		
		\captionsetup{
			font=small,
			skip=2pt
		}
		
		\caption{Test results for CEC 2022 ($D=20$) -- Advanced Group}
		\label{tab:cec2022_results_advanced}
		
		\scriptsize
		\setlength{\tabcolsep}{2.5pt}
		\renewcommand{\arraystretch}{0.72}
		
		\begin{adjustbox}{
				width=0.98\textheight,
				max totalheight=0.80\textwidth,
				keepaspectratio,
				center
			}
			
			\begin{tabular}{c c cccccccccc}
				\toprule
				
				Function & & WOO & CGO & RWGWO & SaDE & SHADE & LSHADE &
				APSM\_jSO & NL\_SHADE\_RSP & LSRTDE & IWOO \\
				
				\midrule
				
				\multirow{5}{*}{F1}
				& min       & 2.6163e+03 & 3.0000e+02 & 6.1185e+02 & 3.8070e+02 & 3.0008e+02 & 3.1900e+02 & 3.0001e+02 & 3.0396e+02 & 3.0054e+02 & 3.2994e+02 \\
				& mean      & 6.2188e+03 & \textbf{3.0000e+02} & 3.6143e+03 & 2.3269e+03 & 5.0779e+03 & 8.8778e+03 & 3.0260e+02 & 3.7091e+02 & 3.0880e+02 & 9.8987e+02 \\
				& std       & 3.1658e+03 & 1.6258e-03 & 2.1408e+03 & 1.5091e+03 & 1.0879e+04 & 1.2133e+04 & 7.2077e+00 & 7.7558e+01 & 8.8639e+00 & 5.0559e+02 \\
				& p-value   & 1.7344e-06 & 1.7344e-06 & 8.4661e-06 & 7.5137e-05 & 2.0589e-01 & 1.4704e-01 & 1.7344e-06 & 2.3534e-06 & 1.7344e-06 & -- \\
				& $+ / = / -$ & + & - & + & + & = & = & - & - & - & -- \\
				\midrule
				
				\multirow{5}{*}{F2}
				& min       & 4.5504e+02 & 4.0008e+02 & 4.4525e+02 & 4.4490e+02 & 4.4490e+02 & 4.4493e+02 & 4.0577e+02 & 4.4490e+02 & 4.4491e+02 & 4.0001e+02 \\
				& mean      & 4.7207e+02 & \textbf{4.4465e+02} & 4.6217e+02 & 4.5203e+02 & 4.4839e+02 & 4.4896e+02 & 4.5252e+02 & 4.4853e+02 & 4.4822e+02 & 4.4950e+02 \\
				& std       & 1.0281e+01 & 1.8215e+01 & 1.8631e+01 & 8.1074e+00 & 1.5878e+00 & 7.6259e-01 & 1.3807e+01 & 1.4477e+00 & 5.0009e+00 & 1.1262e+01 \\
				& p-value   & 2.1266e-06 & 7.5213e-02 & 5.7924e-05 & 1.1973e-03 & 5.5774e-01 & 1.1748e-02 & 2.1336e-01 & 4.4919e-02 & 2.6230e-01 & -- \\
				& $+ / = / -$ & + & = & + & + & = & - & = & - & = & -- \\
				\midrule
				
				\multirow{5}{*}{F3}
				& min       & 6.0258e+02 & 6.0266e+02 & 6.0037e+02 & 6.0000e+02 & 6.0000e+02 & 6.0048e+02 & 6.0005e+02 & 6.0007e+02 & 6.0014e+02 & 6.0000e+02 \\
				& mean      & 6.0388e+02 & 6.0841e+02 & 6.0238e+02 & 6.0001e+02 & 6.0009e+02 & 6.0156e+02 & \textbf{6.0000e+02} & 6.0002e+02 & 6.0033e+02 & \textbf{6.0000e+02} \\
				& std       & 1.1253e+00 & 4.2856e+00 & 2.5679e+00 & 3.2938e-03 & 6.0734e-02 & 4.6455e-01 & 1.2097e-04 & 1.0441e-01 & 1.3612e-01 & 1.6905e-05 \\
				& p-value   & 1.7344e-06 & 1.7344e-06 & 1.7344e-06 & 1.7344e-06 & 1.7344e-06 & 1.7344e-06 & 1.7344e-06 & 1.7344e-06 & 1.7344e-06 & -- \\
				& $+ / = / -$ & + & + & + & + & + & + & + & + & + & -- \\
				\midrule
				
				\multirow{5}{*}{F4}
				& min       & 8.2776e+02 & 8.1890e+02 & 8.3348e+02 & 8.4231e+02 & 8.2965e+02 & 8.3932e+02 & 8.1592e+02 & 8.2197e+02 & 8.7421e+02 & 8.0895e+02 \\
				& mean      & 8.5971e+02 & 8.3389e+02 & 8.4635e+02 & 8.8725e+02 & 8.4595e+02 & 8.6301e+02 & 8.2566e+02 & 8.3783e+02 & 9.0344e+02 & \textbf{8.2486e+02} \\
				& std       & 1.2501e+01 & 1.0047e+01 & 8.1326e+00 & 1.6847e+01 & 1.1735e+01 & 1.2575e+01 & 6.9450e+00 & 1.2124e+01 & 1.1227e+01 & 8.0409e+00 \\
				& p-value   & 1.7344e-06 & 2.1619e-05 & 1.7344e-06 & 1.7344e-06 & 1.9209e-06 & 1.7344e-06 & 8.2901e-01 & 8.4661e-06 & 1.7344e-06 & -- \\
				& $+ / = / -$ & + & + & + & + & + & + & = & + & + & -- \\
				\midrule
				
				\multirow{5}{*}{F5}
				& min       & 9.6695e+02 & 9.3413e+02 & 9.0101e+02 & 9.0000e+02 & 9.0000e+02 & 9.0033e+02 & 9.0118e+02 & 9.0010e+02 & 9.0001e+02 & 9.0000e+02 \\
				& mean      & 1.0545e+03 & 1.0918e+03 & 9.5736e+02 & \textbf{9.0015e+02} & 9.0326e+02 & 9.0234e+02 & 9.4215e+02 & 9.0159e+02 & 9.0071e+02 & 9.0103e+02 \\
				& std       & 7.1918e+01 & 1.3060e+02 & 1.0903e+02 & 3.5587e-01 & 7.9063e+00 & 2.3349e+00 & 8.6591e+01 & 1.5271e+00 & 6.3913e-01 & 1.3175e+00 \\
				& p-value   & 1.7344e-06 & 1.7344e-06 & 2.3534e-06 & 1.4773e-04 & 7.2710e-03 & 8.2167e-03 & 3.1817e-06 & 1.2544e-01 & 9.2626e-01 & -- \\
				& $+ / = / -$ & + & + & + & - & + & + & + & = & = & -- \\
				\midrule
				
				\multirow{5}{*}{F6}
				& min       & 6.7296e+04 & 1.9092e+03 & 3.6193e+03 & 1.8503e+03 & 1.8638e+03 & 1.9207e+03 & 1.8330e+03 & 1.8691e+03 & 1.8147e+03 & 1.8741e+03 \\
				& mean      & 2.4874e+05 & 6.3352e+03 & 1.3652e+04 & 4.4218e+03 & 6.7510e+03 & 3.1351e+03 & 2.2618e+03 & 2.8513e+03 & \textbf{1.8490e+03} & 3.7889e+03 \\
				& std       & 1.2546e+05 & 4.9905e+03 & 1.0304e+04 & 3.3634e+03 & 2.3778e+04 & 2.9165e+03 & 1.1743e+03 & 1.7244e+03 & 2.0202e+01 & 2.7767e+03 \\
				& p-value   & 1.7344e-06 & 2.7029e-02 & 1.7988e-05 & 5.8571e-01 & 2.7653e-03 & 1.2044e-01 & 2.7653e-03 & 1.2044e-01 & 1.7344e-06 & -- \\
				& $+ / = / -$ & + & + & + & = & + & = & - & = & - & -- \\
				\midrule
				
				\multirow{5}{*}{F7}
				& min       & 2.0414e+03 & 2.0301e+03 & 2.0320e+03 & 2.0398e+03 & 2.0284e+03 & 2.0490e+03 & 2.0266e+03 & 2.0275e+03 & 2.0236e+03 & 2.0125e+03 \\
				& mean      & 2.0581e+03 & 2.0550e+03 & 2.0719e+03 & 2.0536e+03 & 2.0469e+03 & 2.0604e+03 & 2.0449e+03 & 2.0411e+03 & \textbf{2.0294e+03} & 2.0306e+03 \\
				& std       & 8.4688e+00 & 1.5708e+01 & 3.1918e+01 & 6.7181e+00 & 1.0567e+01 & 7.9964e+00 & 1.1894e+01 & 6.6732e+00 & 5.1714e+00 & 6.6860e+00 \\
				& p-value   & 1.7344e-06 & 1.7344e-06 & 2.1266e-06 & 1.7344e-06 & 1.4936e-05 & 1.7344e-06 & 2.8434e-05 & 8.4661e-06 & 2.6230e-01 & -- \\
				& $+ / = / -$ & + & + & + & + & + & + & + & + & = & -- \\
				\midrule
				
				\multirow{5}{*}{F8}
				& min       & 2.2264e+03 & 2.2206e+03 & 2.2271e+03 & 2.2261e+03 & 2.2234e+03 & 2.2261e+03 & 2.2215e+03 & 2.2226e+03 & 2.2109e+03 & 2.2226e+03 \\
				& mean      & 2.2301e+03 & 2.2403e+03 & 2.2434e+03 & 2.2284e+03 & 2.2301e+03 & 2.2302e+03 & 2.2375e+03 & 2.2255e+03 & 2.2286e+03 & \textbf{2.2249e+03} \\
				& std       & 2.1480e+00 & 4.0637e+01 & 3.1138e+01 & 1.1882e+00 & 2.1897e+01 & 2.0167e+00 & 3.4460e+01 & 1.3772e+00 & 3.6905e+00 & 1.2439e+00 \\
				& p-value   & 1.7344e-06 & 9.2626e-01 & 1.7344e-06 & 2.1266e-06 & 9.2710e-03 & 1.7344e-06 & 2.7029e-02 & 4.2767e-02 & 3.1123e-05 & -- \\
				& $+ / = / -$ & + & = & + & + & + & + & + & + & + & -- \\
				\midrule
				
				\multirow{5}{*}{F9}
				& min       & 2.4839e+03 & 2.4800e+03 & 2.4809e+03 & 2.4800e+03 & 2.4808e+03 & 2.4800e+03 & 2.4800e+03 & 2.4800e+03 & 2.4808e+03 & 2.4800e+03 \\
				& mean      & 2.4860e+03 & \textbf{2.4800e+03} & 2.4818e+03 & \textbf{2.4800e+03} & 2.4808e+03 & \textbf{2.4800e+03} & \textbf{2.4800e+03} & \textbf{2.4800e+03} & 2.4808e+03 & \textbf{2.4800e+03} \\
				& std       & 1.0833e+00 & 1.4595e-11 & 1.4461e+00 & 3.0437e-05 & 2.7136e-07 & 2.6177e-02 & 1.1087e-08 & 5.3889e-04 & 2.1544e-02 & 2.0562e-05 \\
				& p-value   & 1.7344e-06 & 1.7344e-06 & 1.7344e-06 & 1.2544e-01 & 1.7344e-06 & 1.7344e-06 & 1.7344e-06 & 1.9209e-06 & 1.7344e-06 & -- \\
				& $+ / = / -$ & + & + & + & = & - & + & - & + & + & -- \\
				\midrule
				
				\multirow{5}{*}{F10}
				& min       & 2.5005e+03 & 2.5005e+03 & 2.5005e+03 & 2.5006e+03 & 2.4100e+03 & 2.5005e+03 & 2.4215e+03 & 2.5005e+03 & 2.5005e+03 & 2.4021e+03 \\
				& mean      & 2.5241e+03 & 2.5008e+03 & 3.4140e+03 & 2.5282e+03 & 2.5111e+03 & 2.5009e+03 & 2.6255e+03 & 2.5008e+03 & 2.5007e+03 & \textbf{2.4973e+03} \\
				& std       & 7.5085e+01 & 2.6936e-01 & 8.1086e+02 & 8.8511e+01 & 6.7666e+01 & 7.9703e-01 & 1.8130e+02 & 1.5610e-01 & 1.1047e-01 & 1.7987e+01 \\
				& p-value   & 1.6394e-05 & 2.3704e-05 & 1.2381e-05 & 8.4661e-06 & 4.0483e-01 & 2.8434e-05 & 2.2248e-04 & 6.8923e-05 & 4.4493e-05 & -- \\
				& $+ / = / -$ & + & + & + & + & = & + & + & + & + & -- \\
				\midrule
				
				\multirow{5}{*}{F11}
				& min       & 3.2887e+03 & 2.9000e+03 & 2.9139e+03 & 2.9000e+03 & 2.9000e+03 & 2.9074e+03 & 2.9000e+03 & 2.9000e+03 & 2.9077e+03 & 2.9000e+03 \\
				& mean      & 3.5388e+03 & \textbf{2.9000e+03} & 2.9296e+03 & \textbf{2.9000e+03} & \textbf{2.9000e+03} & 2.9217e+03 & \textbf{2.9000e+03} & 2.9012e+03 & 2.9262e+03 & \textbf{2.9000e+03} \\
				& std       & 1.3231e+02 & 2.6137e-12 & 1.0592e+01 & 4.8399e-03 & 5.0900e-04 & 8.2332e+00 & 1.2348e-04 & 6.0751e-01 & 1.4853e+01 & 4.7017e-13 \\
				& p-value   & 1.7344e-06 & 2.0079e-05 & 1.7344e-06 & 1.7344e-06 & 1.7344e-06 & 1.7344e-06 & 1.7344e-06 & 1.7344e-06 & 1.7344e-06 & -- \\
				& $+ / = / -$ & + & + & + & + & + & + & + & + & + & -- \\
				\midrule
				
				\multirow{5}{*}{F12}
				& min       & 2.9666e+03 & 2.9432e+03 & 2.9358e+03 & 2.9332e+03 & 2.9320e+03 & 2.9349e+03 & 2.9374e+03 & 2.9334e+03 & 2.9353e+03 & 2.9384e+03 \\
				& mean      & 3.0065e+03 & 2.9964e+03 & 2.9533e+03 & 2.9405e+03 & 2.9433e+03 & 2.9400e+03 & 2.9586e+03 & \textbf{2.9397e+03} & 2.9448e+03 & 2.9541e+03 \\
				& std       & 2.4985e+01 & 5.8262e+01 & 8.4110e+00 & 4.0970e+00 & 6.7007e+00 & 3.1285e+00 & 1.5398e+01 & 4.3593e+00 & 5.9743e+00 & 1.0150e+01 \\
				& p-value   & 1.9209e-06 & 1.1499e-04 & 8.9364e-01 & 1.1265e-05 & 2.0515e-04 & 3.5152e-06 & 2.1336e-01 & 5.2165e-06 & 6.8923e-05 & -- \\
				& $+ / = / -$ & + & + & = & - & - & - & = & - & - & -- \\
				\midrule
				Friedman & Mean rank & 8.7500 & 5.3333 & 8.4167 & 5.0833 & 5.0000 & 6.3333 & 5.0833 & 3.6667 & 4.2500 & \textbf{3.0833} \\
				\midrule
				ToL Wilcoxon & $+ / = / -$ & 12/0/0 & 8/2/2 & 11/1/0 & 8/2/2 & 7/3/2 & 8/2/2 & 6/3/3 & 7/2/3 & 6/3/3 & -- \\
				\bottomrule
			\end{tabular}
			
		\end{adjustbox}
		
	\end{sidewaystable*}
	
	\clearpage
	
	\begingroup
	\renewcommand{\tablename}{Table}
	\captionsetup[table]{font=small,skip=2pt}
	
	\begin{sidewaystable*}[p]
		\centering
		
		\caption{Test results for CEC 2017 ($D=50$) -- Advanced Group}
		\label{tab:cec2017_50d_results_advanced}
		
		\scriptsize
		\setlength{\tabcolsep}{2.8pt}
		\renewcommand{\arraystretch}{0.72}
		
		\begin{adjustbox}{
				width=0.98\textheight,
				max totalheight=0.88\textwidth,
				keepaspectratio,
				center
			}
			\begin{tabular}{c c cccccccccc}
				\toprule
				Function & & WOO & CGO & RWGWO & SaDE & SHADE & LSHADE & APSM\_jSO & NL\_SHADE\_RSP & LSRTDE & IWOO \\
				\midrule
				
				\multirow{5}{*}{F1}
				& min       & 2.4631e+09 & 5.2574e+02 & 2.8481e+07 & 1.5194e+05 & 3.6331e+03 & 2.5810e+08 & 2.1401e+05 & 1.3006e+08 & 1.1468e+09 & 1.0002e+02 \\
				& mean      & 3.9074e+09 & 8.2762e+04 & 1.2047e+08 & 1.0784e+06 & 1.1436e+05 & 4.5194e+08 & 2.9620e+06 & 3.2372e+08 & 2.4856e+09 & \textbf{1.7274e+03} \\
				& std       & 8.4623e+08 & 1.6779e+05 & 6.2804e+07 & 1.1284e+06 & 1.1701e+05 & 1.1851e+08 & 3.2257e+06 & 1.1722e+08 & 7.2024e+08 & 2.6363e+03 \\
				& p-value   & 1.7344e-06 & 6.9838e-06 & 1.7344e-06 & 1.7344e-06 & 1.7344e-06 & 1.7344e-06 & 1.7344e-06 & 1.7344e-06 & 1.7344e-06 & -- \\
				& $+ / = / -$ & + & + & + & + & + & + & + & + & + & -- \\
				\midrule
				
				\multirow{5}{*}{F3}
				& min       & 6.9179e+04 & 1.8108e+04 & 1.0299e+05 & 5.1321e+04 & 4.6874e+04 & 7.8304e+04 & 1.9122e+04 & 4.9497e+04 & 4.2667e+04 & 6.4634e+04 \\
				& mean      & 1.1281e+05 & \textbf{3.2809e+04} & 1.5211e+05 & 1.0433e+05 & 1.4713e+05 & 2.1710e+05 & 3.6553e+04 & 6.9681e+04 & 6.8078e+04 & 8.7236e+04 \\
				& std       & 1.8884e+04 & 1.0357e+04 & 3.1910e+04 & 2.2067e+04 & 9.2800e+04 & 6.5901e+04 & 1.0953e+04 & 1.2316e+04 & 1.3163e+04 & 1.2750e+04 \\
				& p-value   & 5.3070e-05 & 1.7344e-06 & 1.7344e-06 & 3.8542e-03 & 7.1903e-02 & 3.5152e-06 & 1.7344e-06 & 1.6394e-05 & 8.4661e-06 & -- \\
				& $+ / = / -$ & + & - & + & + & = & + & - & - & - & -- \\
				\midrule
				
				\multirow{5}{*}{F4}
				& min       & 1.0726e+03 & 4.4326e+02 & 5.6222e+02 & 4.9079e+02 & 4.6210e+02 & 6.8254e+02 & 5.0278e+02 & 6.2182e+02 & 7.7720e+02 & 4.2934e+02 \\
				& mean      & 1.2662e+03 & 5.6495e+02 & 6.5070e+02 & 6.2940e+02 & 5.6608e+02 & 7.4713e+02 & 5.8849e+02 & 6.7828e+02 & 9.3966e+02 & \textbf{5.2829e+02} \\
				& std       & 1.3205e+02 & 6.2262e+01 & 5.0030e+01 & 5.7444e+01 & 3.9905e+01 & 2.9887e+01 & 3.9958e+01 & 2.6808e+01 & 8.9275e+01 & 5.9031e+01 \\
				& p-value   & 1.7344e-06 & 9.8421e-03 & 6.3391e-06 & 1.1265e-05 & 1.8519e-02 & 1.7344e-06 & 1.3595e-04 & 1.7344e-06 & 1.7344e-06 & -- \\
				& $+ / = / -$ & + & + & + & + & + & + & + & + & + & -- \\
				\midrule
				
				\multirow{5}{*}{F5}
				& min       & 7.7035e+02 & 6.7312e+02 & 6.2705e+02 & 7.6999e+02 & 6.7352e+02 & 8.2167e+02 & 6.0173e+02 & 7.1127e+02 & 8.7851e+02 & 5.6169e+02 \\
				& mean      & 8.0656e+02 & 7.2434e+02 & 8.0944e+02 & 8.2259e+02 & 7.3782e+02 & 8.9129e+02 & 6.7357e+02 & 7.5872e+02 & 9.2203e+02 & \textbf{5.8928e+02} \\
				& std       & 1.8782e+01 & 3.5128e+01 & 5.6964e+01 & 2.5815e+01 & 2.6524e+01 & 3.1989e+01 & 3.6737e+01 & 2.9832e+01 & 2.2533e+01 & 1.7086e+01 \\
				& p-value   & 1.7344e-06 & 1.7344e-06 & 1.7344e-06 & 1.7344e-06 & 1.7344e-06 & 1.7344e-06 & 1.7344e-06 & 1.7344e-06 & 1.7344e-06 & -- \\
				& $+ / = / -$ & + & + & + & + & + & + & + & + & + & -- \\
				\midrule
				
				\multirow{5}{*}{F6}
				& min       & 6.1063e+02 & 6.2411e+02 & 6.0658e+02 & 6.0065e+02 & 6.0173e+02 & 6.1081e+02 & 6.0679e+02 & 6.0672e+02 & 6.1689e+02 & 6.0002e+02 \\
				& mean      & 6.1606e+02 & 6.3509e+02 & 6.1378e+02 & 6.0113e+02 & 6.0370e+02 & 6.1671e+02 & 6.1869e+02 & 6.1100e+02 & 6.2596e+02 & \textbf{6.0019e+02} \\
				& std       & 2.0950e+00 & 5.6797e+00 & 4.6112e+00 & 3.7750e-01 & 1.3322e+00 & 3.0039e+00 & 6.7806e+00 & 3.4343e+00 & 3.9725e+00 & 1.5788e-01 \\
				& p-value   & 1.7344e-06 & 1.7344e-06 & 1.7344e-06 & 1.7344e-06 & 1.7344e-06 & 1.7344e-06 & 1.7344e-06 & 1.7344e-06 & 1.7344e-06 & -- \\
				& $+ / = / -$ & + & + & + & + & + & + & + & + & + & -- \\
				\midrule
				
				\multirow{5}{*}{F7}
				& min       & 1.2025e+03 & 1.0378e+03 & 9.7707e+02 & 1.0557e+03 & 1.0159e+03 & 1.1015e+03 & 1.0050e+03 & 9.8730e+02 & 1.1230e+03 & 8.0091e+02 \\
				& mean      & 1.2832e+03 & 1.1468e+03 & 1.0937e+03 & 1.0983e+03 & 1.0607e+03 & 1.1593e+03 & 1.1296e+03 & 1.0293e+03 & 1.2052e+03 & \textbf{8.3478e+02} \\
				& std       & 3.7781e+01 & 8.0052e+01 & 9.1383e+01 & 2.1655e+01 & 3.1608e+01 & 2.6024e+01 & 8.4850e+01 & 2.3215e+01 & 2.6887e+01 & 2.0289e+01 \\
				& p-value   & 1.7344e-06 & 1.7344e-06 & 1.7344e-06 & 1.7344e-06 & 1.7344e-06 & 1.7344e-06 & 1.7344e-06 & 1.7344e-06 & 1.7344e-06 & -- \\
				& $+ / = / -$ & + & + & + & + & + & + & + & + & + & -- \\
				\midrule
				
				\multirow{5}{*}{F8}
				& min       & 1.0561e+03 & 9.7412e+02 & 9.9172e+02 & 1.1072e+03 & 9.8471e+02 & 1.1420e+03 & 9.0111e+02 & 1.0010e+03 & 1.1156e+03 & 8.5373e+02 \\
				& mean      & 1.1067e+03 & 1.0277e+03 & 1.1250e+03 & 1.1305e+03 & 1.0387e+03 & 1.1823e+03 & 9.8040e+02 & 1.0471e+03 & 1.2180e+03 & \textbf{8.8282e+02} \\
				& std       & 2.6073e+01 & 3.3247e+01 & 5.2610e+01 & 1.6194e+01 & 2.3073e+01 & 2.3081e+01 & 4.0297e+01 & 2.8923e+01 & 2.5916e+01 & 1.5138e+01 \\
				& p-value   & 1.7344e-06 & 1.7344e-06 & 1.7344e-06 & 1.7344e-06 & 1.7344e-06 & 1.7344e-06 & 1.7344e-06 & 1.7344e-06 & 1.7344e-06 & -- \\
				& $+ / = / -$ & + & + & + & + & + & + & + & + & + & -- \\
				\midrule
				
				\multirow{5}{*}{F9}
				& min       & 3.0861e+03 & 3.9499e+03 & 3.0656e+03 & 1.1062e+03 & 1.1312e+03 & 1.6080e+03 & 3.0204e+03 & 1.8140e+03 & 2.3360e+03 & 9.4813e+02 \\
				& mean      & 4.9761e+03 & 5.6744e+03 & 7.4437e+03 & 1.7085e+03 & 2.4176e+03 & 2.3390e+03 & 5.7189e+03 & 3.6734e+03 & 3.6205e+03 & \textbf{1.0884e+03} \\
				& std       & 1.3169e+03 & 1.1512e+03 & 1.8859e+03 & 4.4748e+02 & 1.2748e+03 & 5.1950e+02 & 1.9652e+03 & 1.0154e+03 & 8.2630e+02 & 1.3626e+02 \\
				& p-value   & 1.7344e-06 & 1.7344e-06 & 1.7344e-06 & 4.2857e-06 & 1.7344e-06 & 1.7344e-06 & 1.7344e-06 & 1.7344e-06 & 1.7344e-06 & -- \\
				& $+ / = / -$ & + & + & + & + & + & + & + & + & + & -- \\
				\midrule
				
				\multirow{5}{*}{F10}
				& min       & 9.6302e+03 & 5.3545e+03 & 1.0187e+04 & 1.0521e+04 & 8.6062e+03 & 1.2003e+04 & 5.8003e+03 & 9.3261e+03 & 1.4339e+04 & 4.9426e+03 \\
				& mean      & 1.0641e+04 & 8.4548e+03 & 1.1088e+04 & 1.2009e+04 & 9.8651e+03 & 1.2982e+04 & 8.7811e+03 & 1.0808e+04 & 1.5318e+04 & \textbf{6.4364e+03} \\
				& std       & 4.6358e+02 & 1.8240e+03 & 6.5056e+02 & 7.5480e+02 & 4.2250e+02 & 5.8504e+02 & 1.3369e+03 & 5.9667e+02 & 3.8028e+02 & 1.1194e+03 \\
				& p-value   & 1.7344e-06 & 3.1123e-05 & 1.7344e-06 & 1.7344e-06 & 1.7344e-06 & 1.7344e-06 & 1.6394e-05 & 1.7344e-06 & 1.7344e-06 & -- \\
				& $+ / = / -$ & + & + & + & + & + & + & + & + & + & -- \\
				\midrule
				
				\multirow{5}{*}{F11}
				& min       & 1.5805e+03 & 1.1954e+03 & 1.3840e+03 & 1.3927e+03 & 1.2321e+03 & 2.2524e+03 & 1.2342e+03 & 1.3796e+03 & 1.4560e+03 & 1.2116e+03 \\
				& mean      & 1.7529e+03 & 1.3274e+03 & 1.6436e+03 & 1.5849e+03 & 1.5483e+03 & 7.0849e+03 & 1.4014e+03 & 1.4610e+03 & 1.5134e+03 & \textbf{1.2646e+03} \\
				& std       & 1.0331e+02 & 7.1158e+01 & 2.0801e+02 & 1.1497e+02 & 5.3884e+02 & 4.7493e+03 & 7.8512e+01 & 6.6033e+01 & 3.9879e+01 & 3.2788e+01 \\
				& p-value   & 1.7344e-06 & 3.5888e-04 & 1.7344e-06 & 1.7344e-06 & 1.7344e-06 & 1.7344e-06 & 3.1817e-06 & 1.7344e-06 & 1.7344e-06 & -- \\
				& $+ / = / -$ & + & + & + & + & + & + & + & + & + & -- \\
				\bottomrule
			\end{tabular}
		\end{adjustbox}
	\end{sidewaystable*}
	
	\clearpage
	
	\begin{sidewaystable*}[p]
		\centering
		
		\addtocounter{table}{-1}
		\caption[]{Test results for CEC 2017 ($D=50$) -- Advanced Group (continued)}
		
		\scriptsize
		\setlength{\tabcolsep}{2.8pt}
		\renewcommand{\arraystretch}{0.72}
		
		\begin{adjustbox}{
				width=0.98\textheight,
				max totalheight=0.88\textwidth,
				keepaspectratio,
				center
			}
			\begin{tabular}{c c cccccccccc}
				\toprule
				Function & & WOO & CGO & RWGWO & SaDE & SHADE & LSHADE & APSM\_jSO & NL\_SHADE\_RSP & LSRTDE & IWOO \\
				\midrule
				
				\multirow{5}{*}{F12}
				& min       & 1.5120e+08 & 2.9856e+05 & 2.7463e+07 & 1.1735e+06 & 2.6645e+05 & 3.9907e+07 & 7.8211e+05 & 5.4694e+06 & 3.9935e+07 & 2.8973e+05 \\
				& mean      & 3.2558e+08 & 1.2871e+06 & 8.6124e+07 & 5.5123e+06 & 2.6649e+06 & 1.0073e+08 & 3.2447e+06 & 1.8875e+07 & 8.8990e+07 & \textbf{9.7720e+05} \\
				& std       & 9.6369e+07 & 6.9456e+05 & 5.0632e+07 & 2.9978e+06 & 2.4684e+06 & 3.3393e+07 & 2.1177e+06 & 8.3602e+06 & 3.5317e+07 & 5.8517e+05 \\
				& p-value   & 1.7344e-06 & 1.7138e-01 & 1.7344e-06 & 1.7344e-06 & 7.7122e-04 & 1.7344e-06 & 2.8786e-06 & 1.7344e-06 & 1.7344e-06 & -- \\
				& $+ / = / -$ & + & = & + & + & + & + & + & + & + & -- \\
				\midrule
				
				\multirow{5}{*}{F13}
				& min       & 2.7247e+06 & 2.5527e+03 & 3.9681e+05 & 2.0860e+03 & 2.9107e+03 & 4.6584e+06 & 3.0396e+03 & 1.3767e+04 & 1.2865e+05 & 1.4551e+03 \\
				& mean      & 9.9122e+06 & 1.2983e+04 & 1.3509e+06 & 1.5292e+04 & 1.8118e+04 & 1.6842e+07 & 1.1929e+04 & 4.9640e+04 & 4.7249e+05 & \textbf{3.2793e+03} \\
				& std       & 4.9383e+06 & 9.6537e+03 & 1.2788e+06 & 1.1595e+04 & 1.0342e+04 & 9.1657e+06 & 8.5425e+03 & 2.8201e+04 & 3.7566e+05 & 2.5057e+03 \\
				& p-value   & 1.7344e-06 & 3.8822e-06 & 1.7344e-06 & 4.0715e-05 & 5.7517e-06 & 1.7344e-06 & 3.4053e-05 & 1.7344e-06 & 1.7344e-06 & -- \\
				& $+ / = / -$ & + & + & + & + & + & + & + & + & + & -- \\
				\midrule
				
				\multirow{5}{*}{F14}
				& min       & 4.3799e+03 & 2.2477e+03 & 1.0106e+05 & 1.9012e+03 & 1.5470e+03 & 2.1747e+04 & 1.8995e+03 & 2.5996e+03 & 1.6172e+03 & 3.3159e+03 \\
				& mean      & 5.6266e+04 & 1.0629e+04 & 7.9605e+05 & 4.1713e+04 & 2.1880e+03 & 2.0199e+05 & 5.2107e+03 & 1.4613e+04 & \textbf{1.8368e+03} & 3.2732e+04 \\
				& std       & 3.7193e+04 & 1.8802e+04 & 8.4276e+05 & 1.0275e+05 & 9.3775e+02 & 4.3800e+05 & 4.3319e+03 & 1.1718e+04 & 1.5778e+02 & 3.0660e+04 \\
				& p-value   & 2.4308e-02 & 7.1570e-04 & 1.7344e-06 & 3.2857e-01 & 1.7344e-06 & 3.8811e-04 & 6.3391e-06 & 8.7297e-03 & 1.7344e-06 & -- \\
				& $+ / = / -$ & + & - & + & = & - & + & - & - & - & -- \\
				\midrule
				
				\multirow{5}{*}{F15}
				& min       & 1.6764e+05 & 1.8324e+03 & 3.1067e+04 & 3.1632e+03 & 2.2908e+03 & 5.9279e+05 & 2.2739e+03 & 4.5729e+03 & 9.0057e+03 & 1.8875e+03 \\
				& mean      & 4.9154e+05 & 1.1266e+04 & 1.4130e+05 & 8.0729e+03 & \textbf{5.1131e+03} & 2.6483e+06 & 7.6095e+03 & 1.3049e+04 & 1.8240e+04 & 6.6601e+03 \\
				& std       & 2.1511e+05 & 7.9653e+03 & 1.4186e+05 & 3.1733e+03 & 3.4477e+03 & 2.0345e+06 & 7.3966e+03 & 7.5090e+03 & 7.2346e+03 & 3.3723e+03 \\
				& p-value   & 1.7344e-06 & 1.0444e-02 & 1.7344e-06 & 7.5213e-02 & 5.4463e-02 & 1.7344e-06 & 9.5899e-01 & 7.5137e-05 & 1.7344e-06 & -- \\
				& $+ / = / -$ & + & + & + & = & = & + & = & + & + & -- \\
				\midrule
				
				\multirow{5}{*}{F16}
				& min       & 3.5818e+03 & 2.2395e+03 & 2.2245e+03 & 3.4965e+03 & 2.9799e+03 & 4.3615e+03 & 2.1590e+03 & 3.3911e+03 & 4.4465e+03 & 1.9757e+03 \\
				& mean      & 4.0388e+03 & 3.3315e+03 & 3.4881e+03 & 4.0550e+03 & 3.6743e+03 & 4.9535e+03 & 2.8725e+03 & 4.0637e+03 & 5.1544e+03 & \textbf{2.8025e+03} \\
				& std       & 2.1440e+02 & 4.8523e+02 & 4.9904e+02 & 2.6509e+02 & 3.1861e+02 & 3.0884e+02 & 3.0856e+02 & 3.3590e+02 & 2.3502e+02 & 3.5950e+02 \\
				& p-value   & 1.7344e-06 & 2.8308e-04 & 1.0246e-05 & 1.9209e-06 & 2.1266e-06 & 1.7344e-06 & 5.1705e-01 & 1.7344e-06 & 1.7344e-06 & -- \\
				& $+ / = / -$ & + & + & + & + & + & + & = & + & + & -- \\
				\midrule
				
				\multirow{5}{*}{F17}
				& min       & 3.1331e+03 & 2.5088e+03 & 2.1815e+03 & 2.7629e+03 & 2.5314e+03 & 3.4664e+03 & 2.4261e+03 & 2.7378e+03 & 3.2776e+03 & 2.1745e+03 \\
				& mean      & 3.7448e+03 & 3.0142e+03 & 3.0763e+03 & 3.3167e+03 & 3.2082e+03 & 4.0323e+03 & 2.8052e+03 & 3.3249e+03 & 3.9320e+03 & \textbf{2.6682e+03} \\
				& std       & 3.2141e+02 & 3.3127e+02 & 3.5204e+02 & 2.0498e+02 & 2.3161e+02 & 2.4848e+02 & 2.4379e+02 & 3.0545e+02 & 2.3485e+02 & 2.4282e+02 \\
				& p-value   & 1.7344e-06 & 8.9187e-05 & 1.1499e-04 & 1.9209e-06 & 2.8786e-06 & 1.7344e-06 & 2.8486e-02 & 2.3534e-06 & 1.7344e-06 & -- \\
				& $+ / = / -$ & + & + & + & + & + & + & + & + & + & -- \\
				\midrule
				
				\multirow{5}{*}{F18}
				& min       & 1.3469e+05 & 3.0903e+04 & 6.6287e+05 & 5.6198e+04 & 2.1084e+04 & 3.9622e+05 & 2.0992e+04 & 3.5121e+04 & 2.2631e+04 & 7.3185e+04 \\
				& mean      & 4.4874e+05 & 1.6984e+05 & 4.4812e+06 & 3.5437e+05 & 4.6485e+05 & 4.3796e+06 & 1.2132e+05 & 1.7883e+05 & \textbf{4.8147e+04} & 3.6090e+05 \\
				& std       & 3.8079e+05 & 1.2330e+05 & 2.4362e+06 & 3.0195e+05 & 1.7383e+06 & 4.4786e+06 & 1.0334e+05 & 1.4821e+05 & 1.8836e+04 & 4.8685e+05 \\
				& p-value   & 6.2683e-02 & 2.1827e-02 & 1.7344e-06 & 3.0861e-01 & 1.4839e-03 & 2.8786e-06 & 1.2506e-04 & 3.6826e-02 & 1.7344e-06 & -- \\
				& $+ / = / -$ & = & - & + & = & + & + & - & - & - & -- \\
				\midrule
				
				\multirow{5}{*}{F19}
				& min       & 1.5095e+05 & 2.3050e+03 & 1.0102e+05 & 5.2998e+03 & 2.0487e+03 & 3.1262e+05 & 2.1558e+03 & 4.3349e+03 & 5.3078e+03 & 7.5298e+03 \\
				& mean      & 3.5408e+05 & 1.6839e+04 & 1.0395e+06 & 1.5891e+04 & \textbf{4.2516e+03} & 1.8293e+06 & 1.3095e+04 & 1.4943e+04 & 1.4653e+04 & 1.5851e+04 \\
				& std       & 1.3850e+05 & 1.1966e+04 & 7.7479e+05 & 5.9525e+03 & 6.6926e+03 & 1.2669e+06 & 1.0090e+04 & 8.3263e+03 & 8.3847e+03 & 5.4240e+03 \\
				& p-value   & 1.7344e-06 & 7.6552e-01 & 1.7344e-06 & 8.6121e-01 & 9.7110e-05 & 1.7344e-06 & 2.1336e-01 & 7.1889e-01 & 3.7094e-01 & -- \\
				& $+ / = / -$ & + & = & + & = & - & + & = & = & = & -- \\
				\midrule
				
				\multirow{5}{*}{F20}
				& min       & 2.8765e+03 & 2.3990e+03 & 2.6785e+03 & 3.1366e+03 & 2.6520e+03 & 3.6359e+03 & 2.3720e+03 & 3.0946e+03 & 3.6679e+03 & 2.3017e+03 \\
				& mean      & 3.3012e+03 & 2.9741e+03 & 3.2260e+03 & 3.4238e+03 & 3.3144e+03 & 3.9520e+03 & 2.9132e+03 & 3.5042e+03 & 4.0489e+03 & \textbf{2.8064e+03} \\
				& std       & 2.1127e+02 & 3.5022e+02 & 2.5323e+02 & 1.2517e+02 & 2.0520e+02 & 1.6269e+02 & 2.4287e+02 & 1.9984e+02 & 1.4904e+02 & 2.5226e+02 \\
				& p-value   & 7.6909e-06 & 5.1931e-02 & 9.3157e-06 & 1.7344e-06 & 2.8786e-06 & 1.7344e-06 & 1.2044e-01 & 1.9209e-06 & 1.7344e-06 & -- \\
				& $+ / = / -$ & + & = & + & + & + & + & = & + & + & -- \\
				\midrule
				
				\multirow{5}{*}{F21}
				& min       & 2.5655e+03 & 2.4469e+03 & 2.4542e+03 & 2.5850e+03 & 2.4838e+03 & 2.6240e+03 & 2.4138e+03 & 2.4933e+03 & 2.6280e+03 & 2.3448e+03 \\
				& mean      & 2.6069e+03 & 2.5224e+03 & 2.5929e+03 & 2.6272e+03 & 2.5442e+03 & 2.6858e+03 & 2.4737e+03 & 2.5436e+03 & 2.7169e+03 & \textbf{2.3789e+03} \\
				& std       & 2.2686e+01 & 4.4903e+01 & 7.1302e+01 & 2.1499e+01 & 2.8041e+01 & 3.0483e+01 & 3.0177e+01 & 2.9700e+01 & 3.1658e+01 & 1.7626e+01 \\
				& p-value   & 1.7344e-06 & 1.7344e-06 & 1.7344e-06 & 1.7344e-06 & 1.7344e-06 & 1.7344e-06 & 1.7344e-06 & 1.7344e-06 & 1.7344e-06 & -- \\
				& $+ / = / -$ & + & + & + & + & + & + & + & + & + & -- \\
				\bottomrule
			\end{tabular}
		\end{adjustbox}
	\end{sidewaystable*}
	
	\clearpage
	
	\begin{sidewaystable*}[p]
		\centering
		
		\addtocounter{table}{-1}
		\caption[]{Test results for CEC 2017 ($D=50$) -- Advanced Group (continued)}
		
		\scriptsize
		\setlength{\tabcolsep}{2.8pt}
		\renewcommand{\arraystretch}{0.72}
		
		\begin{adjustbox}{
				width=0.98\textheight,
				max totalheight=0.88\textwidth,
				keepaspectratio,
				center
			}
			\begin{tabular}{c c cccccccccc}
				\toprule
				Function & & WOO & CGO & RWGWO & SaDE & SHADE & LSHADE & APSM\_jSO & NL\_SHADE\_RSP & LSRTDE & IWOO \\
				\midrule
				
				\multirow{5}{*}{F22}
				& min       & 4.6563e+03 & 2.3013e+03 & 1.0760e+04 & 2.7390e+03 & 5.1317e+03 & 1.2269e+04 & 2.3243e+03 & 1.1321e+04 & 3.3044e+03 & 2.3000e+03 \\
				& mean      & 1.1390e+04 & 9.6484e+03 & 1.2662e+04 & 1.2464e+04 & 1.1398e+04 & 1.4752e+04 & 9.3746e+03 & 1.2381e+04 & 1.3739e+04 & \textbf{8.6461e+03} \\
				& std       & 2.6345e+03 & 1.9331e+03 & 7.2024e+02 & 3.1892e+03 & 1.2487e+03 & 7.2518e+02 & 2.1983e+03 & 5.4936e+02 & 5.0707e+03 & 2.5337e+03 \\
				& p-value   & 3.5888e-04 & 1.0201e-01 & 1.9209e-06 & 4.5336e-04 & 4.0715e-05 & 1.7344e-06 & 1.8462e-01 & 2.8786e-06 & 4.0715e-05 & -- \\
				& $+ / = / -$ & + & = & + & + & + & + & = & + & + & -- \\
				\midrule
				
				\multirow{5}{*}{F23}
				& min       & 3.0840e+03 & 2.8947e+03 & 2.9568e+03 & 3.0308e+03 & 2.9287e+03 & 3.0754e+03 & 2.8793e+03 & 2.9379e+03 & 3.0890e+03 & 2.7990e+03 \\
				& mean      & 3.1835e+03 & 3.0837e+03 & 3.0767e+03 & 3.0758e+03 & 2.9798e+03 & 3.1463e+03 & 2.9960e+03 & 2.9962e+03 & 3.1688e+03 & \textbf{2.8548e+03} \\
				& std       & 8.3843e+01 & 8.6299e+01 & 4.9009e+01 & 3.0025e+01 & 2.5771e+01 & 2.7034e+01 & 6.9391e+01 & 3.2723e+01 & 4.2726e+01 & 3.1245e+01 \\
				& p-value   & 1.7344e-06 & 1.7344e-06 & 1.7344e-06 & 1.7344e-06 & 1.7344e-06 & 1.7344e-06 & 1.7344e-06 & 1.7344e-06 & 1.7344e-06 & -- \\
				& $+ / = / -$ & + & + & + & + & + & + & + & + & + & -- \\
				\midrule
				
				\multirow{5}{*}{F24}
				& min       & 3.2972e+03 & 3.0985e+03 & 3.1159e+03 & 3.2081e+03 & 2.9779e+03 & 3.2570e+03 & 3.0621e+03 & 3.1255e+03 & 3.2398e+03 & 2.9780e+03 \\
				& mean      & 3.4505e+03 & 3.2447e+03 & 3.2441e+03 & 3.2704e+03 & 3.1416e+03 & 3.3068e+03 & 3.1471e+03 & 3.1654e+03 & 3.3388e+03 & \textbf{3.0138e+03} \\
				& std       & 4.4279e+01 & 8.4375e+01 & 4.1734e+01 & 3.7549e+01 & 4.4932e+01 & 2.3769e+01 & 5.5984e+01 & 1.9967e+01 & 4.6450e+01 & 2.4703e+01 \\
				& p-value   & 1.7344e-06 & 1.7344e-06 & 1.7344e-06 & 1.7344e-06 & 1.9209e-06 & 1.7344e-06 & 1.7344e-06 & 1.7344e-06 & 1.7344e-06 & -- \\
				& $+ / = / -$ & + & + & + & + & + & + & + & + & + & -- \\
				\midrule
				
				\multirow{5}{*}{F25}
				& min       & 3.4013e+03 & 2.9885e+03 & 3.0615e+03 & 3.0476e+03 & 2.9872e+03 & 3.1314e+03 & 3.0083e+03 & 3.0936e+03 & 3.1899e+03 & 3.0121e+03 \\
				& mean      & 3.6173e+03 & 3.0798e+03 & 3.1475e+03 & 3.1303e+03 & \textbf{3.0553e+03} & 3.1745e+03 & 3.0992e+03 & 3.1401e+03 & 3.3026e+03 & 3.0747e+03 \\
				& std       & 1.0846e+02 & 3.1898e+01 & 5.4650e+01 & 3.6047e+01 & 2.9597e+01 & 2.5504e+01 & 4.1647e+01 & 3.1818e+01 & 6.0102e+01 & 2.7085e+01 \\
				& p-value   & 1.7344e-06 & 3.2857e-01 & 4.2857e-06 & 1.1265e-05 & 3.8542e-03 & 1.7344e-06 & 1.8519e-02 & 1.9209e-06 & 1.7344e-06 & -- \\
				& $+ / = / -$ & + & = & + & + & - & + & + & + & + & -- \\
				\midrule
				
				\multirow{5}{*}{F26}
				& min       & 6.9567e+03 & 2.9158e+03 & 5.3866e+03 & 6.5019e+03 & 4.5869e+03 & 6.7937e+03 & 5.6539e+03 & 5.6096e+03 & 7.0802e+03 & 4.3978e+03 \\
				& mean      & 8.7869e+03 & 8.0501e+03 & 6.9265e+03 & 7.1540e+03 & 6.1813e+03 & 7.7055e+03 & 6.9036e+03 & 6.2967e+03 & 7.9663e+03 & \textbf{5.2014e+03} \\
				& std       & 7.9814e+02 & 1.4844e+03 & 5.5626e+02 & 2.8165e+02 & 4.2668e+02 & 3.2188e+02 & 8.5814e+02 & 3.0950e+02 & 3.5956e+02 & 3.7250e+02 \\
				& p-value   & 1.7344e-06 & 3.1817e-06 & 1.7344e-06 & 1.7344e-06 & 2.3534e-06 & 1.7344e-06 & 1.7344e-06 & 2.1266e-06 & 1.7344e-06 & -- \\
				& $+ / = / -$ & + & + & + & + & + & + & + & + & + & -- \\
				\midrule
				
				\multirow{5}{*}{F27}
				& min       & 3.5832e+03 & 3.3087e+03 & 3.3708e+03 & 3.4243e+03 & 3.2241e+03 & 3.4130e+03 & 3.3533e+03 & 3.3727e+03 & 3.6170e+03 & 3.3626e+03 \\
				& mean      & 3.6928e+03 & 3.7007e+03 & 3.4697e+03 & 3.5056e+03 & \textbf{3.3523e+03} & 3.5293e+03 & 3.5533e+03 & 3.4742e+03 & 3.8095e+03 & 3.5101e+03 \\
				& std       & 6.9165e+01 & 2.0067e+02 & 7.6961e+01 & 5.7607e+01 & 8.2094e+01 & 6.2647e+01 & 1.1936e+02 & 6.8163e+01 & 7.9579e+01 & 7.4959e+01 \\
				& p-value   & 2.3534e-06 & 2.0515e-04 & 6.2683e-02 & 7.9710e-01 & 1.6394e-05 & 2.1336e-01 & 1.3591e-01 & 7.1903e-02 & 1.7344e-06 & -- \\
				& $+ / = / -$ & + & + & = & = & - & = & = & = & + & -- \\
				\midrule
				
				\multirow{5}{*}{F28}
				& min       & 3.9649e+03 & 3.2934e+03 & 3.3386e+03 & 3.3921e+03 & 3.2659e+03 & 3.4541e+03 & 3.2949e+03 & 3.3482e+03 & 3.5125e+03 & 3.2795e+03 \\
				& mean      & 4.2719e+03 & 3.5165e+03 & 3.4454e+03 & 3.4593e+03 & 3.3274e+03 & 3.5400e+03 & 3.3960e+03 & 3.5346e+03 & 3.6434e+03 & \textbf{3.3162e+03} \\
				& std       & 2.3259e+02 & 9.1058e+02 & 6.6381e+01 & 4.2688e+01 & 3.5069e+01 & 6.4164e+01 & 1.1883e+02 & 8.5006e+01 & 7.9780e+01 & 2.0925e+01 \\
				& p-value   & 1.7344e-06 & 3.0650e-04 & 2.8786e-06 & 1.7344e-06 & 2.3038e-02 & 1.7344e-06 & 3.1817e-06 & 1.7344e-06 & 1.7344e-06 & -- \\
				& $+ / = / -$ & + & + & + & + & + & + & + & + & + & -- \\
				\midrule
				
				\multirow{5}{*}{F29}
				& min       & 4.8360e+03 & 4.1622e+03 & 3.9796e+03 & 4.1335e+03 & 3.8718e+03 & 5.0975e+03 & 3.7822e+03 & 4.0583e+03 & 4.6374e+03 & 3.3180e+03 \\
				& mean      & 5.9007e+03 & 5.2375e+03 & 4.6952e+03 & 4.5961e+03 & 4.2162e+03 & 5.6473e+03 & 4.4251e+03 & 4.5013e+03 & 5.4556e+03 & \textbf{3.7209e+03} \\
				& std       & 3.5822e+02 & 5.5164e+02 & 3.4550e+02 & 2.2763e+02 & 2.0845e+02 & 2.6099e+02 & 4.3643e+02 & 2.1793e+02 & 2.9113e+02 & 2.6358e+02 \\
				& p-value   & 1.7344e-06 & 1.7344e-06 & 1.9209e-06 & 2.1266e-06 & 1.2381e-05 & 1.7344e-06 & 5.2165e-06 & 1.7344e-06 & 1.7344e-06 & -- \\
				& $+ / = / -$ & + & + & + & + & + & + & + & + & + & -- \\
				\midrule
				
				\multirow{5}{*}{F30}
				& min       & 3.9616e+07 & 7.0255e+05 & 2.1169e+07 & 2.2211e+06 & 7.4070e+05 & 3.4949e+07 & 9.0822e+05 & 3.9346e+06 & 9.5912e+06 & 7.8251e+05 \\
				& mean      & 5.3810e+07 & 1.0593e+06 & 4.1245e+07 & 6.0291e+06 & 1.7456e+06 & 5.8993e+07 & 1.3220e+06 & 8.5992e+06 & 1.5306e+07 & \textbf{9.1707e+05} \\
				& std       & 9.2319e+06 & 2.7666e+05 & 1.4322e+07 & 2.5354e+06 & 6.9031e+05 & 2.6331e+07 & 3.3369e+05 & 2.5530e+06 & 3.5484e+06 & 7.8295e+04 \\
				& p-value   & 1.7344e-06 & 1.7518e-02 & 1.7344e-06 & 1.7344e-06 & 2.6033e-06 & 1.7344e-06 & 2.6033e-06 & 1.7344e-06 & 1.7344e-06 & -- \\
				& $+ / = / -$ & + & + & + & + & + & + & + & + & + & -- \\
				\midrule
				Friedman & Mean rank & 8.0690 & 4.4138 & 6.5517 & 5.6207 & 3.5862 & 8.5862 & 3.4483 & 5.2069 & 7.6897 & \textbf{1.8276} \\
				\midrule
				ToL Wilcoxon & $+ / = / -$ & 28/1/0 & 21/5/3 & 28/1/0 & 24/5/0 & 23/2/4 & 28/1/0 & 20/6/3 & 24/2/3 & 25/1/3 & -- \\
				\bottomrule
			\end{tabular}
		\end{adjustbox}
	\end{sidewaystable*}
	
	\clearpage
	\endgroup
	
\end{document}